\documentclass[10pt, authoryear, final]{elsarticle}

\usepackage{amssymb}
\journal{Elsevier}

\usepackage{pifont}
\usepackage{xcolor}
\newcommand{\cmark}{\ding{51}}
\newcommand{\xmark}{\ding{55}}
\usepackage{booktabs}
\usepackage{hyperref}
\usepackage[ruled,vlined,algosection,linesnumbered]{algorithm2e}
\usepackage{subfig}
\usepackage{amsmath,amsthm}
\usepackage{multirow}
\usepackage{bbm}
\usepackage{calc}
\usepackage[symbol]{footmisc}

\begin{document}

\begin{frontmatter}

%% Title, authors and addresses

%% use the tnoteref command within \title for footnotes;
%% use the tnotetext command for theassociated footnote;
%% use the fnref command within \author or \affiliation for footnotes;
%% use the fntext command for theassociated footnote;
%% use the corref command within \author for corresponding author footnotes;
%% use the cortext command for theassociated footnote;
%% use the ead command for the email address,
%% and the form \ead[url] for the home page:
%% \title{Title\tnoteref{label1}}
%% \tnotetext[label1]{}
%% \author{Name\corref{cor1}\fnref{label2}}
%% \ead{email address}
%% \ead[url]{home page}
%% \fntext[label2]{}
%% \cortext[cor1]{}
%% \affiliation{organization={},
%%            addressline={}, 
%%            city={},
%%            postcode={}, 
%%            state={},
%%            country={}}
%% \fntext[label3]{}

\title{Reconstructing Compact Building Models from Point Clouds Using Deep Implicit Fields}

\author[tud,tum]{Zhaiyu Chen\footnote[2]{Work done at Delft University of Technology}} \ead{zhaiyu.chen@tum.de} 
\author[tud]{Hugo Ledoux} \ead{h.ledoux@tudelft.nl}
\author[tud]{Seyran Khademi} \ead{s.khademi@tudelft.nl}
\author[tud]{Liangliang Nan\footnote[1]{Corresponding author}} \ead{liangliang.nan@tudelft.nl}

\affiliation[tud]{organization={Delft University of Technology}, postcode={2628 BL}, city={Delft}, country={The Netherlands}}

\affiliation[tum]{organization={Technical University of Munich}, postcode={80333}, city={Munich}, country={Germany}}

\begin{abstract}

%% Text of abstract
While three-dimensional (3D) building models play an increasingly pivotal role in many real-world applications, obtaining a compact representation of buildings remains an open problem. In this paper, we present a novel framework for reconstructing compact, watertight, polygonal building models from point clouds.
Our framework comprises three components: \textit{(a)} a cell complex is generated via adaptive space partitioning that provides a polyhedral embedding as the candidate set; 
\textit{(b)} an implicit field is learned by a deep neural network that facilitates building occupancy estimation; 
\textit{(c)} a Markov random field is formulated to extract the outer surface of a building via combinatorial optimization. 
We evaluate and compare our method with state-of-the-art methods in generic reconstruction, model-based reconstruction, geometry simplification, and primitive assembly. 
Experiments on both synthetic and real-world point clouds have demonstrated that, with our neural-guided strategy, high-quality building models can be obtained with significant advantages in fidelity, compactness, and computational efficiency. 
Our method also shows robustness to noise and insufficient measurements, and it can directly generalize from synthetic scans to real-world measurements. The source code of this work is freely available at \url{https://github.com/chenzhaiyu/points2poly}.
\end{abstract}

%%Graphical abstract
% \begin{graphicalabstract}
% \includegraphics{grabs}
% \end{graphicalabstract}

%%Research highlights
% \begin{highlights}
% \item Research highlight 1
% \item Research highlight 2
% \end{highlights}


\end{frontmatter}

%% main text
\section{Introduction}

Three-dimensional (3D) building models have been playing an increasingly important role in various applications, such as urban planning~\citep{herbert2015comparison}, solar potential analysis~\citep{machete2018use}, noise pollution assessment~\citep{stoter2020automated,15_ijgi_3dapps}. 
Recently, with the development of augmented and virtual reality applications, the demand for high-quality 3D models of buildings has grown even faster~\citep{blut2021three}.

\begin{figure}[ht!]
	\centering
	
	\subfloat[Dense triangles]{\includegraphics[width=0.26\linewidth]{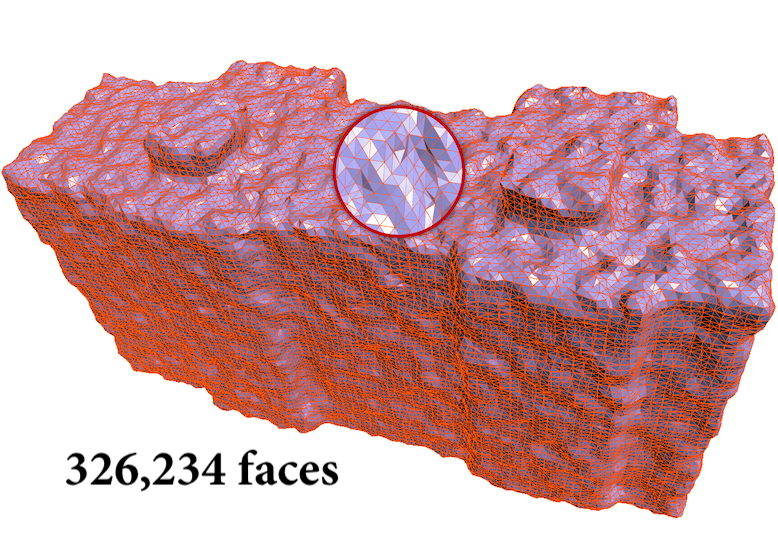}\label{fig:dense_triangles}}
	\hspace{1em}
	\subfloat[Sparse triangles]{\includegraphics[width=0.26\linewidth]{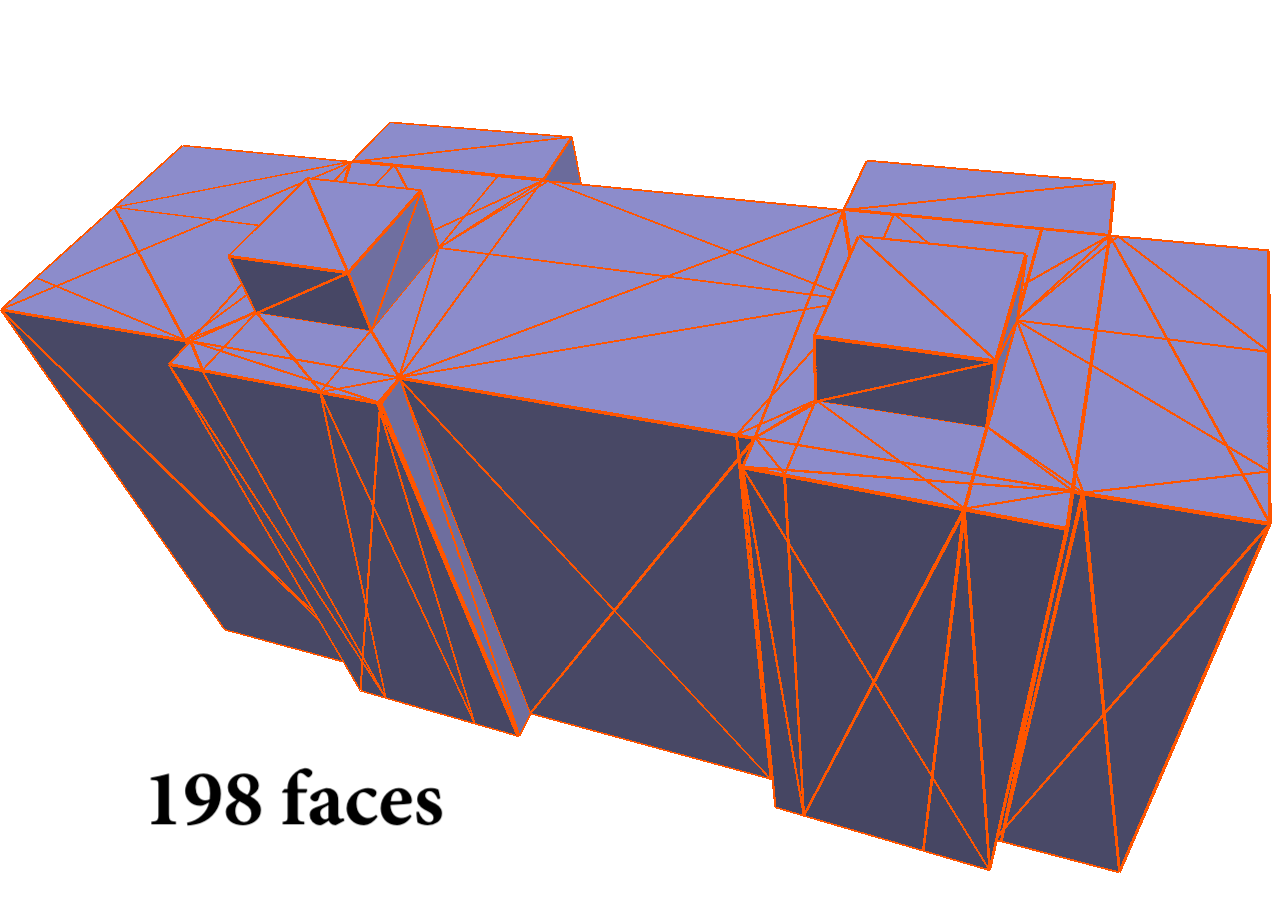}}
	\hspace{1em}
	\subfloat[Polygonal surface]{\includegraphics[width=0.26\linewidth]{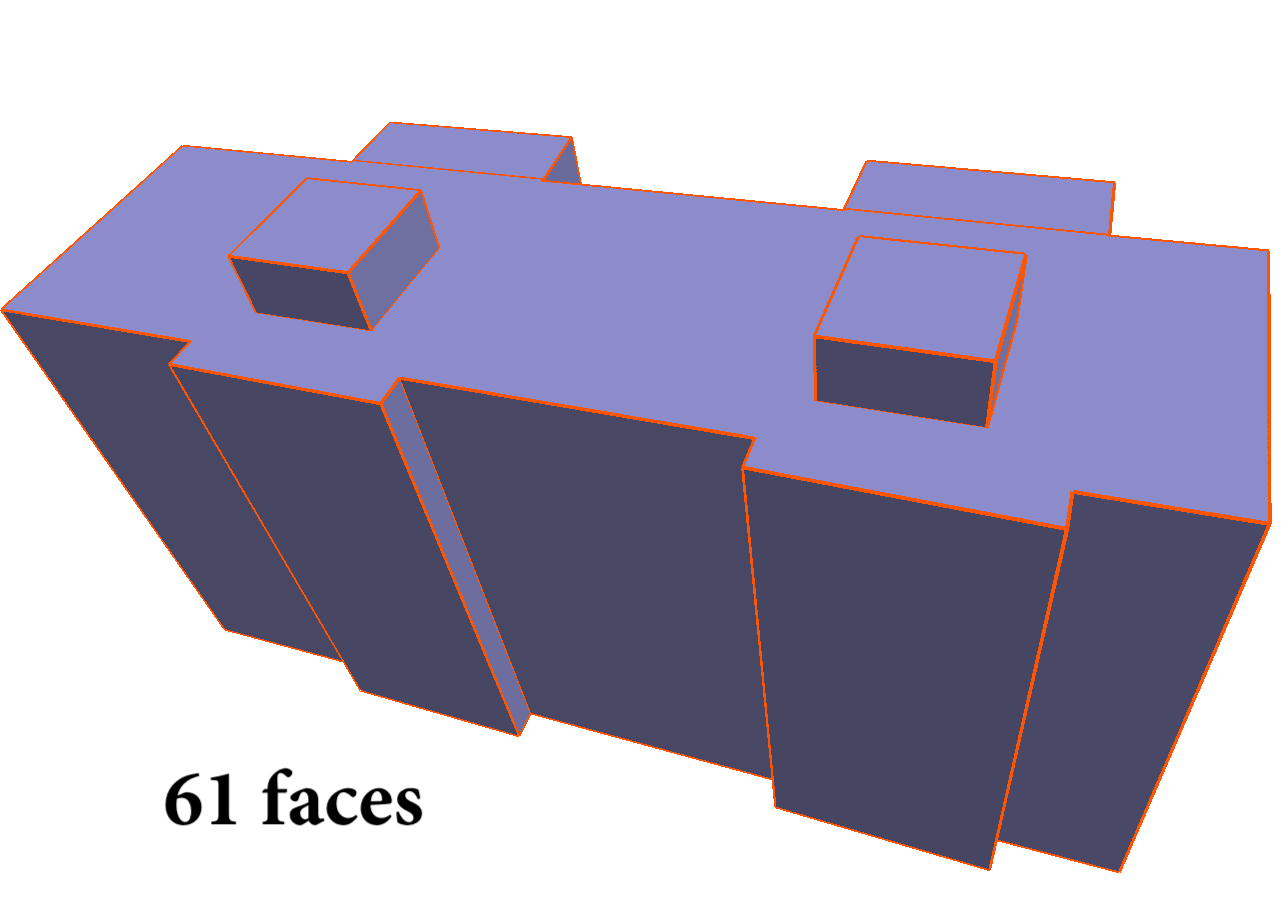}}
	
	\caption{A building in three different surface representations.}
	\label{fig:surface_representations}
\end{figure}

Most existing generic 3D reconstruction methods are dedicated to smooth surfaces represented as dense triangles, irrespective of piecewise planarity exhibited in the urban environment~\citep{kazhdan2006poisson,erler2020points2surf}. 
Compared to the dense triangle mesh representation, a compact surface model has a significantly low number of faces yet can still sufficiently describe the geometry of a building. \autoref{fig:surface_representations} illustrates a building described as a smooth surface and piecewise-planar ones, where an arbitrary-sided polygonal surface exhibits the highest compactness. 
Although some works claim the possibility of reconstructing compact surface models from point clouds~\citep{boulch2014piecewise,mura2016piecewise,li2016reconstructing,nan2017polyfit} or from dense triangle meshes~\citep{bouzas2020structure,li2021feature}, they suffer from serious scalability issues. 
In this paper, we aim at a robust reconstruction of compact building surfaces directly from point clouds.

We exploit the fact that 3D shapes are not confined to explicit representations (e.g., point cloud, surface mesh, voxels), but can be encoded implicitly in a function space.
For example, implicit fields are widely used to characterize 3D shapes, where the surface of a shape is implicitly interpreted as a zero-set of the signed distance field (SDF)~\citep{kazhdan2006poisson}. 
A learnable indicator function of the SDF takes as input a query point and yields an indication of whether the point lies inside or outside the object. 
Then explicit geometry is often extracted from the field via computationally expensive iso-surfacing~\citep{mescheder2019occupancy}. 
Compared with explicit expressions that are heterogeneously distributed, the homogeneous functional representation is particularly favorable for geometric machine learning. 
Recently, \cite{park2019deepsdf} introduced a learning-based scheme with the use of the function space in 3D geometric modeling. An implicit field can be directly learned from the input point cloud and used to extract a smooth surface model of the object. However, extracting a compact polygonal model from the implicit field remains an open problem.

In this paper, we propose a novel framework for reconstructing compact, watertight, polygonal building meshes from point clouds by incorporating learnable implicit surface representation into explicit geometry construction.
The explicit geometry provides a polyhedral embedding as the candidate set, from which the building surface can be obtained from an implicit field learned using a deep neural network (i.e., neural-guided).
We formulate the surface extraction problem as a Markov random field (MRF), which can handle varying surface complexity. Through combinatorial optimization, the occupancy of a building inferred from a deep implicit field is further regularized by penalizing its surface complexity.
Our reconstruction framework inherits the high efficiency and robustness in the inference of deep implicit fields, and the high fidelity of primitive assembly-based reconstruction approaches.

With our neural-guided framework, we demonstrate that high-quality building models can be obtained with significant advantages in terms of fidelity, compactness, and computational efficiency compared to state-of-the-art methods. The main contributions of this paper are as follows:
\begin{enumerate}[(i)]
	\item A learning-based framework for compact building model reconstruction from point clouds. 
	To the best of our knowledge, this is the first work where a deep implicit field is used for building reconstruction.  
	Our method shows significant performance and quality advantage over state-of-the-art methods, especially for complex building models.
	\item An MRF formulation for surface extraction from the occupancy learned by a neural network. Our formulation allows complexity control and favors compactness in the final reconstruction, and is far more efficient than the existing integer programming formulation.
	\item The first point cloud dataset of synthetic buildings (with their corresponding surface models) for cultivating learning-based building reconstruction methods. The dataset contains 768 buildings with varying complexity and styles. Noise and scanning artifacts (e.g., occlusion) are simulated and added to the point clouds.

\end{enumerate}

\section{Related work}

There is a large volume of literature on shape reconstruction from point clouds. In this section, we mainly review approaches aiming at obtaining compact polygonal models, namely model-based reconstruction, geometric simplification, and primitive assembly. We also discuss recent development in deep implicit fields, since it is one of the key components in our learning-based reconstruction framework.

\subsection{Model-based reconstruction}
Urban buildings typically demonstrate specific types of structures, such as roofs,  walls, and dormers, which allows the use of strong shape priors and generic templates of such structures to ease the reconstruction. This type of approach is commonly referred to as model-based reconstruction.

The Manhattan-world assumption~\citep{coughlan2000manhattan} restricts the orientation of faces in only three orthogonal directions and represents the 3D scene with axis-aligned non-uniform polycubes~\citep{ikehata2015structured,li2016manhattan,li2016fitting}. 
This simplification drastically reduces the geometric complexity and the solution space to explore. 
Another common assumption is restricting the output surface to specific disk topologies. 
The 2.5D view-dependent representation~\citep{zhou20102} can generate arbitrarily shaped roofs with vertical walls connecting them, from airborne light detection and ranging (LiDAR) measurements. \citet{xiong2014graph,xiong2015flexible} exploit roof topology graphs for reconstructing LoD2 buildings. 
Similarly, \citet{li2016reconstructing} present a workflow to reconstruct building roofs for urban scenes.
\citet{kelly2017bigsur} formulate a global optimization to produce structured urban reconstruction from street-level imagery, GIS footprint, and coarse 3D mesh.
The model-based approaches can maintain the uniformity of the reconstruction and is thus efficient to implement. 
However, they only apply to specific domains as a limited variety of the models constrains the generalization of these methods. 
Our reconstruction method, instead, imposes no geometric assumption except for piecewise planarity, thus remaining generic.

\subsection{Geometric simplification}
This type of method obtains compact surface models by the simplification of given dense triangle meshes. 
Dense triangle meshes can be obtained from point clouds with generic surface reconstruction techniques such as the Poisson reconstructions~\citep{kazhdan2006poisson, kazhdan2013screened}, which address the problem by establishing an indicator function underpinned by the points with their oriented normals. 
The output surface is acquired by extracting an iso-surface of this function. 
We refer to \citet{berger2017survey} for a survey of generic surface reconstruction algorithms.

Dense triangles on a smooth surface can be simplified into concise polygons via various approaches.
\citet{garland1997surface} propose to iteratively contract vertex pairs under quadric error metrics (QEM), such that the number of faces is reduced to an expected number. 
To preserve the piecewise-planar structure during contraction, \citet{salinas2015structure} propose structure-aware mesh decimation (SAMD), which incorporates the planar proxies detected from pre-processing analysis into an adjacency graph and then approximates the mesh as well as the proxies. 
However, these contraction operators are inclined to more triangles than desired for piecewise-planar building representation. 
Alternatively, variational shape approximation (VSA) proposed by~\citet{cohen2004variational} approaches the approximation through repeatedly partitioning based on an error tolerance.
Furthermore, \citet{verdie2015lod} propose an approach for reconstructing urban scenes with various LoD configurations from dense meshes through classification, abstraction, and reconstruction.
Similarly, \citet{zhu2018large} present a simplification framework for urban scene modeling at different LODs, which is subsequently extended by \citet{han2021urban} to more general scenes.
\citet{bouzas2020structure} incorporates structure awareness with the recovery and preservation of the initial mesh primitives and their adjacencies. \citet{li2021feature} further exploit the preservation of piecewise-planar structures and sharp features.
These methods demand the input smooth surface be accurate in both geometry and topology for a faithful surface approximation. 
However, the requirement is rarely satisfied for real-world measurements. 
In contrast, our proposed reconstruction method can directly output a concise polygonal mesh without approximating an intermediate.

\subsection{Primitive assembly}
This type of method reconstructs compact building models by pursuing the optimal assembly of a set of basic geometric primitives (e.g., polygons, boxes).

Connectivity-based methods address the primitive assembly by extracting proper geometric primitives from an adjacency graph built on planar shapes~\citep{chen2008architectural,schindler2011classification,van2011shape}. 
Though the graph analysis can be efficiently executed, these methods are sensitive to the quality of the adjacency graph. Incorrect connectivity contaminated by linkage errors is prone to an incomplete reconstruction. 
\citet{arikan2013snap} propose an interactive solution that enables the user to complete the surface through an optimization-based snapping, which requires laborious human interventions for complex scenes.
\cite{labatut2009hierarchical,lafarge2013surface} propose a mixed strategy where the confident areas are represented by polygonal shapes and the complex regions by dense triangles. 

Slicing-based methods show stronger robustness to imperfect data with the divide-and-conquer strategy~\citep{chauve2010robust,mura2016piecewise,nan2017polyfit}. 
They partition the 3D space into polyhedral cells using the supporting planes of the detected planar primitives, where the polyhedral cells consist of polygonal faces. 
The reconstruction is therefore transformed into a labeling problem where the polyhedral cells are labeled as either inside or outside the shape or equivalently with labeling other primitives. 
The main limitation of slicing-based methods is the scalability of their data structure. Specifically, the pairwise intersection of supporting planes results in an over-complex tessellation, which is computationally expensive to compute, commonly via a binary tree updated upon each primitive's insertion~\citep{murali1997consistent}. 
When many planar primitives contribute to the intersection, the resulting tessellation may hamper the surface extraction. 
Moreover, since many anisotropic cells are generated regardless of their spatial hierarchy, the resulting surface is inclined to geometric artifacts. 
For instance, PolyFit~\citep{nan2017polyfit} formulates polygonal surface reconstruction as a binary integer program with hard constraints ensuring the generated surface is watertight and manifold. 
However, it suffers from scalability issues due to unnecessary pairwise intersections, and it thus can only process simple models with limited complexity.
In this work, we address the computation bottleneck by exploiting an adaptive space partitioning strategy, which significantly reduces the algorithmic complexity. Our adaptive strategy is similar to the kinetic data structure in KSR~\citep{bauchet2020kinetic} that creates a partition by growing planar primitives at a constant speed until they collide and form polyhedra, avoiding exhaustive partitioning of the space. Instead of relying on a sophisticated kinetic data structure, our method only requires a simple data structure that stems from binary space partitioning (BSP). Moreover, our occupancy indicator does not rely on normal information that is required by KSR, which enables a broader spectrum of inputs.

Recently, \citet{li2021relation} extend PolyFit to further exploit the inter-relation of the primitives into procedural modeling for building reconstruction in the CityGML format. 
Similarly, \citet{xie2021combined} propose to combine the rule-based and the hypothesis-based strategies for efficient building reconstruction. \citet{fang2020connect} introduce a hybrid approach for reconstructing 3D objects by successively connecting and slicing planes detected from 3D data.
In contrast to these works that use hand-crafted features, our method exploits automatically learned deep features.

\subsection{Deep implicit field}

Recent advances in deep implicit fields have revealed their potential for 3D reconstruction. 
The crux of these methods is to learn a mapping from the input (e.g., a point cloud) to a continuous scalar field. Then, the surface of the object can be extracted via iso-surfacing techniques such as Marching Cubes~\citep{lorensen1987marching}. 
Iso-surfacing is powerful in extracting smooth surfaces but has the limitation of preserving sharp features. Inevitably, it introduces discretization errors. Deep implicit fields are thus not natively suitable for reconstructing compact polygonal models.

By incorporating constructed solid geometry (CSG), \citet{chen2020bsp} introduce an end-to-end neural network, BSP-Net, to reconstruct a shape from a set of convexes obtained via binary space partitioning. 
Similarly, \citet{deng2020cvxnet} propose an architecture to represent a low-dimensional family of convexes. 
These two methods both learn to divide and conquer the 3D space with implicit fields. 
However, the inputs to these two neural networks are either images or voxels, instead of point clouds that our work aims to address. 

The single latent feature vector used by most deep implicit fields methods implies strong priors dependent on the training data. 
While this allows plausible surface reconstruction even with highly contaminated data, it significantly limits the generalization ability of these methods. 
With one feature vector encoding the whole shape, the feature space inevitably overfits the shapes in the training set, which may fail for shapes from unseen categories.
\citet{erler2020points2surf} propose the Points2Surf architecture to estimate an SDF with both local and global feature vectors, which has shown great generalization capabilities in implicit field learning. 
In this work, we utilize the Points2Surf architecture as an initialization step for 3D reconstruction.

\subsection{Summary}

\autoref{tab:relatedwork} summarizes the existing works related to ours, including generic reconstruction methods~\citep{kazhdan2006poisson,erler2020points2surf}, model-based approaches~\citep{zhou20102,li2016manhattan}, geometry simplification methods~\citep{garland1997surface,cohen2004variational,salinas2015structure}, and primitive assembly~\citep{nan2017polyfit,bauchet2020kinetic}.
Verdicts are based on whether each method can produce compact and watertight surfaces, whether it consumes unorganized points as input (i.e., without normal information and geometric approximation or reconstruction), whether it applies to generic 3D objects, and its scalability. 
Among these competitors, PolyFit and KSR are both capable of reconstructing compact, watertight, generic surfaces from point clouds and thus are considered the closest to ours. Our learning-based reconstruction framework combines the strengths of primitive assembly and deep implicit fields. It not only inherits the theoretical guarantee of primitive assembly to obtain compact building models but also integrates the data-driven features of deep implicit fields such that various types of buildings can be represented with strong robustness.
\begin{table*}[hpt]
	\centering
	\caption{Overview of the most related works. GR, MR, GS, and PA stand for generic reconstruction, model-based reconstruction, geometry simplification, and primitive assembly, respectively. All the listed works are compared with our method in the experiments.}
	
	{
		\tiny
		\begin{tabular}[width=1.0\linewidth]{lcccccc}
			
			\toprule
			\textbf{Related} & \multicolumn{5}{c}{\textbf{Characteristics}} \\
			% & \multirow{2}{*}{\textbf{Compared}} \\
			\cmidrule(lr){2-6}
			\textbf{work} & \textbf{Category} & \textbf{Compact} & \textbf{Watertight} & \textbf{Raw} & \textbf{Generic} & \textbf{Scalable} \\
			\midrule
			
			Poisson~\citep{kazhdan2006poisson} & GR & \xmark & \xmark & \xmark & \cmark & \cmark
			\\
			
			Points2Surf~\citep{erler2020points2surf} & GR & \xmark & \xmark & \cmark & \cmark & \xmark
			\\
			
			2.5D DC~\citep{zhou20102} & MR & \xmark & \cmark & \cmark & \xmark & \cmark
			\\
			
			Manhattan-world~\citep{li2016manhattan} & MR & \cmark & \cmark & \xmark & \xmark & \cmark
			\\
			
			QEM~\citep{garland1997surface} & GS & \cmark & \xmark & \xmark & \cmark & \xmark
			\\

			VSA~\citep{cohen2004variational} & GS & \cmark & \xmark & \xmark & \cmark & \xmark
			\\
	
			SAMD~\citep{salinas2015structure} & GS & \cmark & \xmark & \xmark & \cmark & \xmark
			\\
			
			FPS~\citep{li2021feature} & GS & \cmark & \xmark & \xmark & \cmark & \xmark
			\\
		
			PolyFit~\citep{nan2017polyfit} & PA & \cmark & \cmark & \cmark & \cmark & \xmark
			\\
			KSR~\citep{bauchet2020kinetic} & PA & \cmark & \cmark & \xmark & \cmark & \cmark			\\
			Ours & PA & \cmark & \cmark & \cmark & \cmark & \cmark
			\\
			
			\bottomrule             
		\end{tabular}
	}
	\label{tab:relatedwork}
\end{table*}

\section{Methodology}

\subsection{Overview}

Our neural-guided approach for building reconstruction from point clouds utilizes the learned implicit representation as an occupancy indicator for extracting the final surface model.
The indicator can be intuitively interpreted as a shape-conditioned binary classifier for which the decision boundary represents the outer surface of a building. 
With the learned implicit field, a compact surface model is extracted using an MRF formulation that accounts for reconstruction fidelity and compactness. 
\autoref{fig:overview_concept} demonstrates the workflow of our neural-guided reconstruction framework that consists of three steps:

\begin{figure*}[th!]
	\centering
	\includegraphics[width=0.85\linewidth]{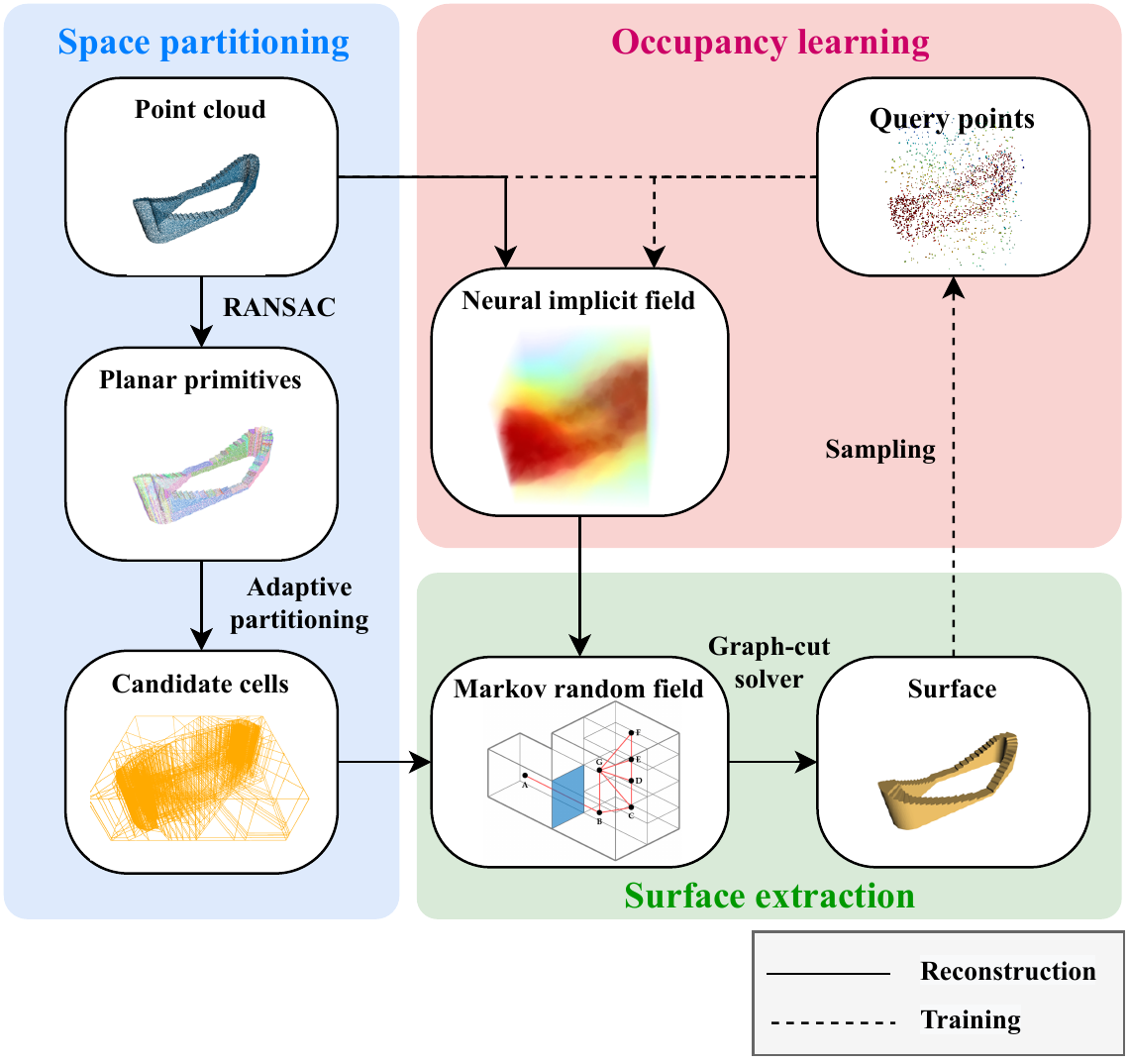}
	\caption{Overview of our framework.
		The framework comprises three functional blocks: within the explicit block (in blue), a linear cell complex is generated from a point cloud by adaptive space partitioning; within the implicit block (in red), a neural implicit field is learned to indicate spatial occupancy of the building; within the optimization block (in green), an MRF is formulated to extract the final surface.}
	\label{fig:overview_concept}
\end{figure*}

\begin{itemize}
	\item \textbf{Adaptive space partitioning}. We partition the ambient 3D space to generate a linear cell complex that complies with planar primitives detected from the point cloud. 
	The partitioning is spatially adaptive therefore being efficient and respective to the building's geometry. 
	The non-overlapping cells in the complex serve as the candidates whose outer shell constitutes the final surface.
	\item \textbf{Occupancy learning}. We utilize a deep neural network to learn a shape-conditioned implicit field that characterizes the building object represented by the point cloud. 
	The implicit field describes the spatial occupancy of the object given any query point in the 3D space.
	\item \textbf{Surface extraction}. This step takes the learned implicit field as an occupancy indicator and outputs the boundary representation of the building's surface. 
	We formulate surface extraction as a binary classification problem and solve it using MRF optimization that encourages compactness and guarantees the final model to be watertight.
	
\end{itemize}

\subsection{Adaptive space partitioning}
\begin{figure*}[th!]
	\centering
	\includegraphics[width=0.97\linewidth]{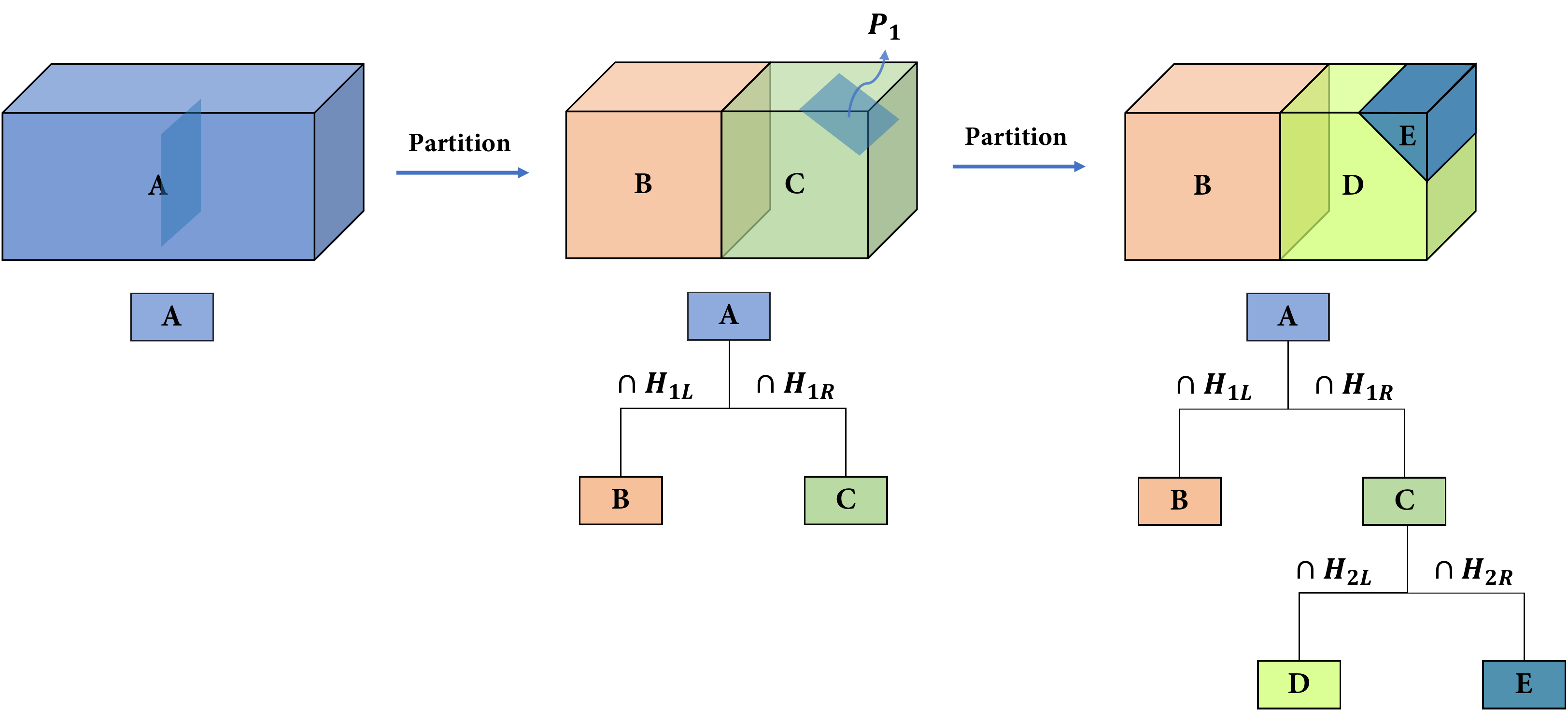}
	\caption{An illustration of one step in space partition. After inserting a planar primitive $P_1$, cell $C$ is split into two cells $D$ and $E$.}
	\label{fig:adjacency}
\end{figure*}

We introduce an adaptive partitioning algorithm that adaptively subdivides the 3D space into a set of cells (i.e., a linear cell complex), where each cell is a convex polyhedron.
Our algorithm partitions the space by extracting planes from the point cloud followed by BSP using the refined planar primitives. 
Specifically, we apply the RANSAC algorithm of \citet{schnabel2007efficient} to detect planes.  
Considering noise and outliers in the data, we perform a refinement procedure that iteratively merges planes under specific proximity conditions (see Algorithm~\ref{alg:refinement}).  
During the partitioning, a binary tree structure is dynamically and locally updated upon insertion of a primitive, and the cell adjacency information is maintained. 
A typical space partitioning step is illustrated in Figure~\ref{fig:adjacency}. 

\begin{figure*}[th!]
	\centering
	\subfloat[Exhaustive partitioning]{\includegraphics[width=0.3\linewidth]{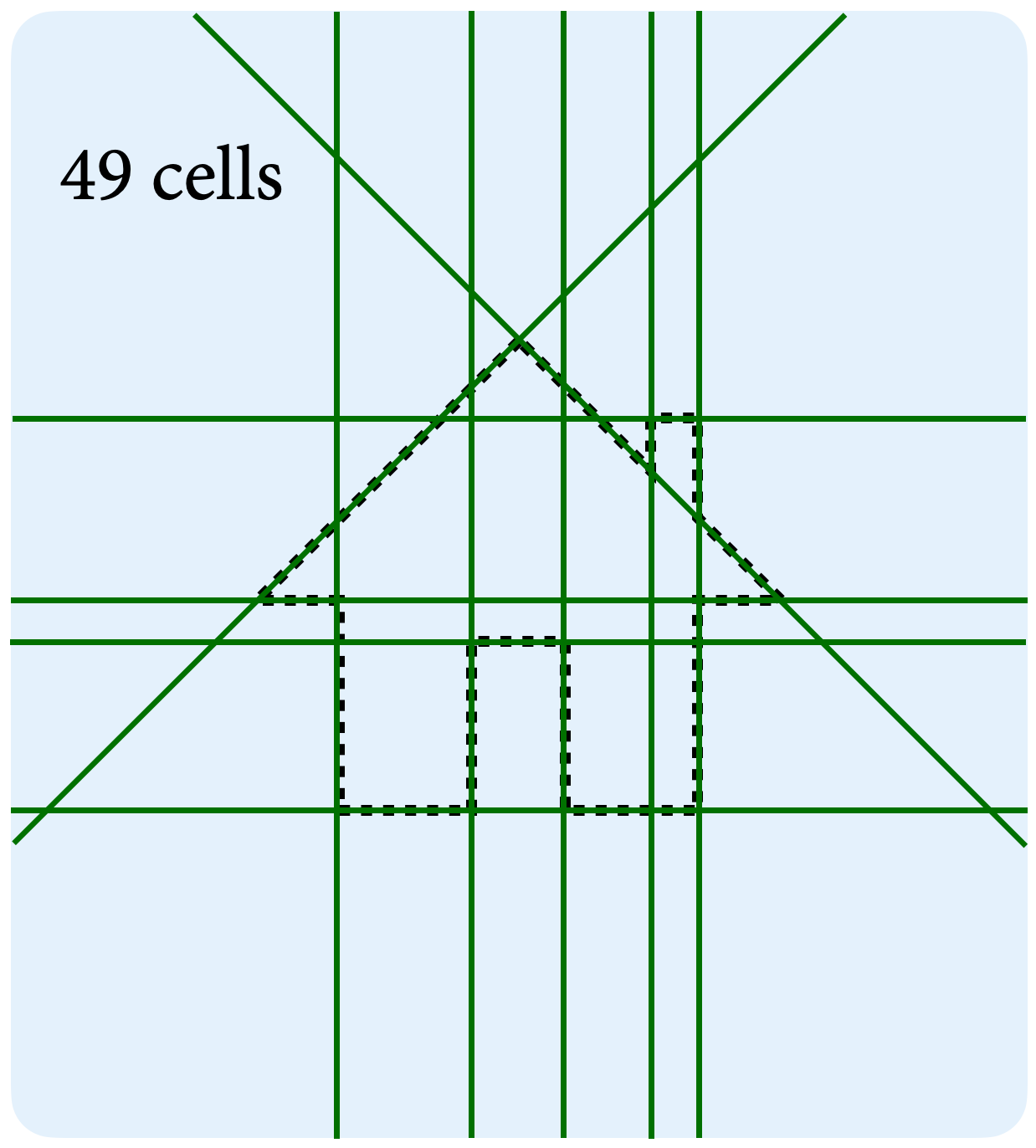}}
	\hspace{2.0em}
	\subfloat[Adaptive partitioning]{\includegraphics[width=0.3\linewidth]{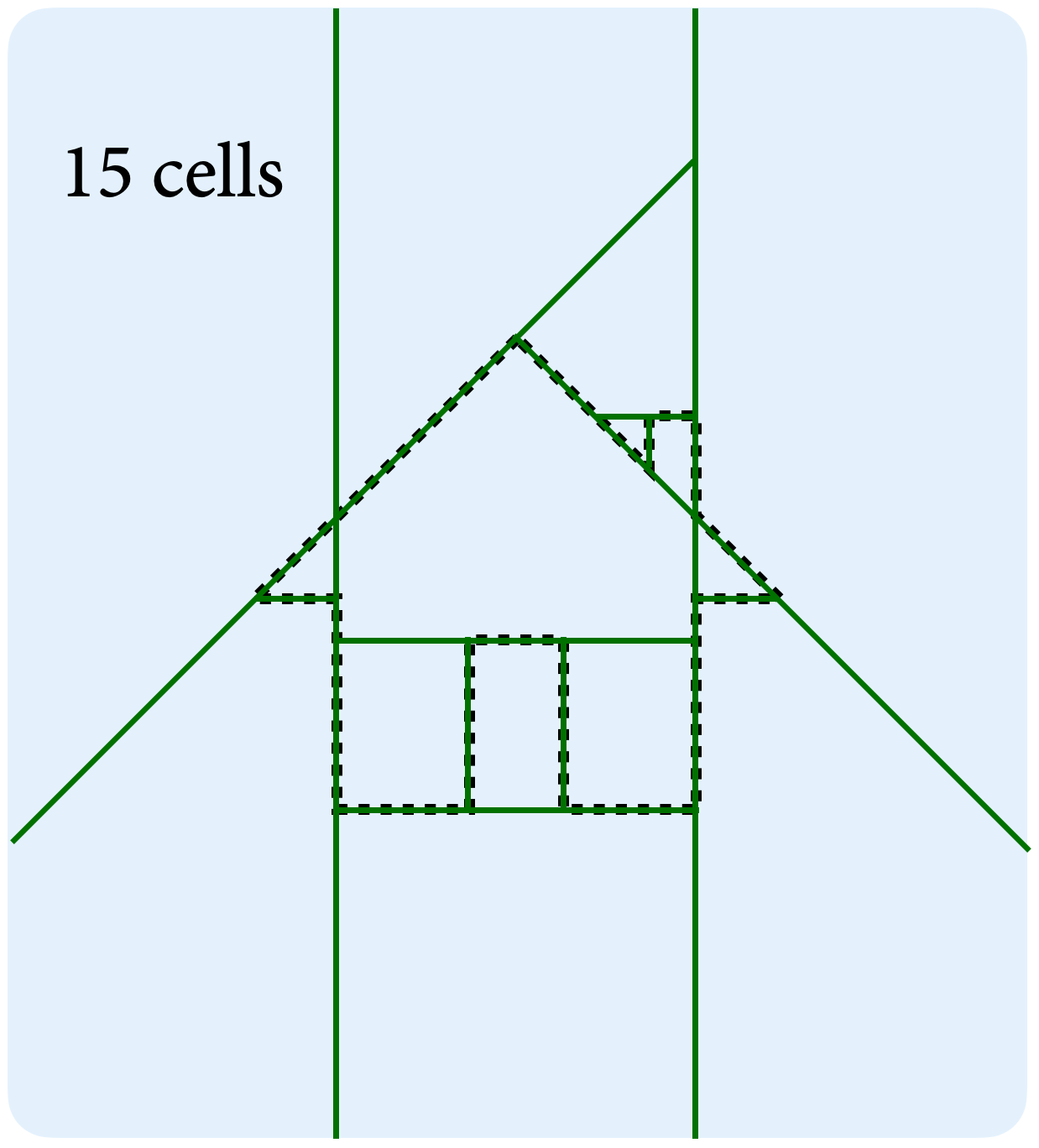}}
	\caption{Comparison of the exhaustive and adaptive partitioning strategies (a 2D example). Note the difference in the numbers of the resulted cells.}
	\label{fig:exhaustiveadaptive}
\end{figure*}

To avoid redundant partitioning, our adaptive strategy only allows intersecting spatially correlated primitives, as illustrated in Figure~\ref{fig:exhaustiveadaptive} (b).
We describe the spatial correlation by intersection test between the axis-aligned bounding box (AABB) of a primitive and the cells in the leaf nodes of the BSP tree.
Compared to exhaustive partitioning strategies (e.g., the pairwise intersection of planar primitives in~\cite{nan2017polyfit}), our adaptive partitioning can avoid unnecessary intersections between primitives and generate fewer yet more meaningful cells, which would otherwise result in computational overhead or potential artifacts in the final model. 
Figure~\ref{fig:exhaustiveadaptive} shows an illustrative comparison of the exhaustive and our adaptive partitioning strategies. 
It is worth noting that in the 3D space, our adaptive space partitioning can significantly reduce the number of cells, allowing an efficient process in the subsequent steps.

\begin{figure*}[th!]
	\centering
	\subfloat[Input]{\includegraphics[width=0.3\linewidth]{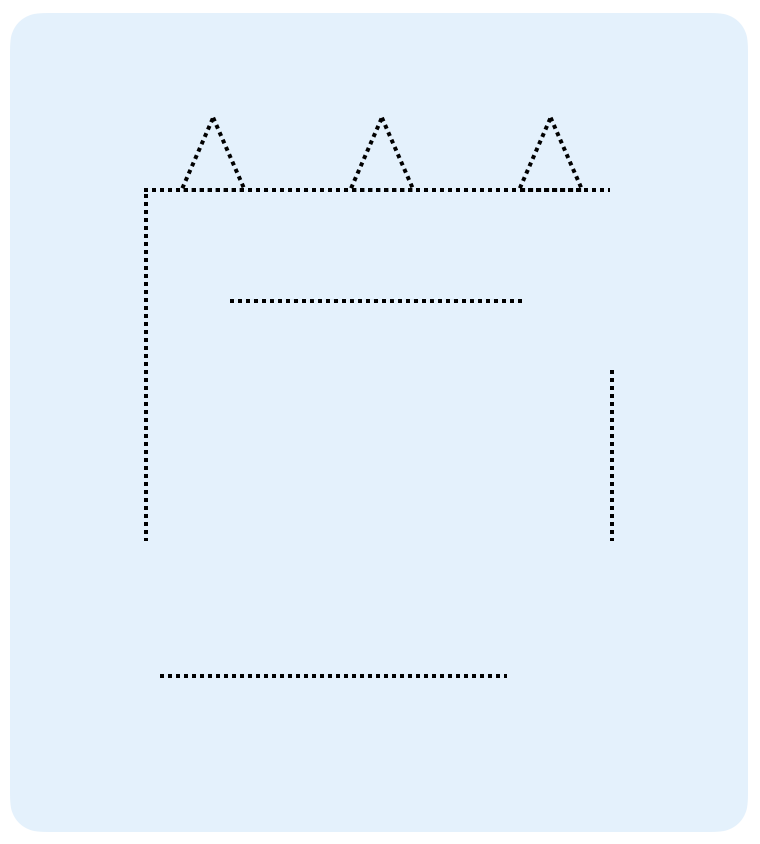}}
		\hspace{1em}
	\subfloat[Without priority]{\includegraphics[width=0.3\linewidth]{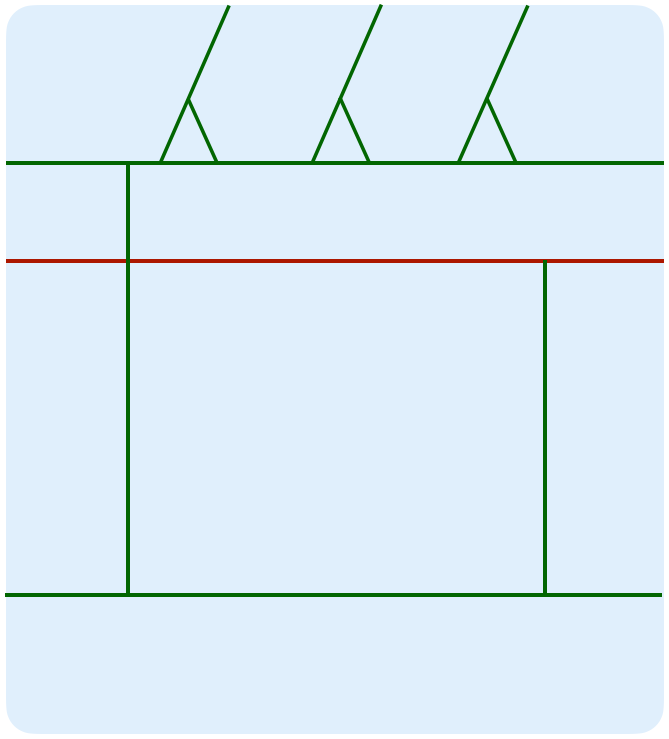}\label{fig:priorityimpact_missing}}
	\hspace{1em}
	\subfloat[With priority]{\includegraphics[width=0.3\linewidth]{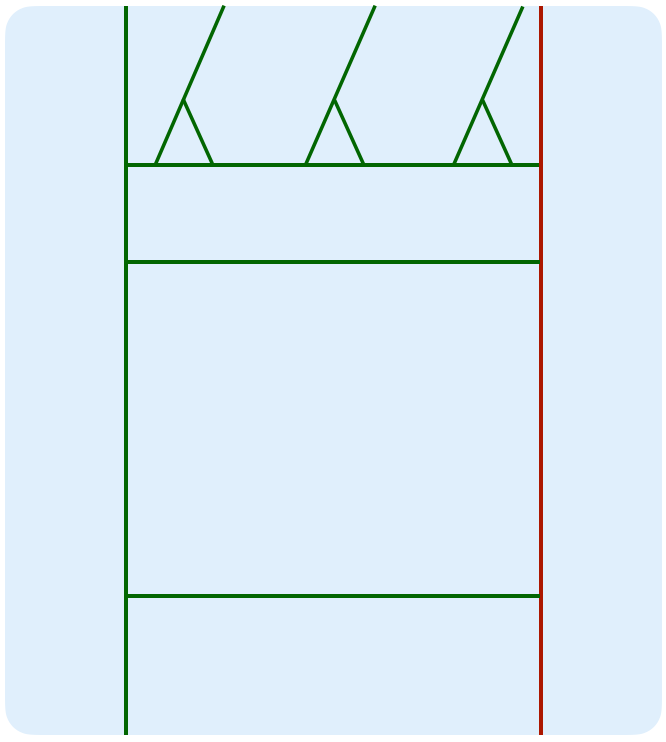}\label{fig:priorityimpact_optimal}}

%	\subfloat[Random]{\includegraphics[width=0.2\linewidth]{figures/methodology/priority_random.png}}
	
	\caption{An illustration of the effect of the verticality priority in space partitioning. 
		(a) Incomplete input. 
		(b) Without the verticality priority. Inserting the planar primitive denoted by the red line results in a missing wall in the upper-right part of the building.
		(c) With the verticality priority. By first inserting the vertical planar primitive (in red), a plausible vertical wall is inferred.}
	\label{fig:priorityimpact}
\end{figure*}

During the partitioning, the BSP tree structure and adjacency information of the cells are incrementally obtained, as shown in Figure~\ref{fig:adjacency}. Thus the partitioning result depends highly on the order of inserting the planes. To this end, we prioritize the partitioning process such that planes conforming to building structural priors play dominant roles in partitioning the space.
Considering that buildings typically demonstrate vertical walls, our space partitioning algorithm gives vertical planes a higher priority to be involved in partitioning. 
We define the verticality of a plane as $\mathcal{V} = 1 - |n_z|$, where $n_z$ denotes the $z$ coordinate of the plane's normal vector. 
A zero verticality value indicates a perfectly horizontal plane and a value of 1 indicates a perfectly vertical plane. 
Note that the verticality priority is only applied to planes whose verticality is greater than a threshold value (0.9 in our work). 
For the remaining planes, the priority is determined by the number of points in each plane because a larger number of points often implies a higher confidence value. 
The verticality priority ensures that vertically oriented planes partition the space before other planes do, which has an advantage in handling partially occluded facades. \autoref{fig:priorityimpact} (b) shows such an example.

\begin{algorithm}[th!]
	\KwIn{Raw planar segments $\mathcal{S}$, angle tolerance $\theta$ and distance tolerance $\epsilon$}
	\KwOut{Refined planar segments $\widetilde{\mathcal{S}}$}
	\BlankLine
	
	$\mathcal{Q}$ $\leftarrow$ initialize priority queue\;
	
	\For{$(i, j) \in \mathcal{S}$}
	{
		$\alpha_{i, j}$ $\leftarrow$ compute angle between $\mathcal{S}_{i}$ and $\mathcal{S}_{j}$\;
		$\mathcal{Q}$ $\leftarrow$ push $(\alpha_{i, j}, i, j)$ ordered by $\alpha_{i, j}$\;
	}
	
	\While{$\mathcal{Q}$ not empty}
	{
		$(\alpha_{i, j}, i, j)$ $\leftarrow$ pop from $\mathcal{Q}$ with the smallest $\alpha_{i, j}$\;
		$d_{i, j}$ $\leftarrow$ compute distance between $\mathcal{S}_{i}$ and $\mathcal{S}_{j}$\;
		% \eIf{$d_{i, j} < \epsilon$ and $\mathcal{S}_{i} \notin \widetilde{\mathcal{S}}
			\eIf{$\alpha_{i, j} < \theta$ and $d_{i, j} < \epsilon$}
			% $ and $\mathcal{S}_{j} \notin \widetilde{\mathcal{S}}$}
		{   
			$m \leftarrow $ merge $\mathcal{S}_{i}$ and $\mathcal{S}_{j}$ with new plane parameters by PCA\;
			$\alpha_{m, \mathbf{n}} \leftarrow$ compute angle between $m$ and every plane $\mathbf{n}$ in $\mathcal{Q}$\;
			$\mathcal{Q} \leftarrow$ push $(\alpha_{m, \mathbf{n}}, m, \mathbf{n})$ ordered by $\alpha_{m, \mathbf{n}}$\;
		}
		{
%			\tcp{no more coplanar pairs can exist in the queue}
			$\widetilde{\mathcal{S}} \leftarrow$ extract planar segments from $\mathcal{Q}$\;
			break\;
		}
		
		\Return{$\widetilde{\mathcal{S}}$}
		
	}
	\caption[Refine]{P\textsc{lane refinement} ($\mathcal{S}$)}
	\label{alg:refinement}
\end{algorithm}

\begin{figure*}[th!]
	\centering
	\includegraphics[width=0.75\linewidth]{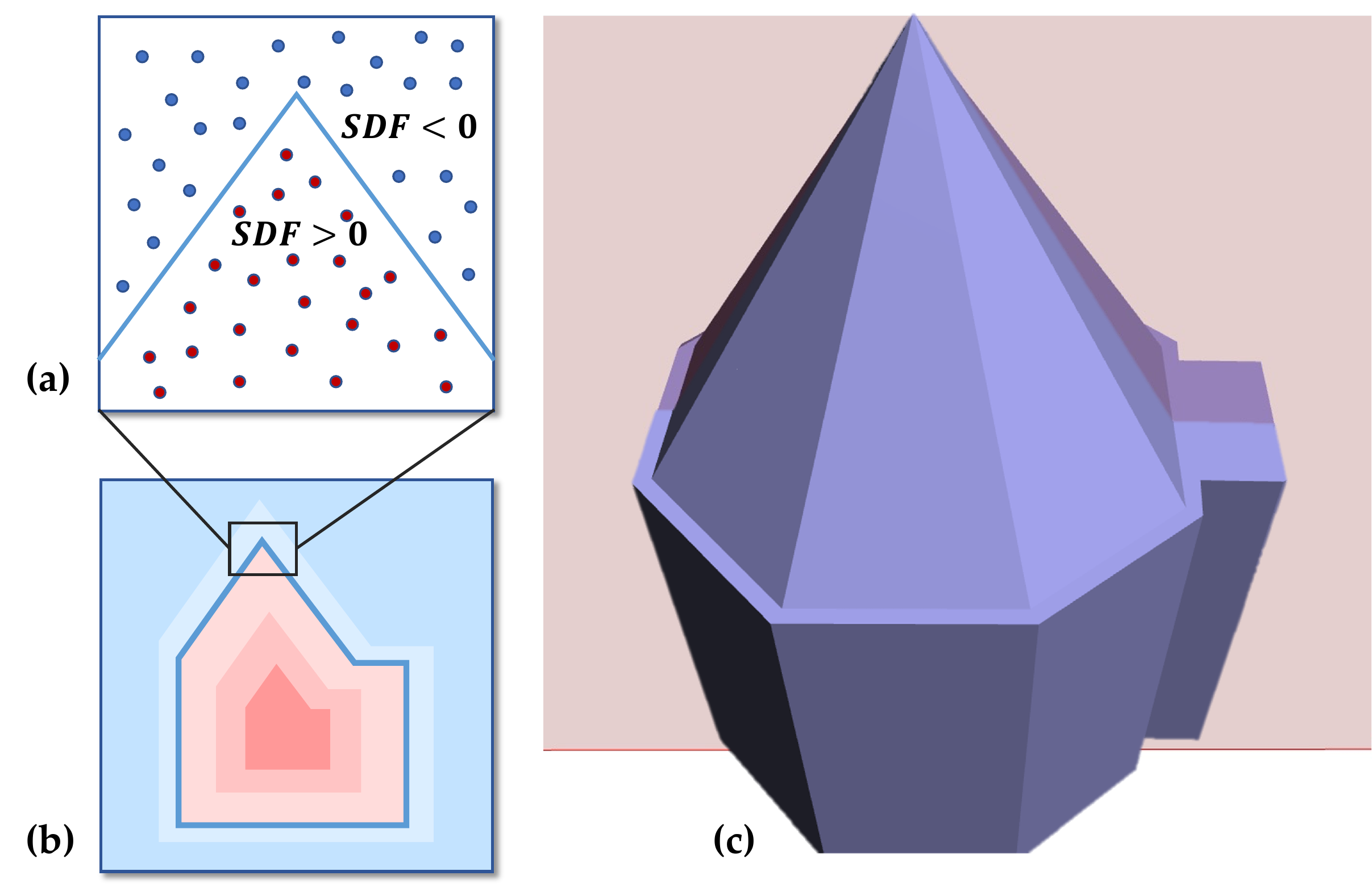}
	\caption{An example of a signed distance field defined on a cross-section of a building model. 
	(a) A visualization of the signed distance field, where the red color indicates the building's interior and the blue color indicates its exterior. 
	(b) The signs of the signed distance field at a set of point samples. The color intensity reflects the distance values at a few discrete levels. 
	(c) The 3D surface model. The cutting plane indicates where the signed distance field is defined.
}	\label{fig:sdf}
\end{figure*}

\subsection{Occupancy learning}

From the cell complex (denoted by $C$) obtained from adaptive binary space partitioning, we would like to extract a subset of the cells, i.e., a subcomplex $L \in C $, such that its occupancy $O_{L}$ indicates the interior space enclosed by the outer surface of the building. 
To this end, we learn the occupancy of the building as a signed distance field (SDF), where the value is the distance $d$ from a point $\mathbf{x}$ to the building surface and its sign indicates whether the point lies inside (with a positive sign) or outside (with a negative sign) the outer surface of the building:
\begin{equation}
	SDF(\mathbf{x}) = d: \mathbf{x} \in \mathbb{R}^3, d \in \mathbb{R}.
	\label{eq:sdf}
\end{equation}

\noindent \autoref{fig:sdf} illustrates an example of the signed distance field at a series of cross-sections of a 3D building model.

Inspired by the work of \citet{erler2020points2surf}, we exploit a deep learning-based approach to learn a signed distance field from a point cloud. 
Specifically, we train a neural network that can predict the signed distance value for any point $\mathbf{x} \in \mathbb{R}^3$, i.e.,
\begin{equation}
	f(\mathbf{x}) \approx \tilde{f}(\mathbf{x})=s_{\theta}(\mathbf{x}\mid \mathbf{z}), \text { with } \mathbf{z}=e_{\phi}(P),
\end{equation}
where $\mathbf{z}$ is a latent representation of the building surface encoded from the input point cloud $P$ by an encoder $e$, and $s$ represents the neural network. 
The encoder $e$ and neural network $s$ are parameterized by $\theta$ and $\phi$, respectively. 
Following the neural network architecture of \citet{erler2020points2surf}, we decompose the signed distance field into two components: the absolute distance $f^{d}$ and its sign $f^{s}$. 

\begin{itemize}
	\item \textbf{The absolute distance value}. The estimated absolute distance $\tilde{f}^{d}(\mathbf{x})$ can be determined from only the neighborhood of the query point:
		\begin{equation}
			\tilde{f}^{d}(\mathbf{x})=s_{\theta}^{d}\left(\mathbf{x} \mid \mathbf{z}_{\mathbf{x}}^{d}\right), \text { with } \mathbf{z}_{\mathbf{x}}^{d}=e_{\phi}^{d}\left(\mathbf{p}_{\mathbf{x}}^{d}\right)
		\end{equation}
		where $\mathbf{p}_{\mathbf{x}}^{d} \in P$ is a set of neighboring points around query point $\mathbf{x}$.
		
	\item \textbf{The sign}. To estimate the sign $\tilde{f}^{s}(\mathbf{x})$ at $\mathbf{x}$, local sampling does not suffice because the occupancy information cannot be reliably estimated from the local neighborhood only. 
		Therefore, a global uniform sub-sample $\mathbf{p}_{\mathbf{x}}^{s} \in P$ is taken as input:
	\begin{equation}
		\tilde{f}^{s}(\mathbf{x})=\operatorname{sgn}\left(\tilde{g}^{s}(\mathbf{x})\right)=\operatorname{sgn}\left(s_{\theta}^{s}\left(\mathbf{x} \mid \mathbf{z}_{\mathbf{x}}^{s}\right)\right), \text { with } \mathbf{z}_{\mathbf{x}}^{s}=e_{\psi}^{s}\left(\mathbf{p}_{\mathbf{x}}^{s}\right)
	\end{equation}
where $\psi$ parameterizes the encoder, and $\tilde{g}^{s}(\mathbf{x})$ is a logit that expresses the confidence of $\mathbf{x}$ having a positive distance to the surface. 

\end{itemize}

The two latent descriptions $\mathbf{z}_{\mathbf{x}}^{s}$ and $\mathbf{z}_{\mathbf{x}}^{d}$ share information, resulting in the formulation for signed distance learning, i.e.,
\begin{equation}
	\left(\tilde{f}^{d}(\mathbf{x}), \tilde{g}^{s}(\mathbf{x})\right)=s_{\theta}\left(\mathbf{x} \mid \mathbf{z}_{\mathbf{x}}^{d}, \mathbf{z}_{\mathbf{x}}^{s}\right), \text { with } \mathbf{z}_{\mathbf{x}}^{d}=e_{\phi}^{d}\left(\mathbf{p}_{\mathbf{x}}^{d}\right) \text { and } \mathbf{z}_{\mathbf{x}}^{s}=e_{\psi}^{s}\left(\mathbf{p}_{\mathbf{x}}^{s}\right)
\end{equation}

% cross-sections
\begin{figure*}[th!]
	\centering
	\subfloat[Volume]{\includegraphics[width=.32\linewidth]{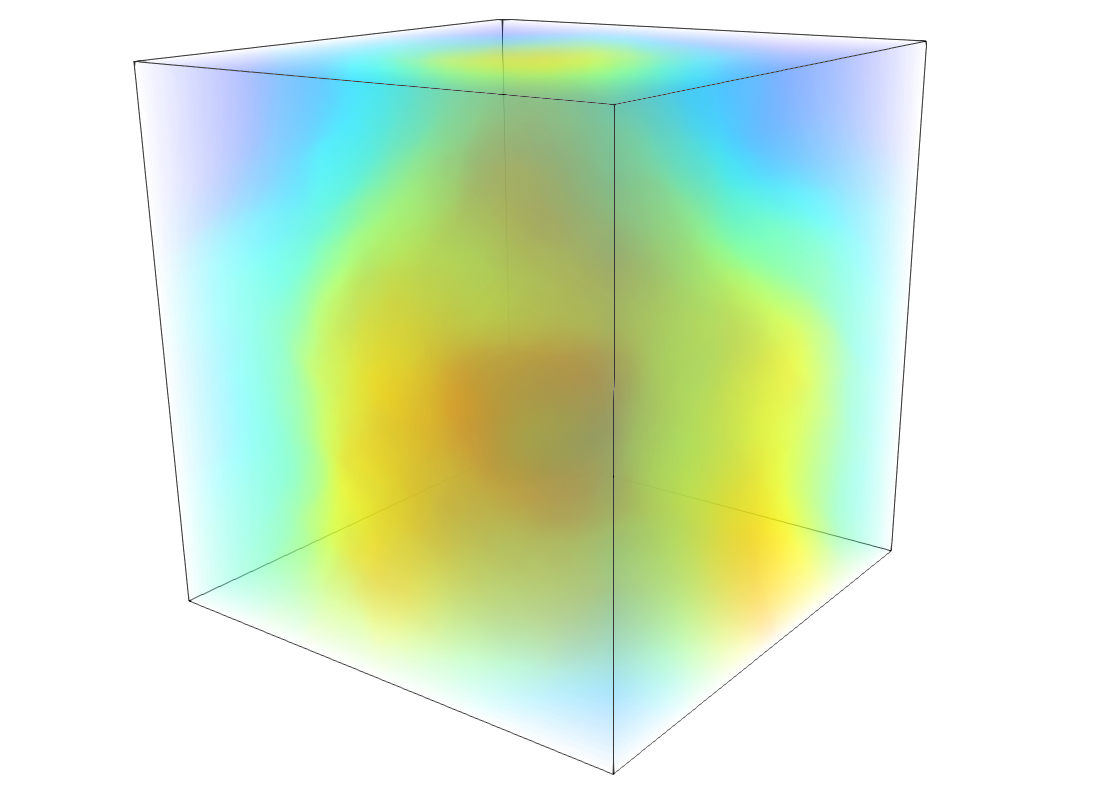}}
	\subfloat[10\%]{\includegraphics[width=.32\linewidth]{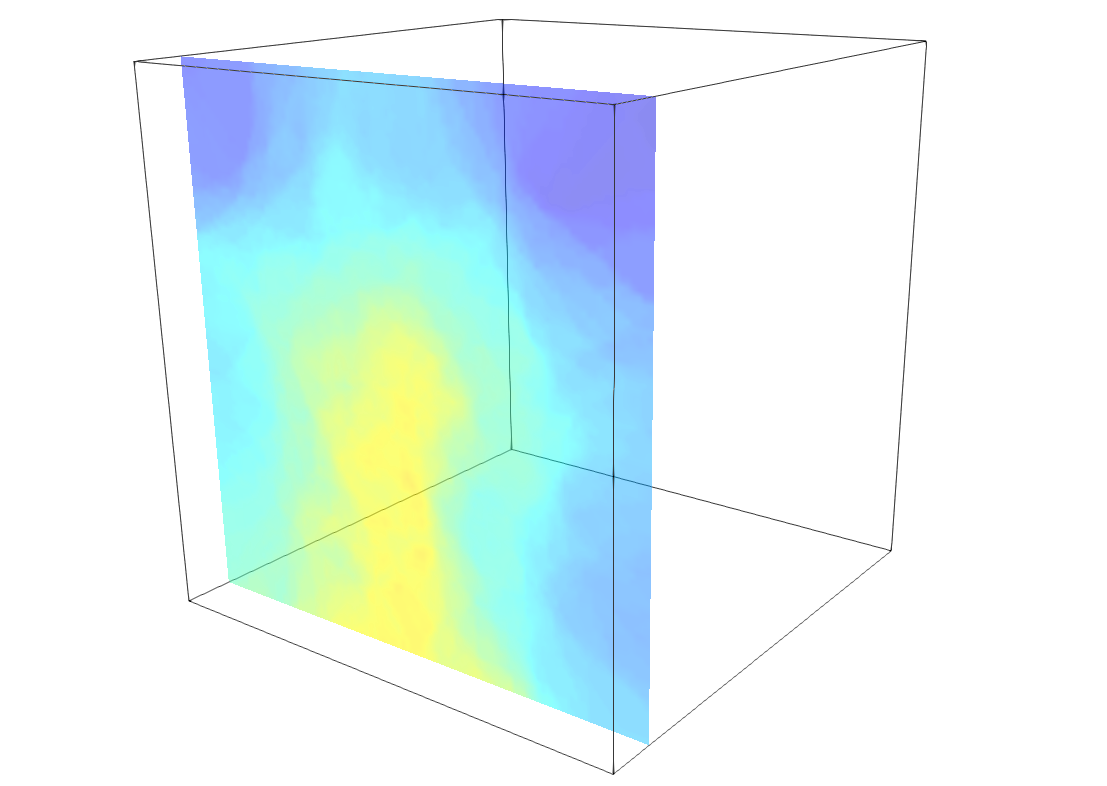}}
	\subfloat[30\%]{\includegraphics[width=.32\linewidth]{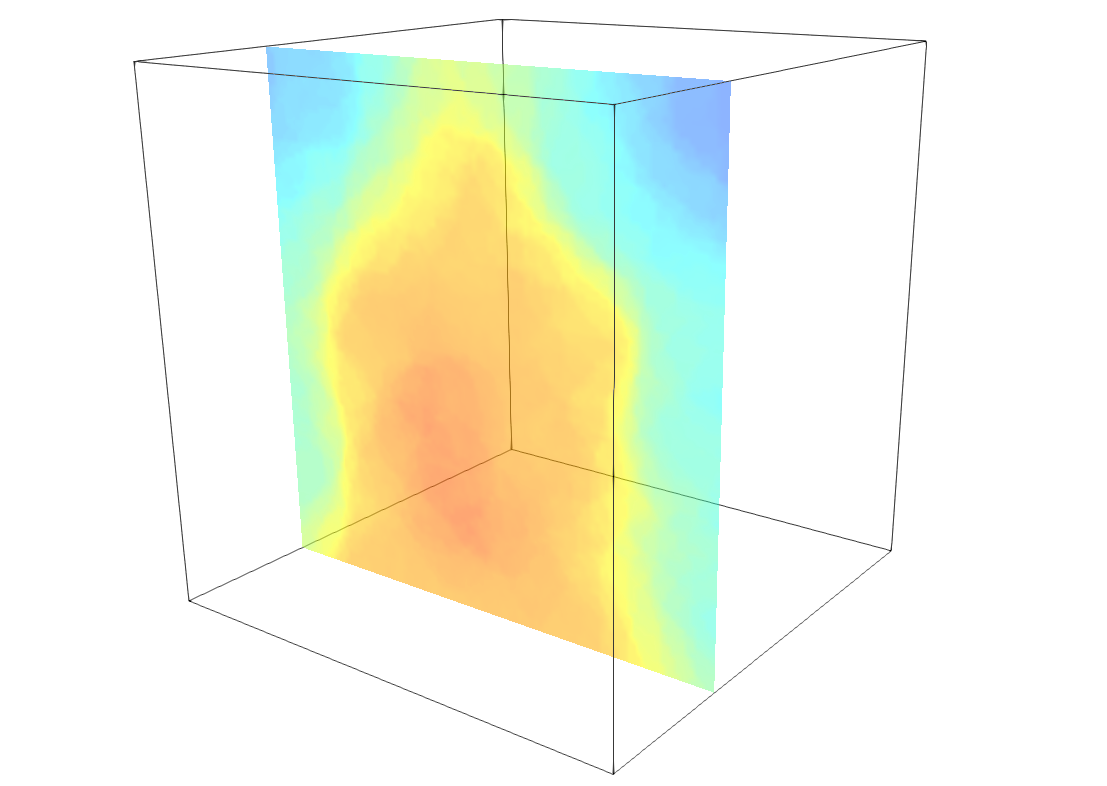}}
	\vspace{1em}
	
	\subfloat[50\%]{\includegraphics[width=.32\linewidth]{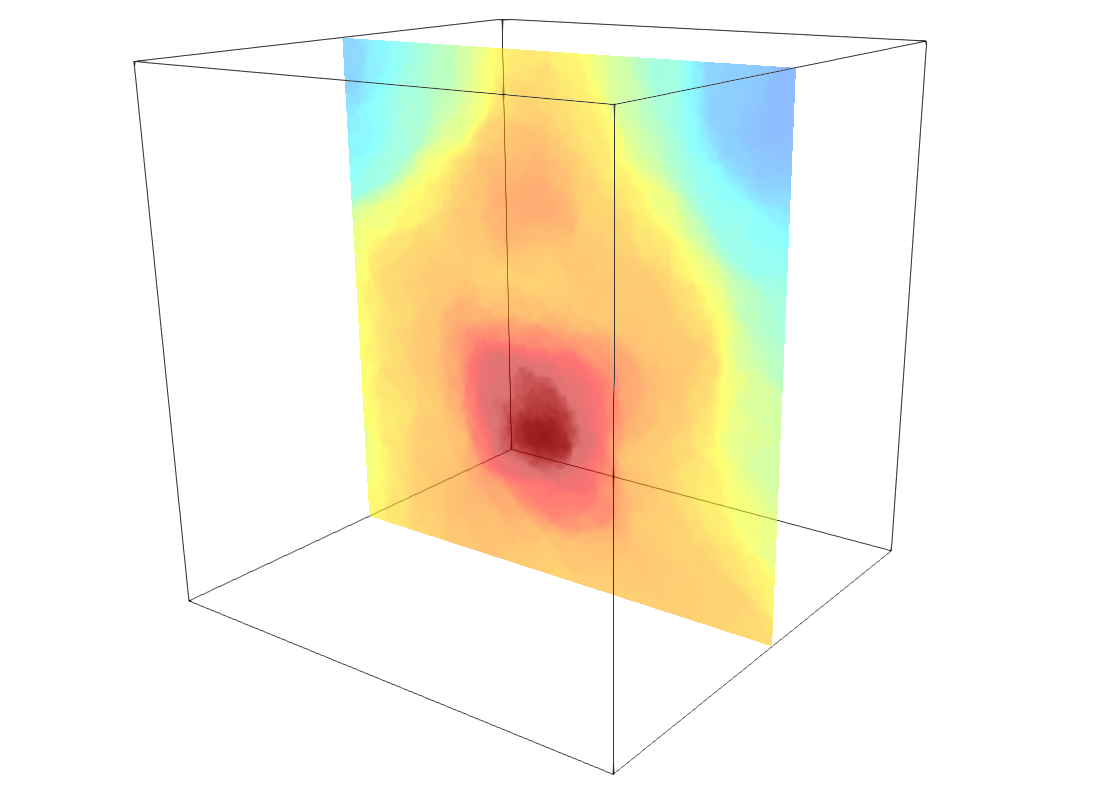}	}
	\subfloat[70\%]{\includegraphics[width=.32\linewidth]{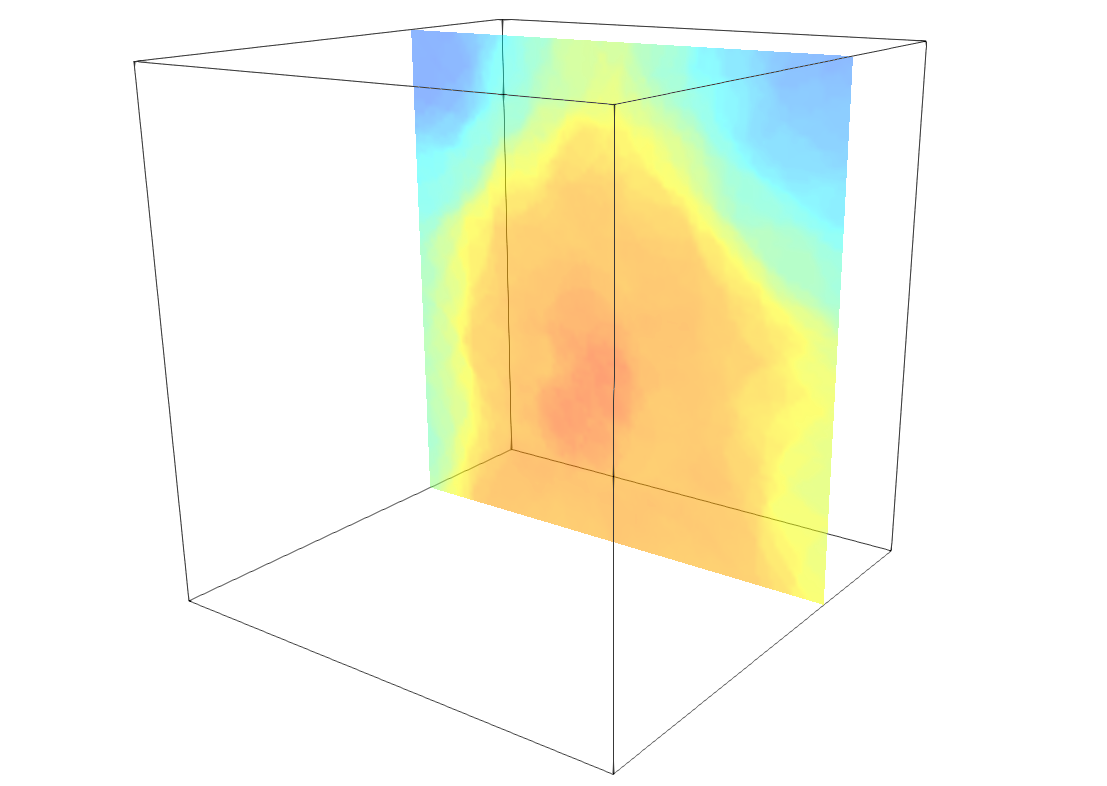}}
	\subfloat[90\%]{\includegraphics[width=0.32\linewidth]{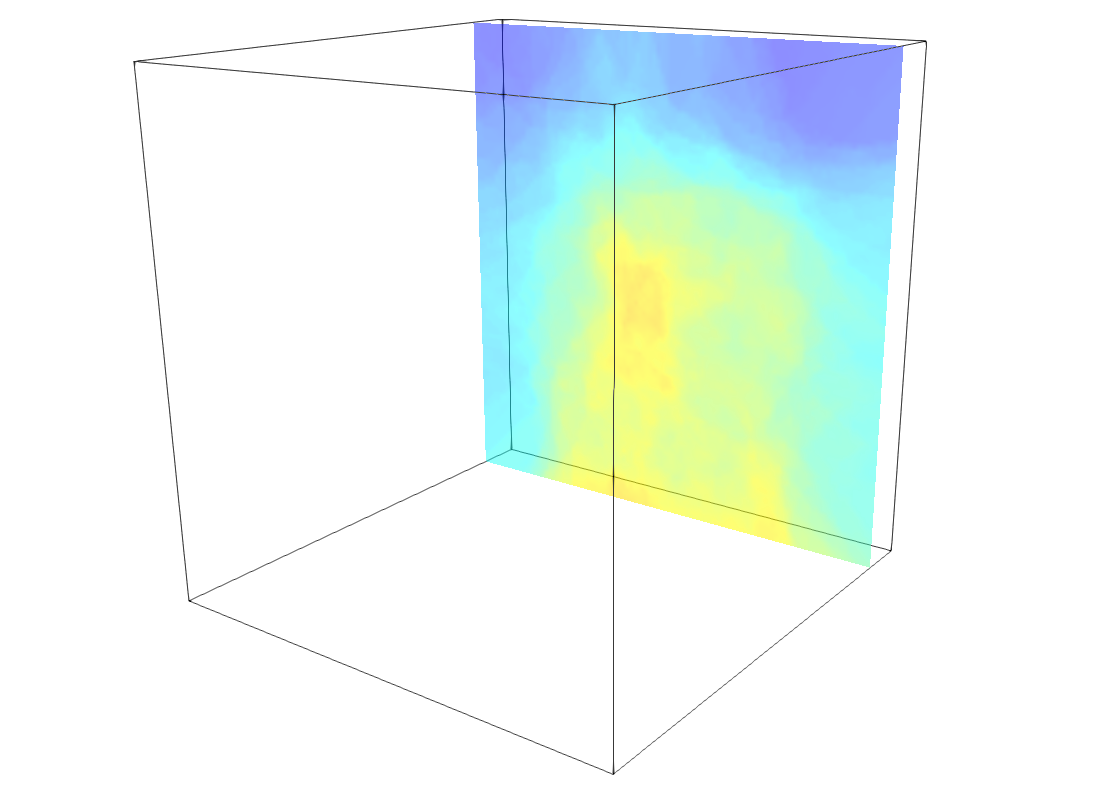}}
	\includegraphics[height=0.22\linewidth]{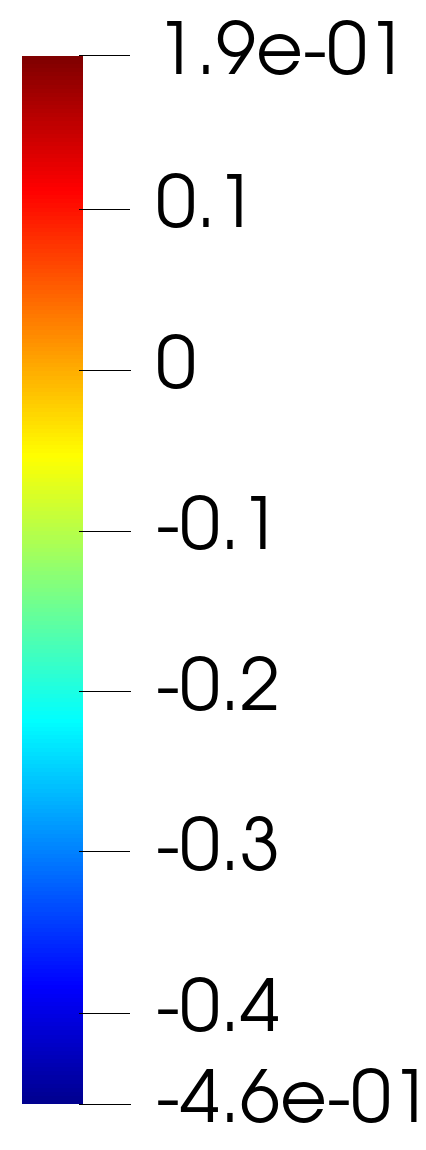}
	\caption{A series of cross-sections of a signed distance field learned for the building shown in \autoref{fig:sdf}. (a) Volume rendering of the signed distance field. (b) - (f) are a few cross-sections of (a) where the percentages represent relatively where the sections are taken.}
	\label{fig:cross_section}
\end{figure*}

\autoref{fig:cross_section} illustrates a series of cross-sections of the signed distance field predicted by the neural network for the 3D model shown in \autoref{fig:sdf}.
With the learned signed distance field, instead of performing Marching Cubes, we assign each cell $c_i \in C $ a signed distance value at its centroid. This sparse query can significantly reduce the computation for occupancy evaluation. %(e.g., by 8 times for the building model in Figure 6).

\subsection{Surface extraction}

\begin{figure*}[th!]
	\centering
	\subfloat[Cell complex]{\includegraphics[width=0.28\linewidth]{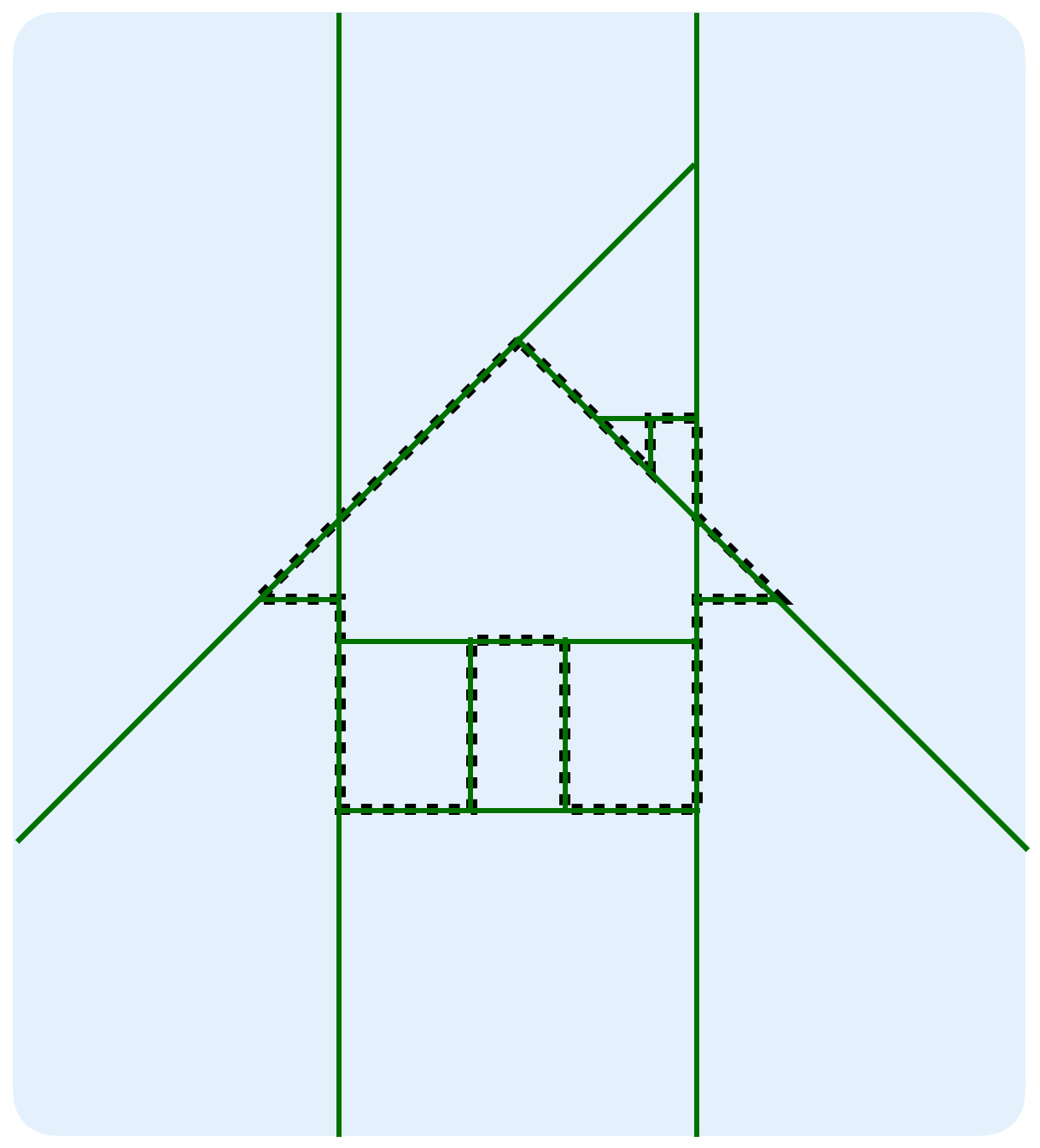}}
	\hspace{1em}
	\subfloat[Interior cells]{\includegraphics[width=0.28\linewidth]{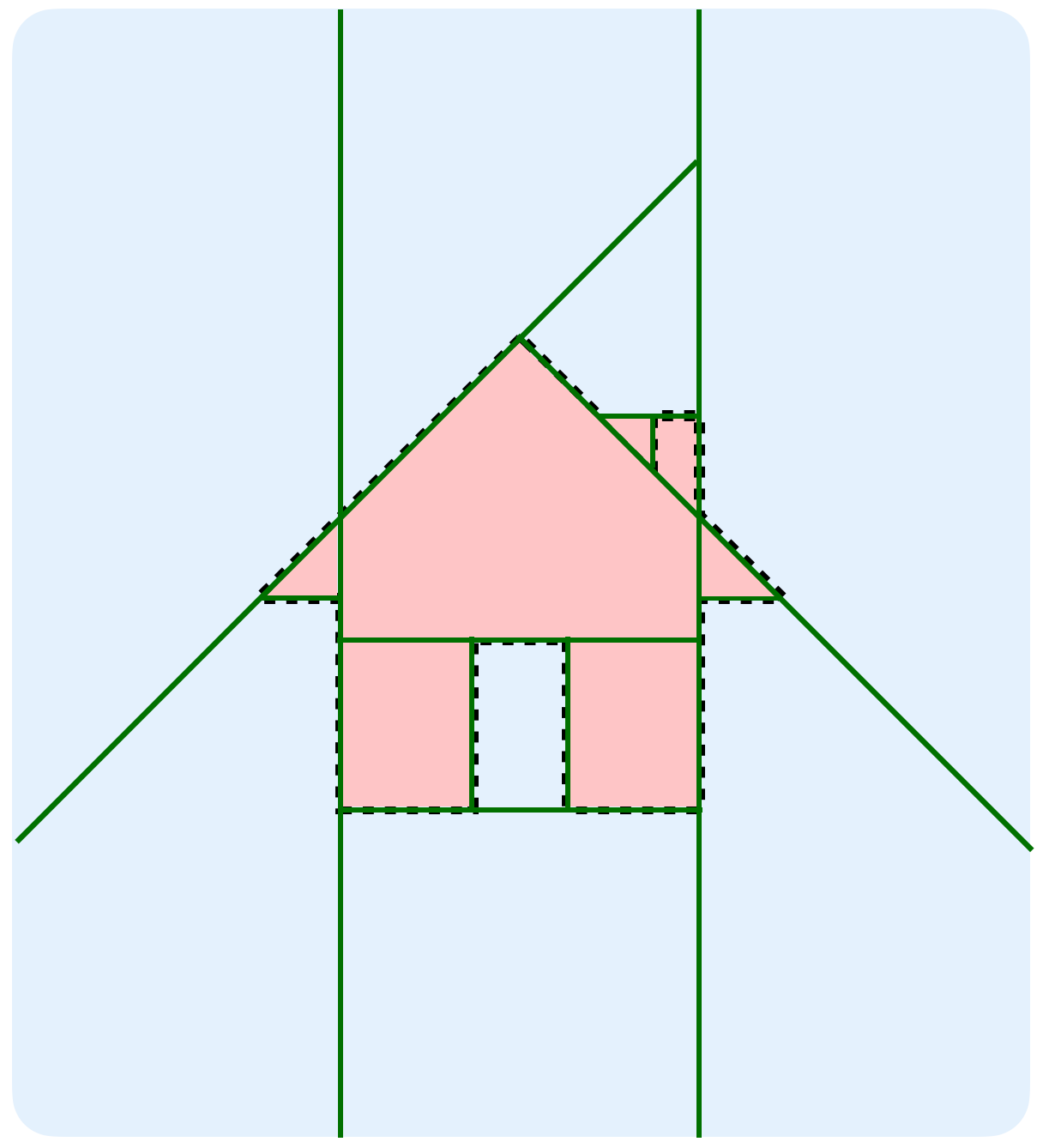}}
	\hspace{1em}
	\subfloat[Surface model]{\includegraphics[width=0.28\linewidth]{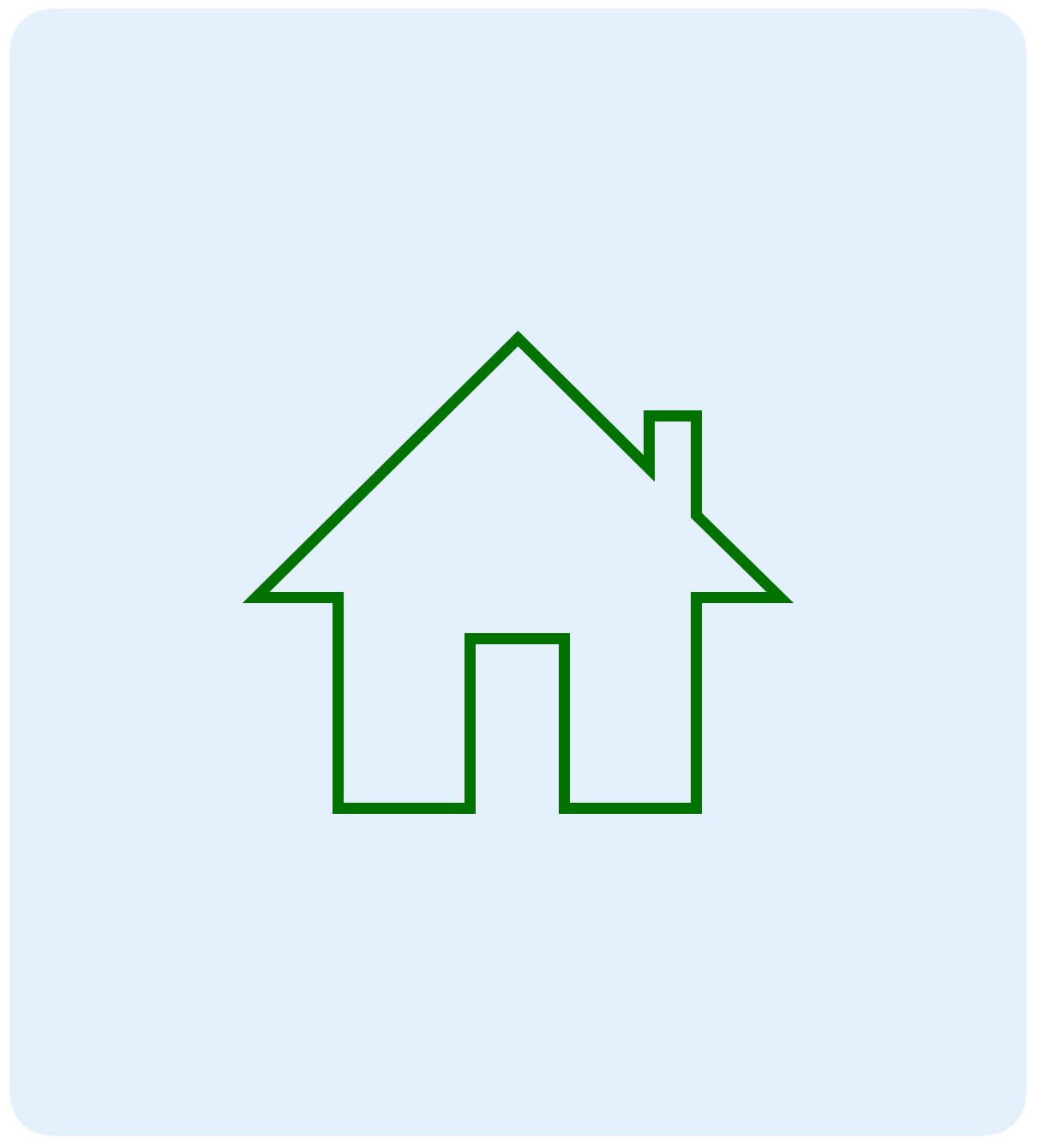}}
	\caption{An illustration of the surface extraction step. (a) The cell complex with the learned occupancy (the red and blue colors indicate interior and exterior, respectively). (b) The union of the interior cells. (c) The surface model was obtained by extracting the outer shell of the union of the interior cells.}
	\label{fig:surface_extraction}
\end{figure*}

With the cell complex and the occupancy of its cells obtained in the previous steps, surface reconstruction can be addressed by obtaining a consistent classification of the cells into \textit{interior} and \textit{exterior} categories, followed by an outer shell extraction step, as shown in Figure \ref{fig:surface_extraction}.
We address the interior/exterior cell classification using a Markov Random Field (MRF) formulation.

%Notice that, however, the output surface may be non-manifold. 
%As shown in Figure \ref{fig:labellingexamples} (a), two inside polyhedra can be connected along one non-manifold edge or at one non-manifold vertex. 
%We argue this non-manifold abstraction can be seen in real-world building structures and therefore should be allowed in reconstruction.

%\begin{figure*}[th!]
%	\centering
%%	\subfloat[Manifold]
%{\includegraphics[height=0.30\linewidth]{figures/methodology/labelling_manifold.png}\label{fig:labelling_manifold}}
%
%	\caption{delete this figure after finishing Fig.~\ref{fig:surface_extraction} 
%	\label{fig:to_delete}
%\end{figure*}

%A naive reconstruction can be achieved by extracting the surface between each pair of polyhedra where one is classified as inside while the other one as outside. 
%However, this solution does not take into consideration the regularization needed for compact building models.
%Instead, we propose an efficient surface extraction method based on an MRF formulation, and solve the optimization problem using an efficient graph-cut solver.

Given the cell complex $C = \{c_i\}$, we denote the binary label to be assigned to a cell $c_i$ by $x_{i} \in \{in, out\}$. Our energy function is written as the weighted sum of two energy terms, i.e., 
\begin{equation}
	E(x) = D(x) + \lambda V(x)
	\label{eq:energy}
\end{equation}
where $D(x)$ and $V(x)$ denote the data cost term and the smoothness cost term, respectively, and $\lambda$ is the weight that balances the two terms. 

\begin{figure*}[th!]
	\centering
	\subfloat[]{\includegraphics[width=0.28\linewidth]{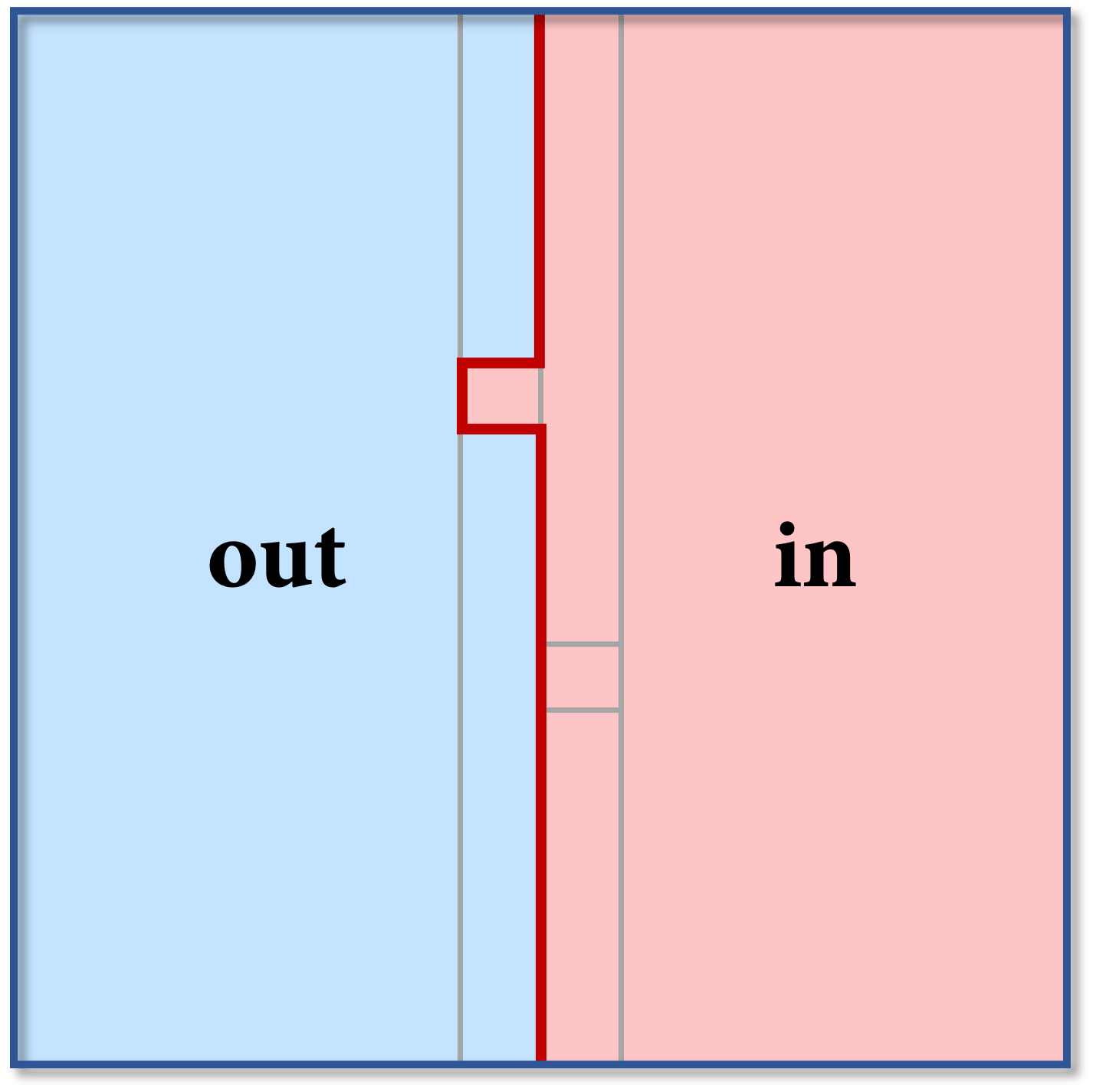}}
	\hspace{1em}
	\subfloat[]{\includegraphics[width=0.28\linewidth]{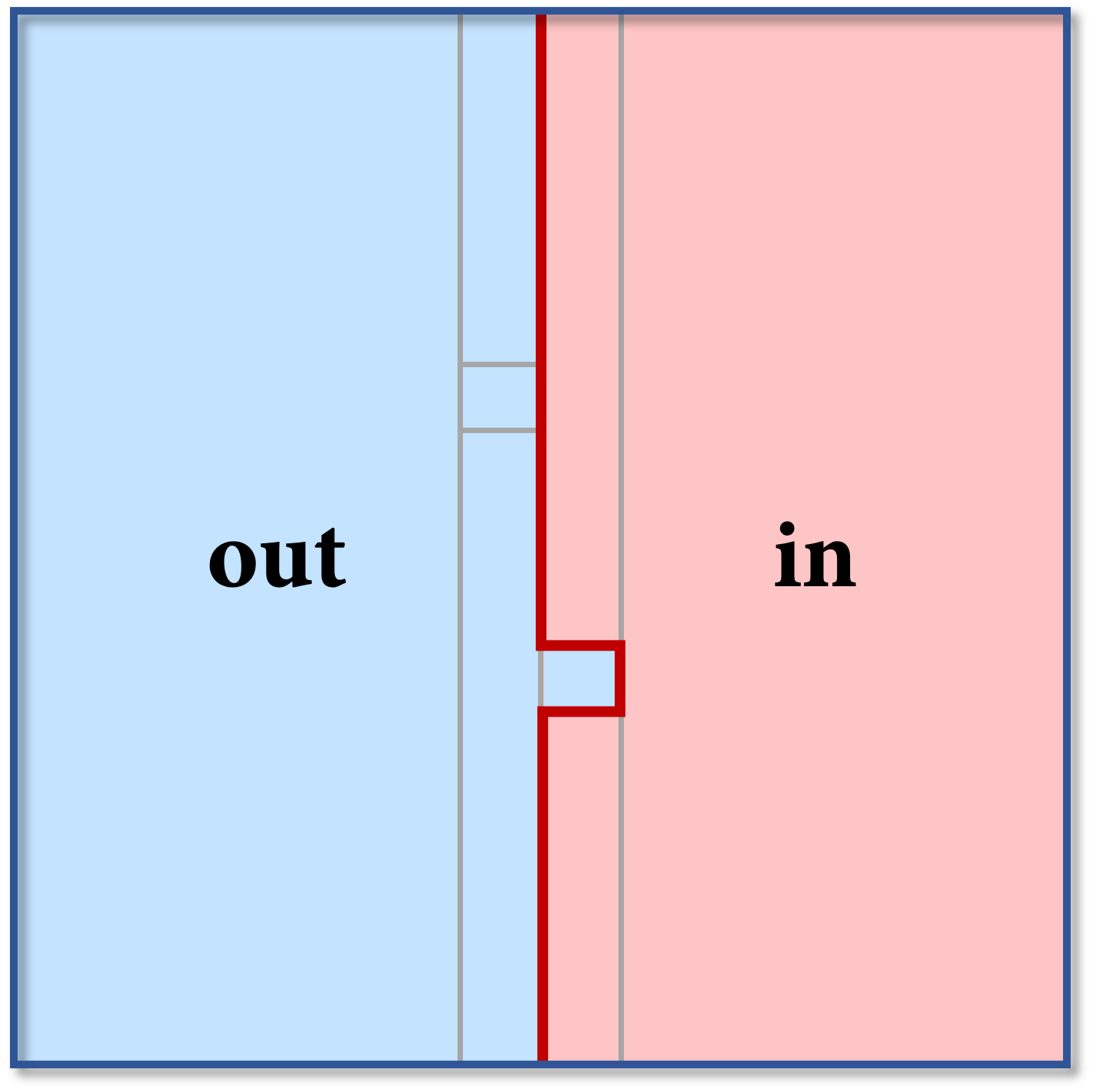}}
	\hspace{1em}
	\subfloat[]{\includegraphics[width=0.28\linewidth]{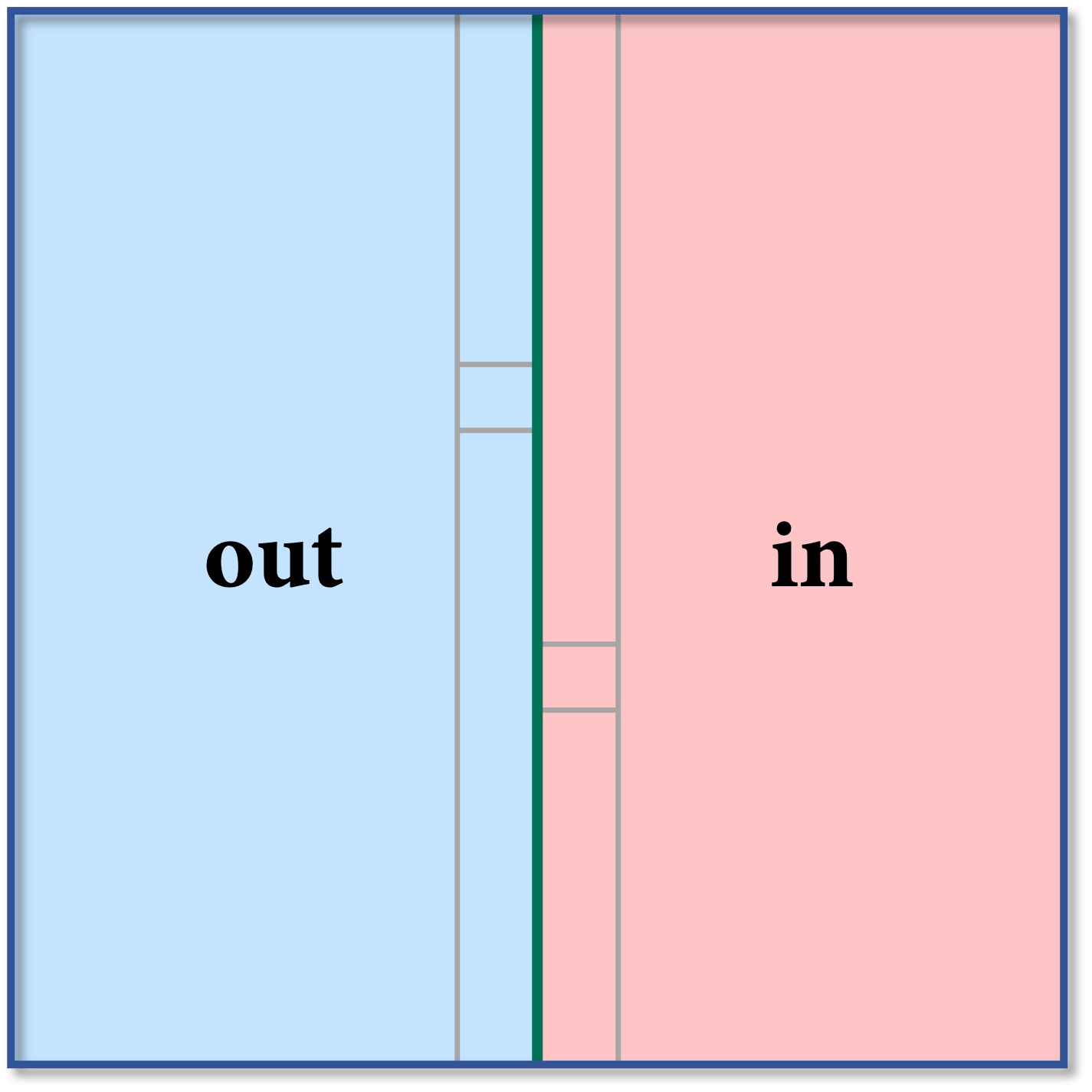}}
	\caption{An illustration of the effect of the smoothness cost energy term. (a) and (b) Interior-exterior classification result without the smoothness cost energy term. (c) Interior-exterior classification result with the smoothness cost energy term. The smoothness cost energy term penalizes zigzag artifacts and encourages a compact surface model.}
	\label{fig:zigzag}
\end{figure*}

\begin{itemize}
	\item \textbf{Data cost}. We define the data cost term in a way such that it reflects the confidence of the classification. Specifically, it is defined to measure how much the classification differs from the previously estimated occupancy of the cells in the cell complex, i.e.,
\begin{equation}
	D(X) = \frac{1}{\left | C \right |} \sum_{c_i\in C} \left |x_{i} - occupancy(c_{i}) \right |,
\end{equation}
where $occupancy(c_{i})$ denotes the learned occupancy of cell $c_i$, which can be computed as 
\begin{equation}
	occupancy(c_{i}) = sigmoid(SDF(c_i) \cdot volume(c_i)),
\end{equation}
where $SDF(c_i)$ is the signed distance value of the query point at the centroid of $c_{i}$, predicted by the neural network, and $volume(c_i)$ is the volume of $c_{i}$. 
Intuitively, a cell with a larger volume should weigh higher regardless of its predicated signed distance.
The sigmoid function $sigmoid(x) = \frac{1}{1+e^{-x}}$ normalizes the signed distance to the range $(0, 1)$. 

	\item \textbf{Smoothness cost}. This energy term encourages assigning similar labels to the adjacent cells. We design this term in a way such that it penalizes the complexity of the output building surface model. Since the final surface will be eventually extracted as the outer shell of the interior cells, lowering its complexity is equivalent to limiting its surface area. Thus, our smoothness cost is defined as 
\begin{equation}
	V(X) = \frac{1}{A} \sum_{\{c_i,c_j\} \in C}a_{ij} \cdot \mathbbm{1}(c_i, c_j),
\end{equation}
where $\{c_i, c_j\} \in C$ represents a pair of adjacent cells in the complex. $a_{ij}$ denotes the surface area of the common face of the two cells. $A$ is a normalization factor that is chosen as the maximum area of all faces in the cell complex. $\mathbbm{1}(c_i, c_j)$ is an indicator function, which has the value of 1 if $c_i$ and $c_j$ receive different labels, i.e, 
\begin{equation}
  \mathbbm{1}(c_i, c_j) =
    \begin{cases}
      0, &x_i = x_j\\
      1, &x_i \neq x_j
    \end{cases}       
\end{equation}

Intuitively, the smoothness cost serves as a regularization term that penalizes zigzag artifacts on the final surface model, whose effect is illustrated in Figure~\ref{fig:zigzag}.  

\end{itemize}

By minimizing the energy function given in \autoref{eq:energy} using the graph cut algorithm~\citep{boykov2006graph}, the interior cells can be obtained, as shown in \ref{fig:surface_extraction} (b). 
The final surface model is then obtained by extracting the outer surface of the union of the interior cells, which is illustrated in \ref{fig:surface_extraction} (c).
As our adaptive binary space partitioning produces a valid polyhedral embedding, the surface is inherently guaranteed to be watertight.

\section{Datasets}

Sufficient high-quality point clouds and the corresponding surface models of real-world buildings are necessary for training and evaluating learning-based reconstruction methods. Due to the lack of such data, we have created a synthetic dataset on which we train our neural network for occupancy learning. We then evaluate the performance of the proposed method on both the synthetic dataset and a real-world dataset.

\subsection{Synthetic dataset}

To create the synthetic dataset for training our neural network and further evaluating our reconstruction method, we have created a synthetic dataset by simulating the scanning process based on the Helsinki LoD2 CityGML models~\citep{helsinki}.
Specifically, we pick 678 watertight building meshes for training, 45 for validation, and another 45 for testing.
Because of our patch-based architecture for occupancy learning, a large number of diverse patches are produced from each mesh as training samples.

The point clouds are generated by simulated scanning on the building mesh models. 
In pre-processing, the building meshes are translated to the origin and scaled uniformly to a unit length.
To simulate the acquisition of a point cloud $P$ on a building $S$, a LiDAR sensor is configured from random perspectives. 
We use Blensor~\citep{gschwandtner2011blensor} to simulate the scanning, intentionally with various levels of Gaussian noise and artifacts such as occlusions and light reflections.

\begin{figure}[ht]
	\centering
	\subfloat[Full-view scanning]{\includegraphics[width=0.46\linewidth]{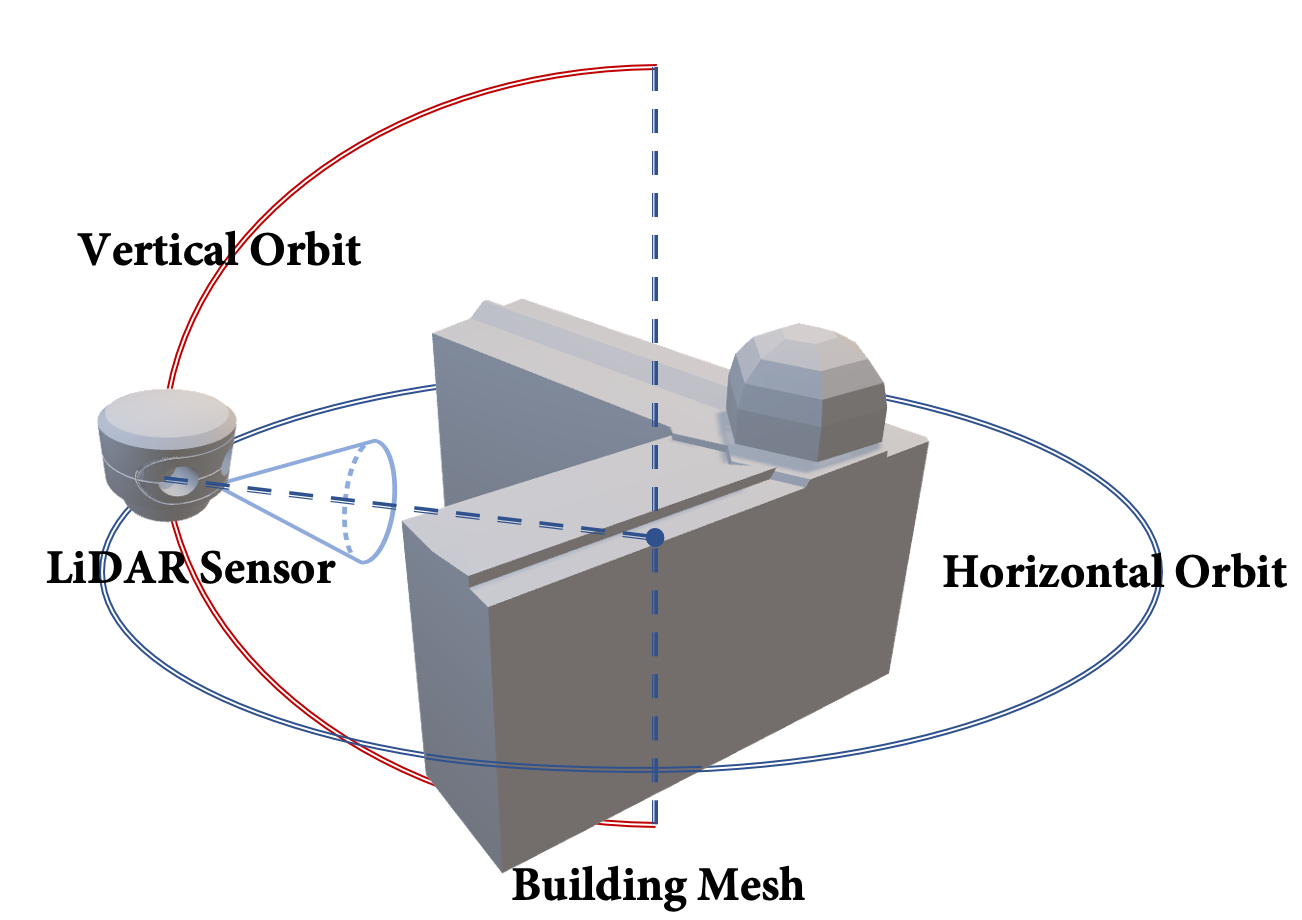}\label{fig:simulation_fullview}}
	\hspace{1em}
	\subfloat[Without-bottom scanning]{\includegraphics[width=0.46\linewidth]{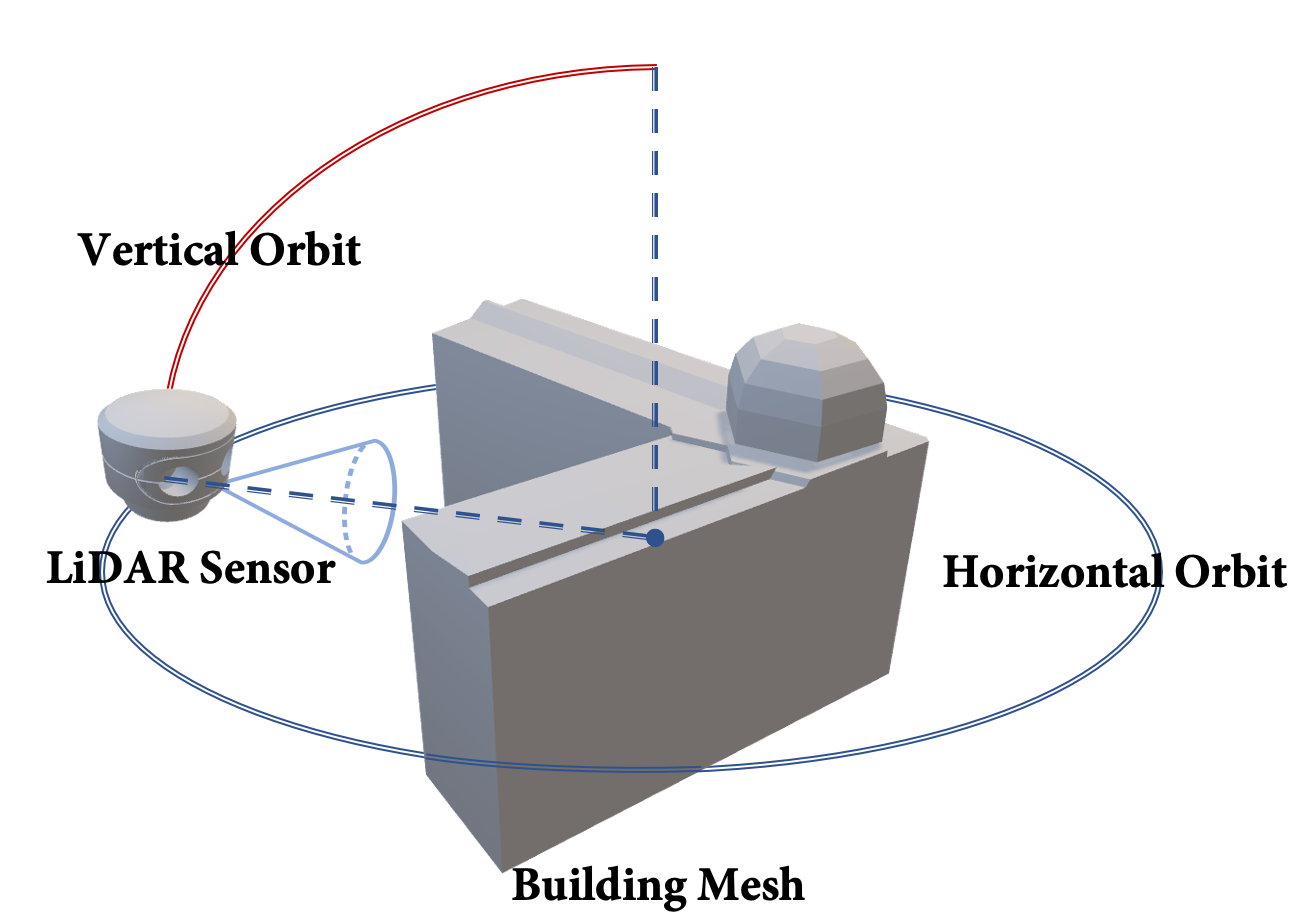}\label{fig:simulation_nobottom}}
	\caption{Simulation of point cloud acquisition. (a) A LiDAR scanner rotates on a sphere around the building mesh to generate a full-view point cloud. (b) The scanner's position is constrained on the upper half of the sphere to simulate real-world scanning where the bottom of a building will not be captured.}
	\label{fig:simulation}
\end{figure}

To handle noise, we created multiple versions of the Helsinki dataset with different levels of noise, which is achieved by intentionally adding artificial  Gaussian noise to the depth values during the virtual scanning process to augment the training data.
For the training of the neural network, we obtain point clouds with Gaussian noise whose standard deviation is randomly chosen from the range $U[0, 0.005R]$, where $R$ is the largest side of the building's bounding box, and $U$ stands for uniform distribution.
We also create multiple versions of point clouds for the same building sets, where the Gaussian noise is set to four different levels $\{0, 0.001R, 0.005R, 0.010R, 0.050R\}$, for evaluating the robustness of the proposed method and its competitors.
In addition to the simulated point clouds, the training set also contains a set of query points $X_{S}$ for each building. The query points are pre-sampled with known signed distance to the building's surfaces.
Since query points with smaller absolute distances are of higher importance for detailed surface reconstruction, we randomly sample 1000 points on the surface and then apply perturbation in their normal direction by a random displacement in $U[-0.02R, 0.02R]$. 
In addition, we sample 1000 query points randomly distributed in the bounding box of the building, mounting up to 2000 query points in total per building. 
To enhance the robustness of the learned SDF, we randomly drop out 1000 query points and use the other 1000 samples for training. 
We denote this full-view dataset as \textit{Helsinki full-view}.

To mimic the real-world scanning process and simulate occlusions in the dataset, we also constrain the scanner's position such that no bottom view is used (see Figure~\ref{fig:simulation} (b)), to create another set of point clouds, denoted as \textit{Helsinki no-bottom}. 
Similar to \textit{Helsinki full-view}, we generate the point clouds with various Gaussian noise along with the query points for training, and a series of point clouds with fixed noise levels for evaluation.

\subsection{Real-world dataset}

Besides the synthetic data, we further evaluate our method on the real-world dataset consisting of photogrammetric point clouds of six buildings~\citep{xie2021combined}. These point clouds are obtained from aerial images using multi-view stereo techniques.
The original images were captured by Unmanned Aerial Vehicles (UAV) from top and lateral perspectives. 
The main bodies of buildings are visible with well-captured roof structures, while the facades of the buildings are only partially visible with missing areas due to occlusions and poor lighting conditions in the lower part of the buildings. The point clouds in this dataset are demonstrated in the first column of Figure~\ref{fig:results_shenzhen}.

\subsection{Summary}
\autoref{tab:datasets} summarizes the characteristics of the datasets used for evaluation.
Since global shape priors are dataset-dependent, the neural network trained on one dataset may not capture the characteristics of another. To this end, we train our neural network for SDF estimation on the \textit{Helsinki full-view} dataset and \textit{Helsinki no-bottom} dataset with various levels of noise, respectively. 
The trained model on the former is used for evaluation on the \textit{full-view} set, while that of the latter on the \textit{no-bottom} set.

\begin{table}[ht]
	\caption{Datasets overview}
	
	{
		\tiny
		\centering
		\begin{tabular}{lcccccc}
			\toprule
			\multirow{2}{*}{\textbf{Name}}     & \multirow{2}{*}{\textbf{Type}}   & \multicolumn{3}{c}{\textbf{Perspective}} & \multirow{2}{*}{\textbf{Quantity}} & \multirow{2}{*}{\textbf{Usage}} \\
			\cmidrule(lr){3-5}
			& & \textbf{Top} & \textbf{Bottom} & \textbf{Lateral} \\
			
			\midrule
			
			\textit{Helsinki full-view} & Simulated LiDAR       & \cmark & \cmark & \cmark            & 768               & Training + evaluation \\
			\textit{Helsinki no-bottom} & Simulated LiDAR      & \cmark  & \xmark & \cmark      & 768               & Training + evaluation \\
			\textit{Shenzhen}          & Real-world MVS & \cmark  & \xmark & \cmark     & 6                 & Evaluation            \\
			
			\bottomrule             
		\end{tabular}
	}
	
	\label{tab:datasets}
\end{table}

\section{Results and analysis}

\subsection{Reconstruction results}

\begin{figure*}[ht!]
	\centering
			(a)
	\subfloat{\includegraphics[width=0.15\linewidth]{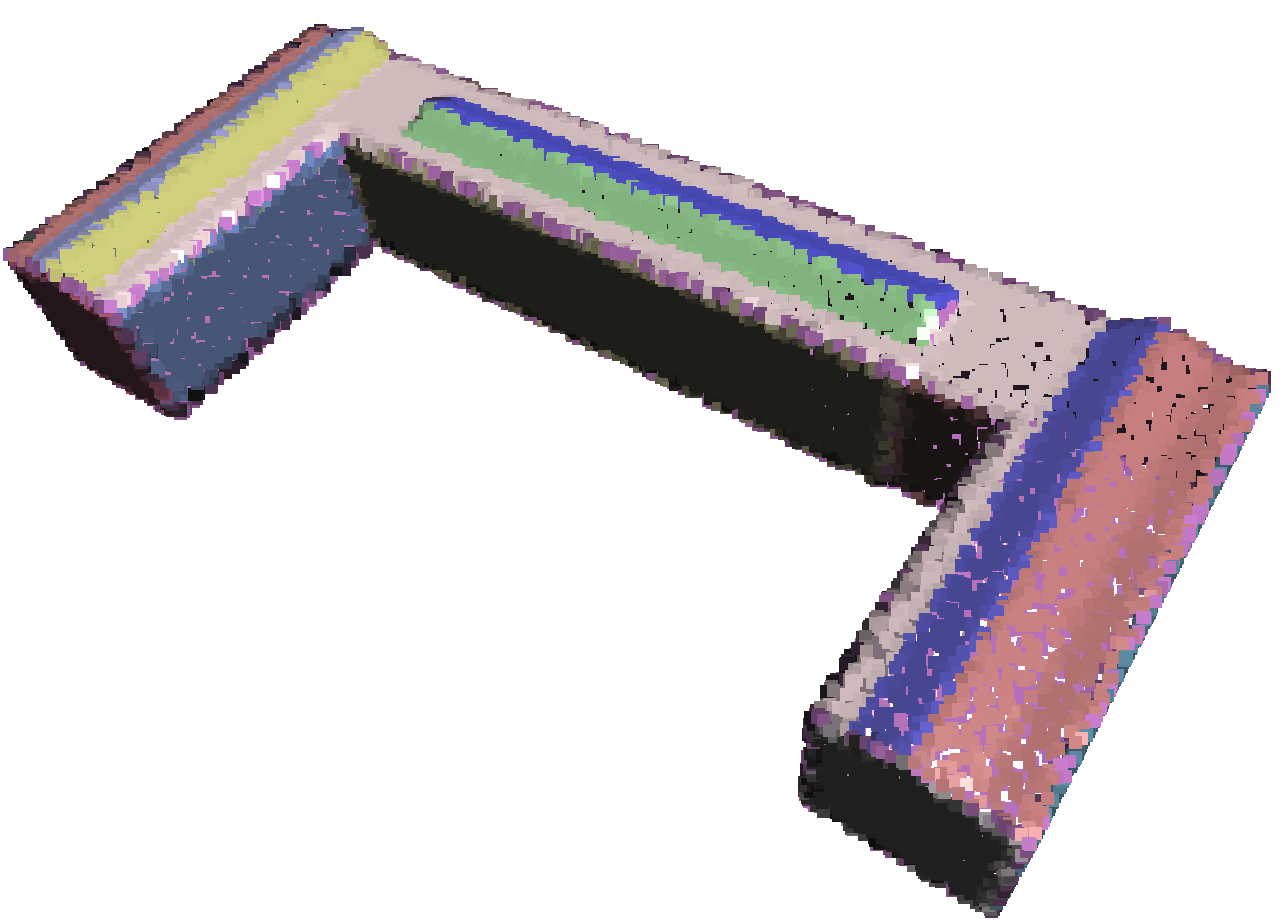}}
	\hspace{1em}
	\subfloat{\includegraphics[width=0.15\linewidth]{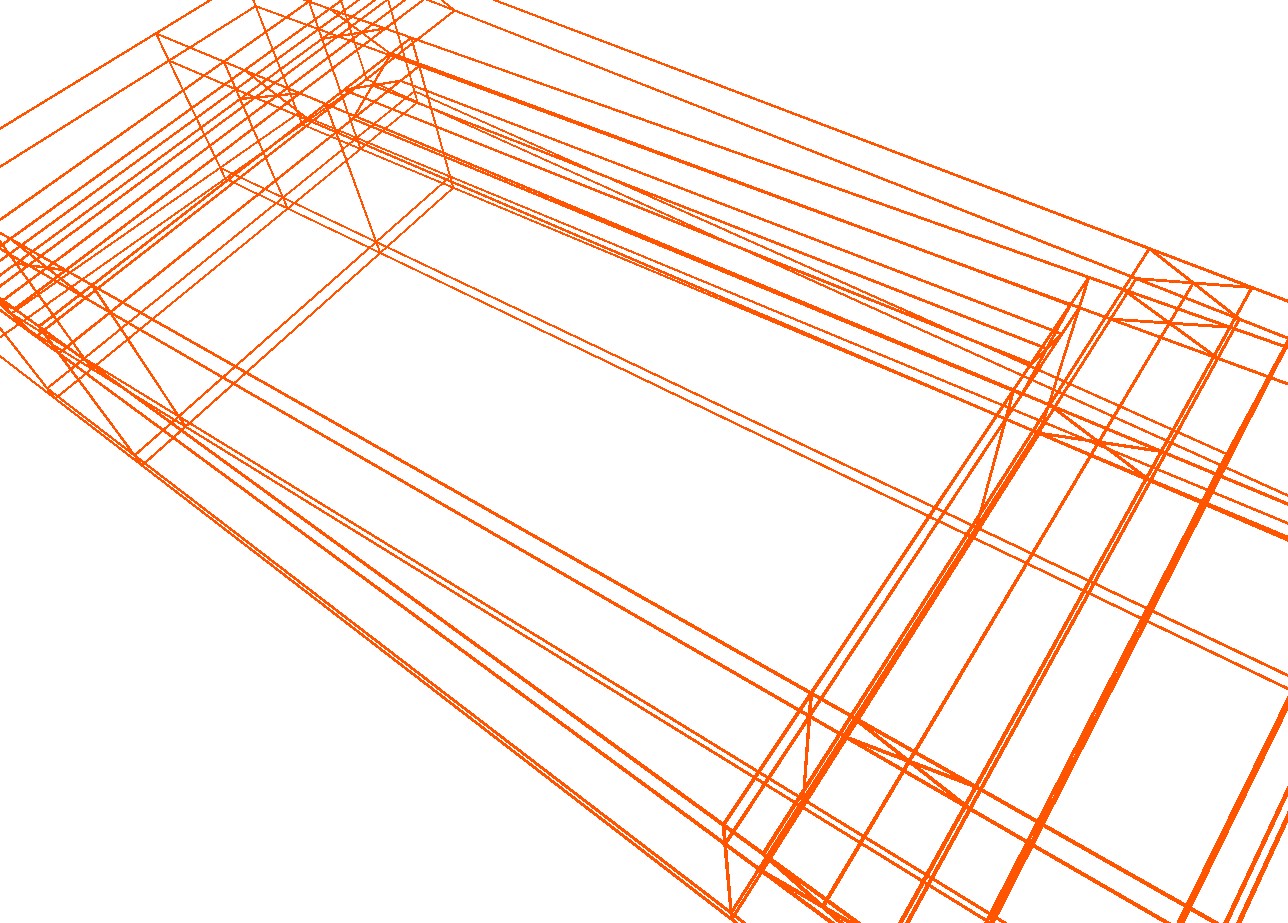}}
	\hspace{1em}
	\subfloat{\includegraphics[width=0.15\linewidth]{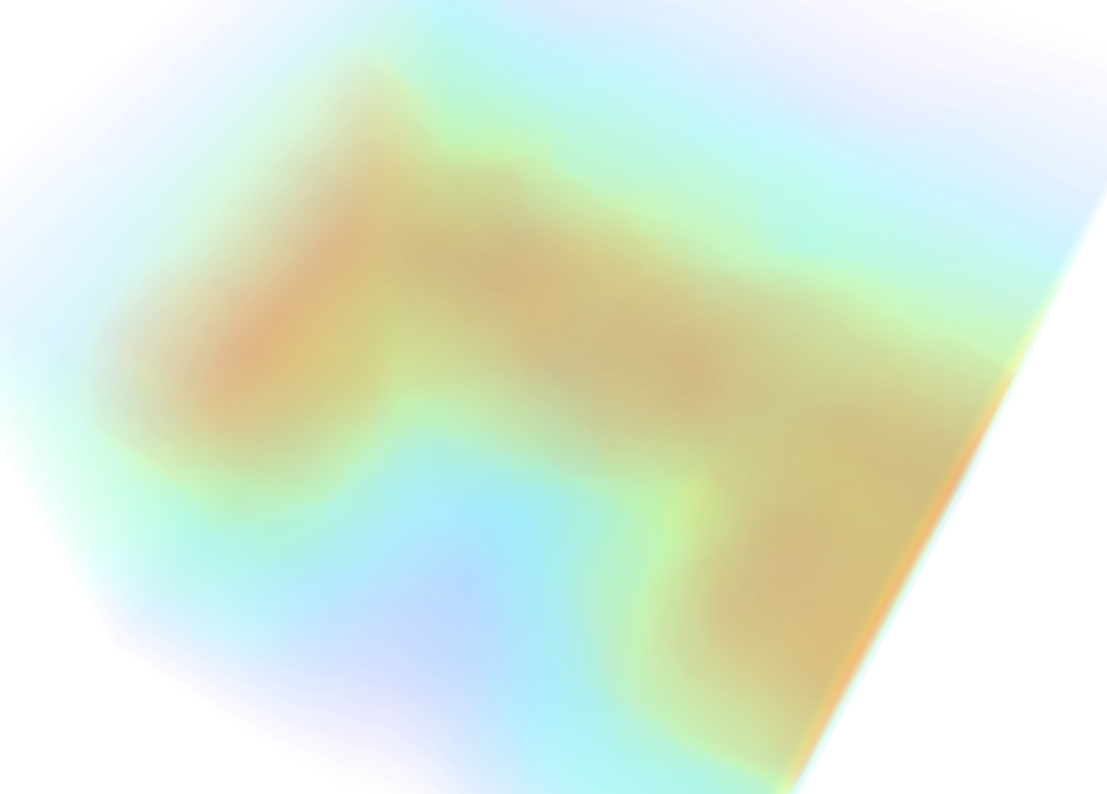}}
	\hspace{1em}
	\subfloat{\includegraphics[width=0.15\linewidth]{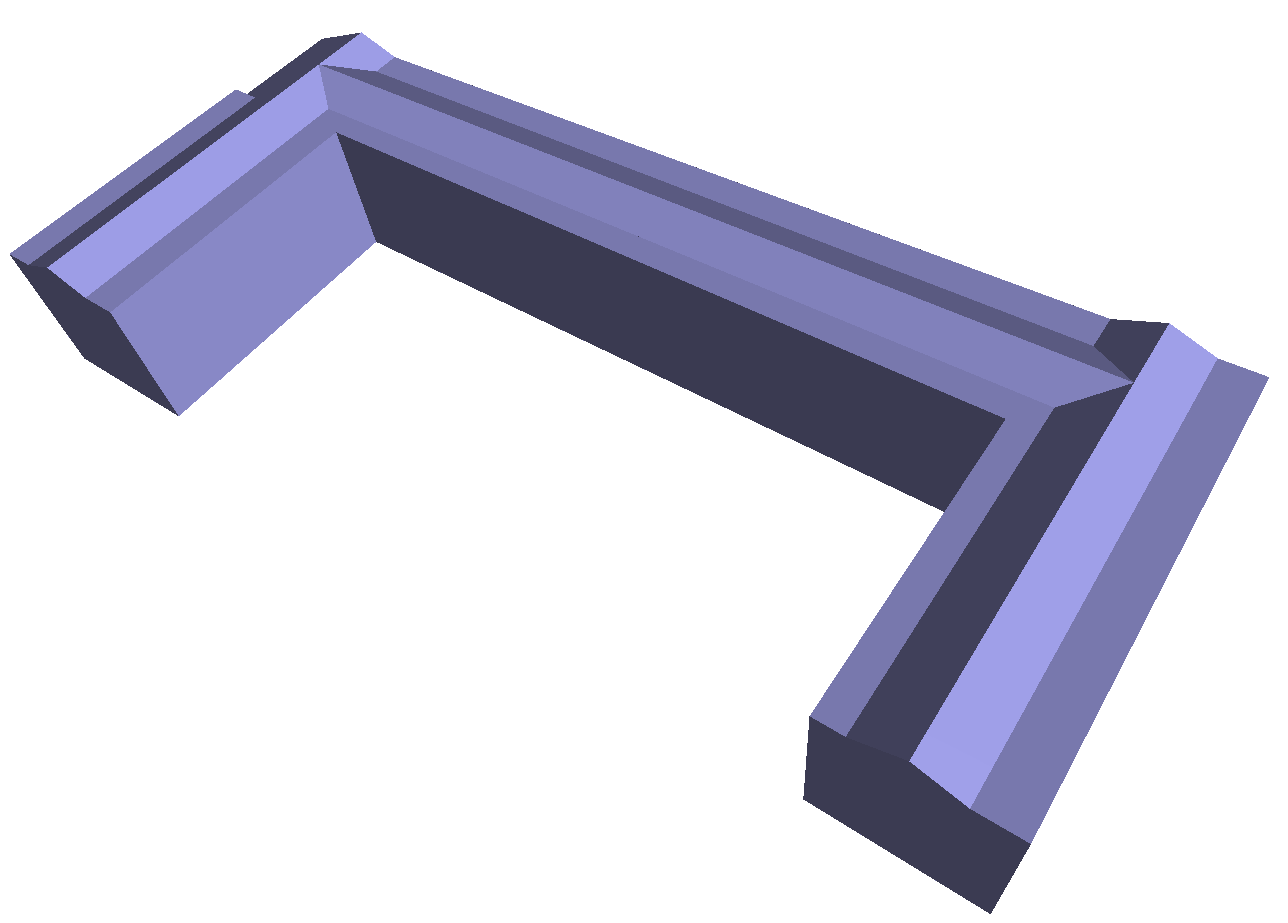}}
	\hspace{1em}
	\subfloat{\includegraphics[width=0.15\linewidth]{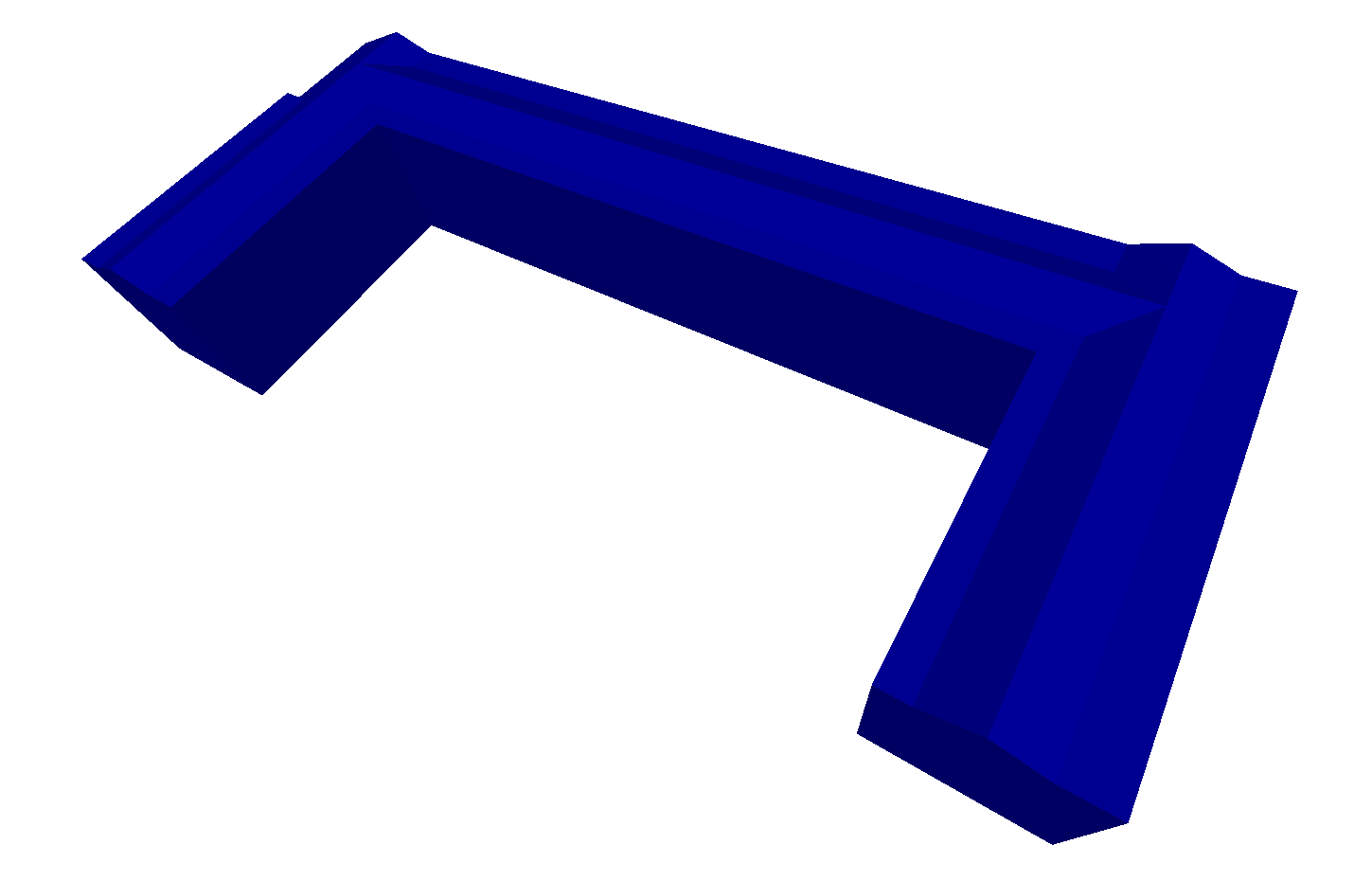}}    
	
	\vspace{0em}
	
	(b)
	\subfloat{\includegraphics[width=0.15\linewidth]{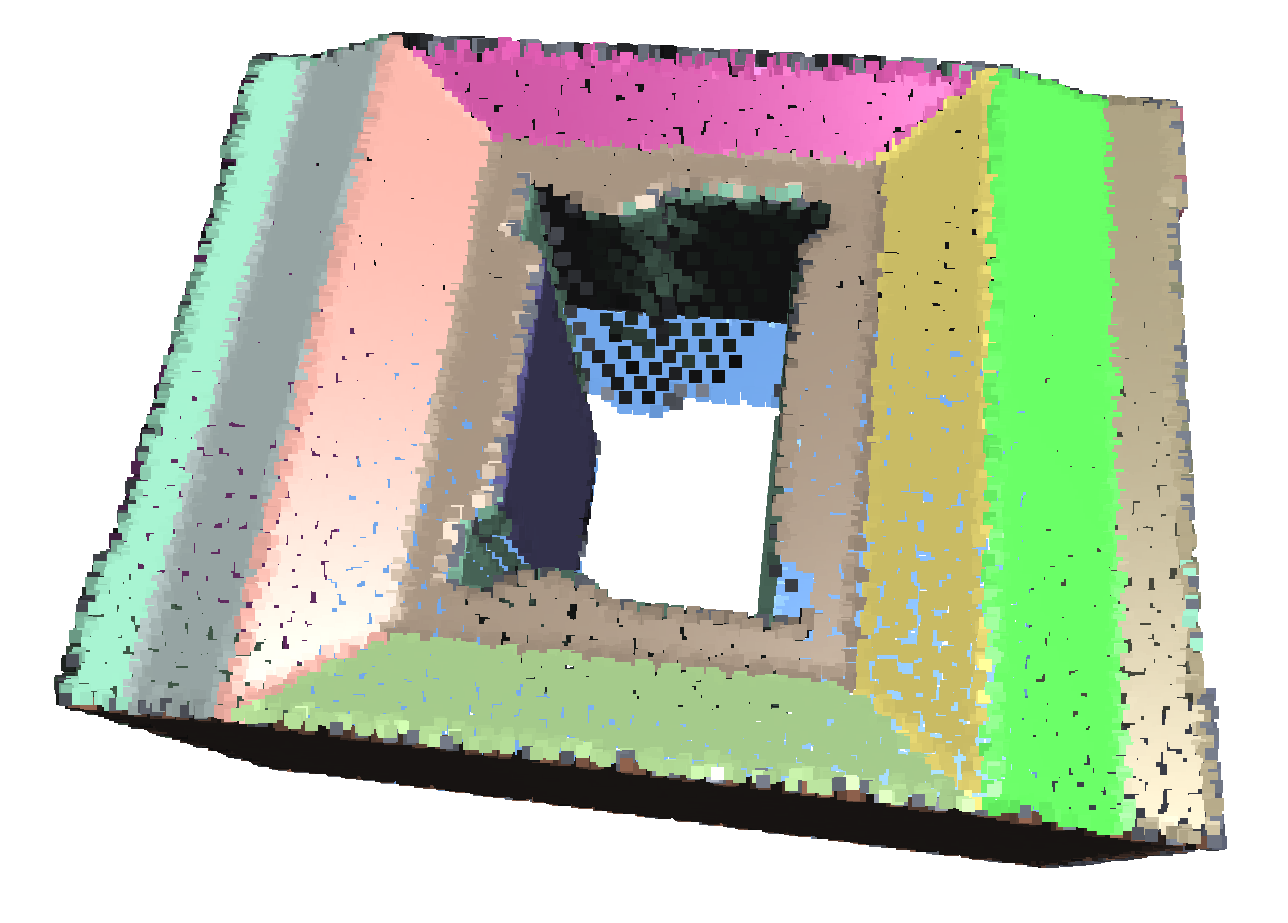}}
	\hspace{1em}
	\subfloat{\includegraphics[width=0.15\linewidth]{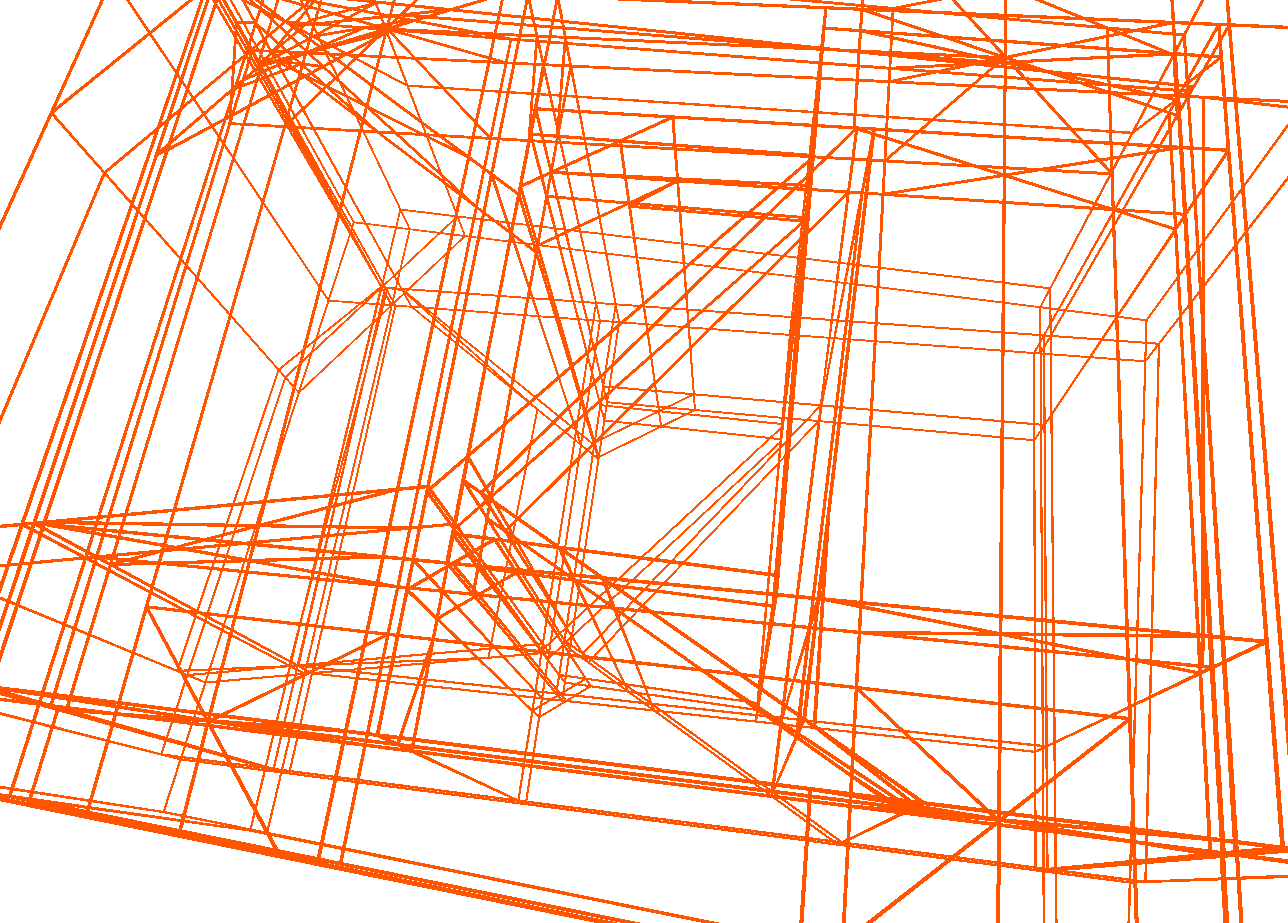}}
	\hspace{1em}
	\subfloat{\includegraphics[width=0.15\linewidth]{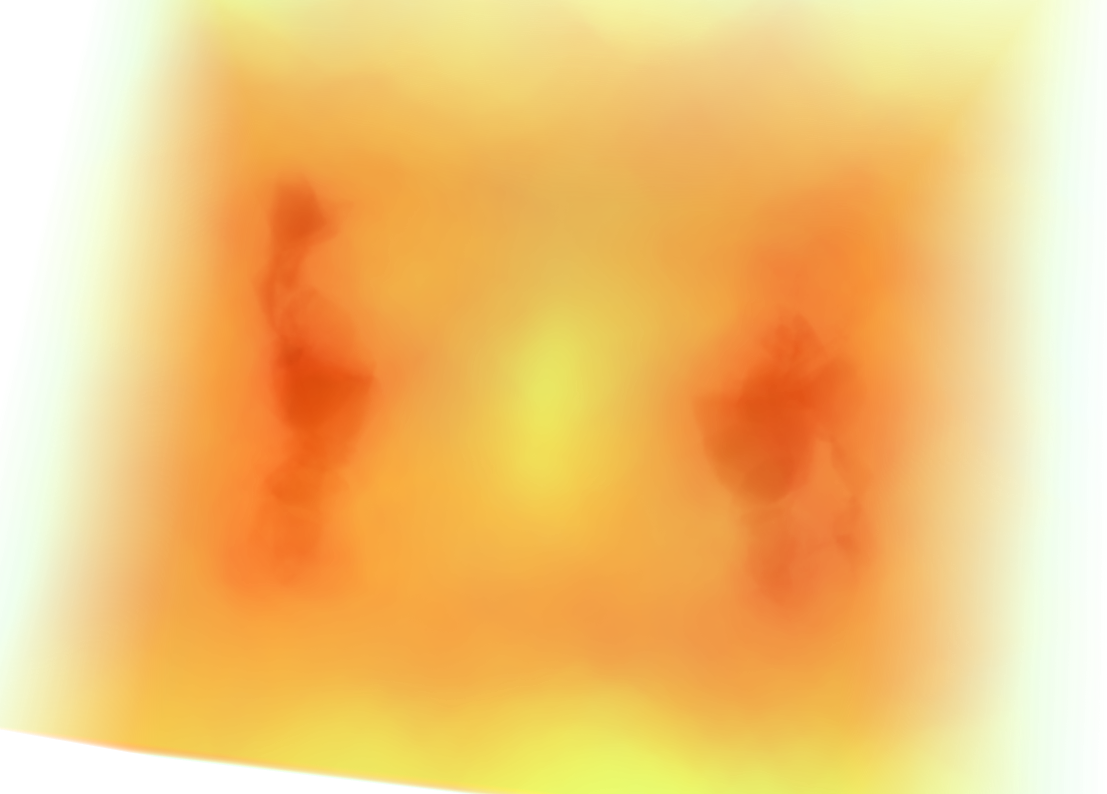}}
	\hspace{1em}
	\subfloat{\includegraphics[width=0.15\linewidth]{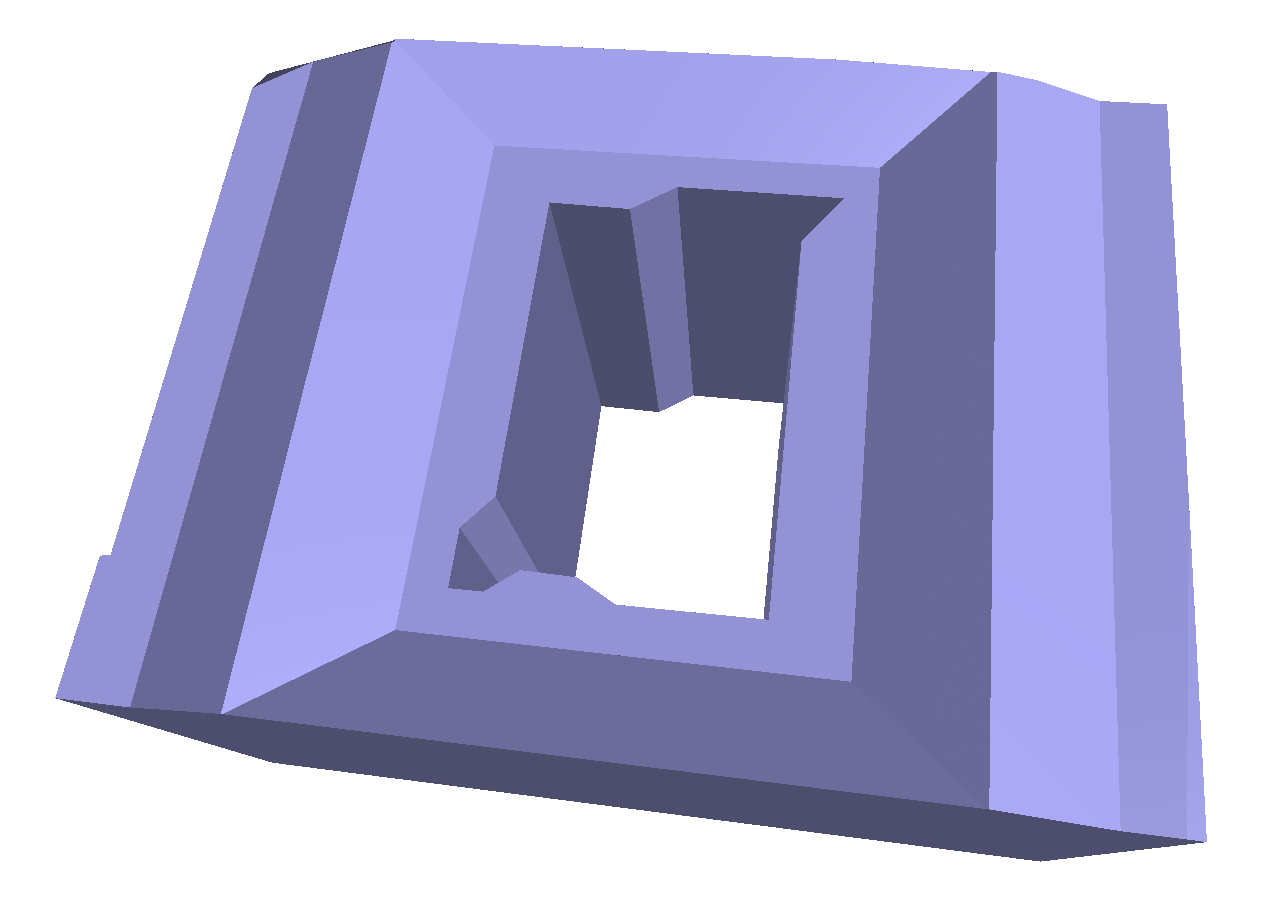}}
	\hspace{1em}
	\subfloat{\includegraphics[width=0.15\linewidth]{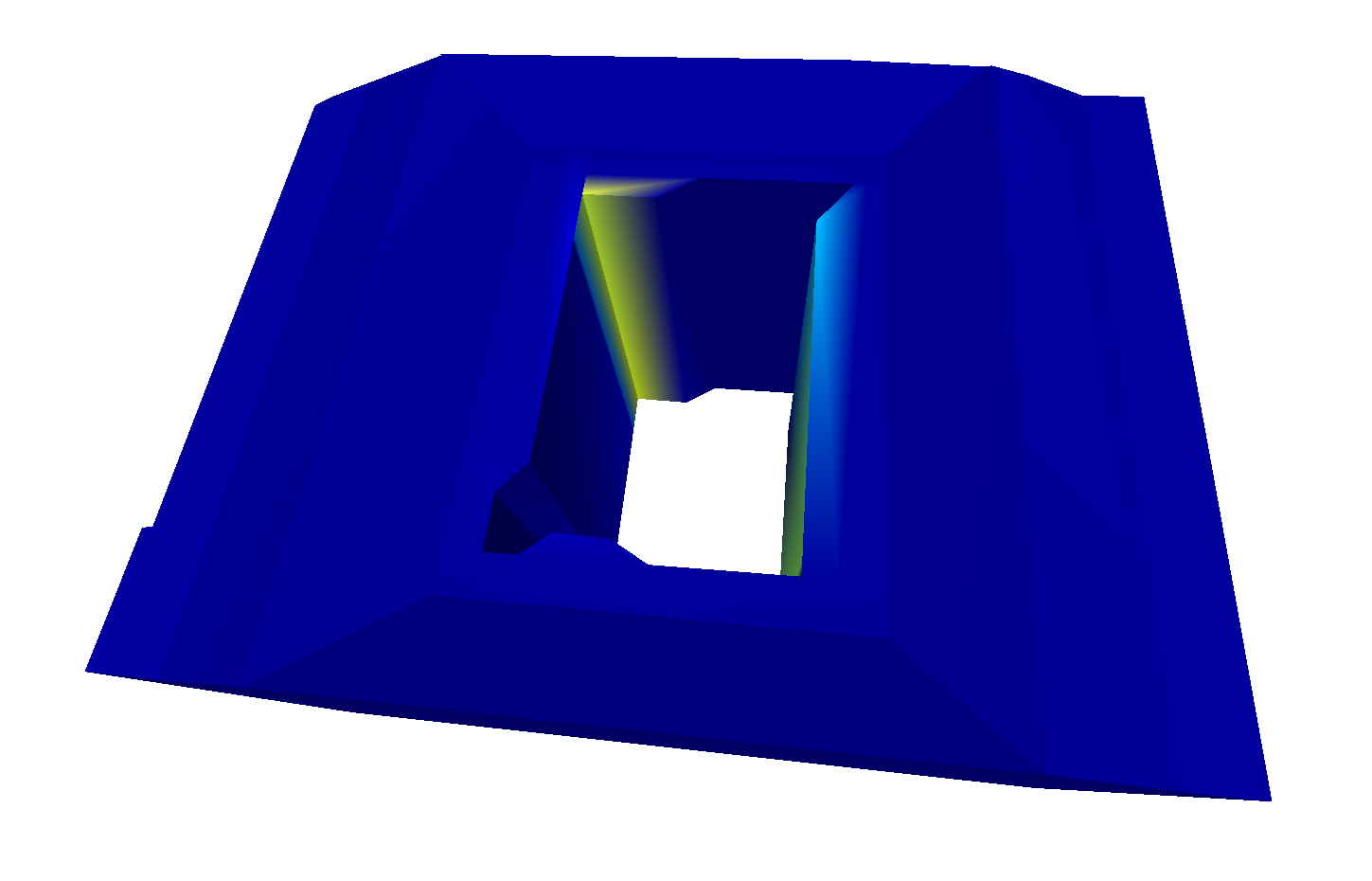}} 
	
	\vspace{0em}
	
	(c)
	\subfloat{\includegraphics[width=0.15\linewidth]{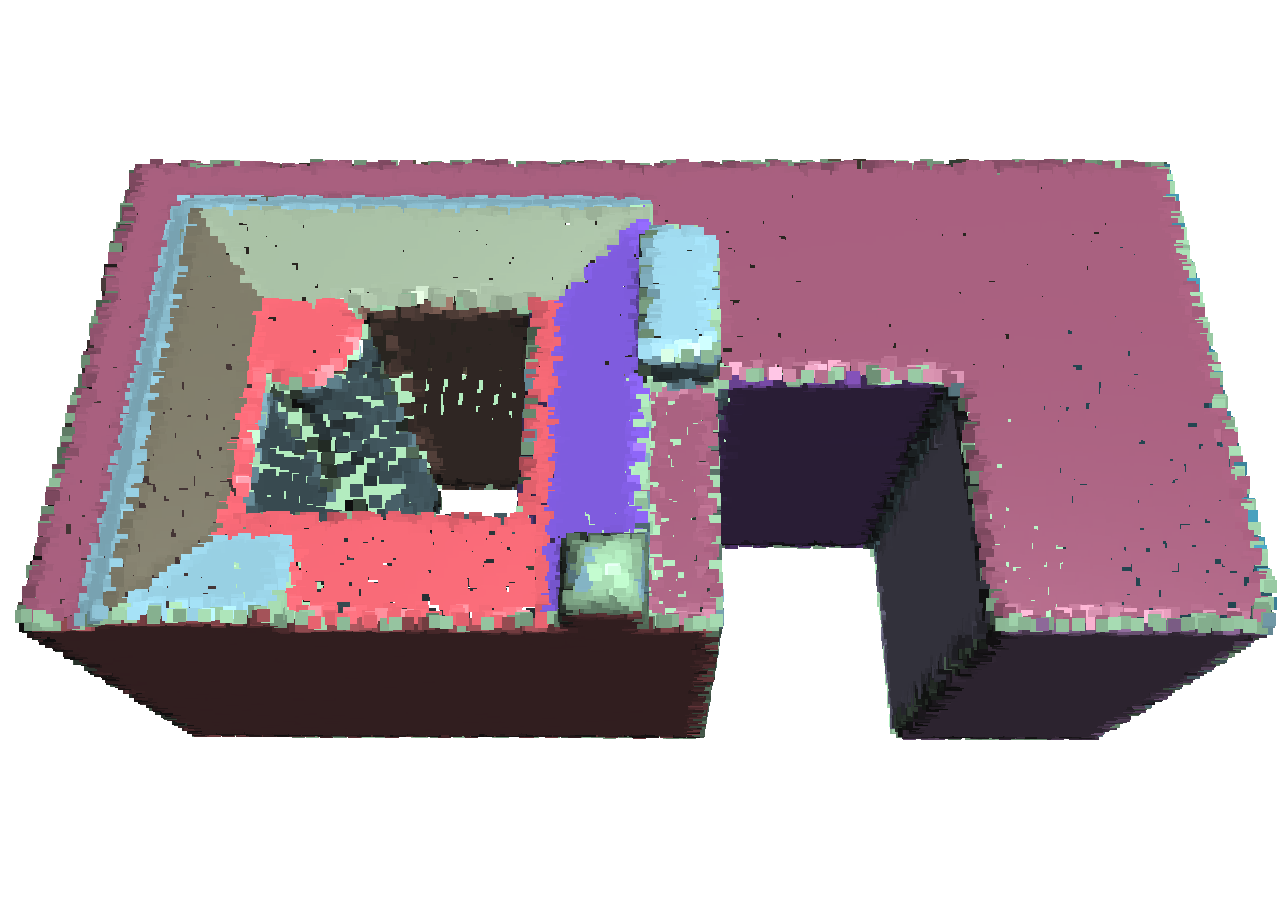}}
	\hspace{1em}
	\subfloat{\includegraphics[width=0.15\linewidth]{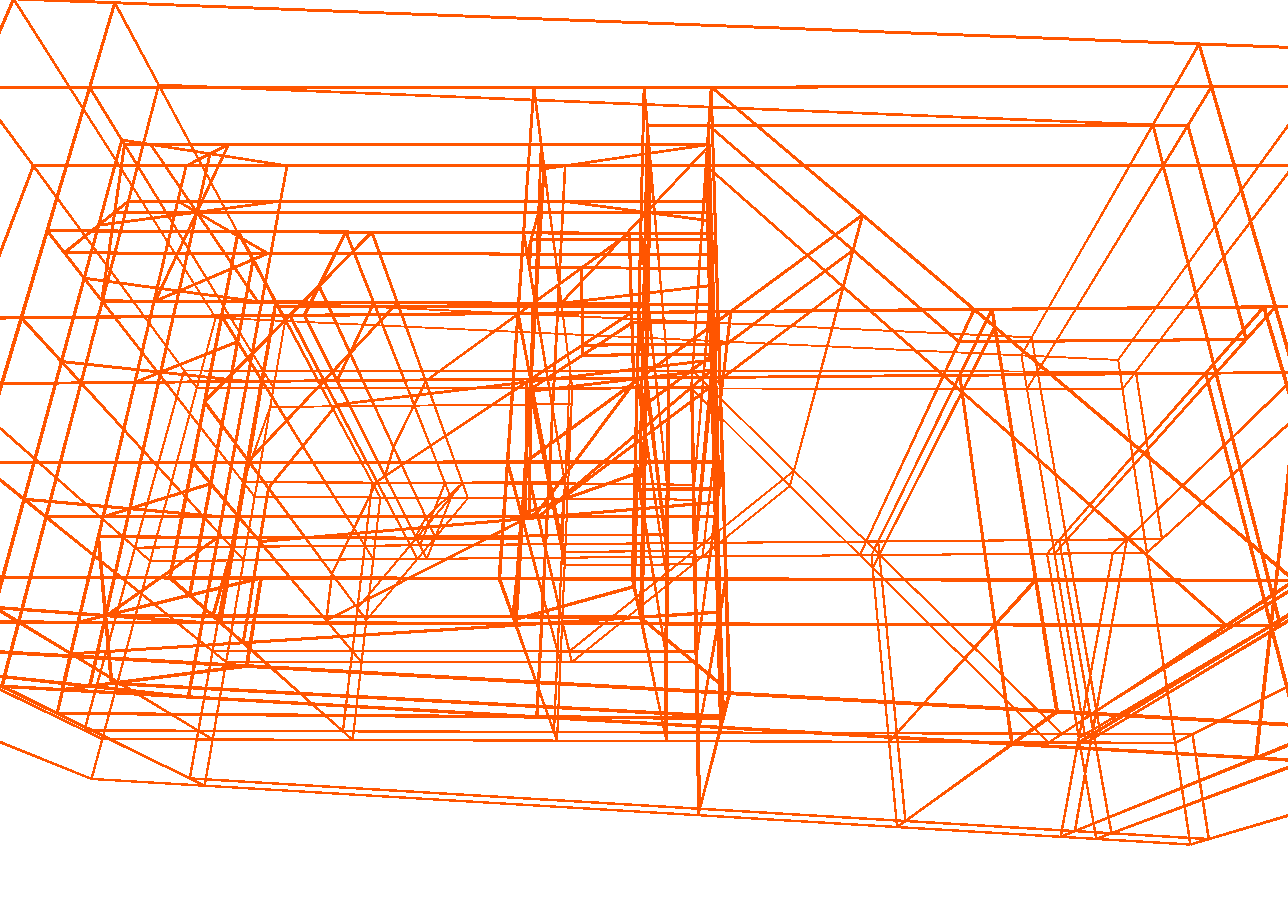}}
	\hspace{1em}
	\subfloat{\includegraphics[width=0.15\linewidth]{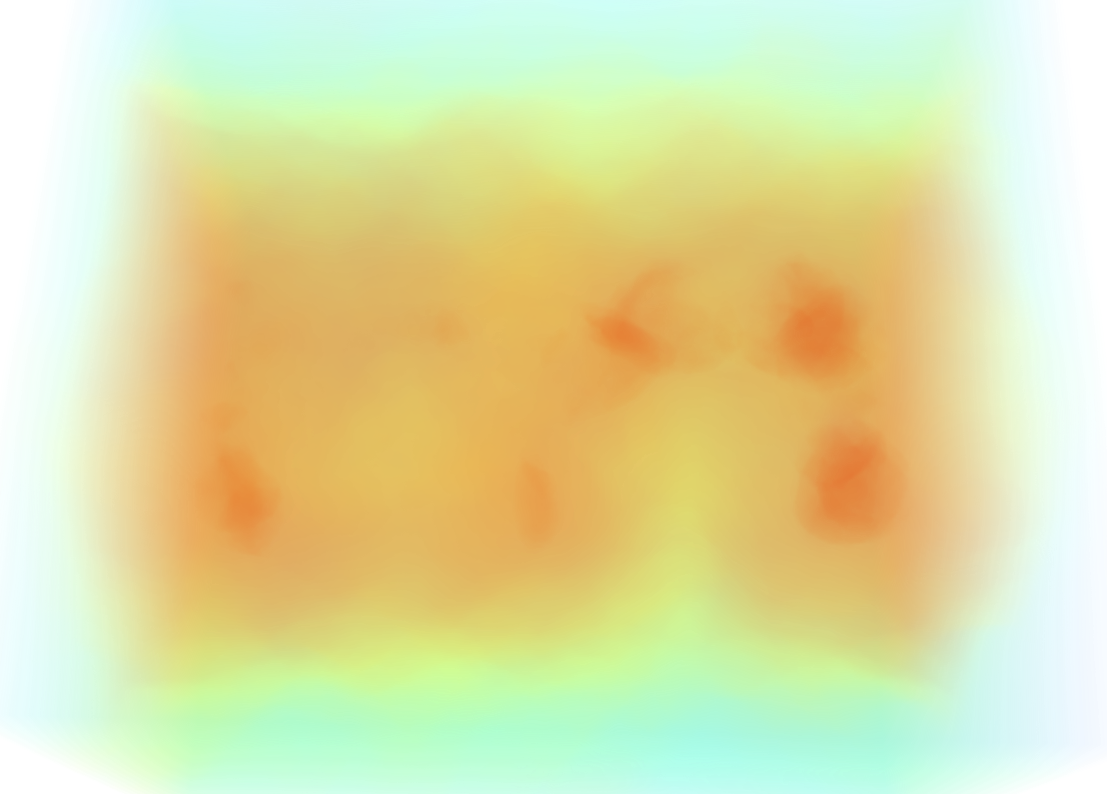}}
	\hspace{1em}
	\subfloat{\includegraphics[width=0.15\linewidth]{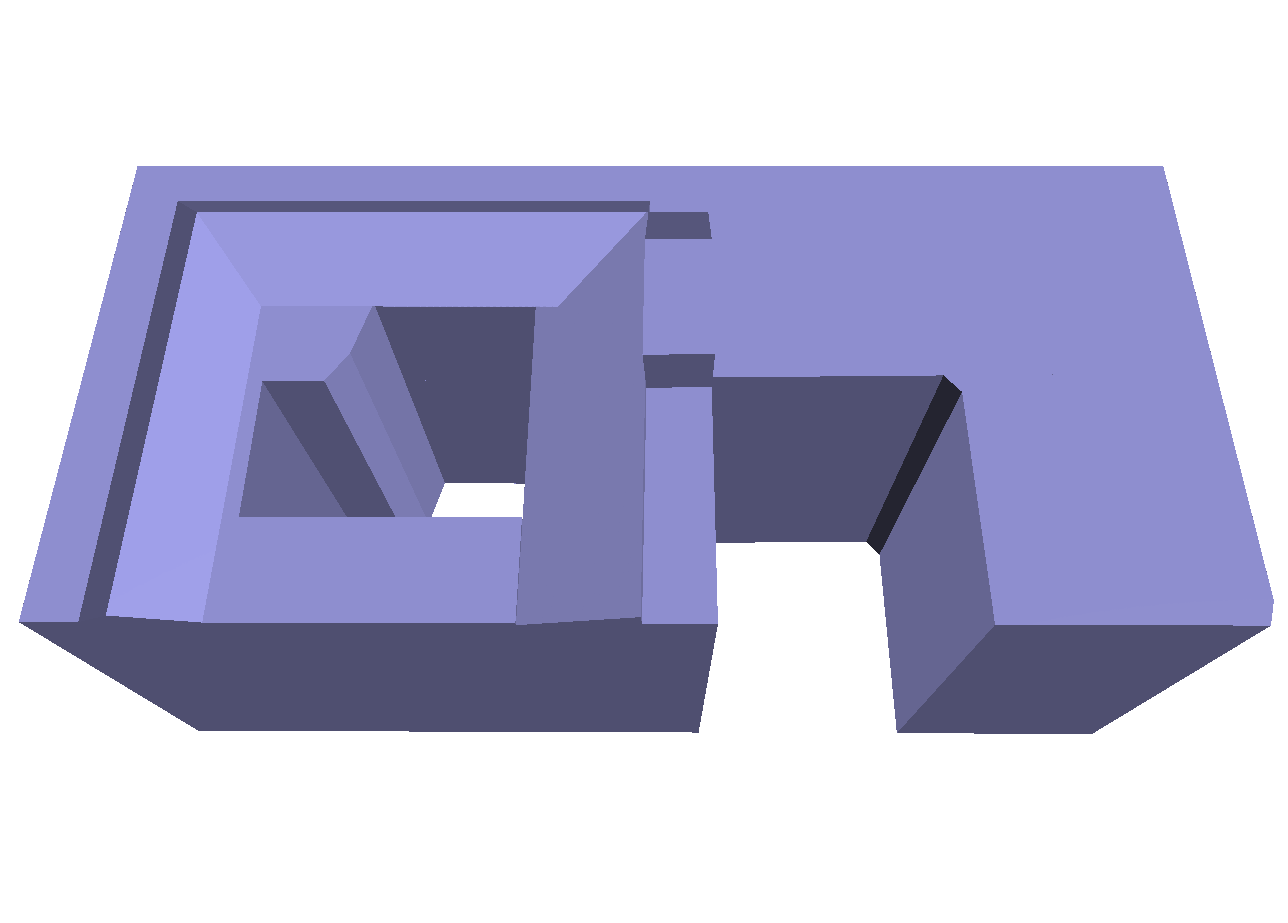}}
	\hspace{1em}
	\subfloat{\includegraphics[width=0.15\linewidth]{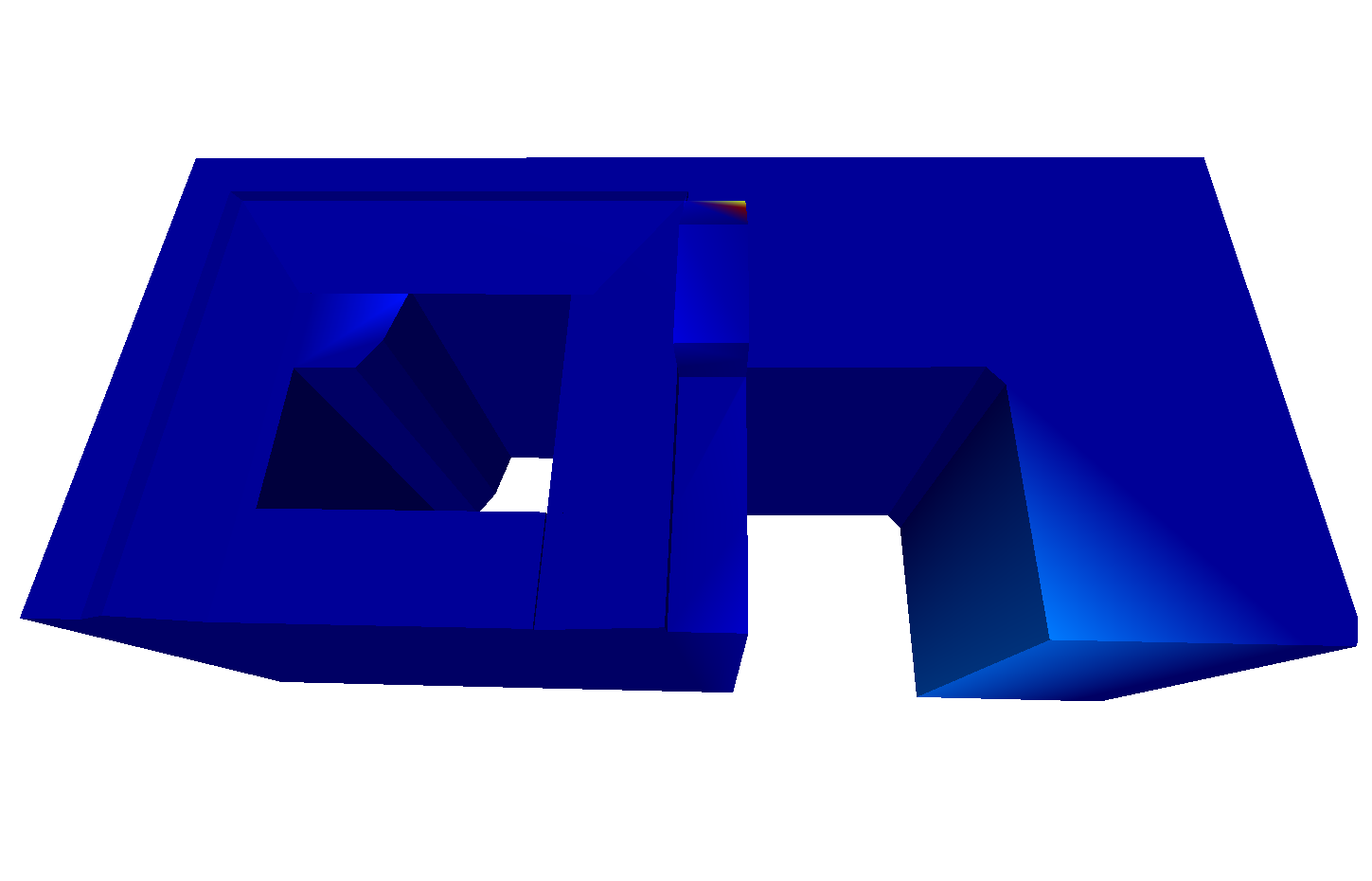}} 
	
	\vspace{0em}
	
	(d)
	\subfloat{\includegraphics[width=0.15\linewidth]{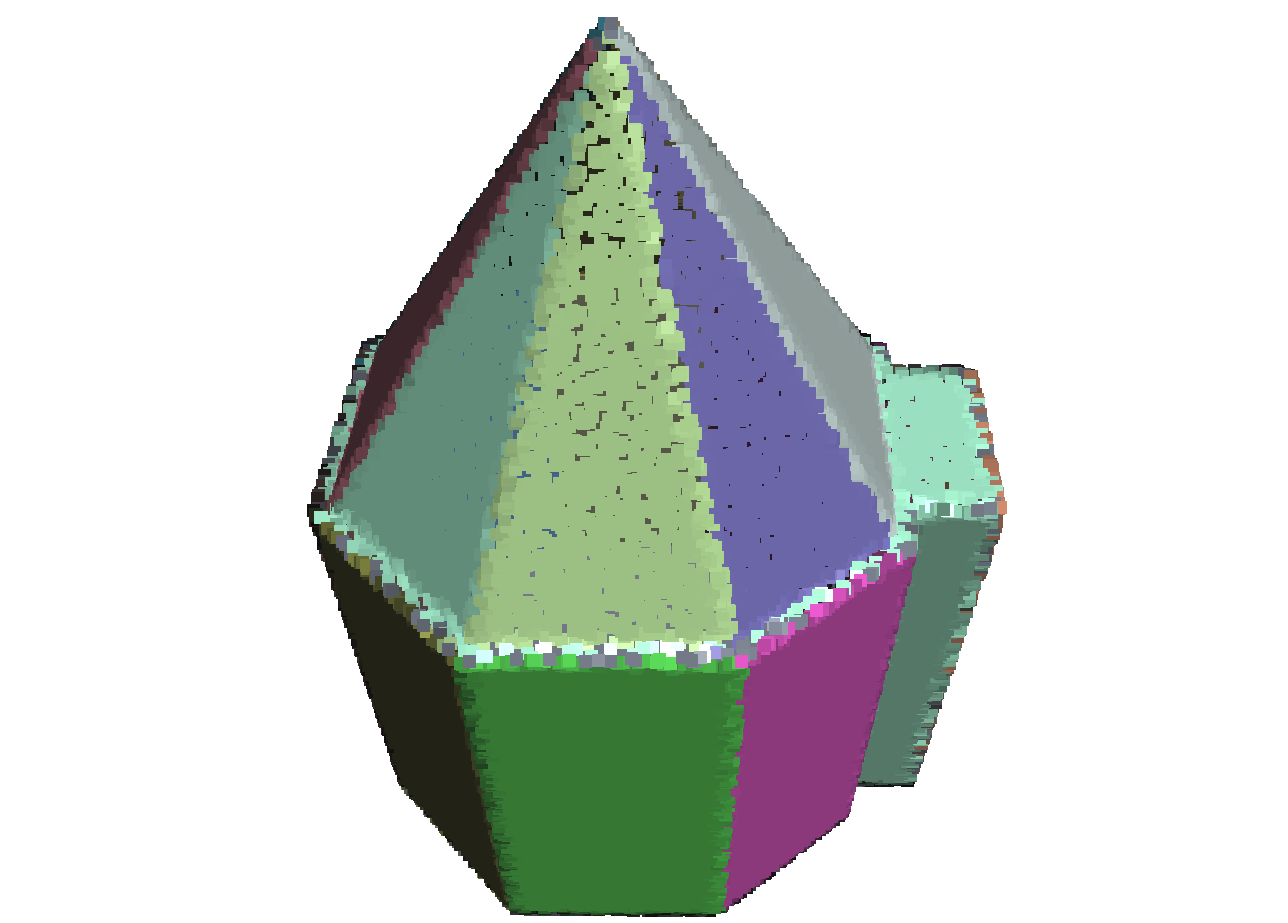}}
	\hspace{1em}
	\subfloat{\includegraphics[width=0.15\linewidth]{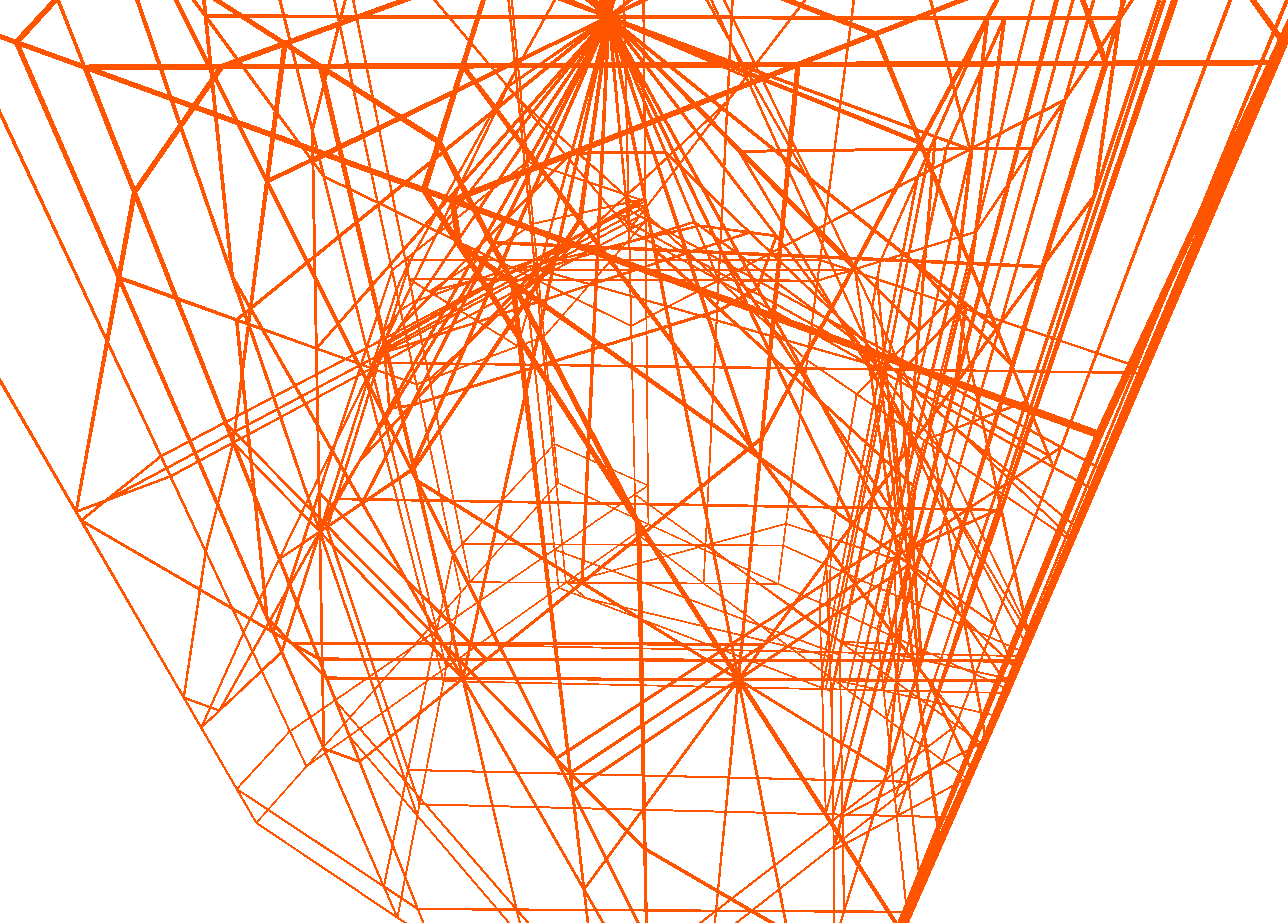}}
	\hspace{1em}
	\subfloat{\includegraphics[width=0.15\linewidth]{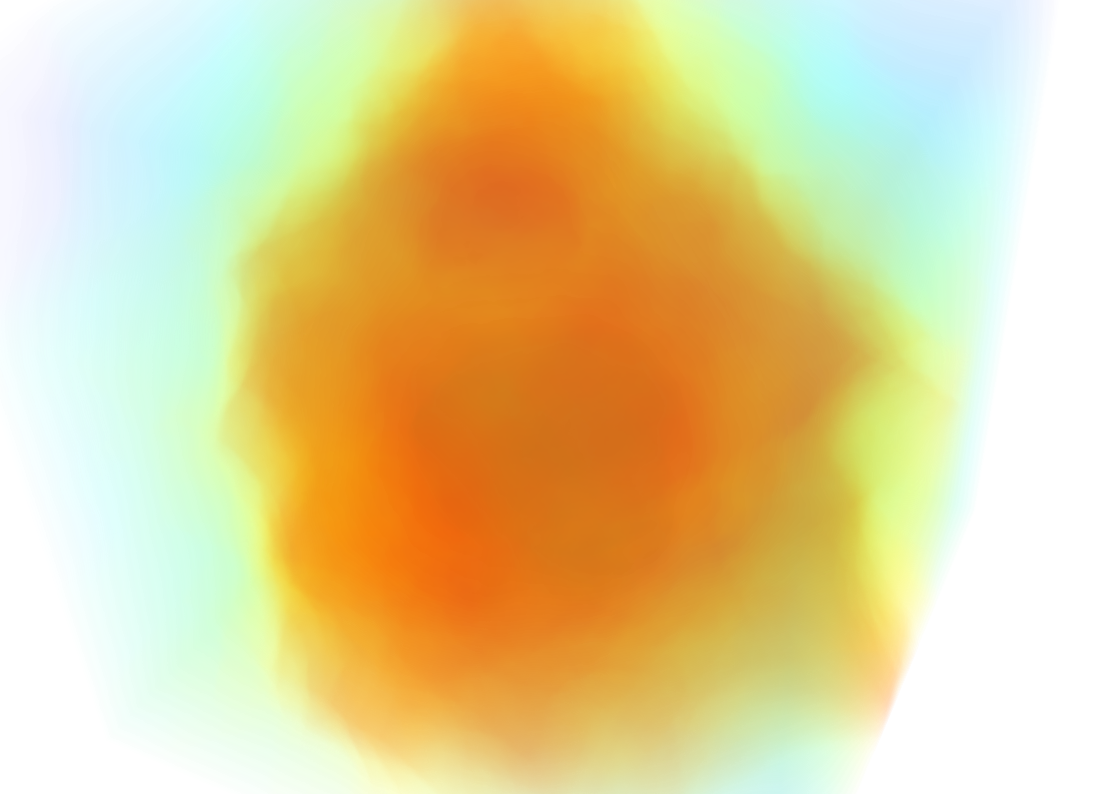}}
	\hspace{1em}
	\subfloat{\includegraphics[width=0.15\linewidth]{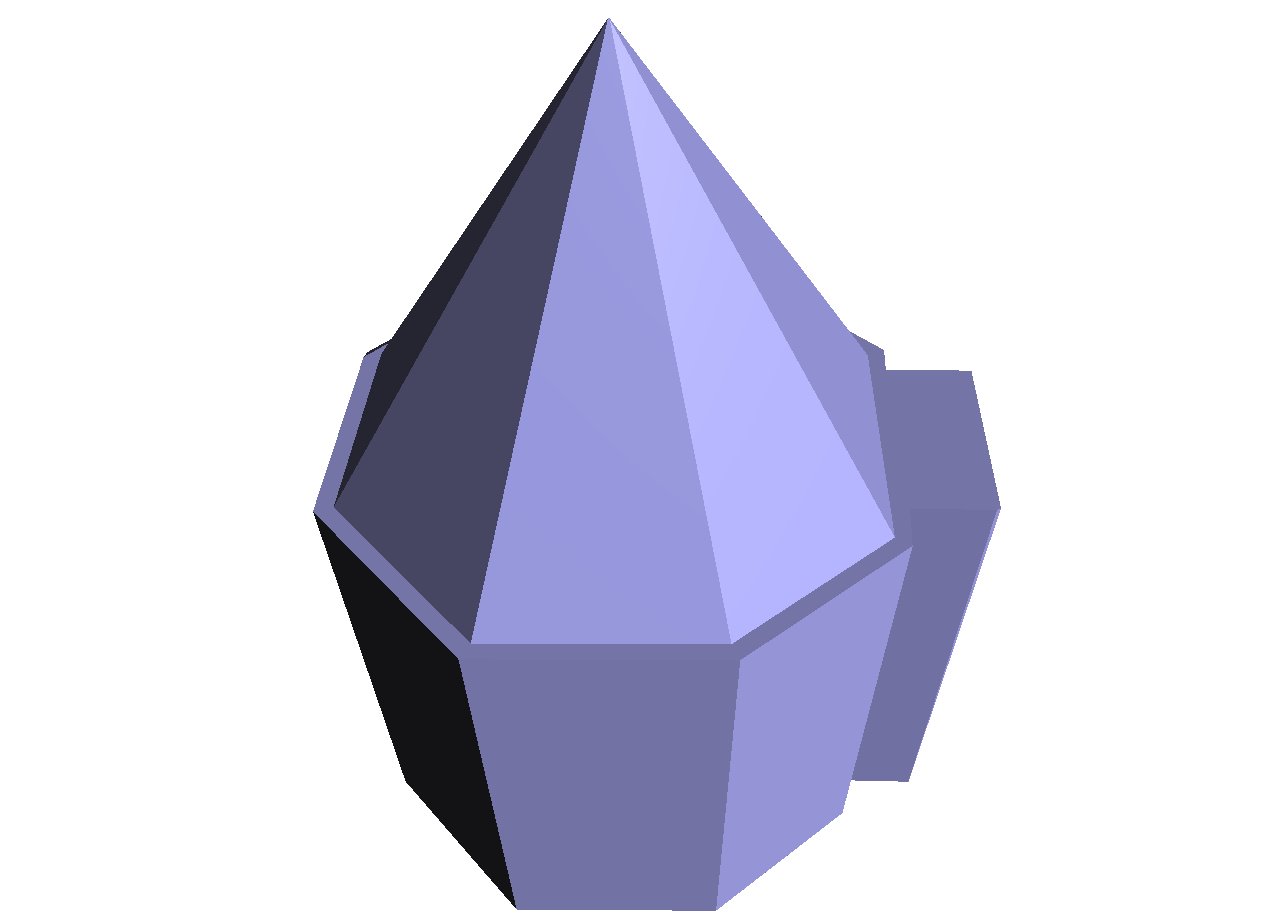}}
	\hspace{1em}
	\subfloat{\includegraphics[width=0.15\linewidth]{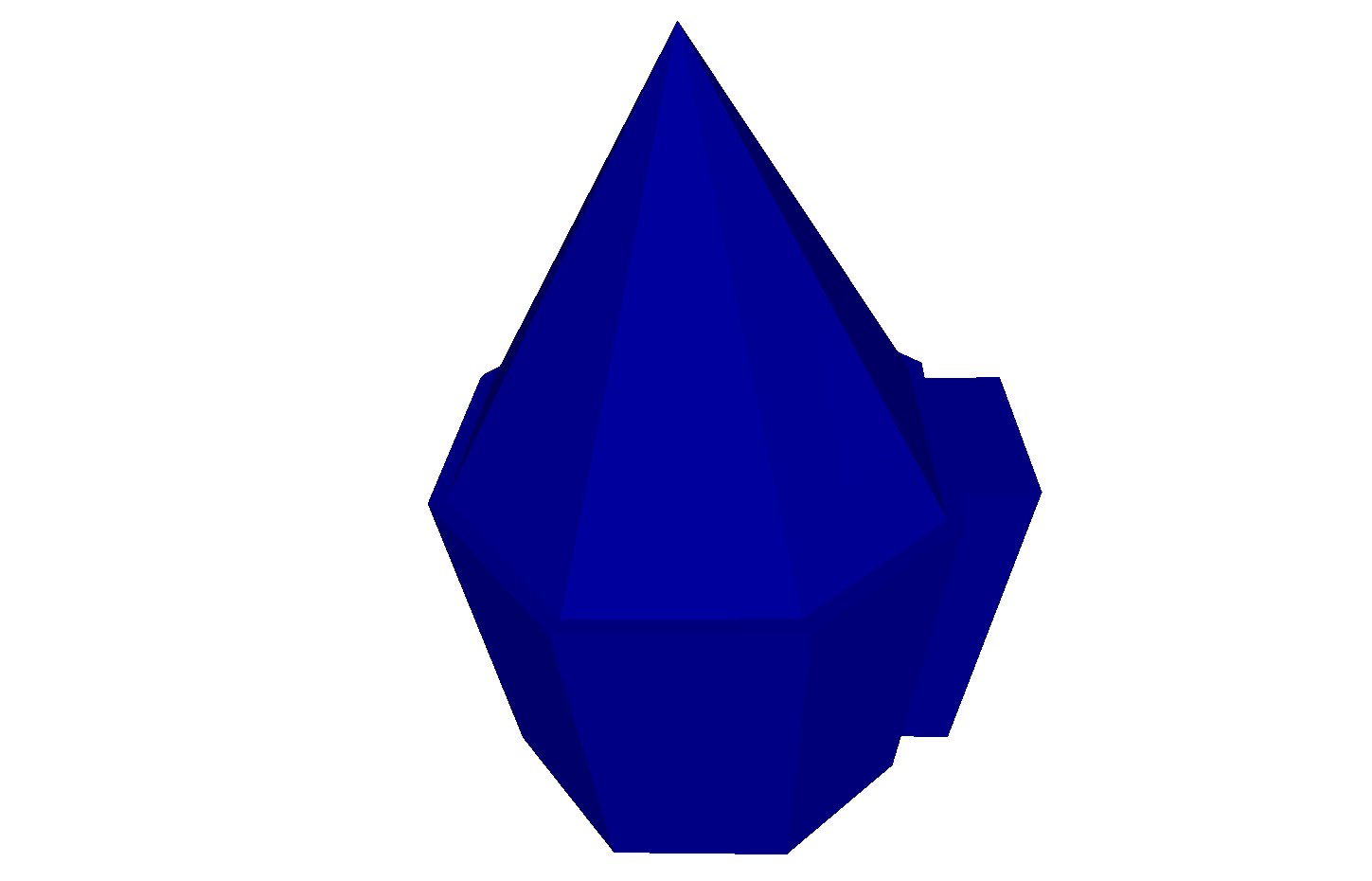}} 
	
	\vspace{0em}

	(e)
	\subfloat{\includegraphics[width=0.15\linewidth]{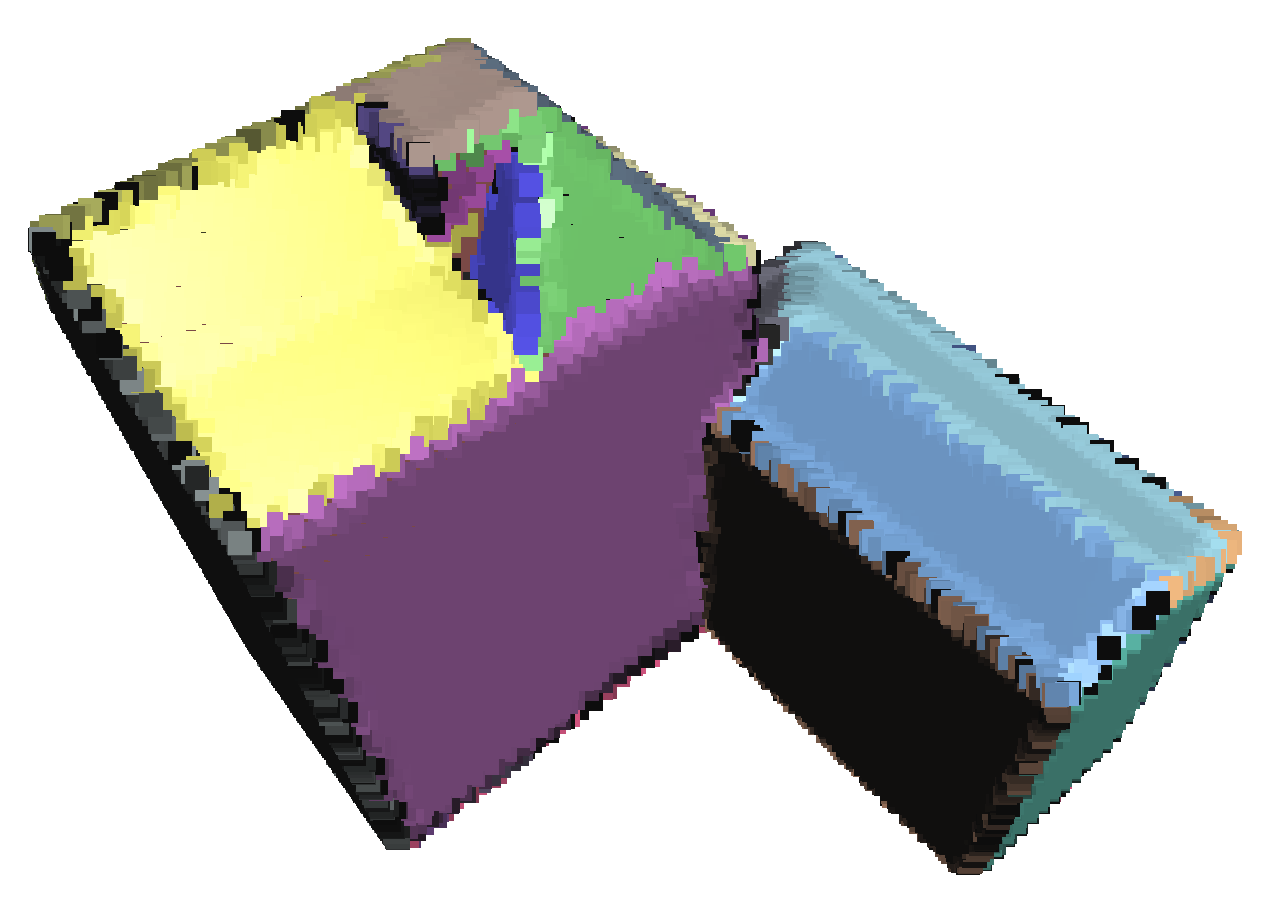}}
	\hspace{1em}
	\subfloat{\includegraphics[width=0.15\linewidth]{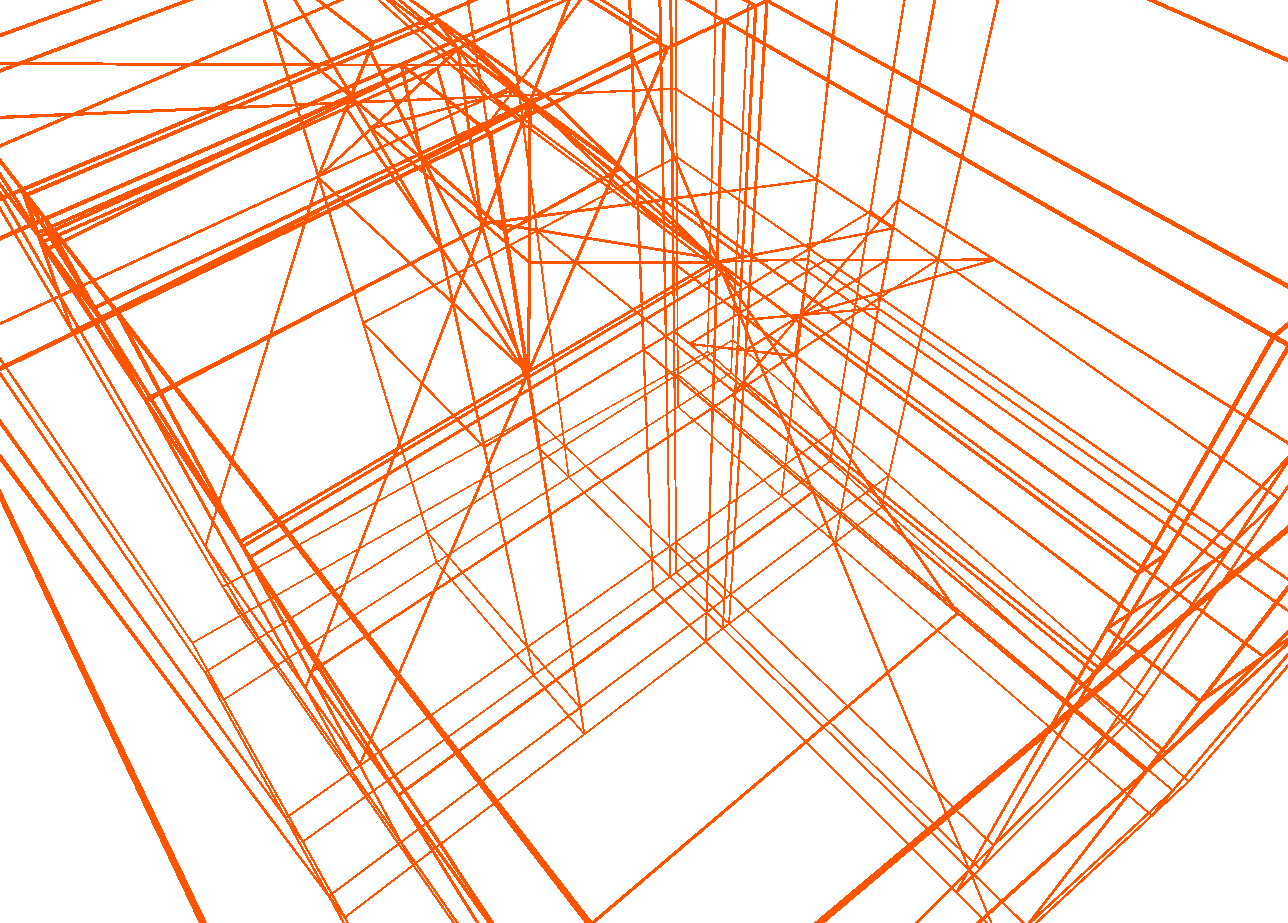}}
	\hspace{1em}
	\subfloat{\includegraphics[width=0.15\linewidth]{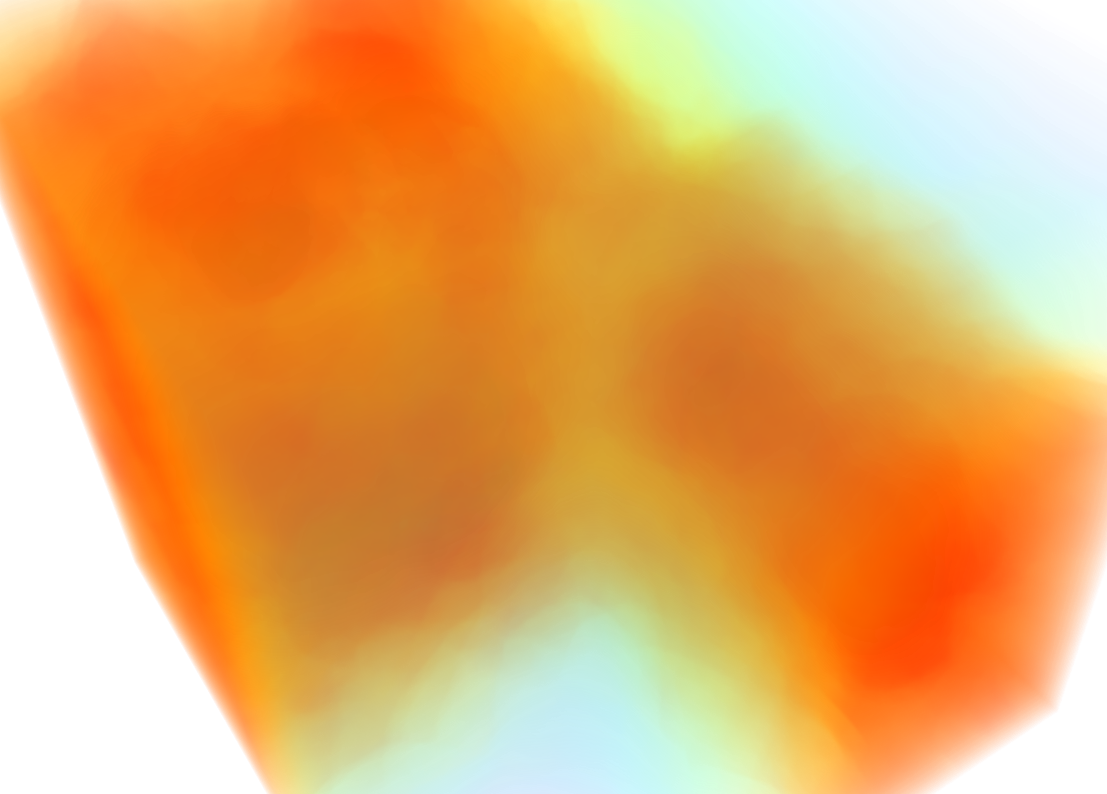}}
	\hspace{1em}
	\subfloat{\includegraphics[width=0.15\linewidth]{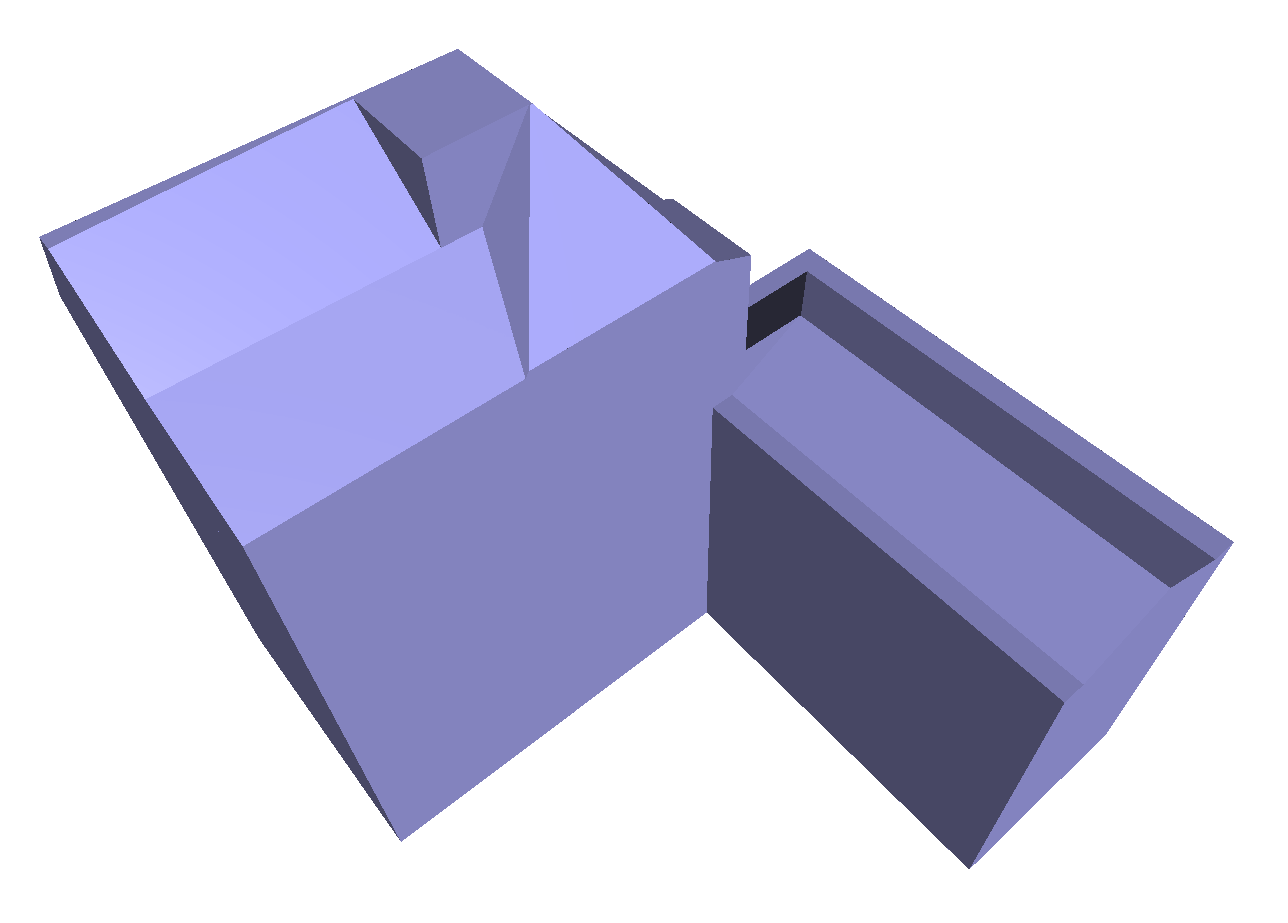}}
	\hspace{1em}
	\subfloat{\includegraphics[width=0.15\linewidth]{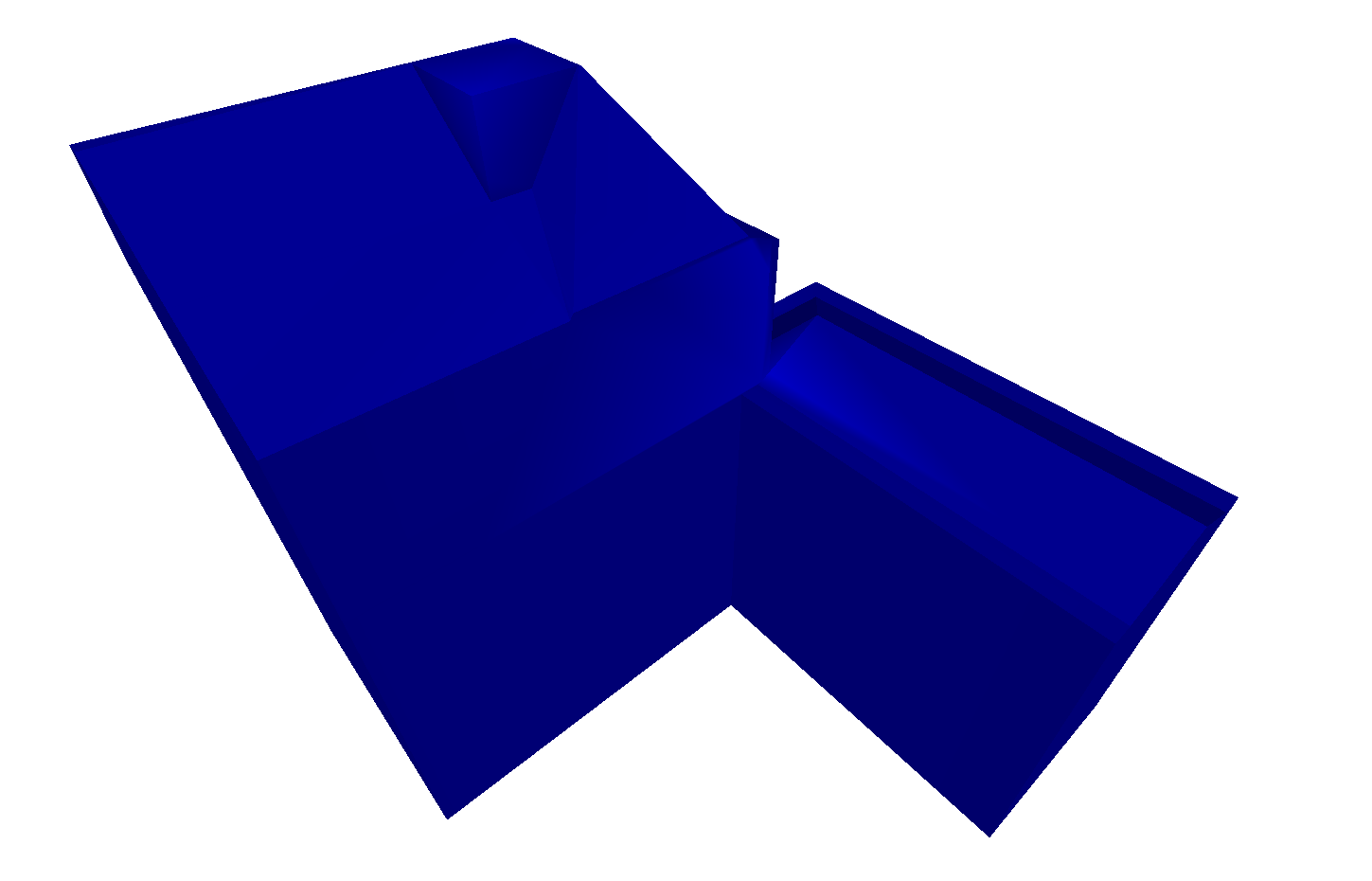}} 
	
	\vspace{0em}
	
	(f)
	\subfloat{\includegraphics[width=0.15\linewidth]{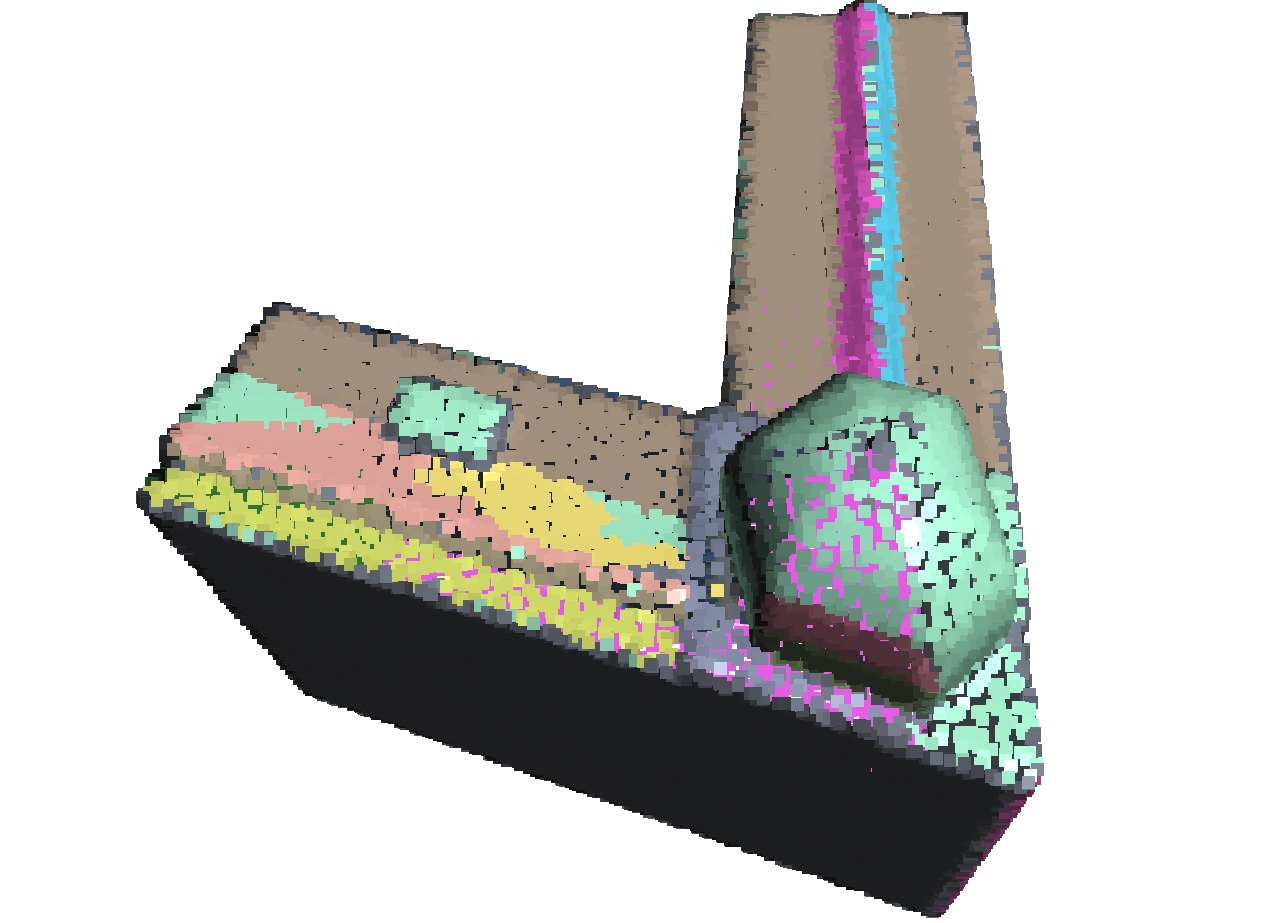}}
	\hspace{1em}
	\subfloat{\includegraphics[width=0.15\linewidth]{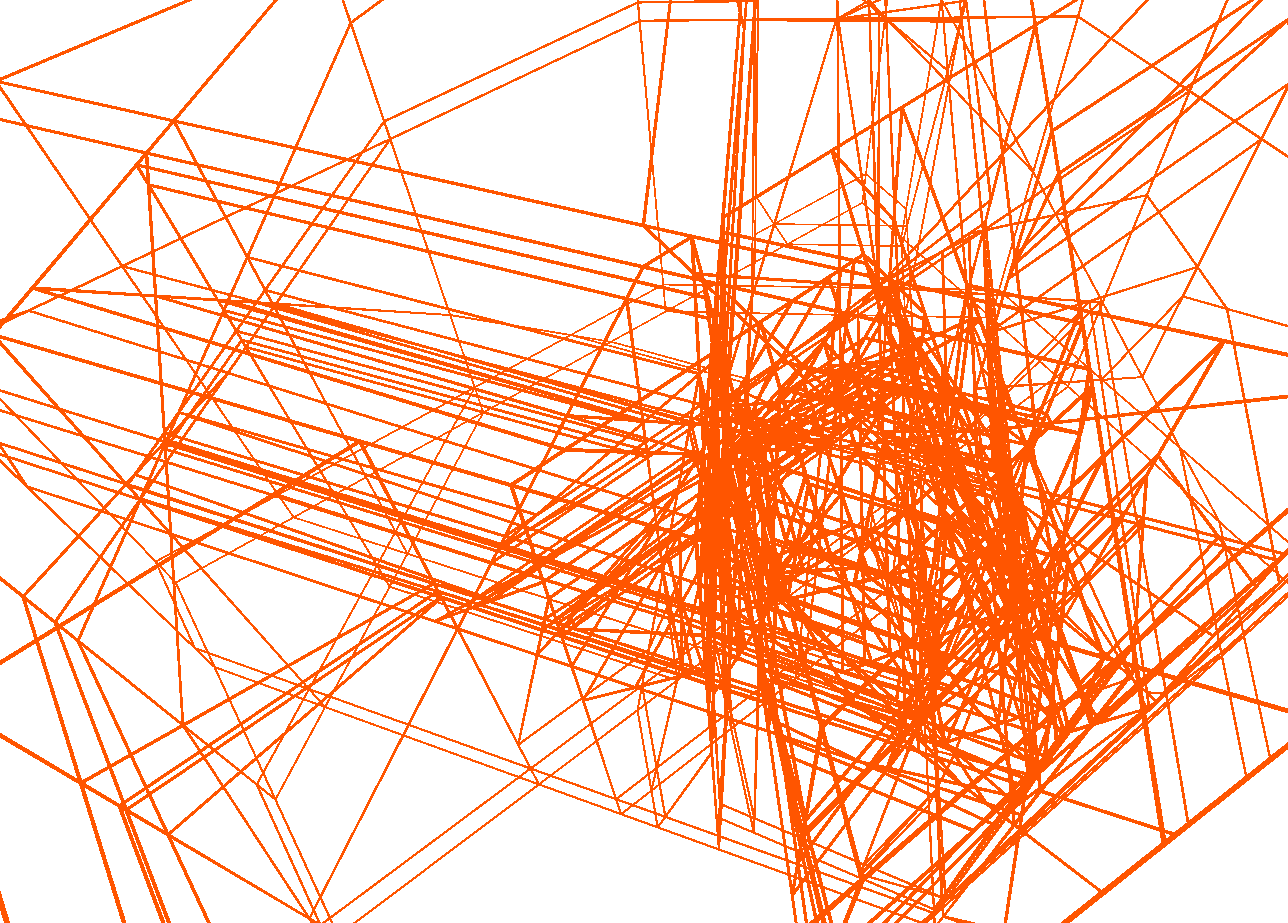}}
	\hspace{1em}
	\subfloat{\includegraphics[width=0.15\linewidth]{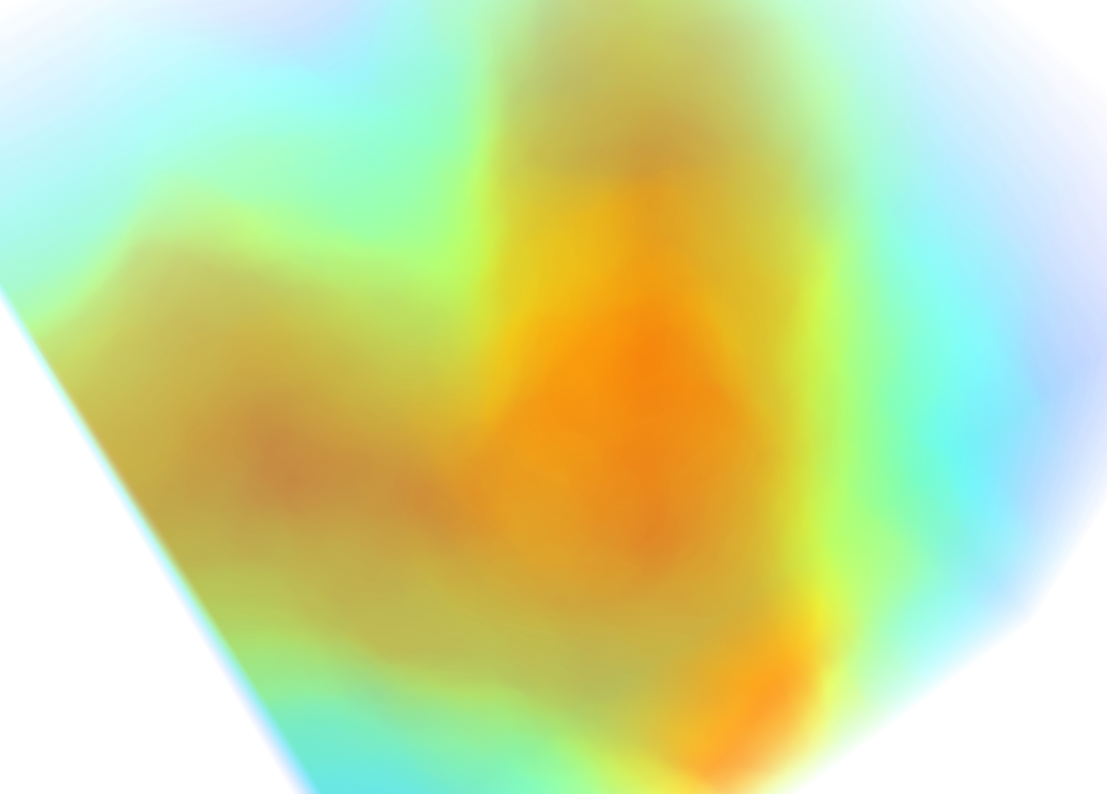}}
	\hspace{1em}
	\subfloat{\includegraphics[width=0.15\linewidth]{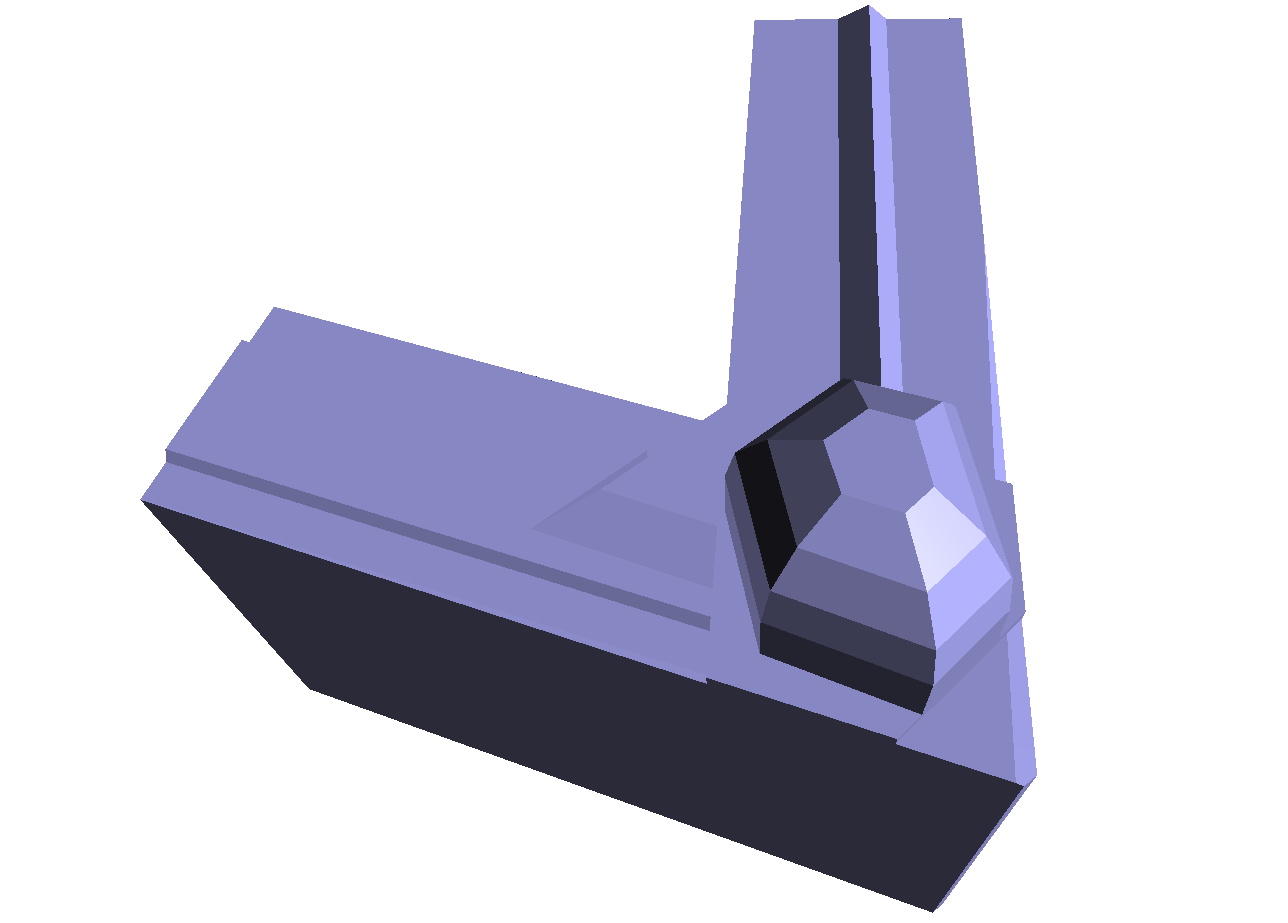}}
	\hspace{1em}
	\subfloat{\includegraphics[width=0.15\linewidth]{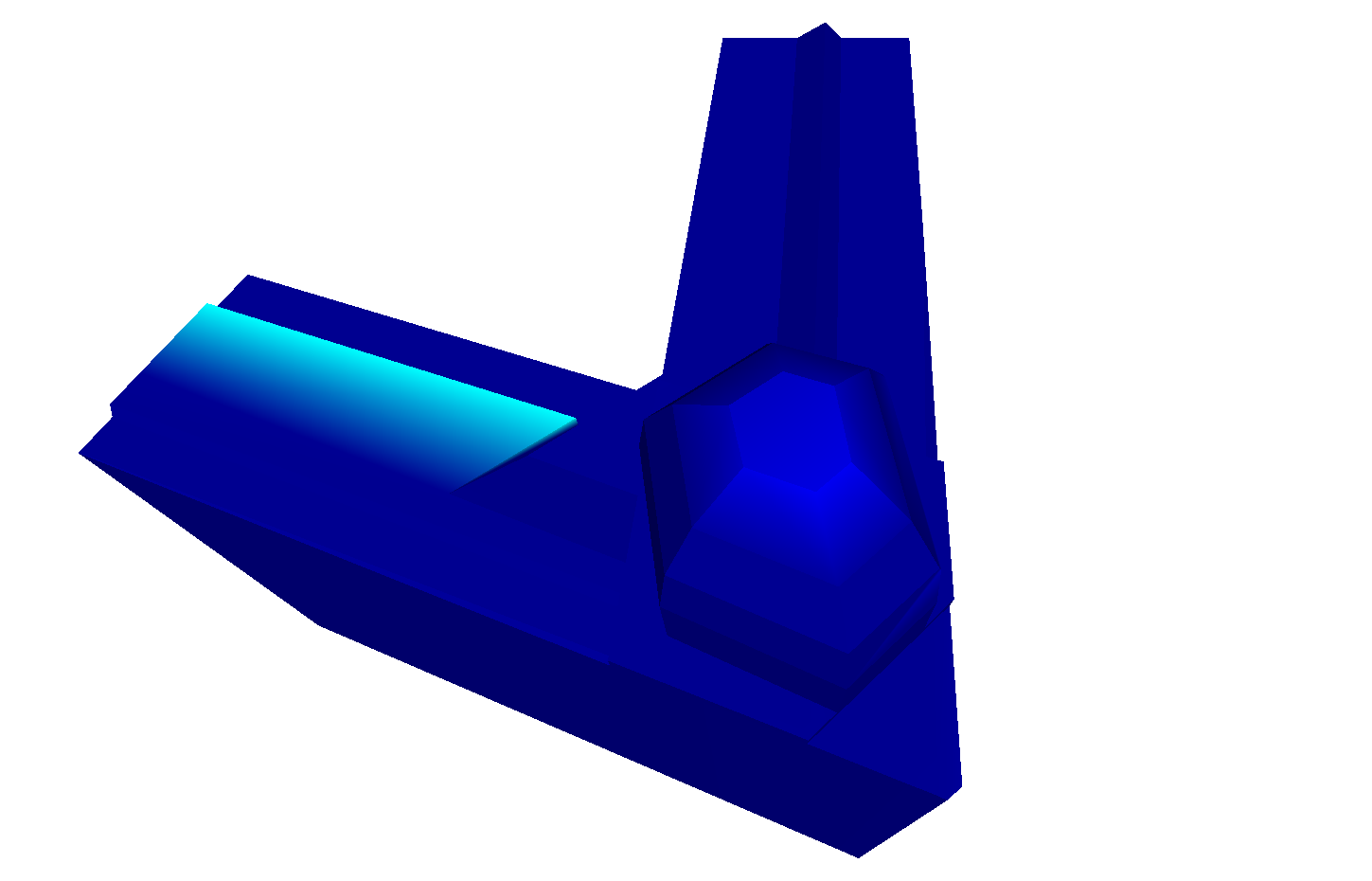}} 
	
	\vspace{0em}

	(g)
	\subfloat{\includegraphics[width=0.15\linewidth]{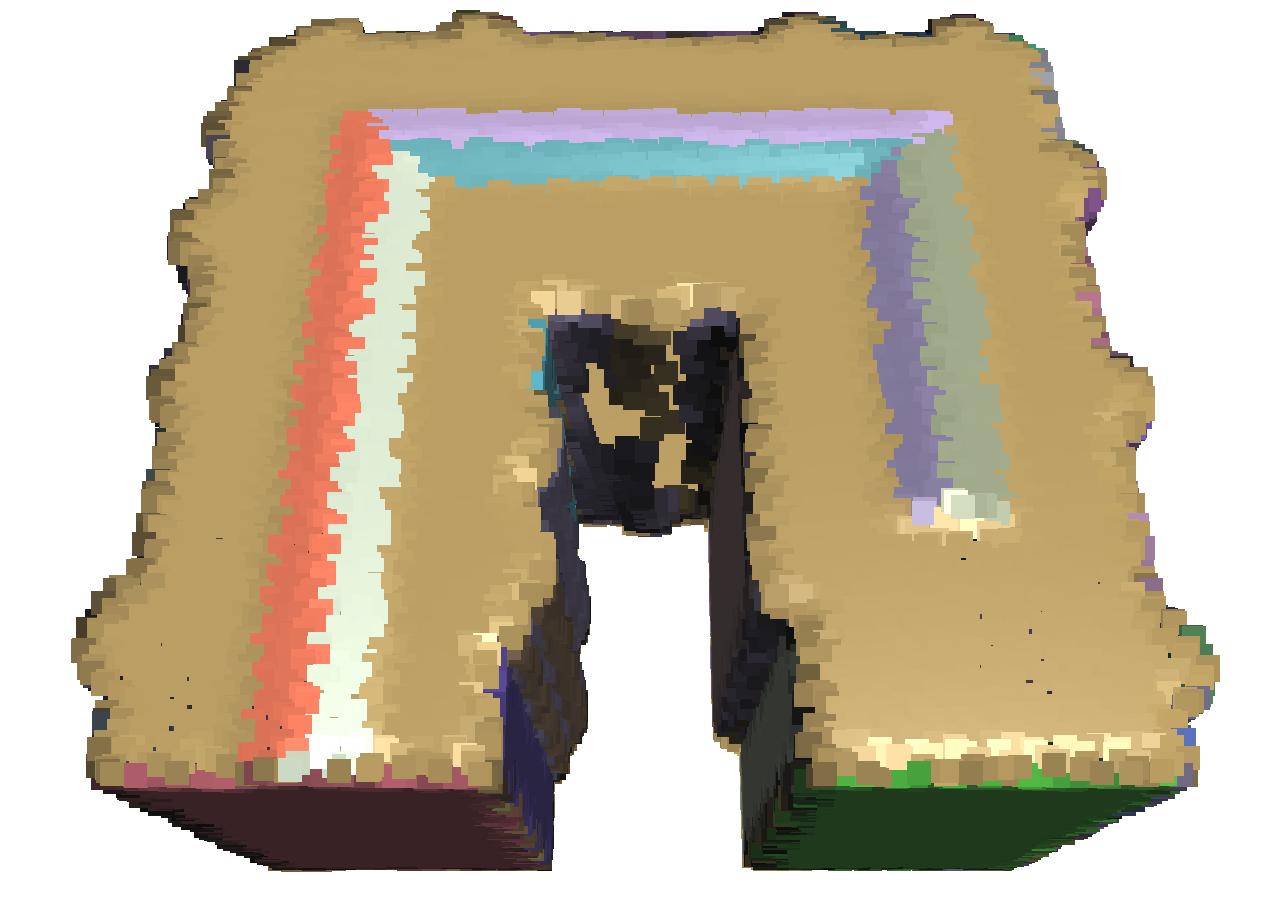}}
	\hspace{1em}
	\subfloat{\includegraphics[width=0.15\linewidth]{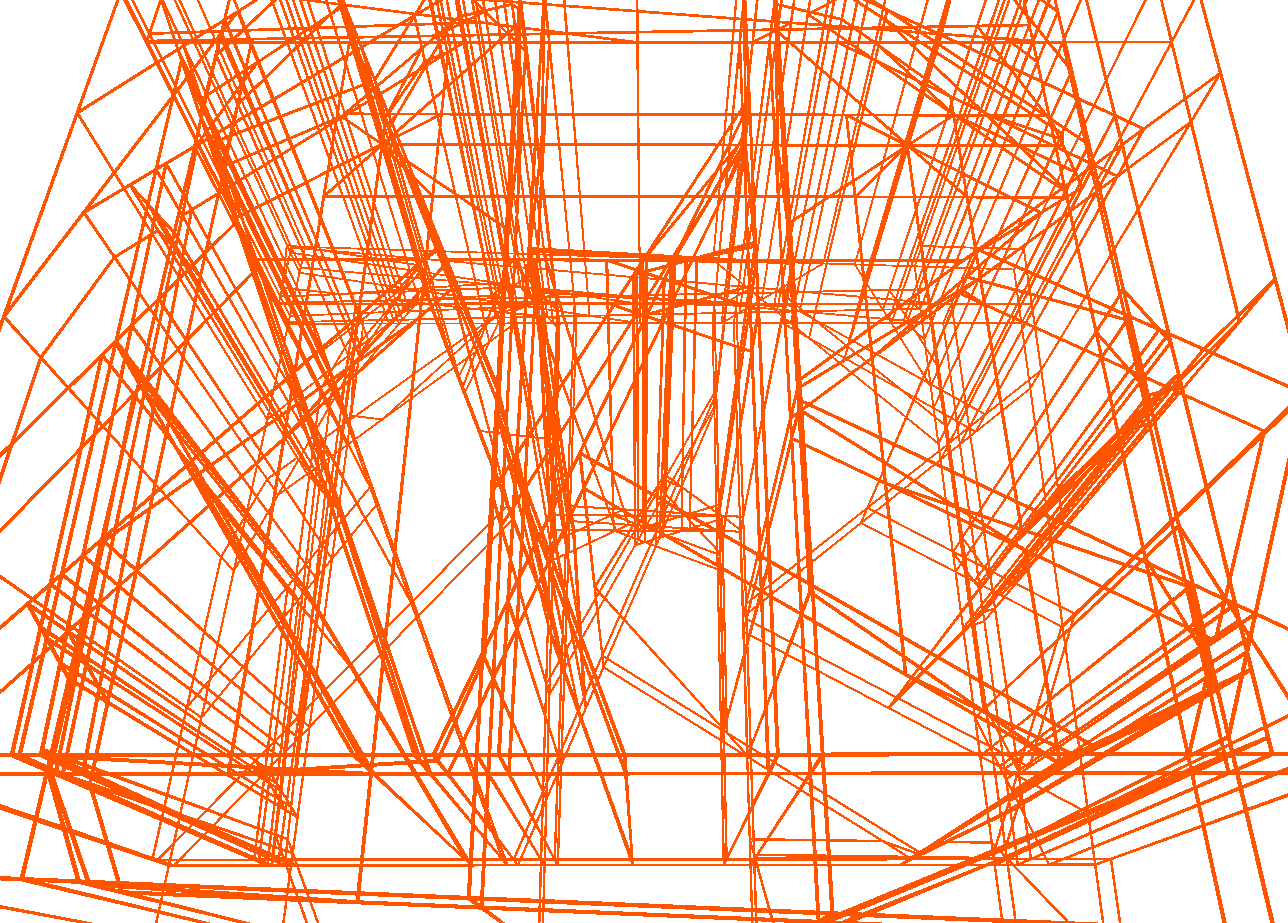}}
	\hspace{1em}
	\subfloat{\includegraphics[width=0.15\linewidth]{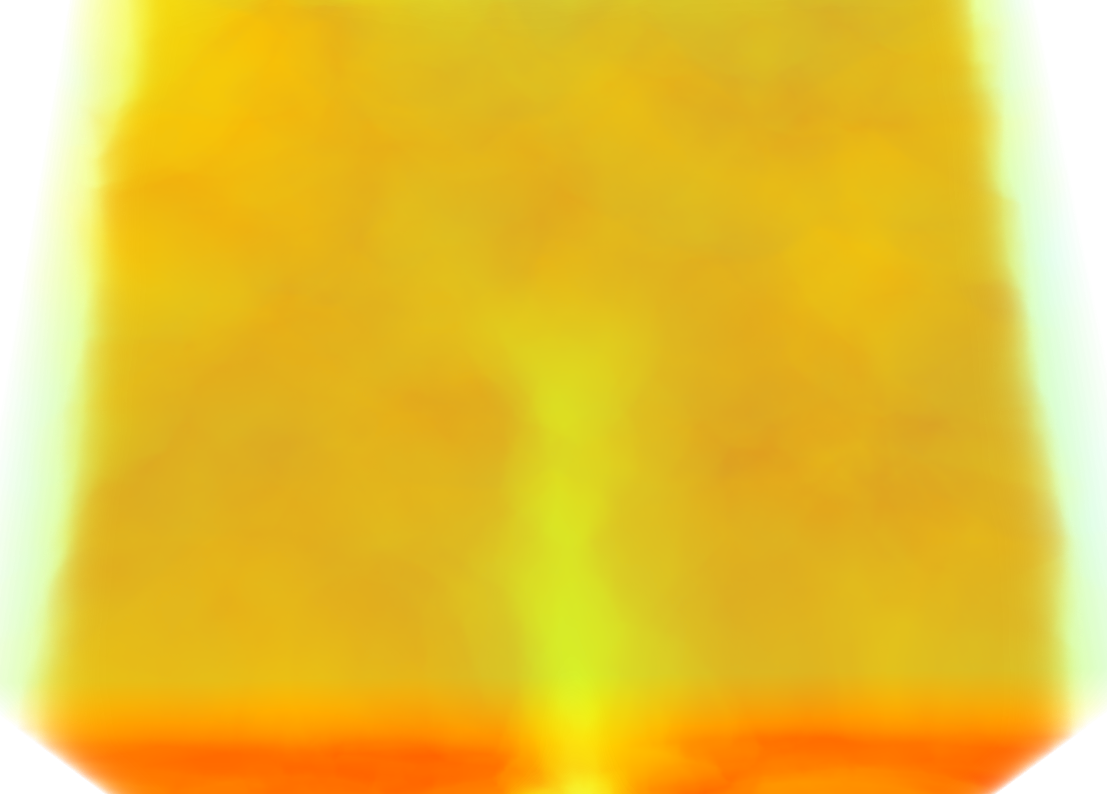}}
	\hspace{1em}
	\subfloat{\includegraphics[width=0.15\linewidth]{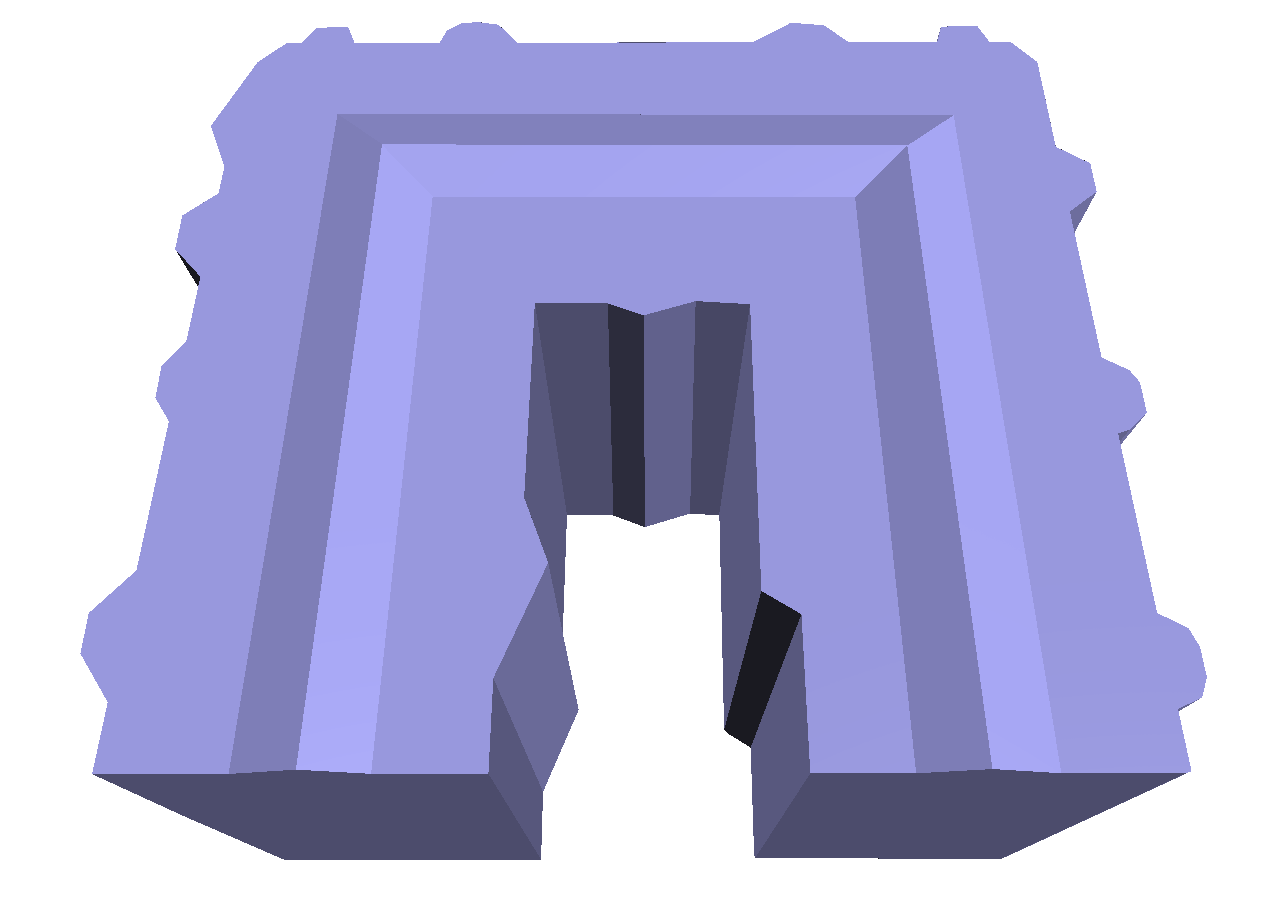}}   
	\hspace{1em}
	\subfloat{\includegraphics[width=0.15\linewidth]{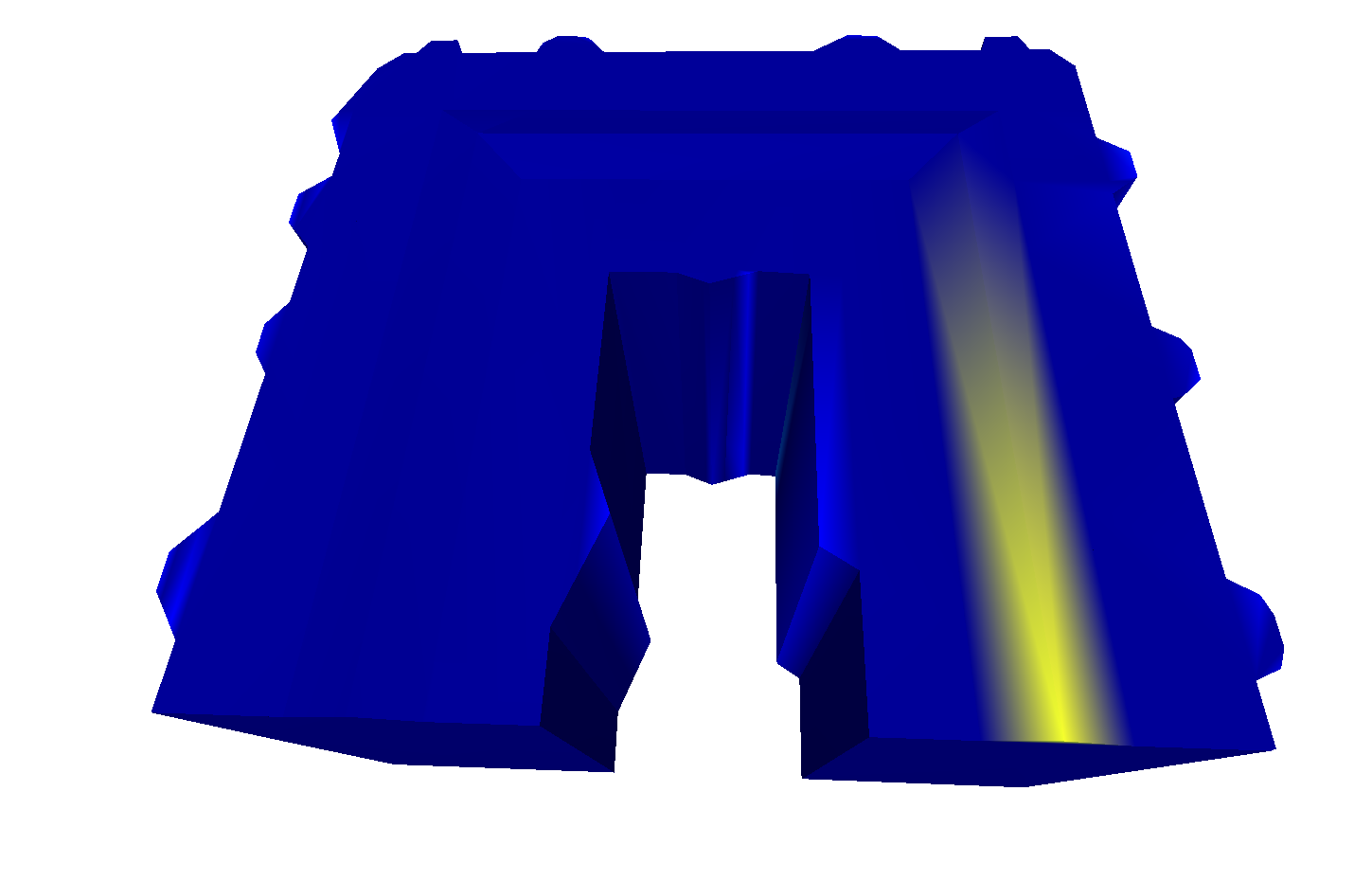}} 
	\vspace{0em}

	(h)
	\subfloat{\includegraphics[width=0.15\linewidth]{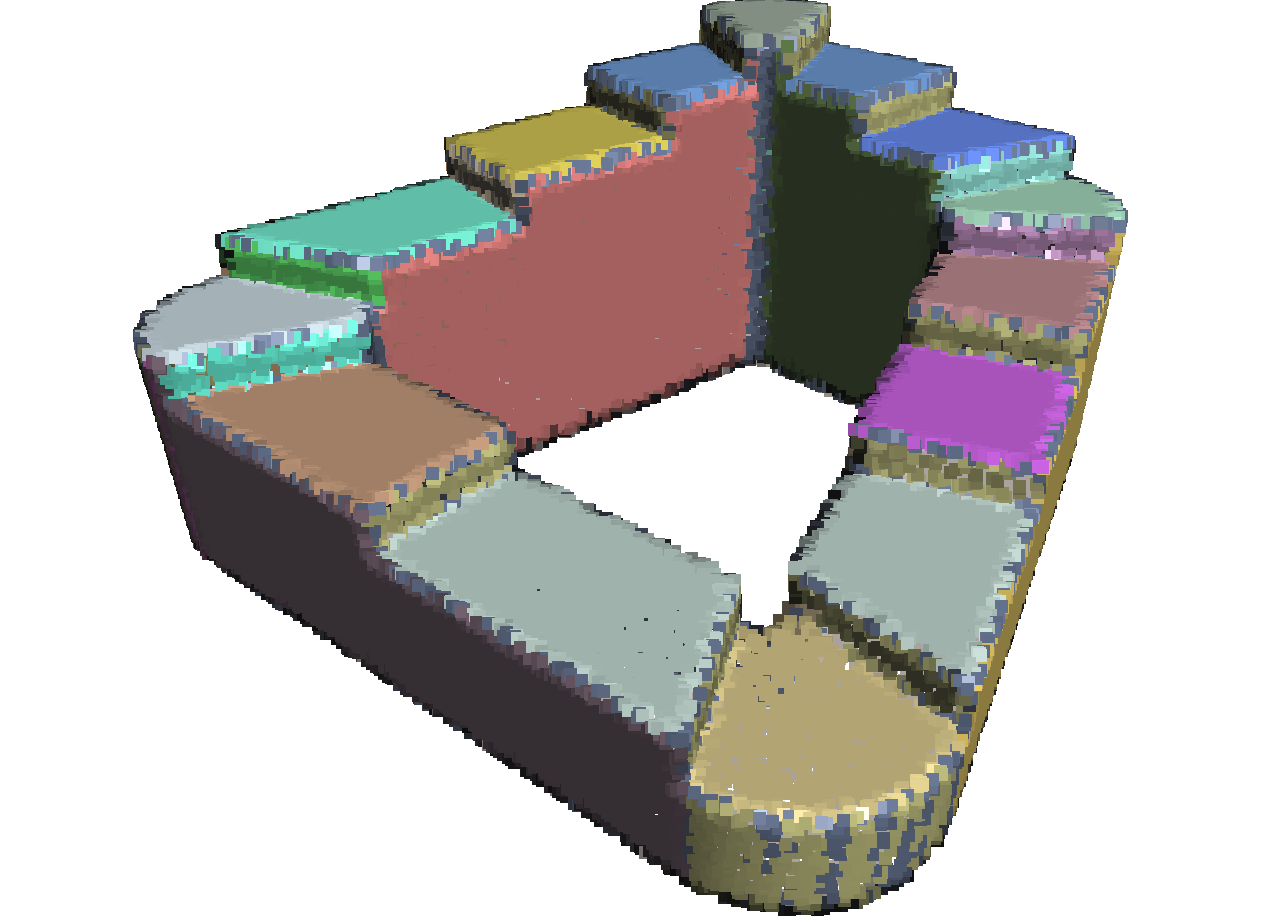}}
	\hspace{1em}
	\subfloat{\includegraphics[width=0.15\linewidth]{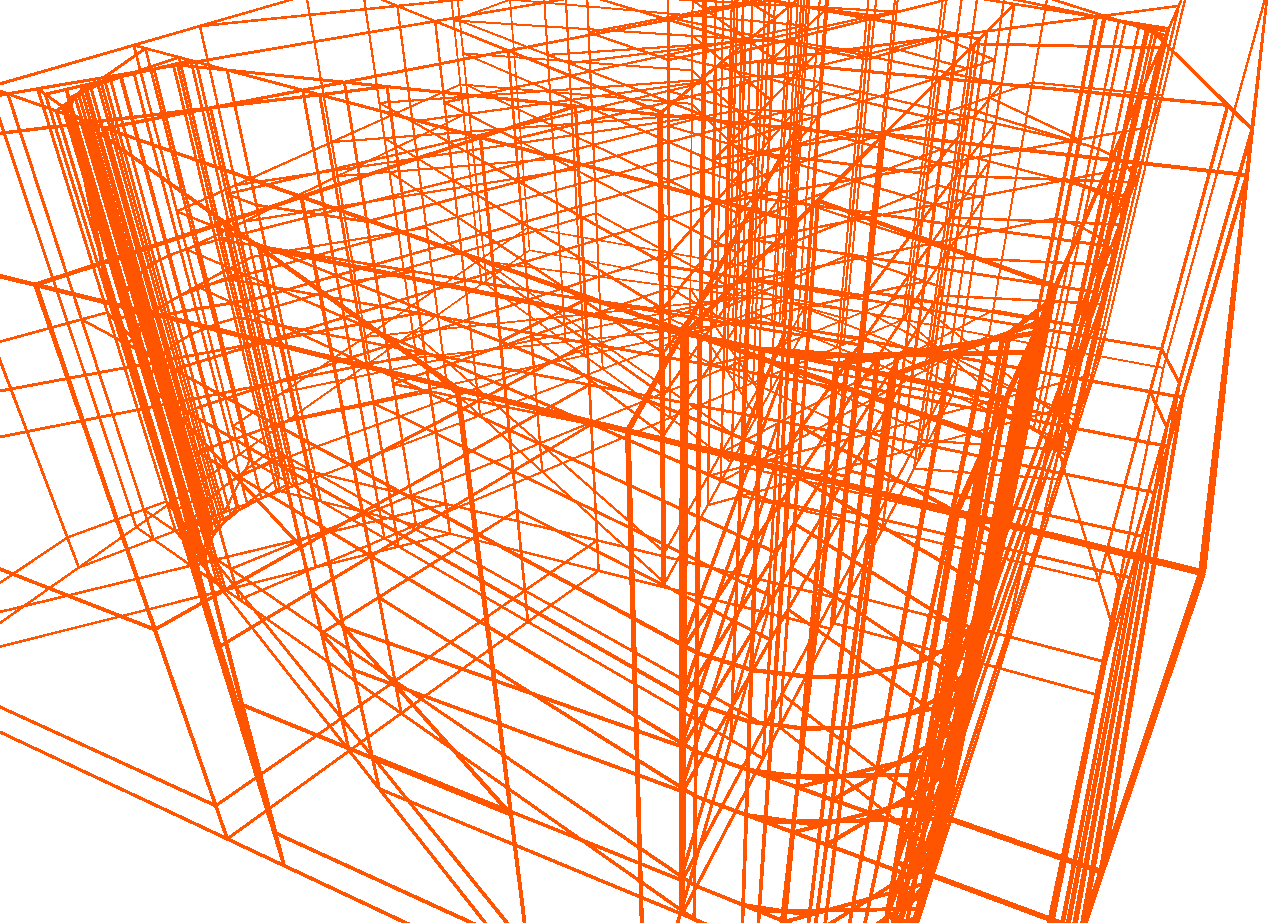}}
	\hspace{1em}
	\subfloat{\includegraphics[width=0.15\linewidth]{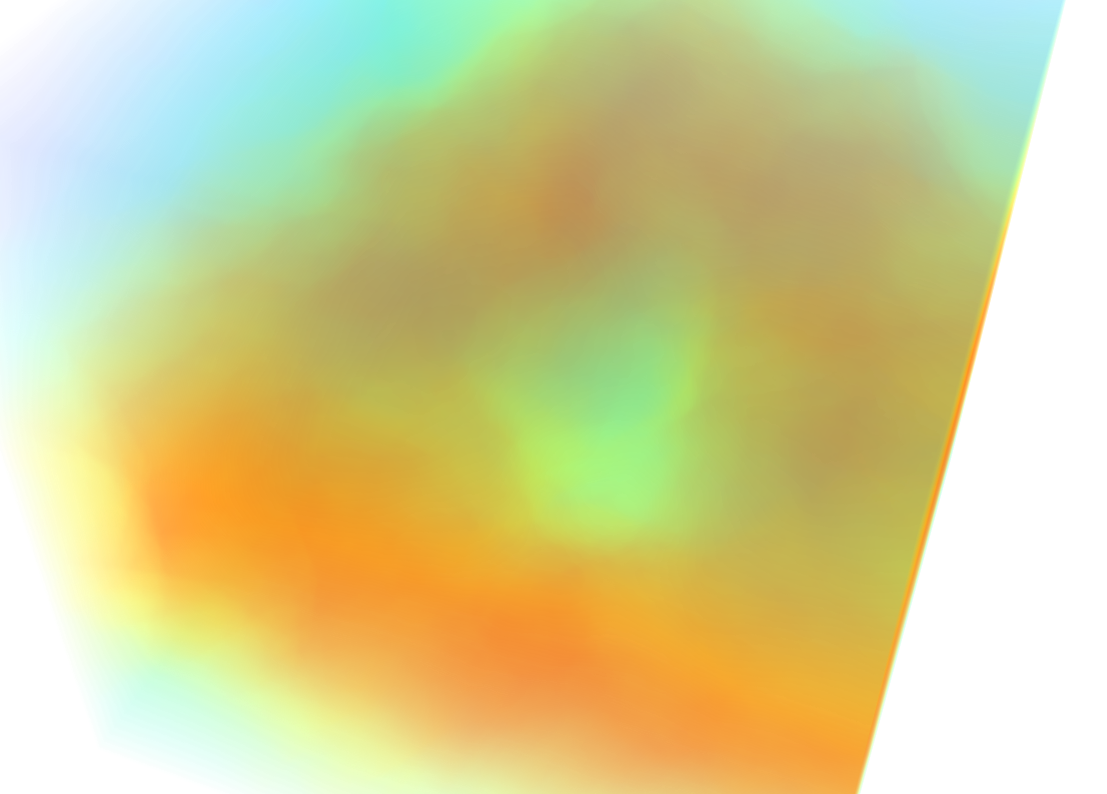}}
	\hspace{-0.5em}
	\subfloat{\includegraphics[height=0.1\linewidth]{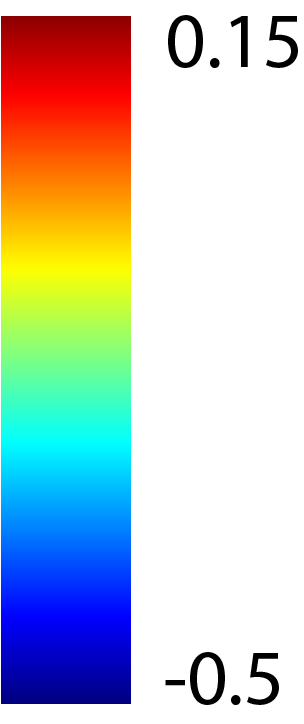}} 
	\subfloat{\includegraphics[width=0.15\linewidth]{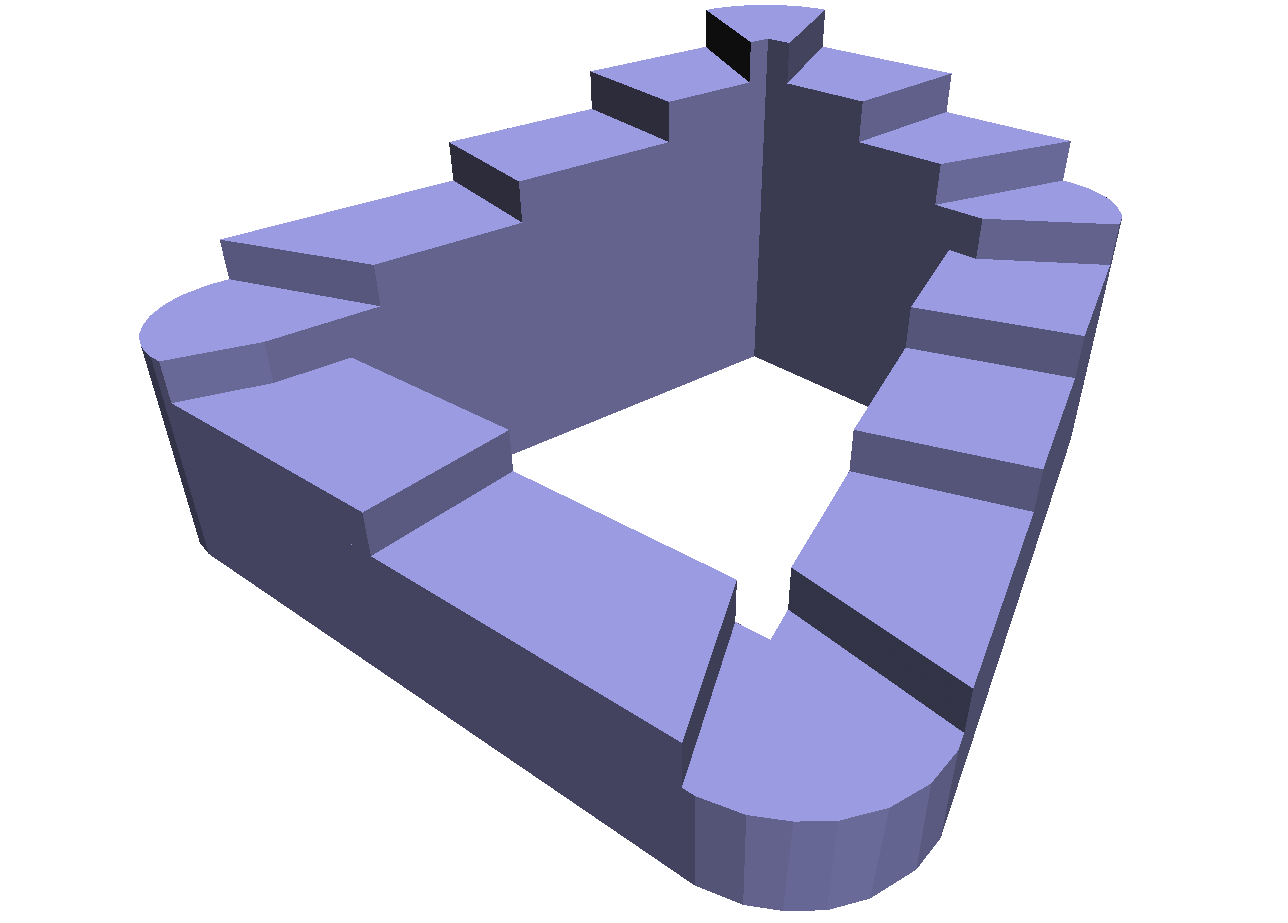}}  
	\hspace{0.em}
	\subfloat{\includegraphics[width=0.15\linewidth]{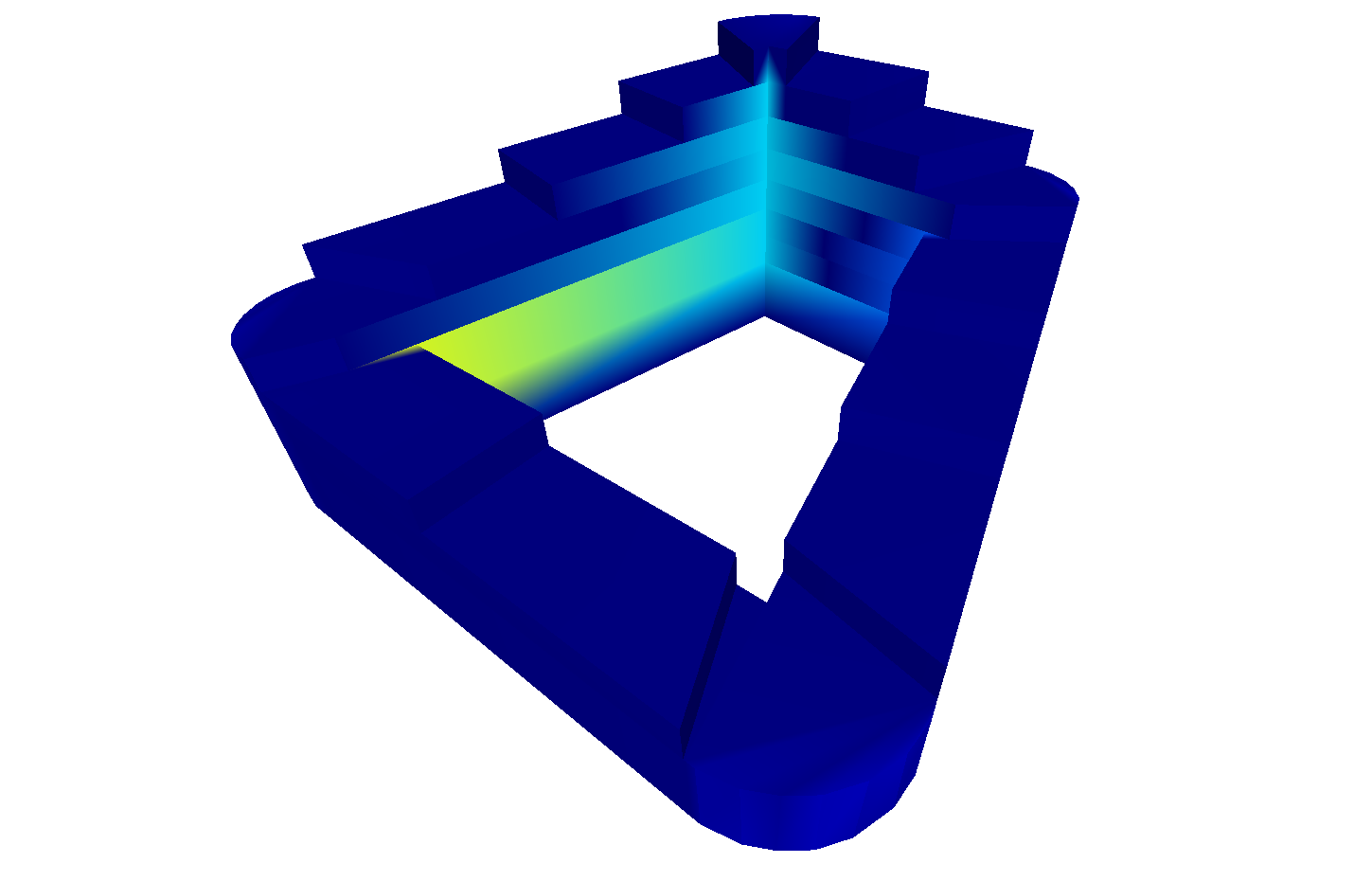}}
	\hspace{-1.1em}   
	\subfloat{\includegraphics[height=0.1\linewidth]{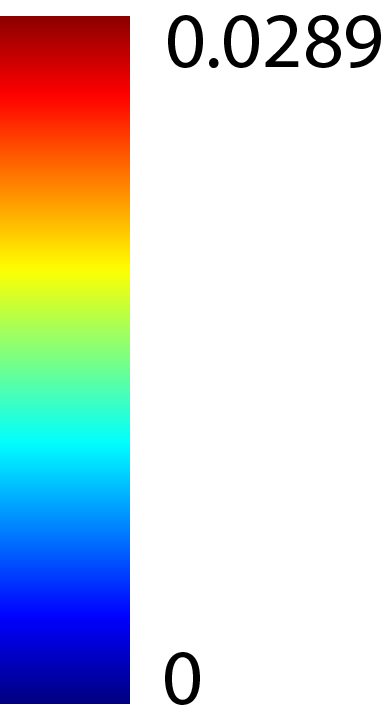}} 	
	
	\caption{Reconstruction results on the \textit{Helsinki full-view} point clouds. From left to right: input point cloud (colored randomly per planar primitive), wireframe of the cell complex, volume rendering of the SDF, reconstructed building model, and error map.
	}
	\label{fig:results_fullview}
\end{figure*}

With the neural network trained on the \textit{Helsinki full-view} point clouds, the proposed method can reconstruct buildings of various architectural styles.
\autoref{fig:results_fullview} presents the reconstruction results of eight buildings from the \textit{Helsinki full-view} set.
The SDF can accurately describe the occupancy of the buildings even though their styles may be distinct from the data on which the neural network was trained.
The errors occur only when subtle structures exist, which is due to the uncertainty in planar primitive detection and the regularization imposed to surface extraction.
Nonetheless, all building surfaces can be effectively retrieved from the point clouds with plausible visual quality and symmetric mean Hausdorff (SMH) distance of less than 0.3\%.

\begin{figure}[ht]
	\centering
%	\includegraphics[width=0.21\linewidth]{figures/results/robustness/helsinki_2_points.png}
%	\hspace{1em}
%	\includegraphics[width=0.21\linewidth]{figures/results/robustness/helsinki_1_points.png}
%	\hspace{1em}
%	\includegraphics[width=0.21\linewidth]{figures/results/robustness/shenzhen_2_points.png}	
%	\hspace{1em}
%	\includegraphics[width=0.21\linewidth]{figures/results/robustness/shenzhen_1_points.png}	

	\includegraphics[width=0.21\linewidth]{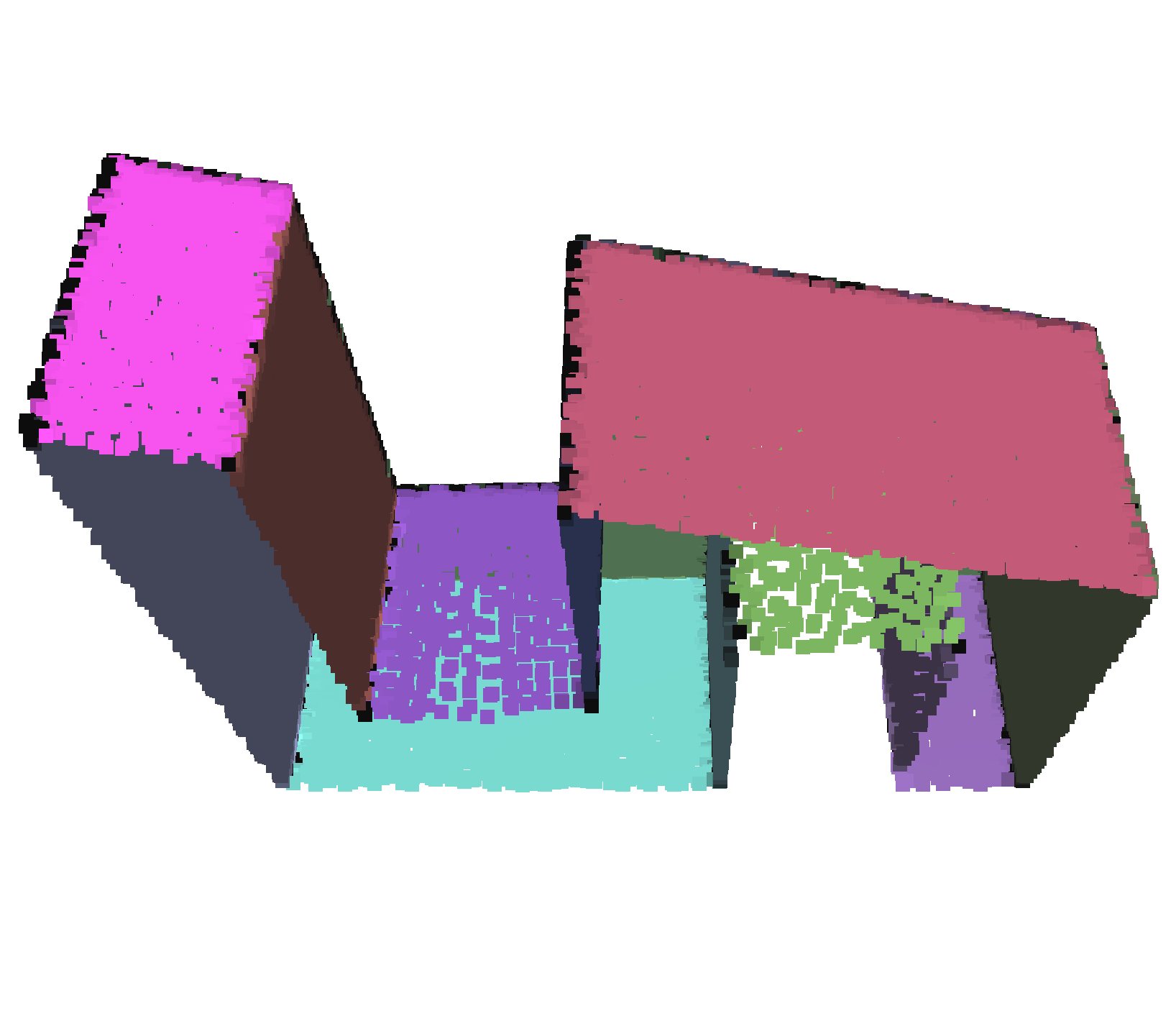}
	\hspace{1em}
	\includegraphics[width=0.21\linewidth]{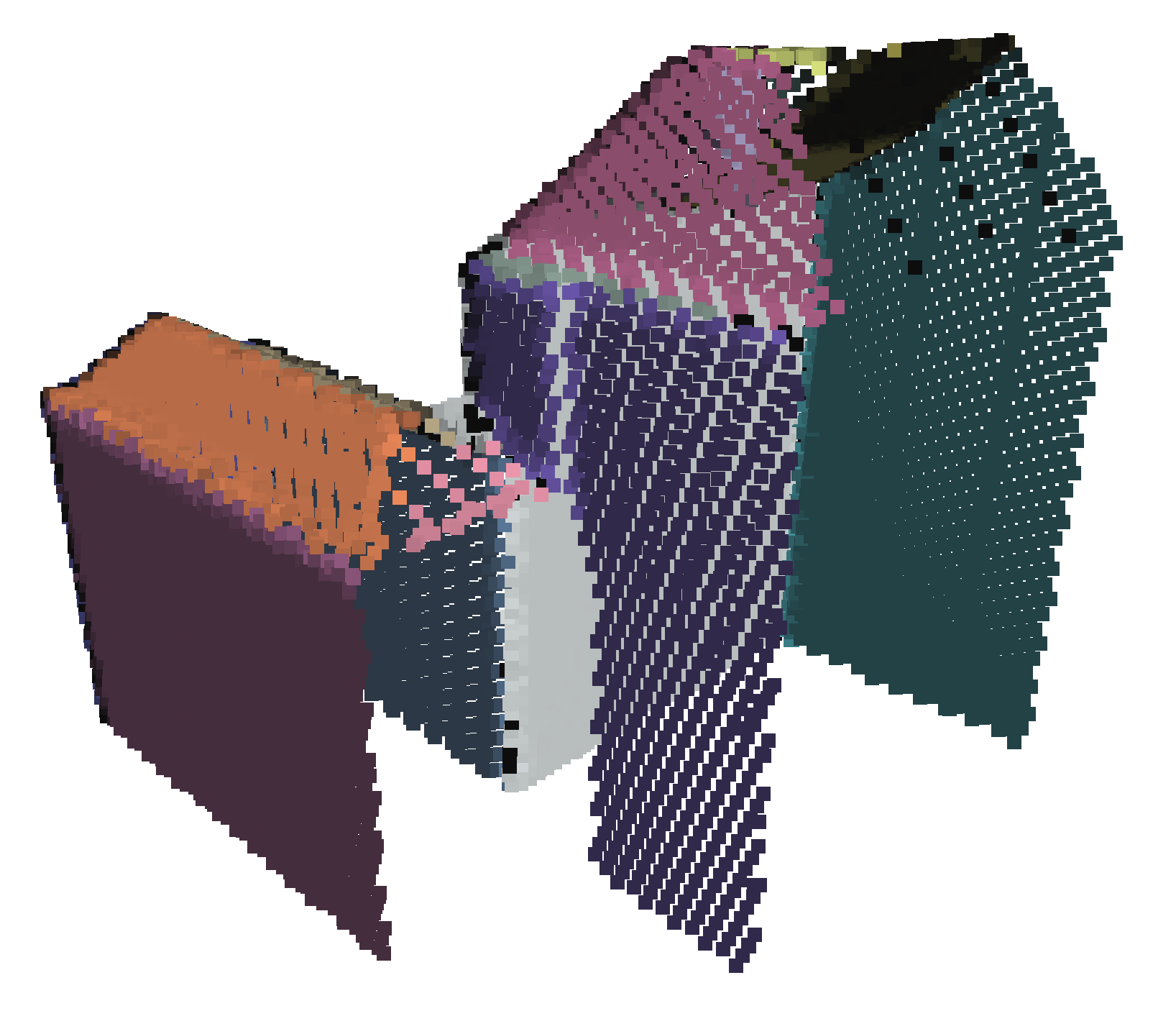}
	\hspace{1em}
	\includegraphics[width=0.21\linewidth]{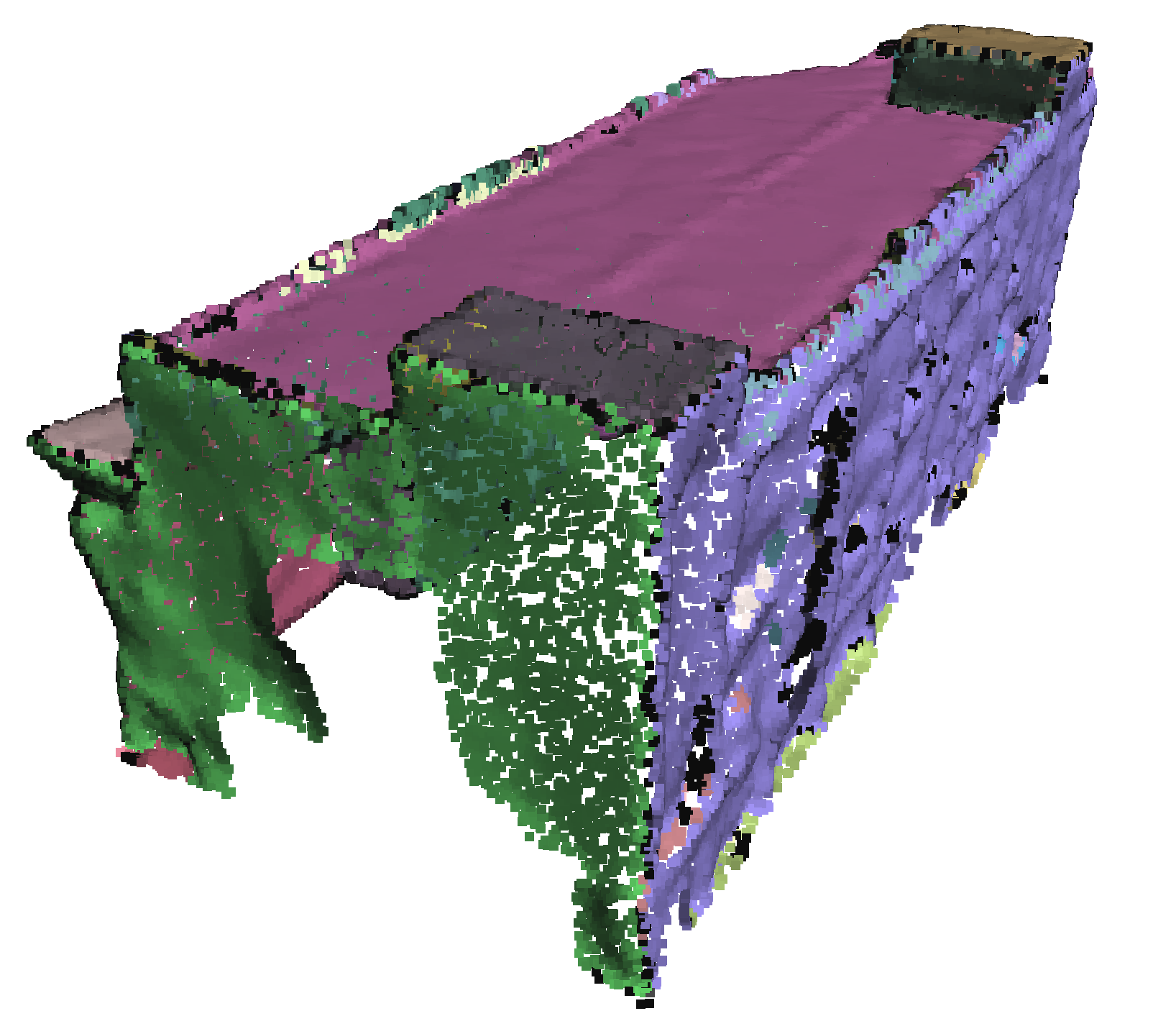}	
	\hspace{1em}
	\includegraphics[width=0.21\linewidth]{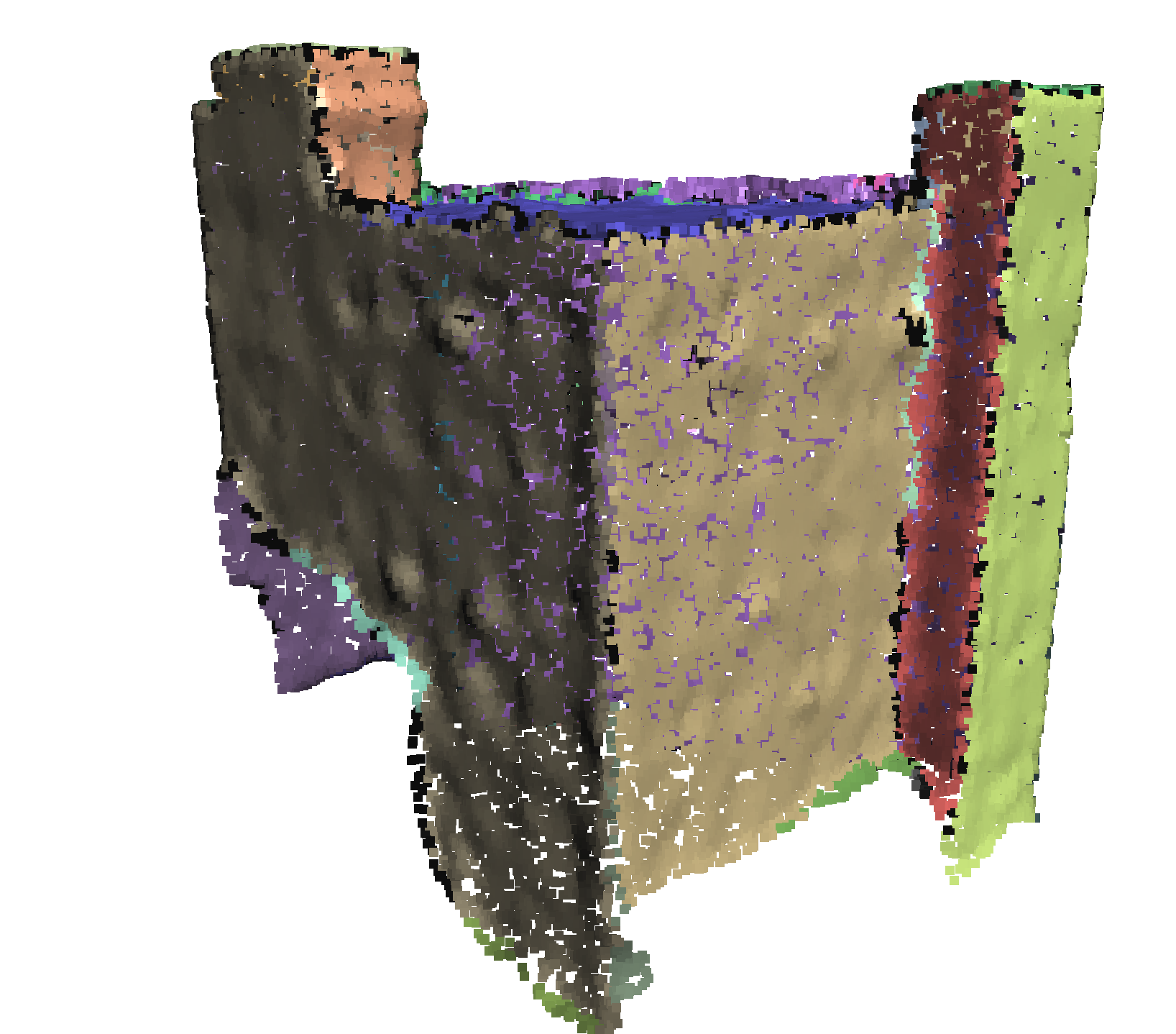}	
	
	\includegraphics[width=0.21\linewidth]{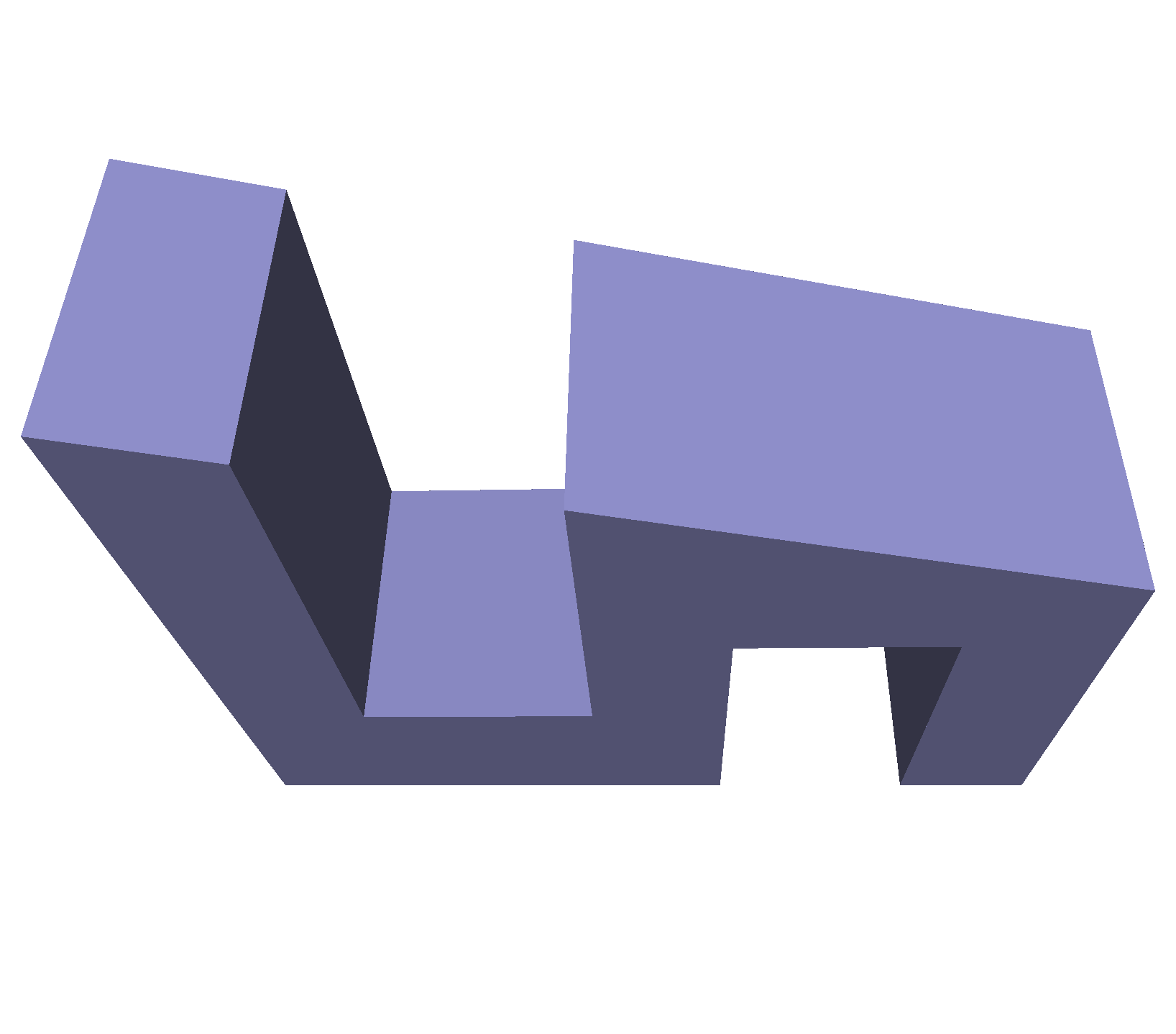}
	\hspace{1em}
	\includegraphics[width=0.21\linewidth]{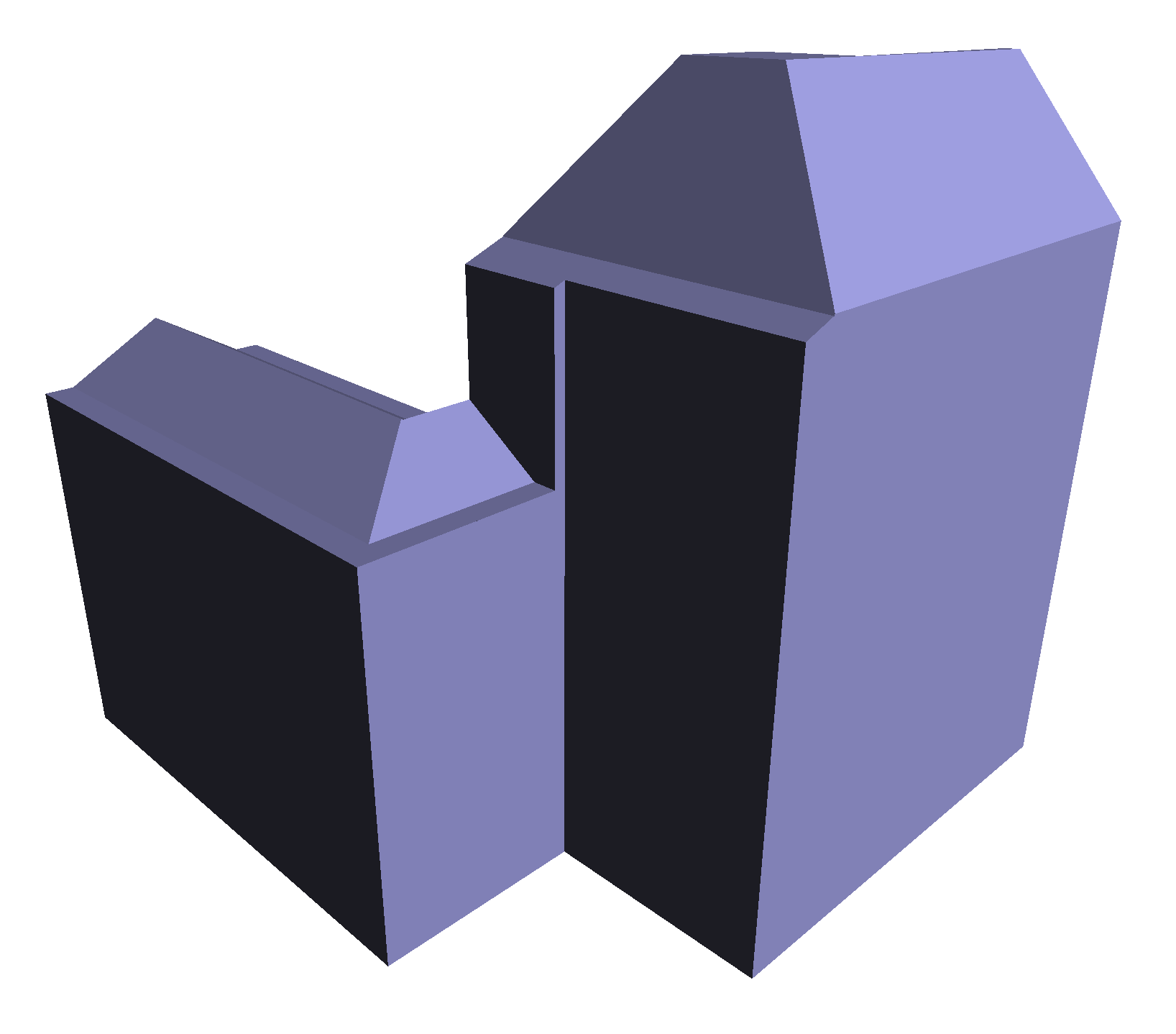}
	\hspace{1em}
	\includegraphics[width=0.21\linewidth]{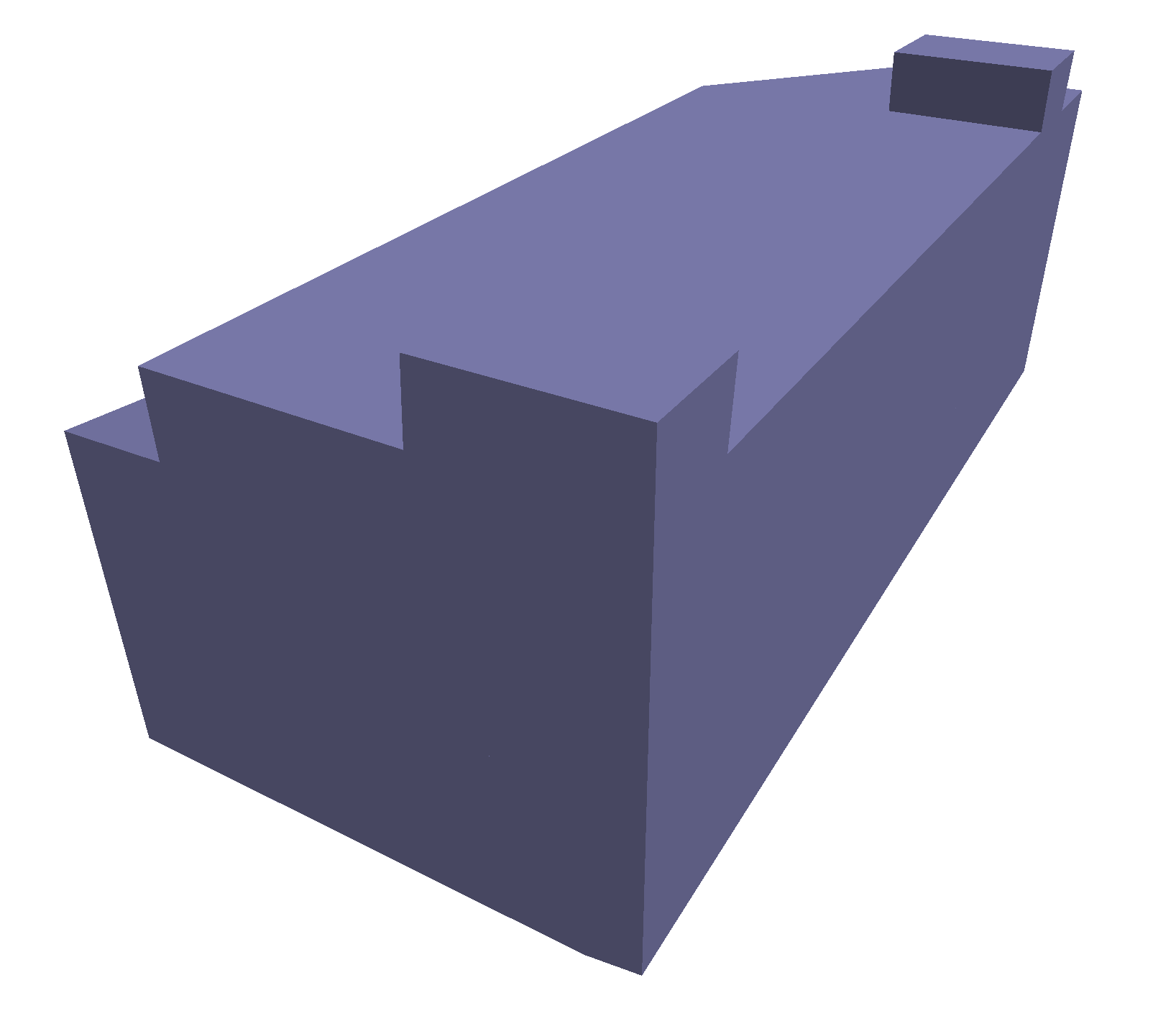}	
	\hspace{1em}
	\includegraphics[width=0.21\linewidth]{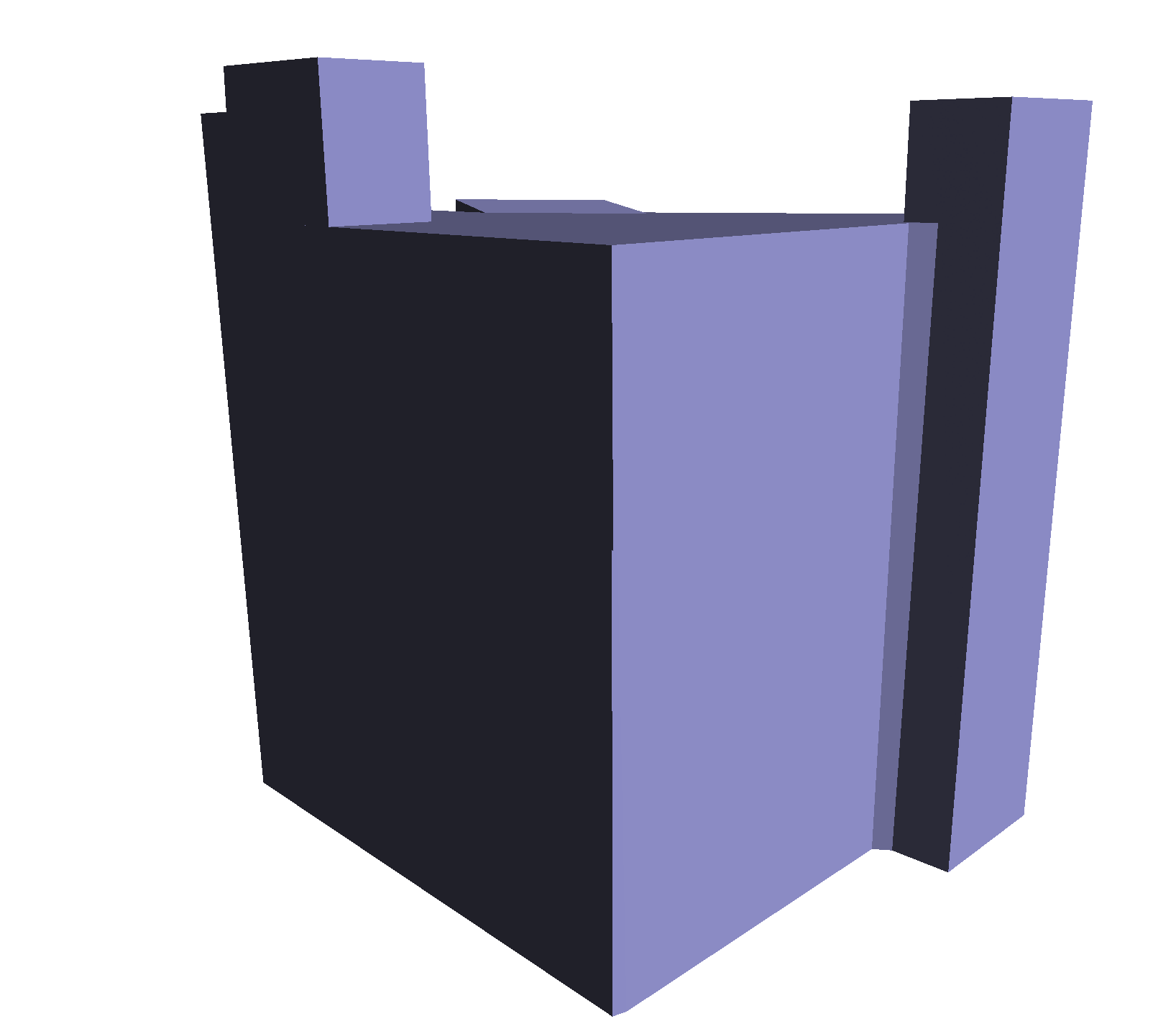}	

	\includegraphics[width=0.21\linewidth]{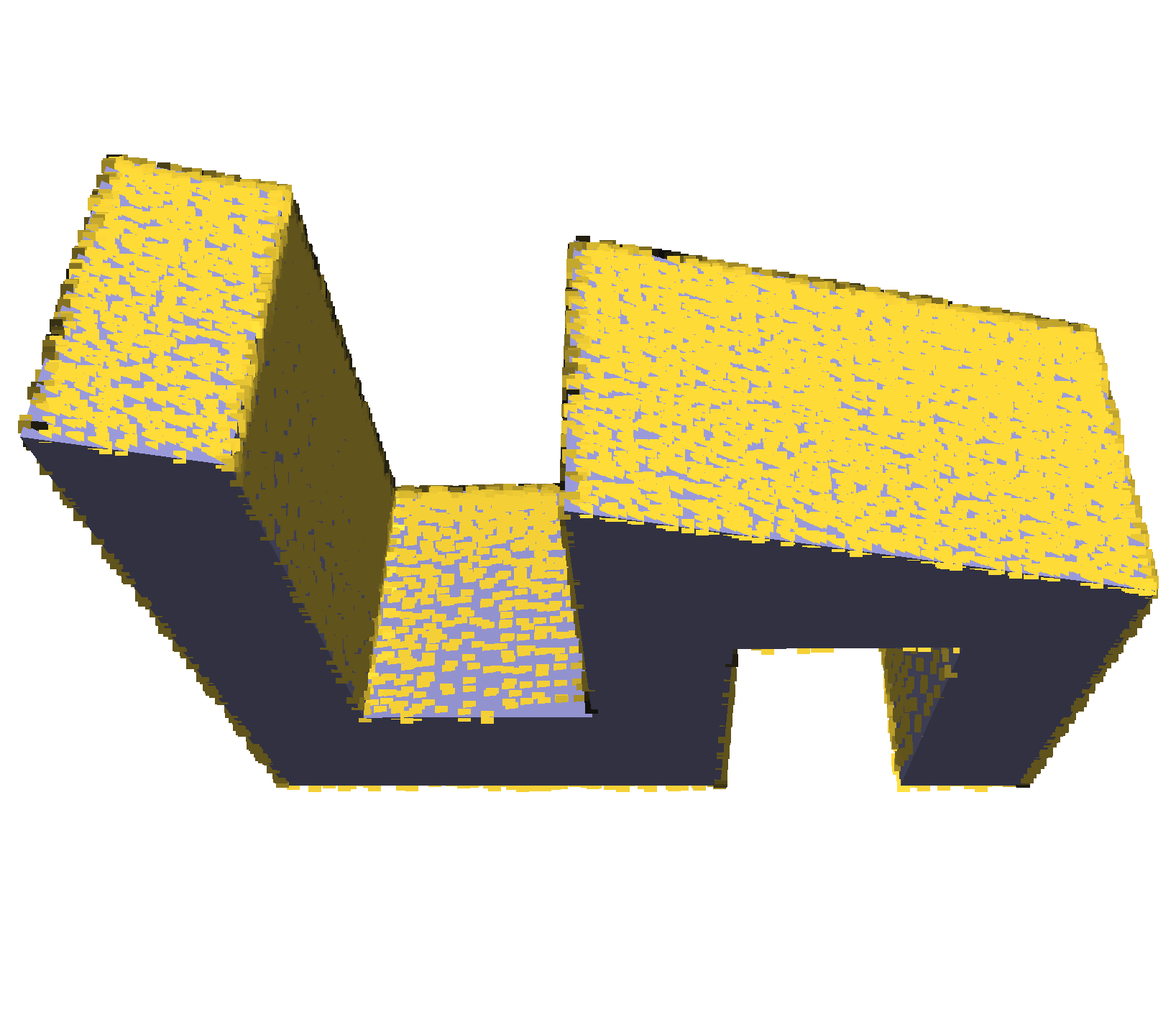}
	\hspace{1em}
	\includegraphics[width=0.21\linewidth]{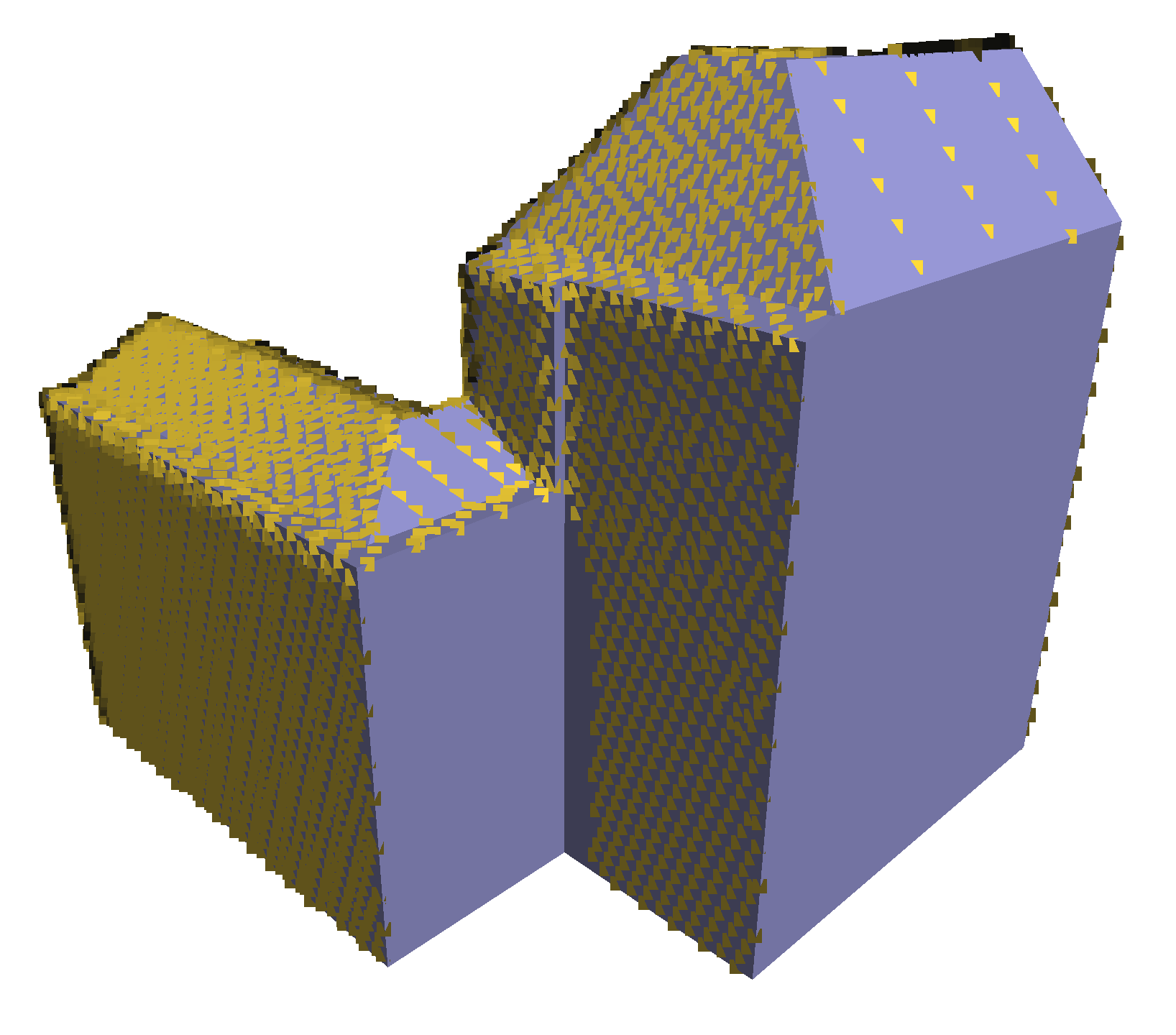}
	\hspace{1em}
	\includegraphics[width=0.21\linewidth]{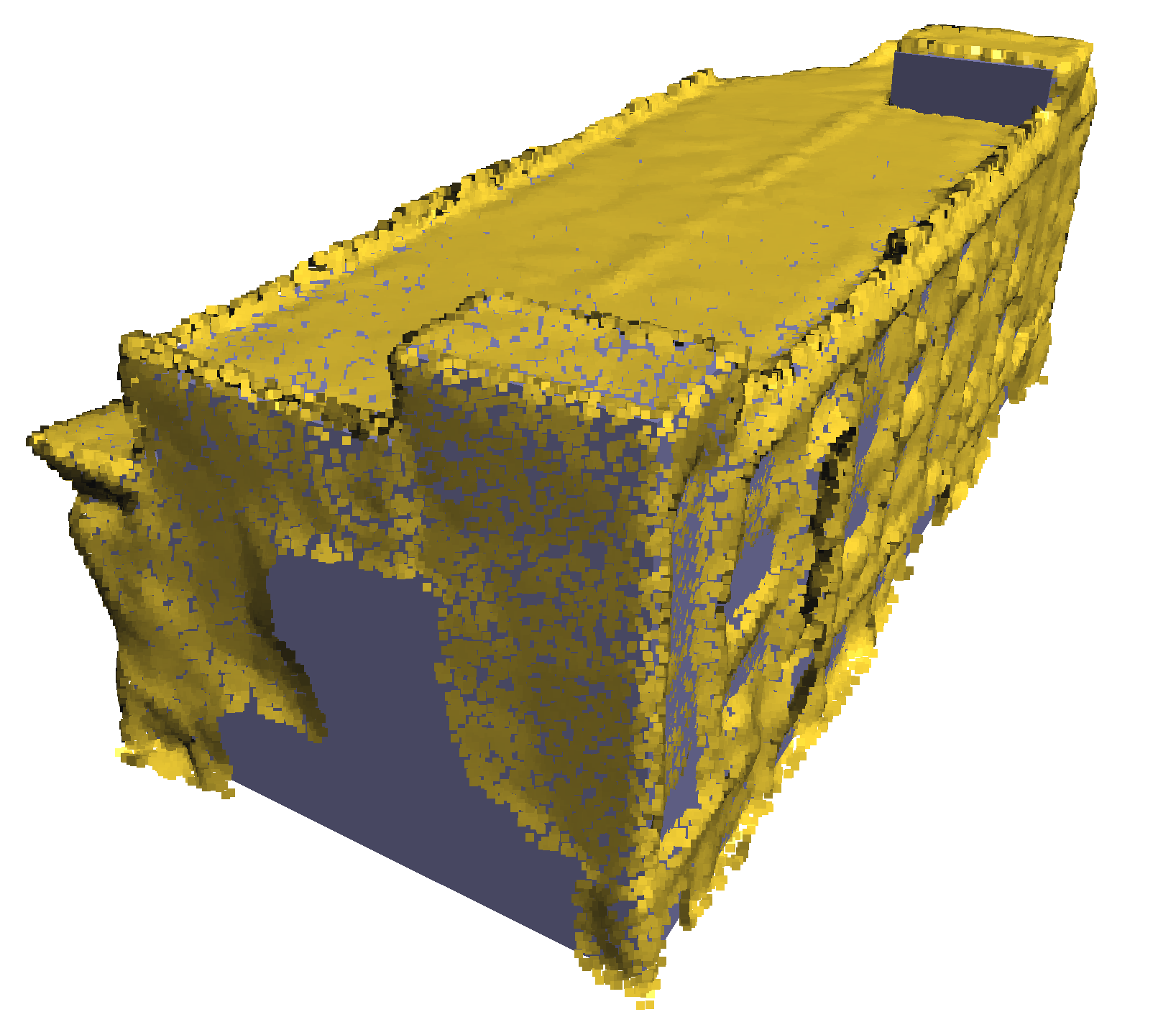}	
	\hspace{1em}
	\includegraphics[width=0.21\linewidth]{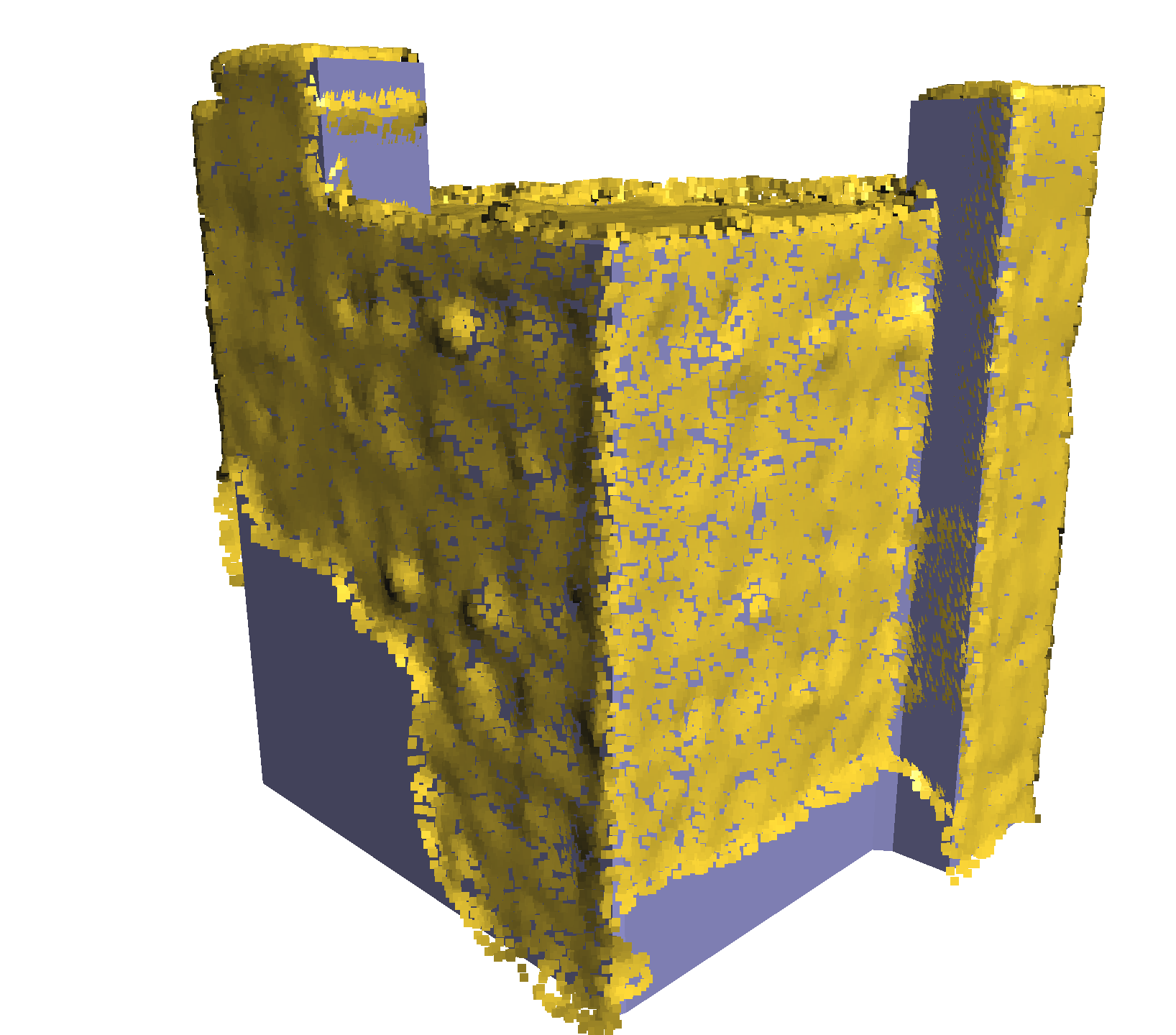}	
	
	\caption{Surface reconstruction from incomplete point clouds. Our method can reconstruct complete building models even if the input point clouds have missing regions on the bottom or facades. From top to bottom: input point clouds colored with the extracted planar primitives, reconstructed models, and an overlay of the two. The left two buildings are from \textit{Helinki no-bottom}, and the right two are from \textit{Shenzhen}.}
	\label{fig:results_nobottom}
\end{figure}

We additionally train the neural network and evaluate it on the \textit{Helsinki no-bottom} point clouds.
As shown in \autoref{fig:results_nobottom}, our method can still correctly infer the occupancy of the entire buildings regardless of the missing ground planes, resulting in complete reconstruction results. 
Unlike PolyFit~\citep{nan2017polyfit} that relies on a complete set of planes, our method can make use of the additional faces of the AABB of the point clouds, which guarantees to complete the missing surface models\footnote{This is specially designed for point clouds where no ground is captured.}.
Besides, with the learned SDF, facades with few supporting points can be reliably reconstructed. 

\begin{figure*}[th]
	\centering

	(a)
	\subfloat{\includegraphics[width=0.15\linewidth]{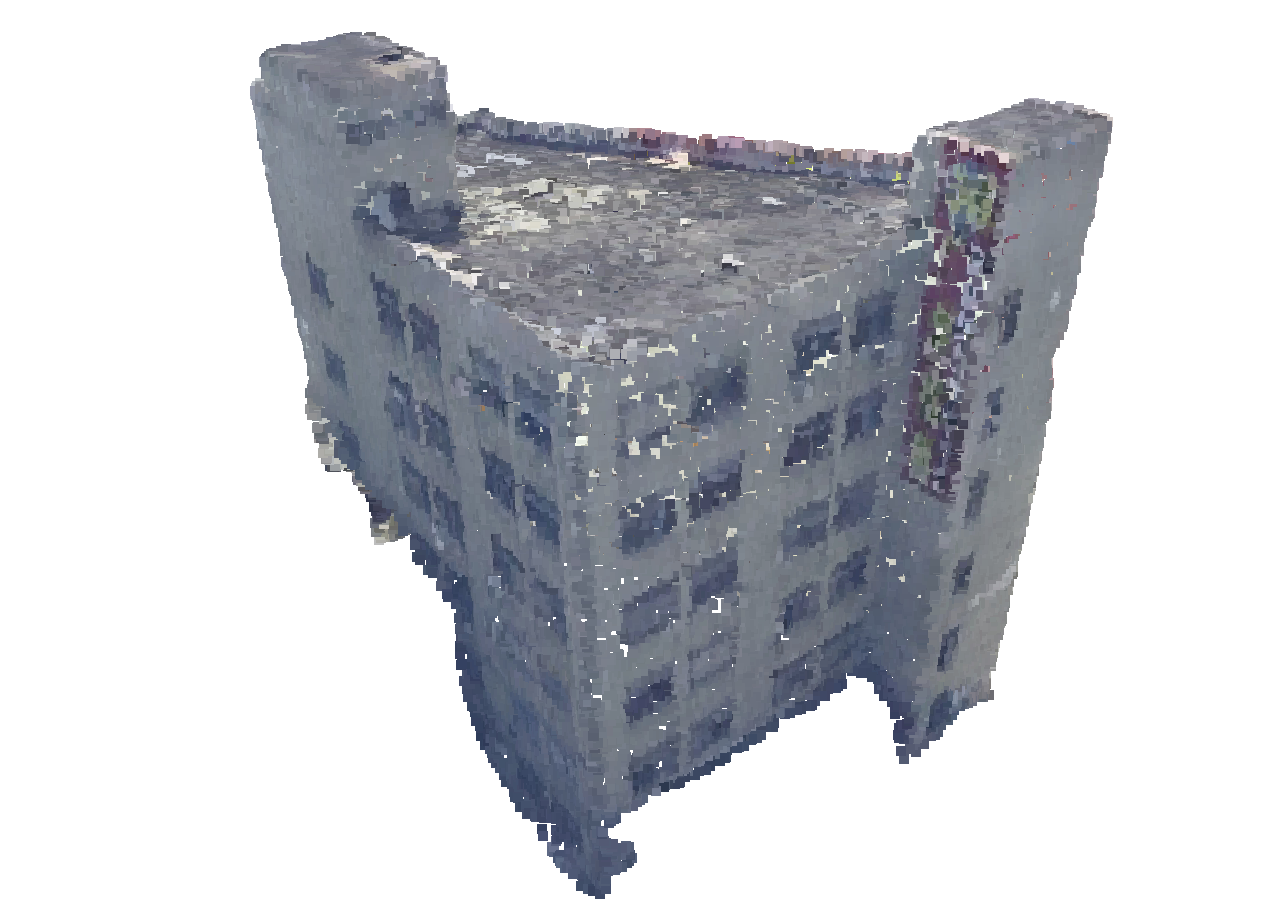}}
	\hspace{1em}
	\subfloat{\includegraphics[width=0.15\linewidth]{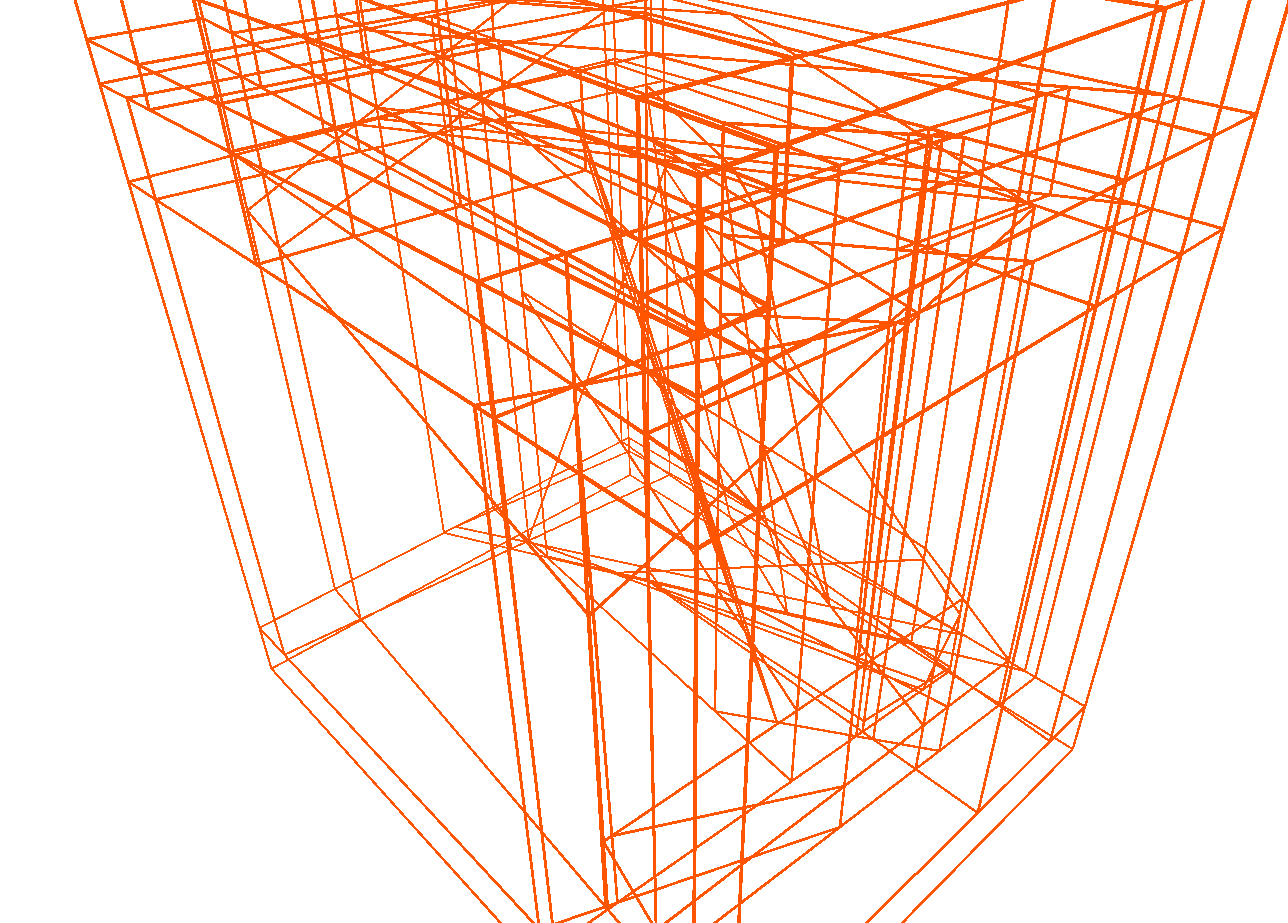}}
	\hspace{1em}
	\subfloat{\includegraphics[width=0.15\linewidth]{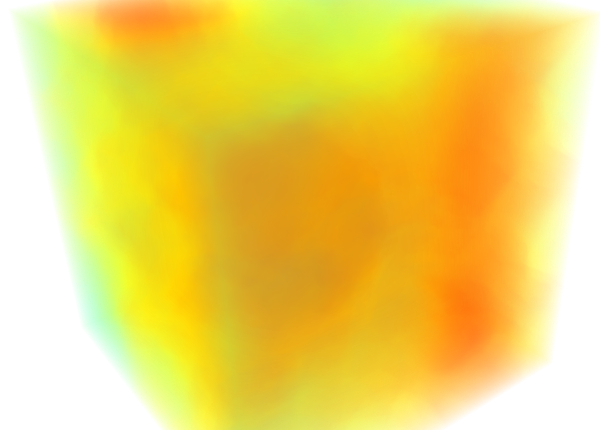}}
	\hspace{1em}
	\subfloat{\includegraphics[width=0.15\linewidth]{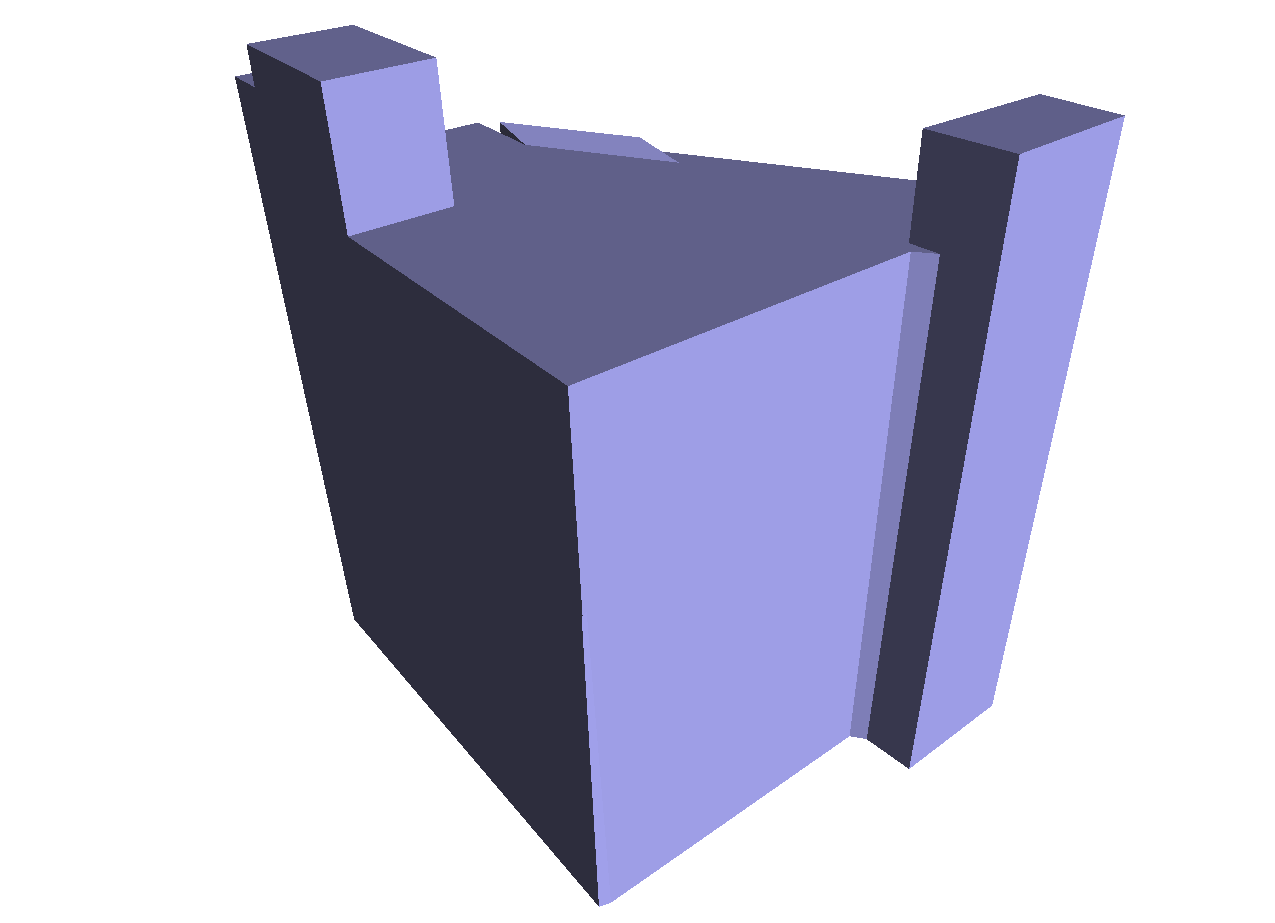}}
	\hspace{1em}
	\subfloat{\includegraphics[width=0.15\linewidth]{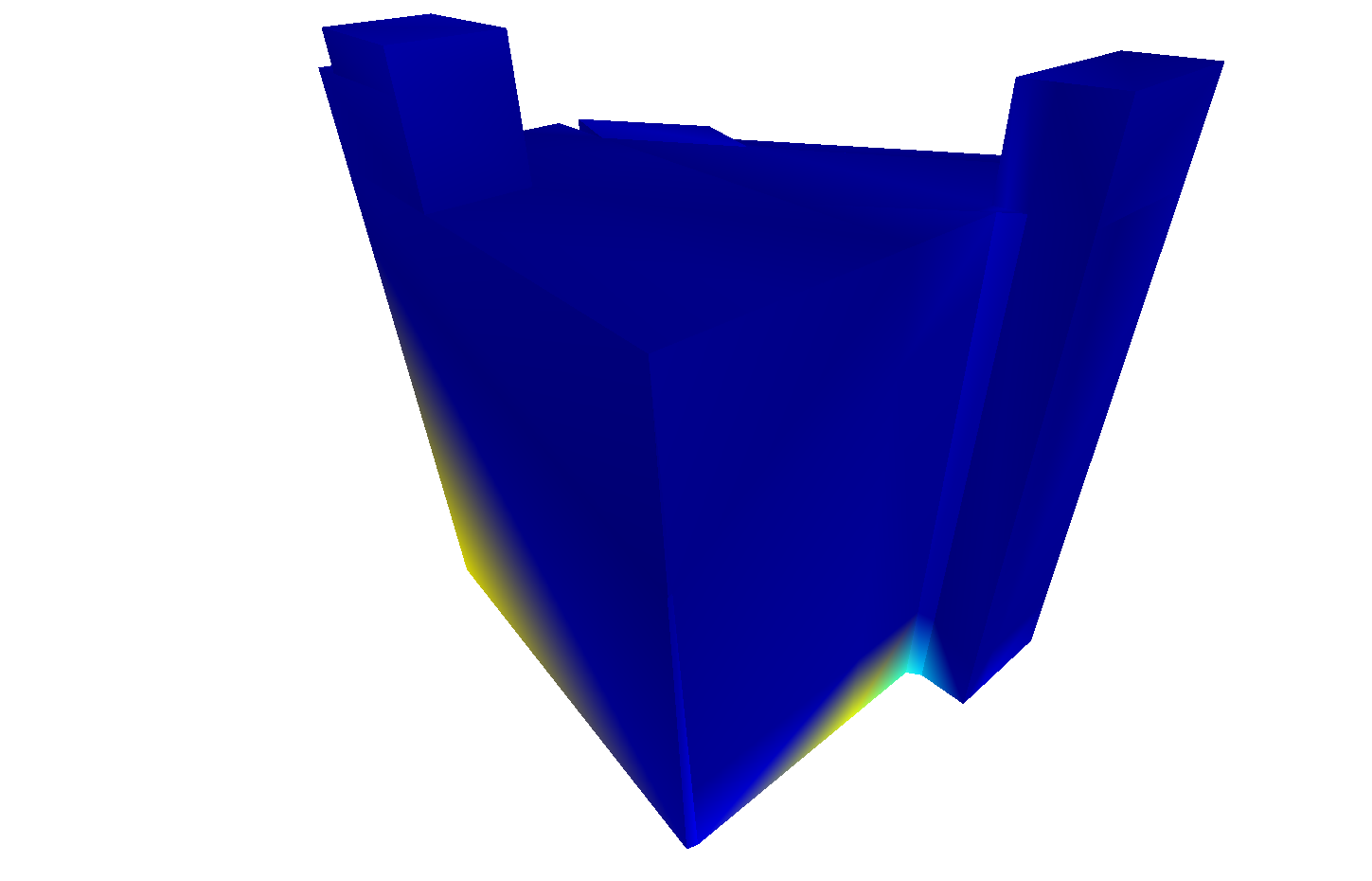}}

	(b)	
	\subfloat{\includegraphics[width=0.15\linewidth]{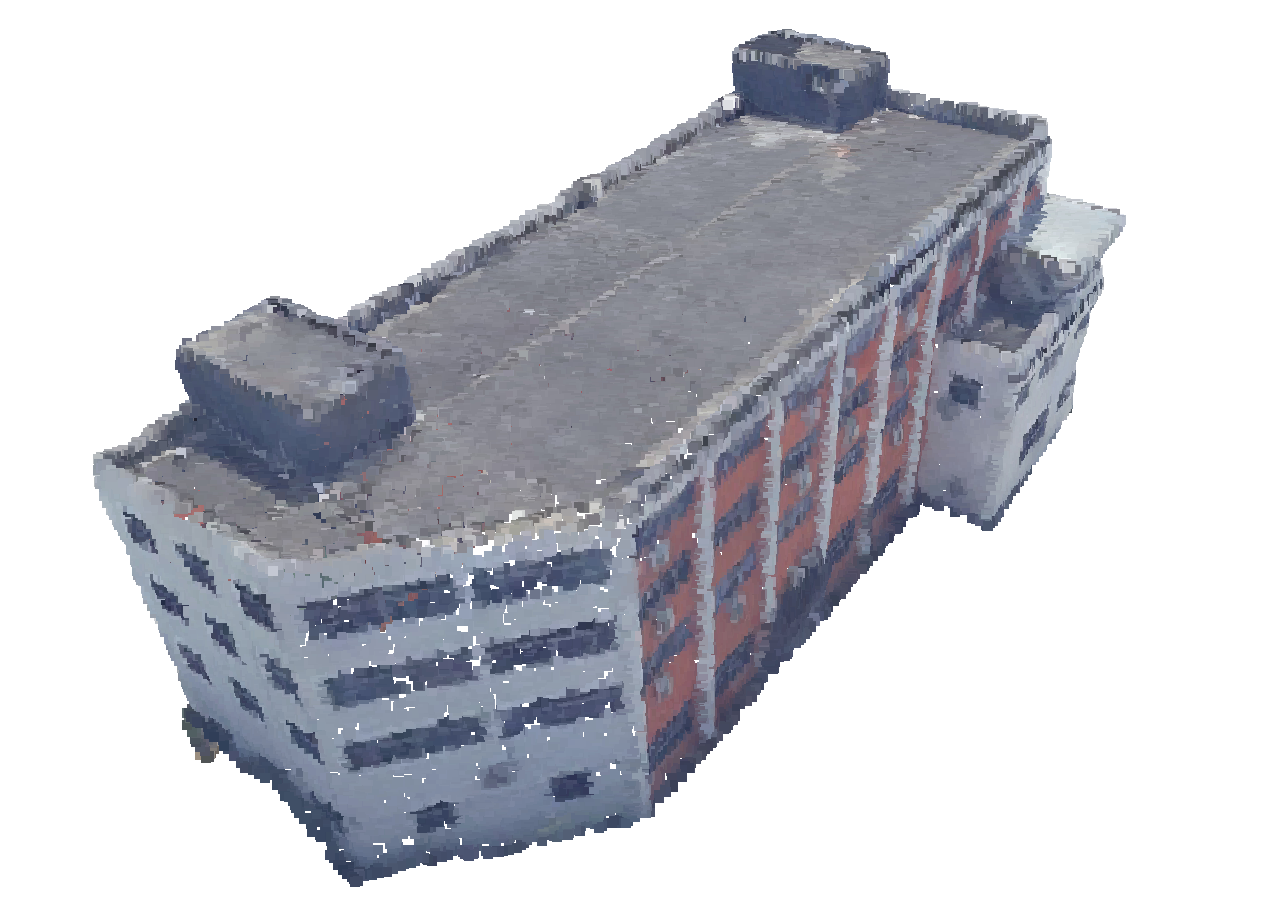}}
	\hspace{1em}
	\subfloat{\includegraphics[width=0.15\linewidth]{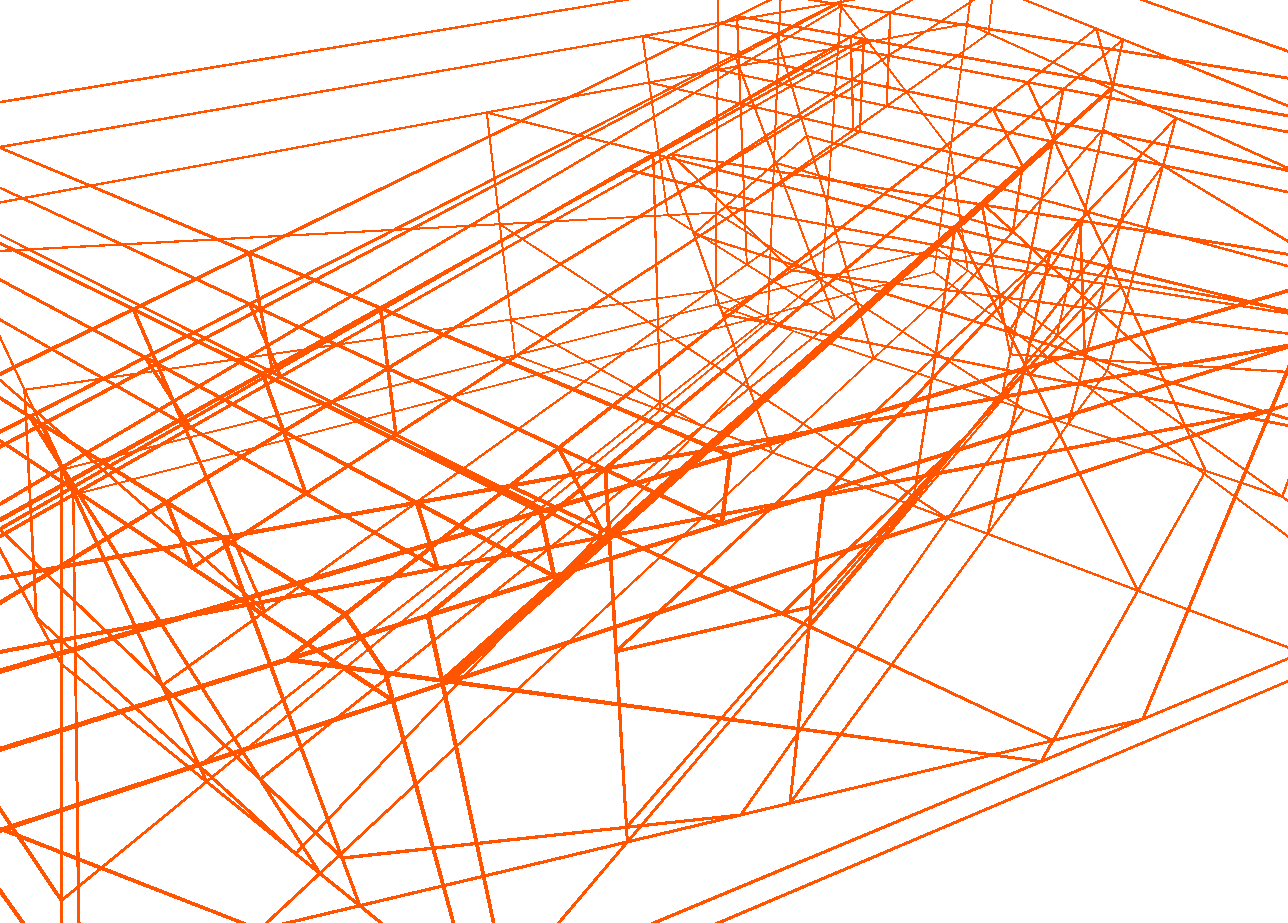}}
	\hspace{1em}
	\subfloat{\includegraphics[width=0.15\linewidth]{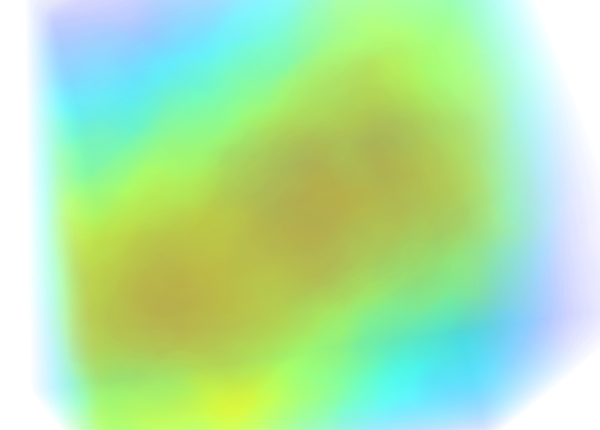}}
	\hspace{1em}
	\subfloat{\includegraphics[width=0.15\linewidth]{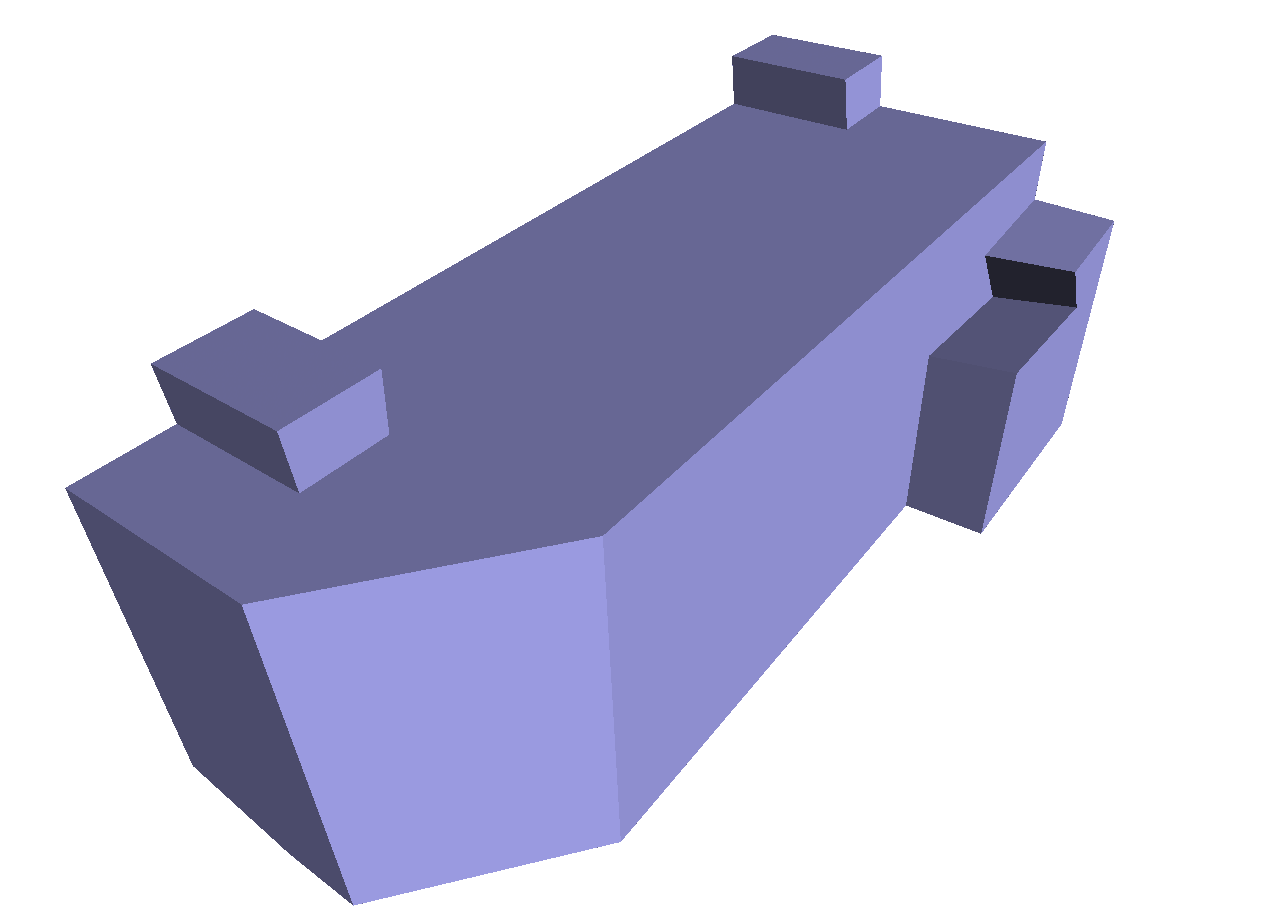}}
	\hspace{1em}
	\subfloat{\includegraphics[width=0.15\linewidth]{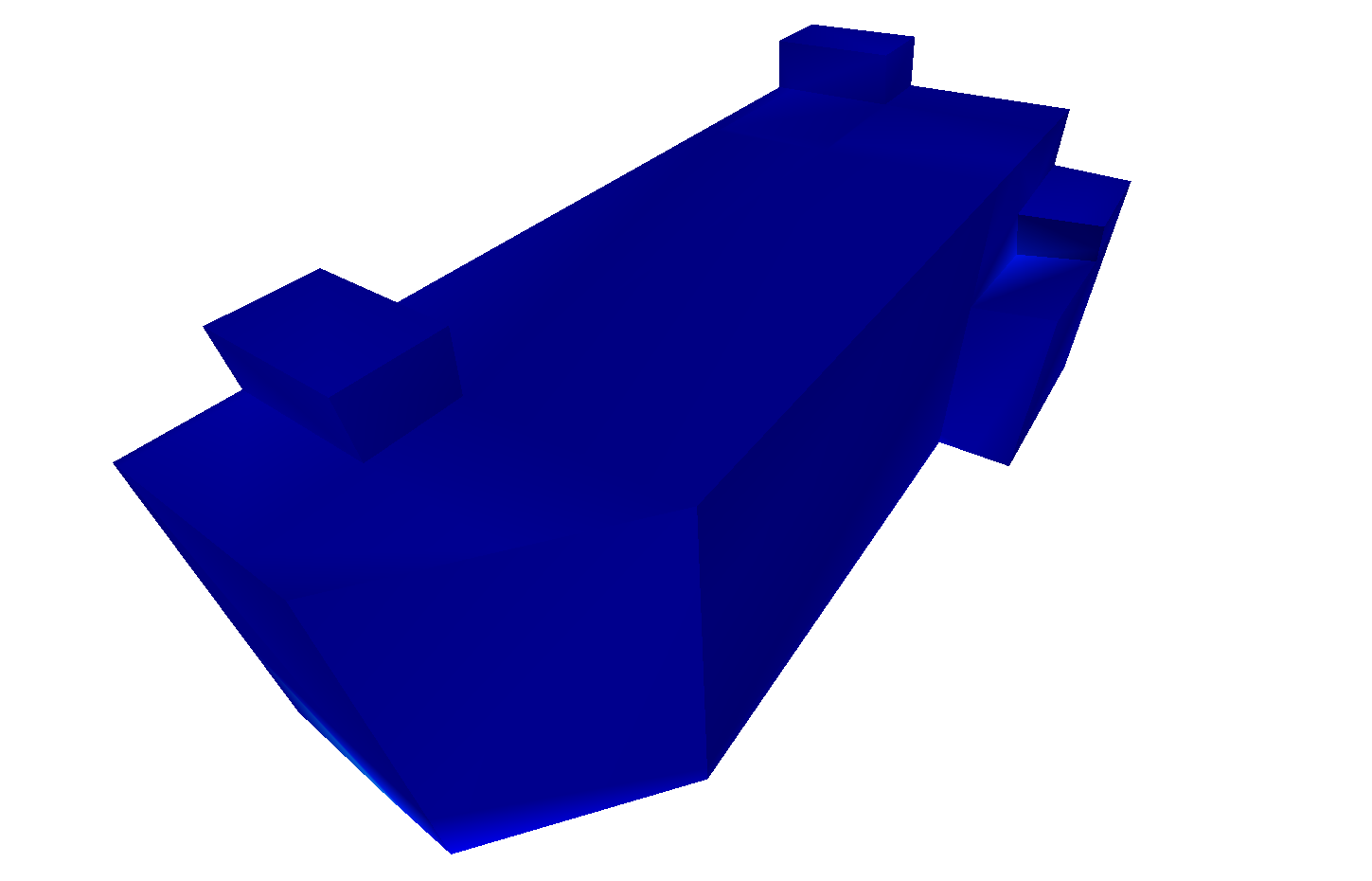}}

	(c)	
	\subfloat{\includegraphics[width=0.15\linewidth]{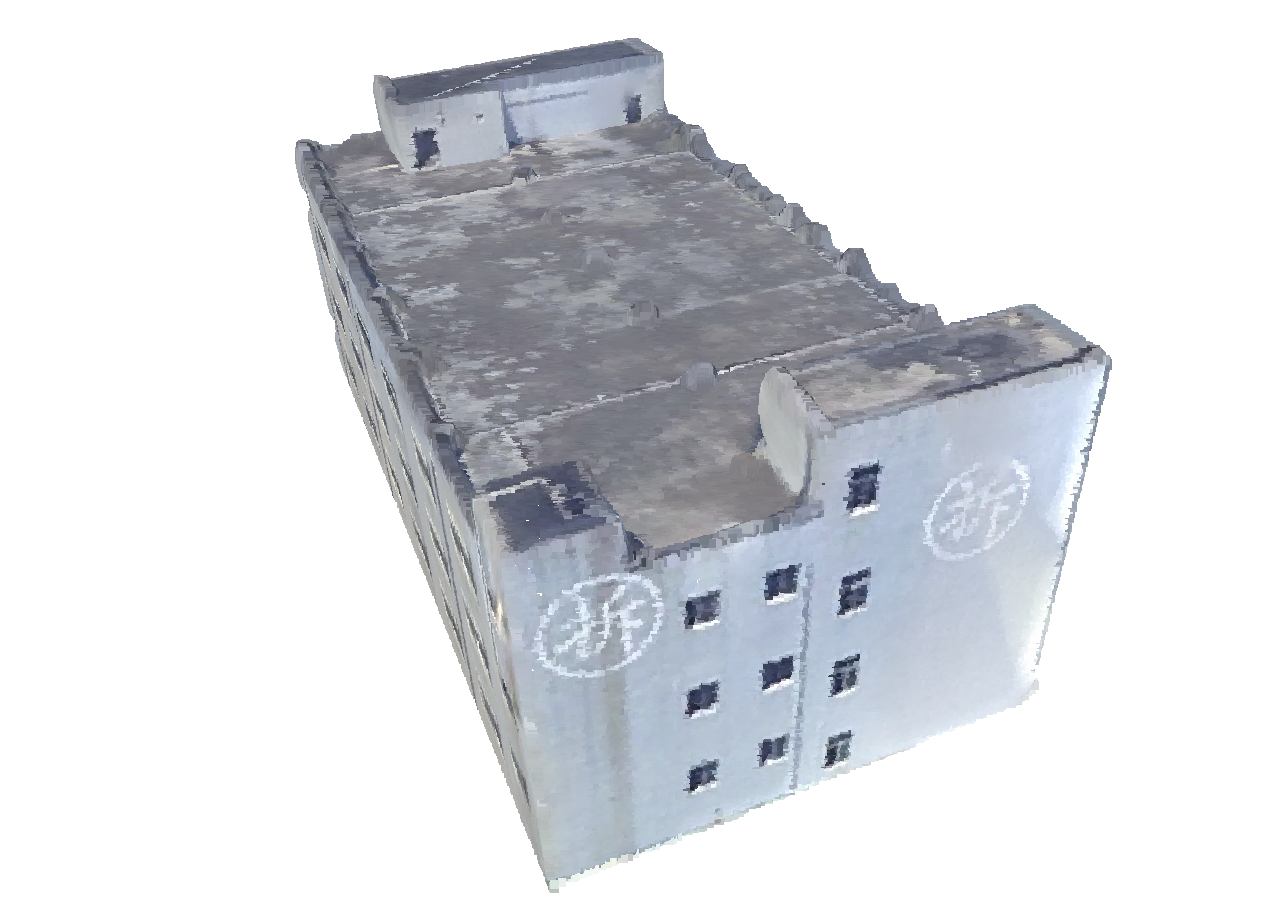}}
	\hspace{1em}
	\subfloat{\includegraphics[width=0.15\linewidth]{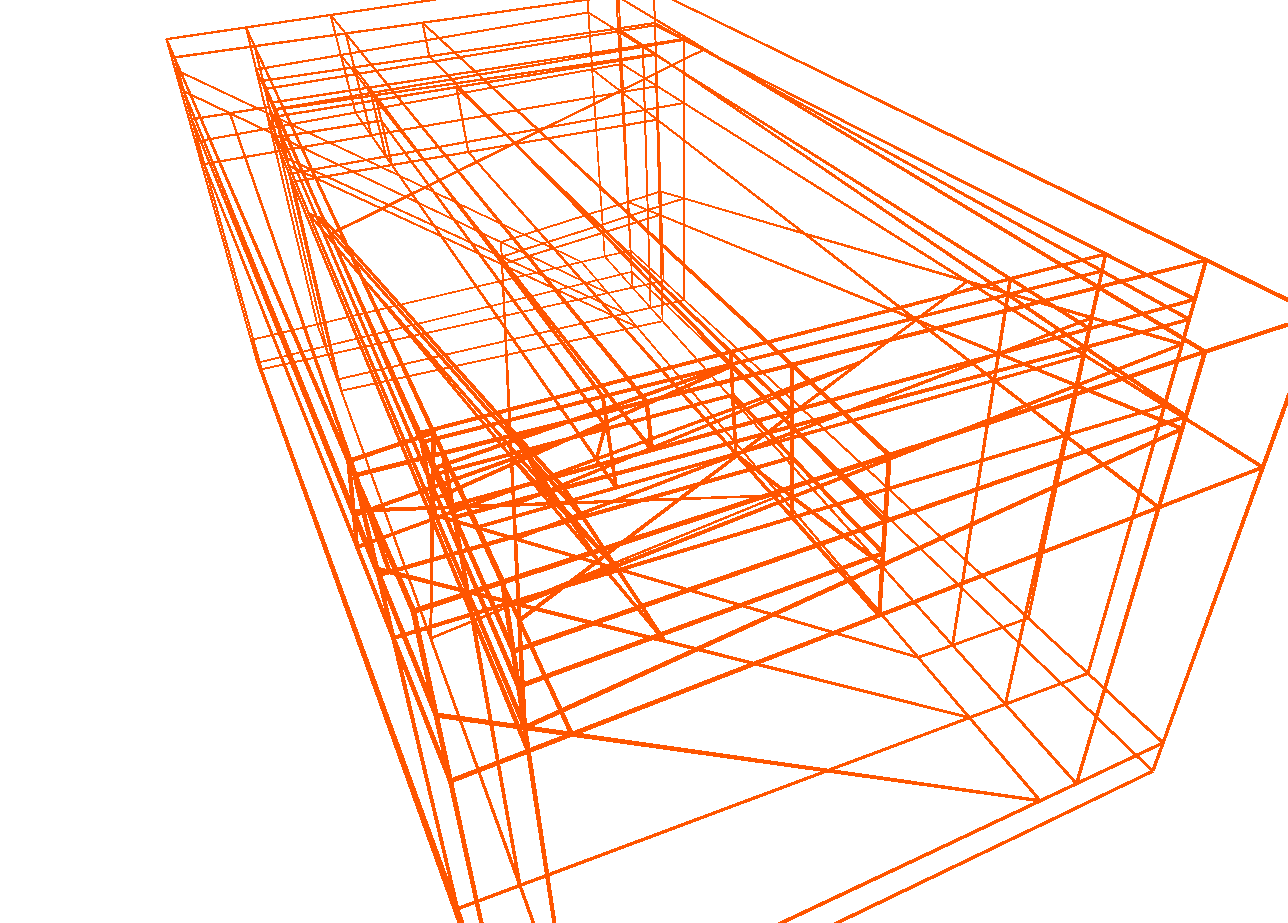}}
	\hspace{1em}
	\subfloat{\includegraphics[width=0.15\linewidth]{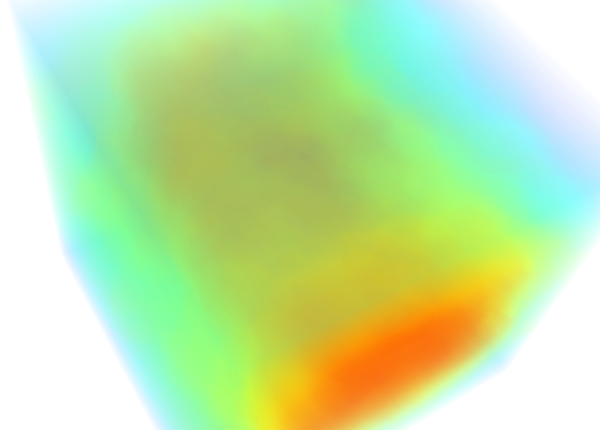}}
	\hspace{1em}
	\subfloat{\includegraphics[width=0.15\linewidth]{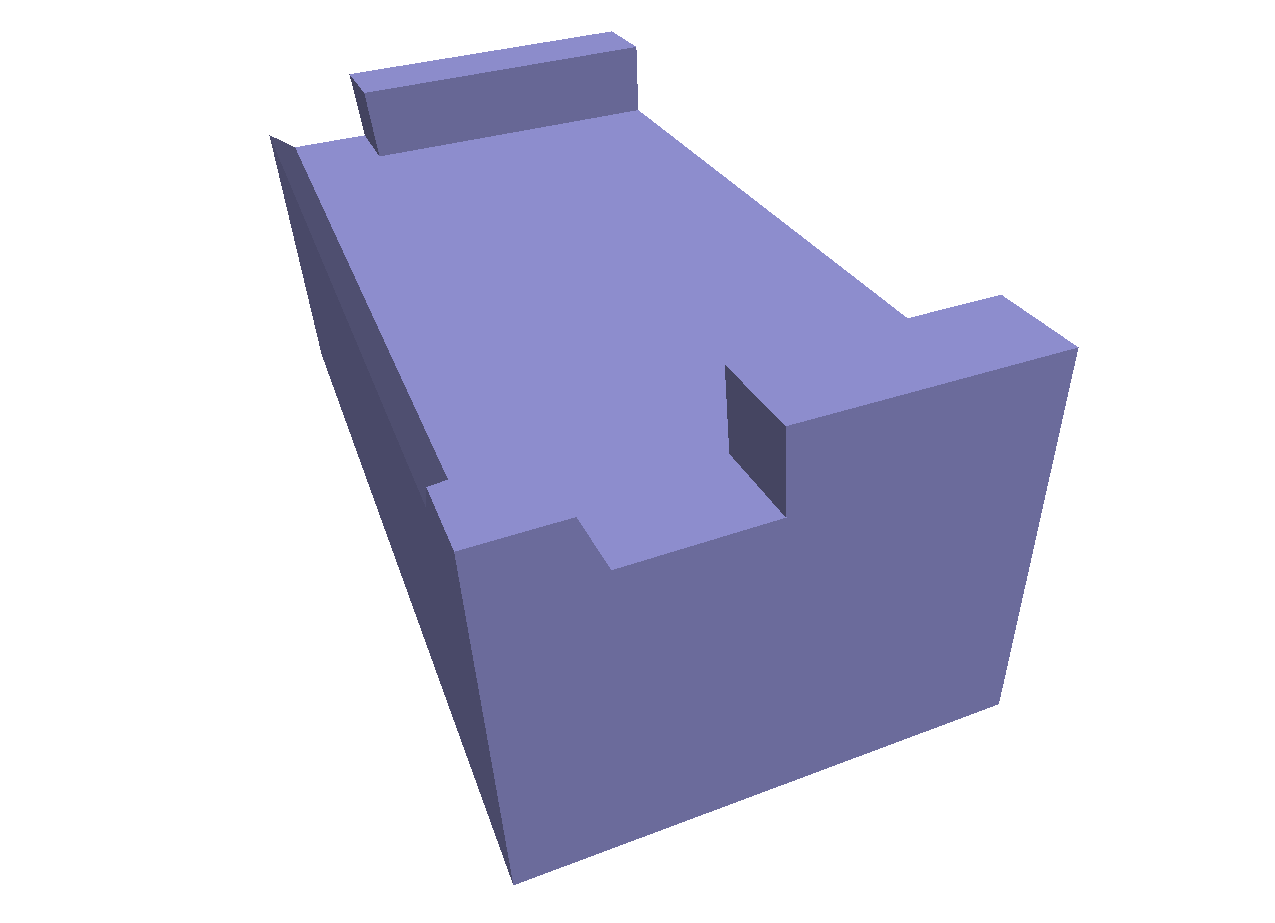}}
	\hspace{1em}
	\subfloat{\includegraphics[width=0.15\linewidth]{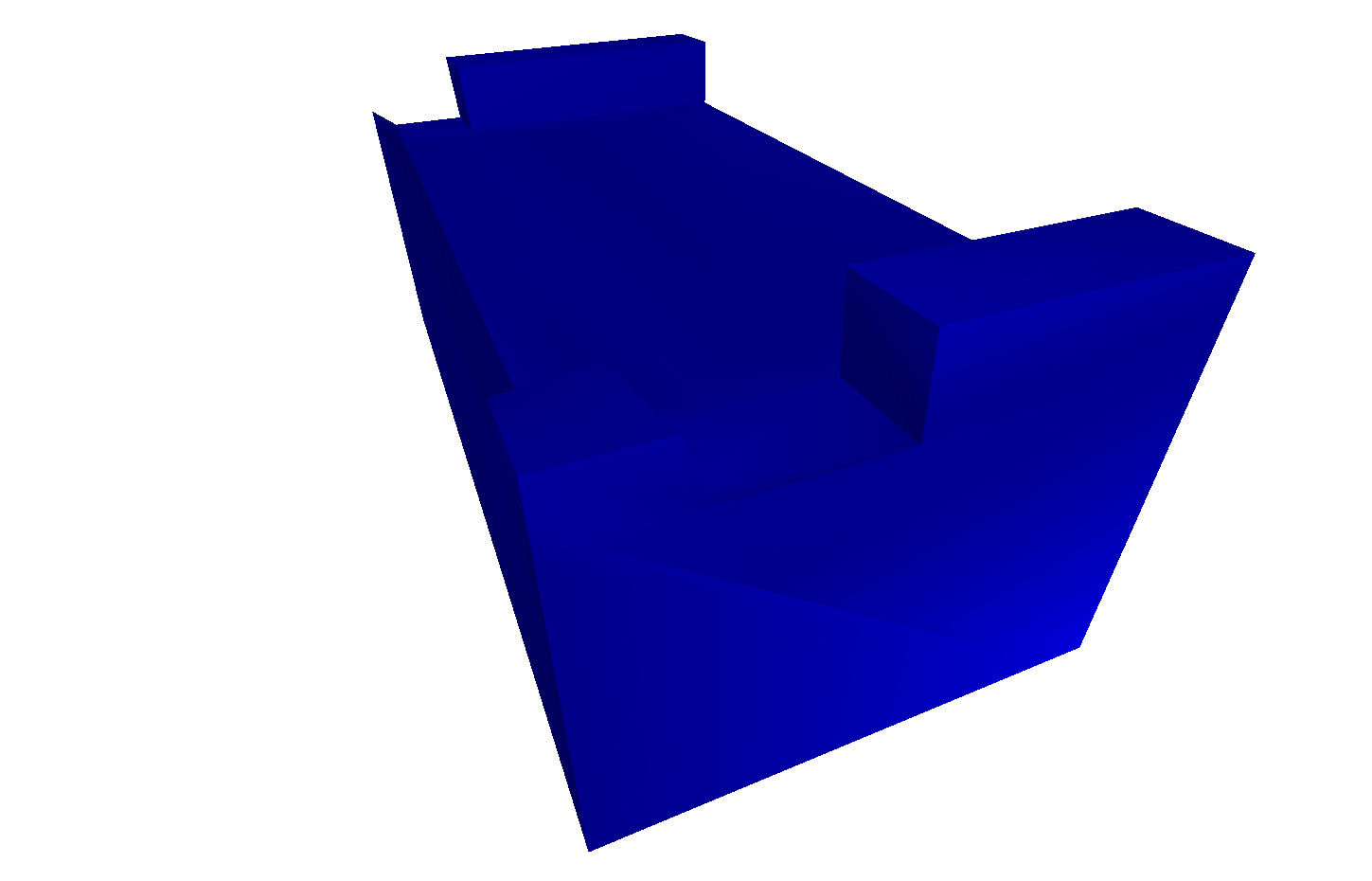}} 
	
	(d)	
	\subfloat{\includegraphics[width=0.15\linewidth]{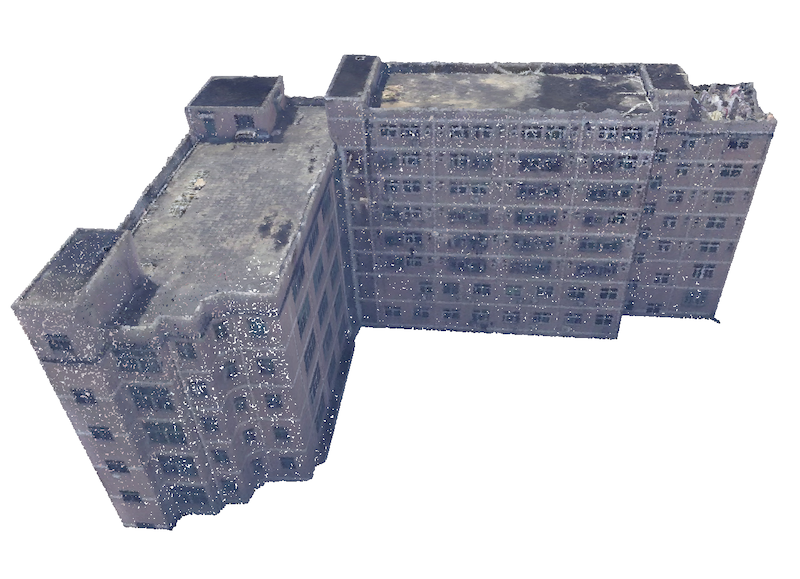}}
	\hspace{1em}
	\subfloat{\includegraphics[width=0.15\linewidth]{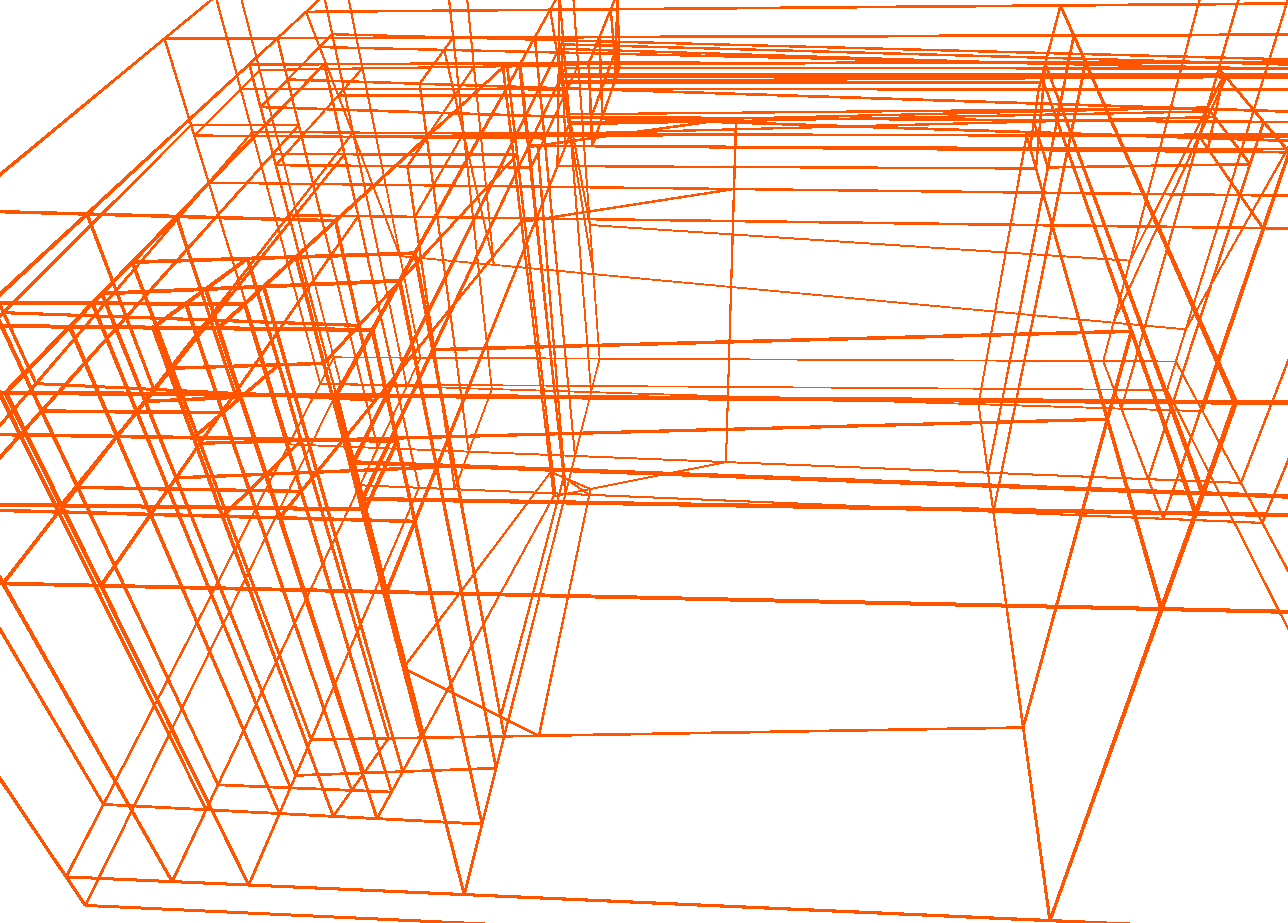}}
	\hspace{1em}
	\subfloat{\includegraphics[width=0.15\linewidth]{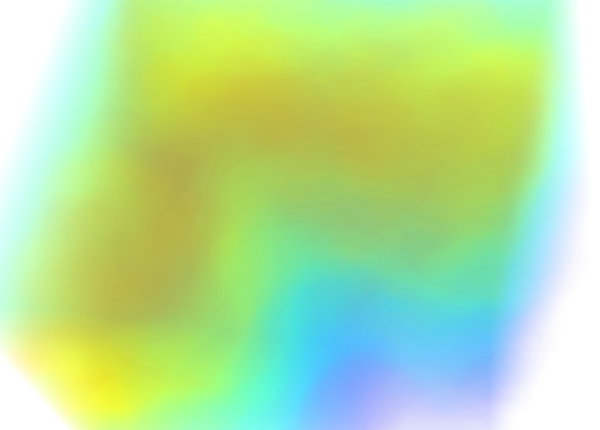}}
	\hspace{1em}
	\subfloat{\includegraphics[width=0.15\linewidth]{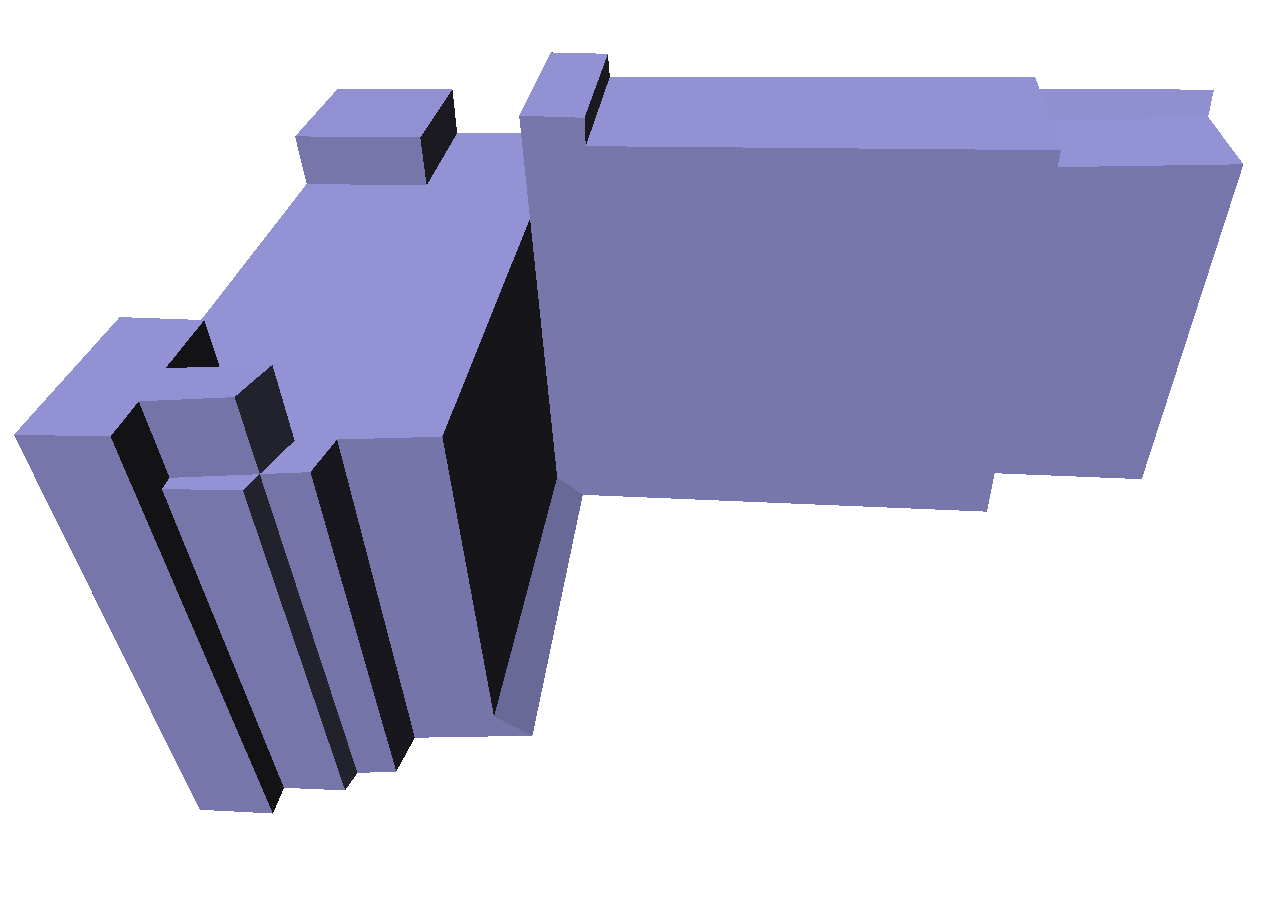}}
	\hspace{1em}
	\subfloat{\includegraphics[width=0.15\linewidth]{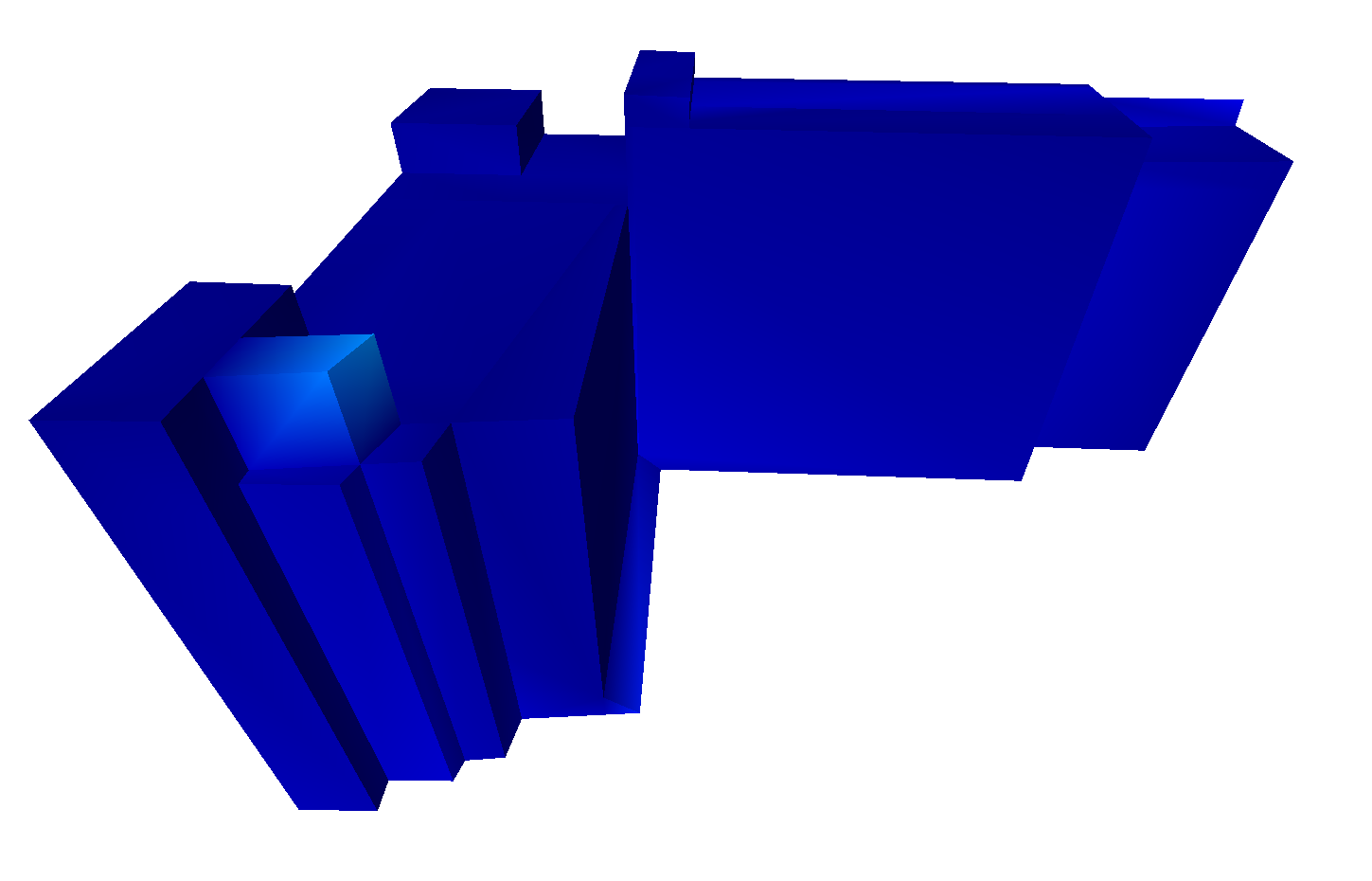}}

	(e)	
	\subfloat{\includegraphics[width=0.15\linewidth]{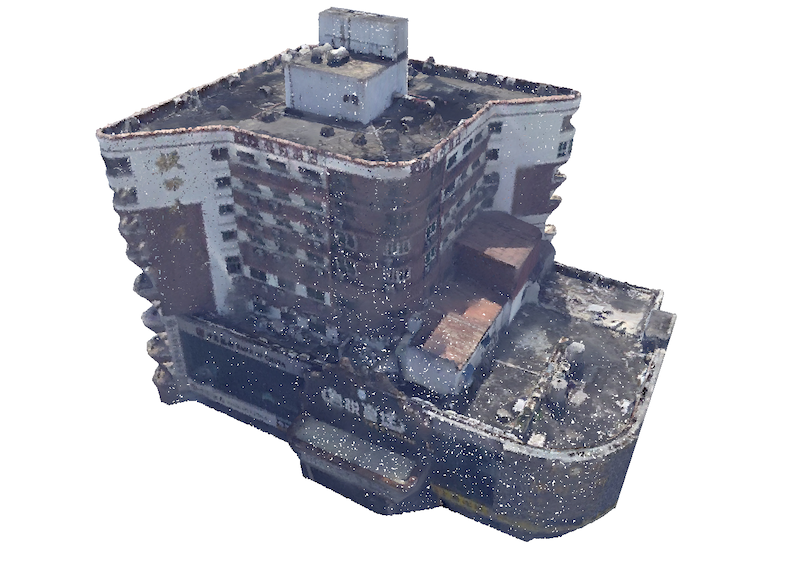}}
	\hspace{1em}
	\subfloat{\includegraphics[width=0.15\linewidth]{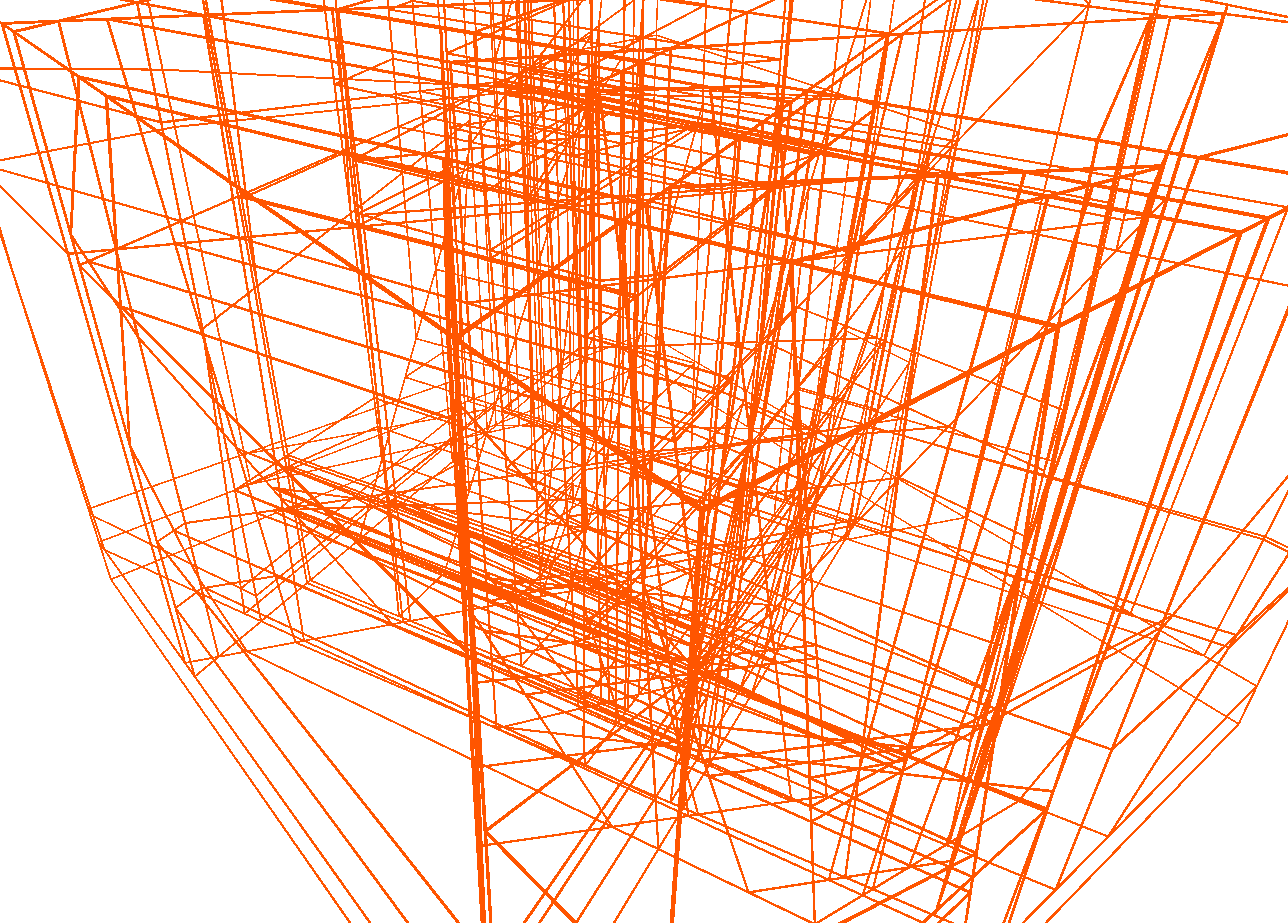}}
	\hspace{1em}
	\subfloat{\includegraphics[width=0.15\linewidth]{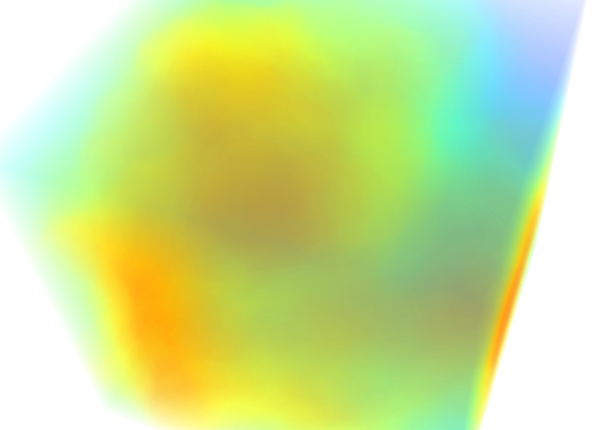}}
	\hspace{1em}
	\subfloat{\includegraphics[width=0.15\linewidth]{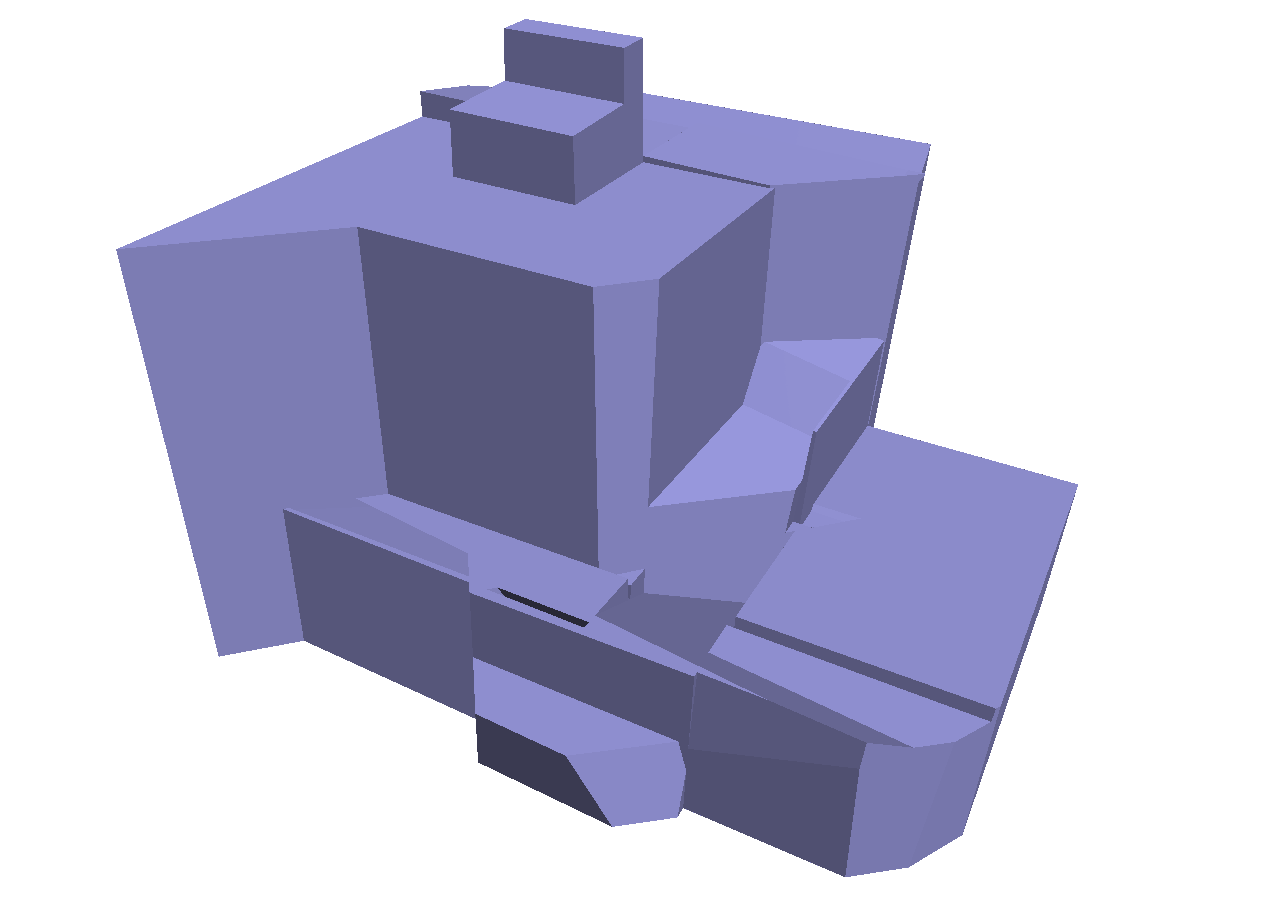}}
	\hspace{1em}
	\subfloat{\includegraphics[width=0.15\linewidth]{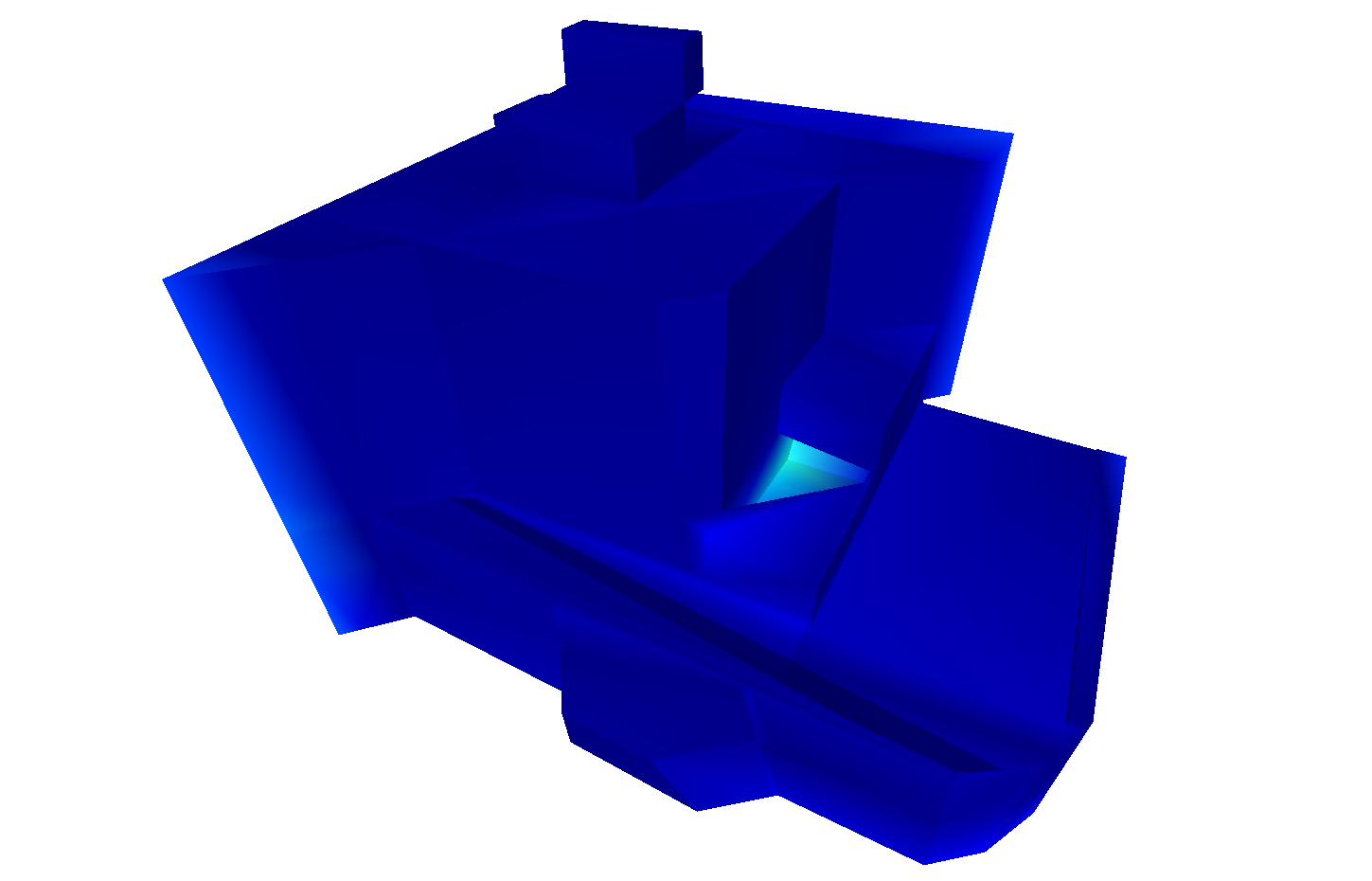}} 

	(f)	
	\subfloat{\includegraphics[width=0.15\linewidth]{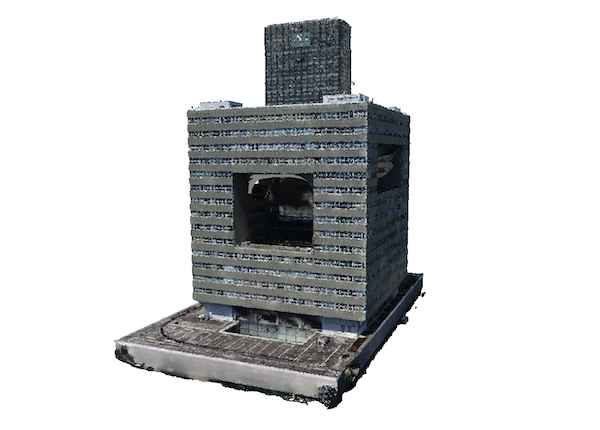}}
	\hspace{1em}
	\subfloat{\includegraphics[width=0.15\linewidth]{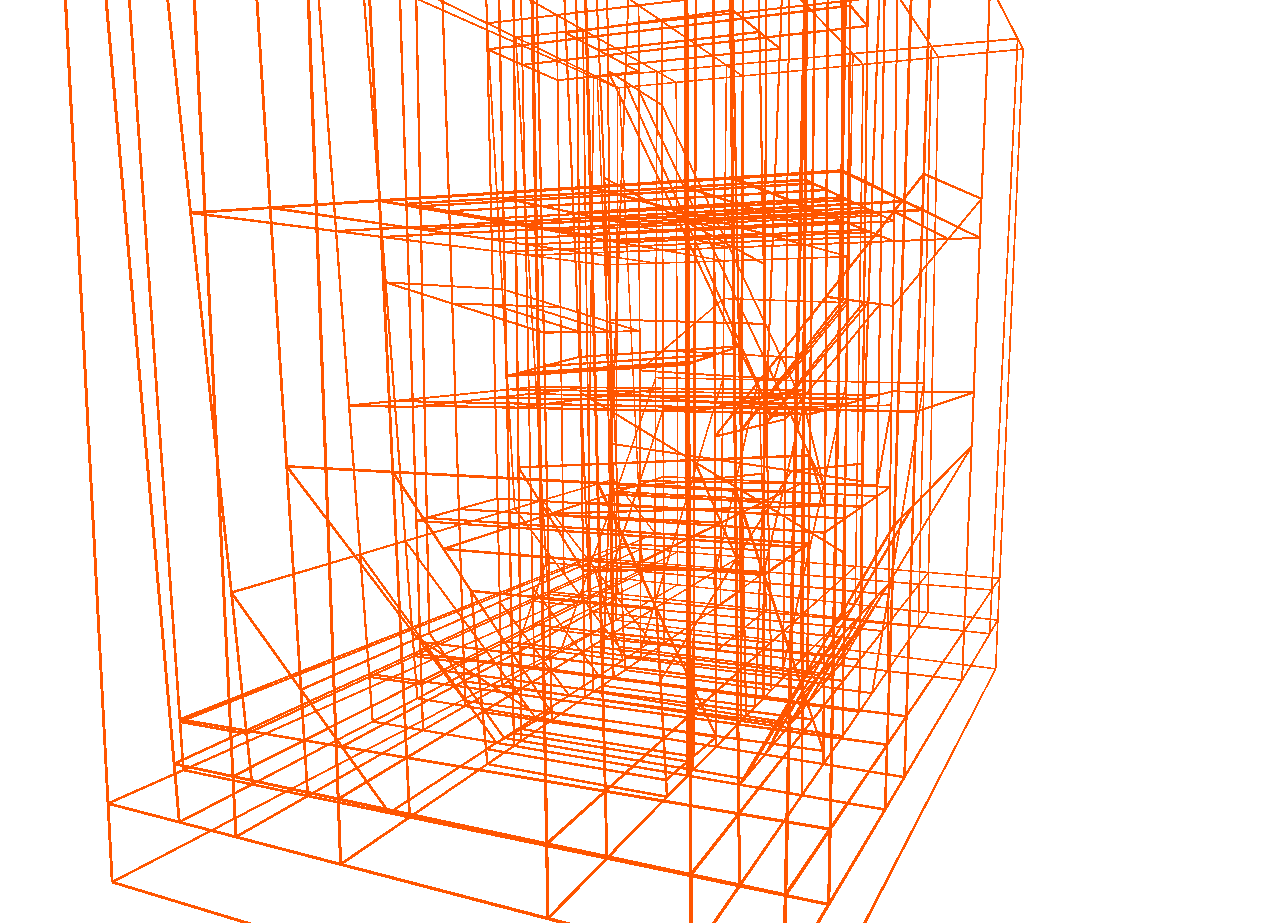}}
	\hspace{1em}
	\subfloat{\includegraphics[width=0.15\linewidth]{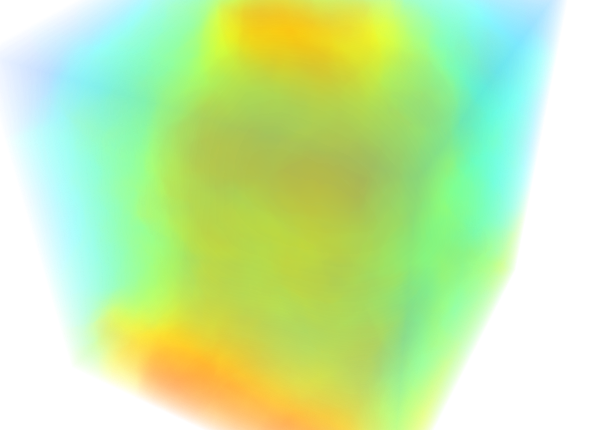}}
	\hspace{-0.5em}
	\subfloat{\includegraphics[height=0.1\linewidth]{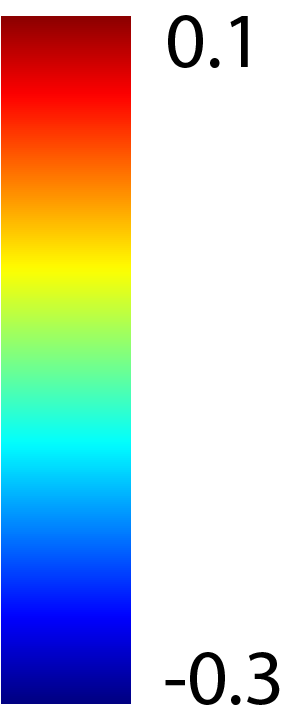}}
%	\hspace{0.5em}
	\subfloat{\includegraphics[width=0.15\linewidth]{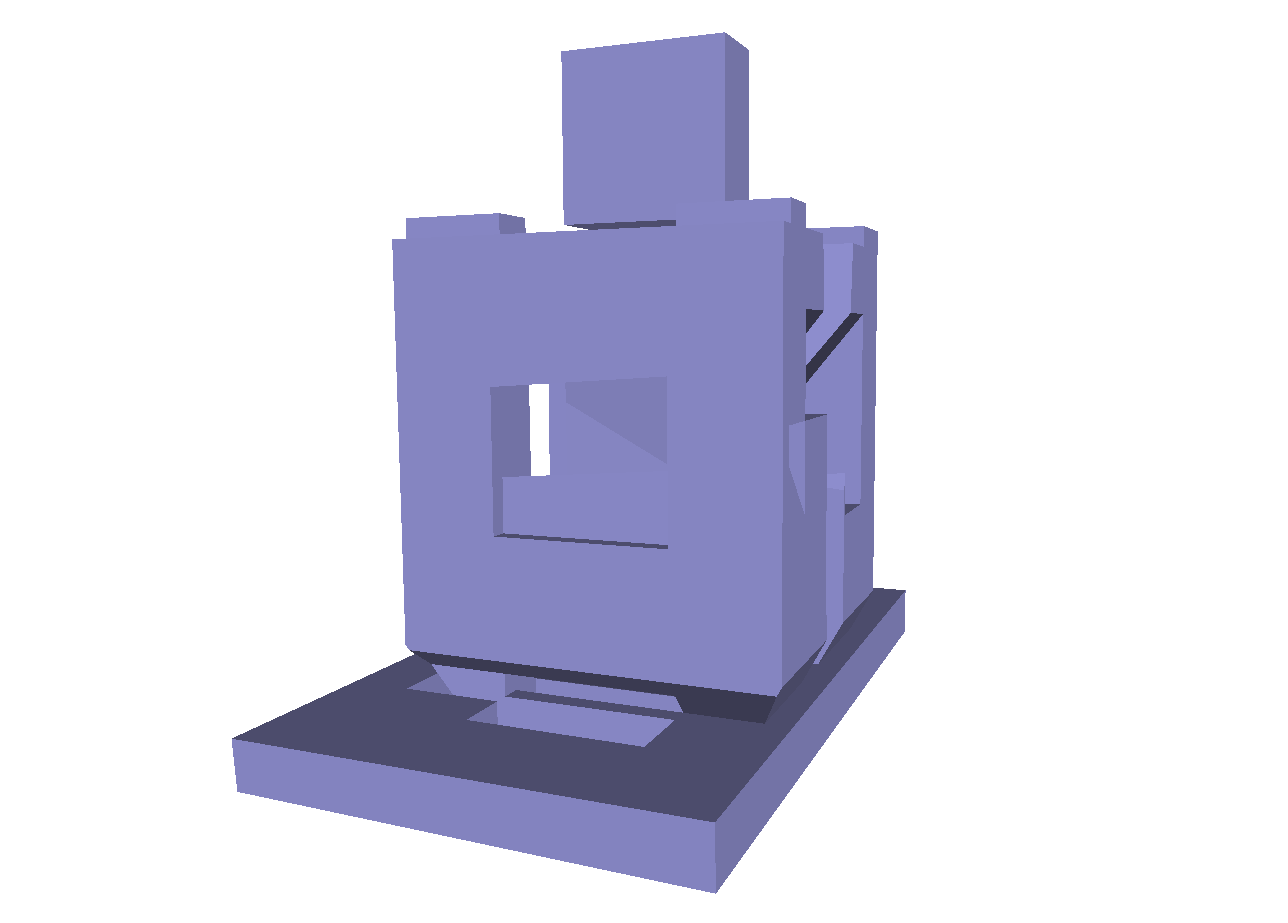}}
	\hspace{0em}
	\subfloat{\includegraphics[width=0.15\linewidth]{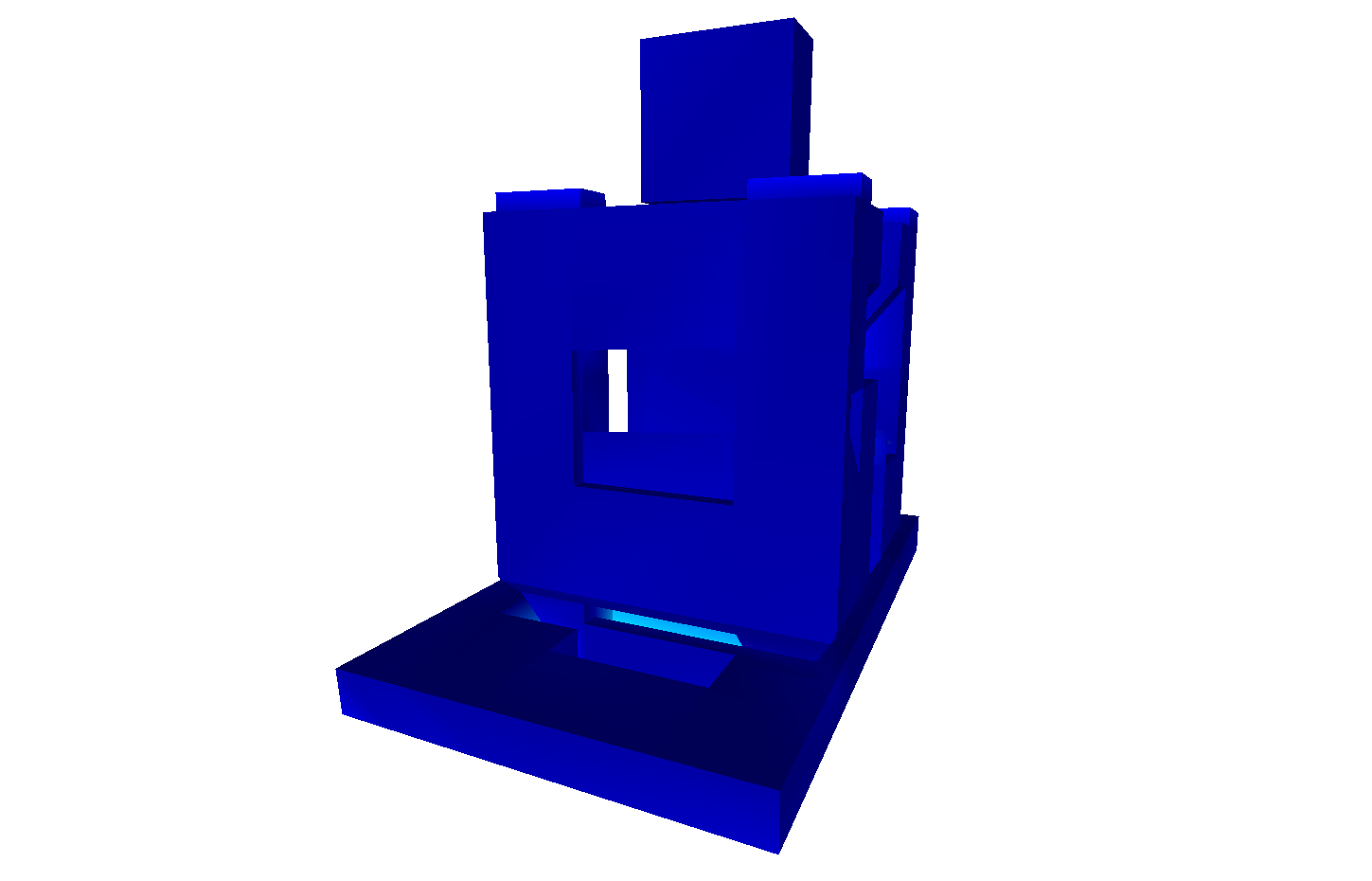}}
	\hspace{-1.1em}
	\subfloat{\includegraphics[height=0.1\linewidth]{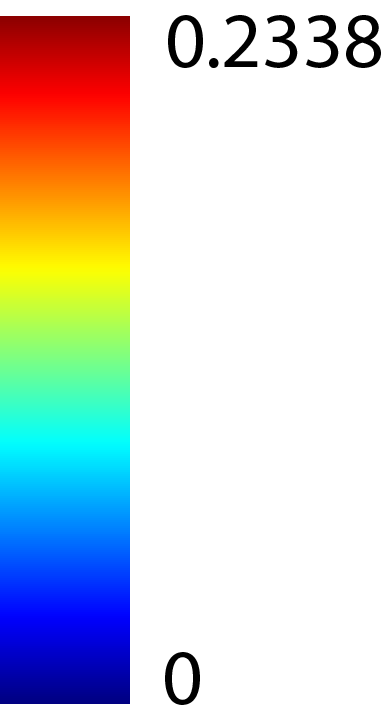}}
	
	\caption{Reconstruction results on the \textit{Shenzhen} point clouds. From left to right: input point cloud (colored randomly per planar primitive), wireframe of the cell complex, volume rendering of the SDF, reconstructed building model, and error map.}
	\label{fig:results_shenzhen}
\end{figure*}

Since local features are used for occupancy estimation, and the training data is augmented with intentionally added noise and occlusions, our reconstruction can reasonably generalize from synthetic data to real-world point clouds.
\autoref{fig:results_shenzhen} shows the reconstruction results on the \textit{Shenzhen} point clouds directly using the neural network trained on the \textit{Helsinki no-bottom} dataset. 
The inferred distance fields still conform to the real-world buildings with styles unseen by the network during training.
Though lower parts of the buildings are insufficiently measured, our method can still successfully reconstruct complete buildings.
For the test on the \textit{Shenzhen} dataset, since no ground truth surface models are available for quantitative evaluation of the reconstruction, we quantify reconstruction accuracy by measuring the Hausdorff distances from the reconstructed surfaces to the corresponding input point clouds.
It is worth noting that the most prominent error seen from each error map lies in where the measurement is missing (i.e., some lower parts of the building were occluded in data acquisition).
In addition, the complex roof structures occasionally introduce tangible errors when approximated with piecewise-planar abstraction. \autoref{fig:results_shenzhen} (d) shows such an example, where the protuberance on top of the building is approximated with a superfluous prism.

\subsection{Evaluation}

\subsubsection{Fidelity and complexity}
We have compared our method with various generic reconstruction, model-based approaches, geometry simplification, and primitive assembly methods.

\begin{figure*}[ht]
	\centering
%	\subfloat{
%	\vtop{%
%  	\vskip-100pt
%  	\hbox{%
	\raisebox{\height-6ex}{
	\includegraphics[width=0.26\linewidth]{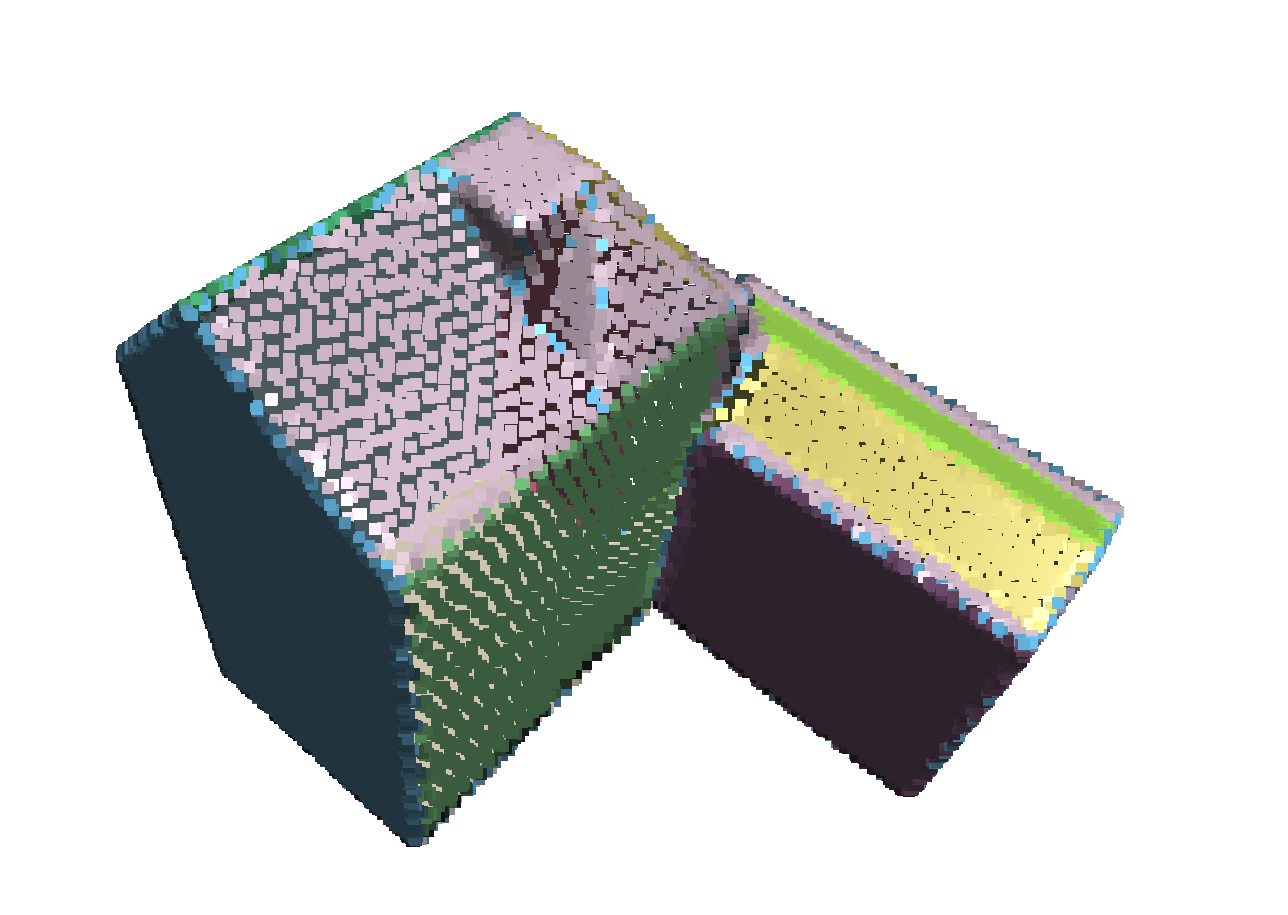}}
	\includegraphics[width=0.22\linewidth]{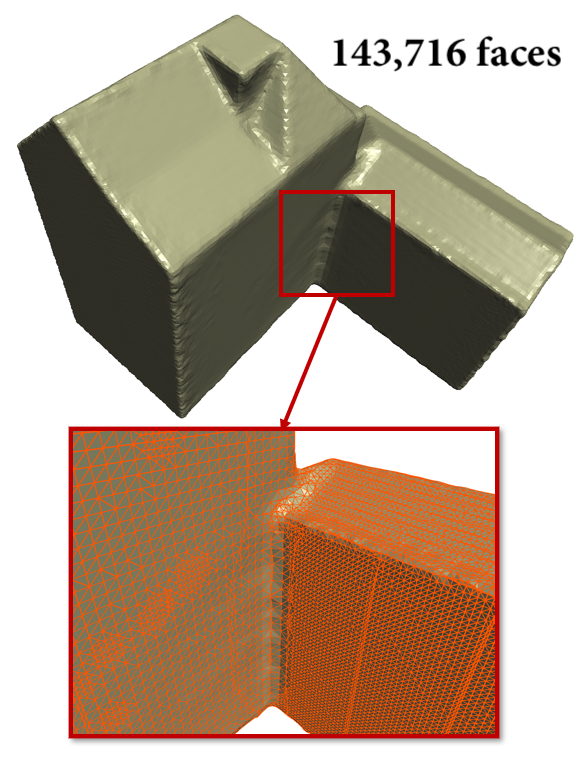}
	\includegraphics[width=0.22\linewidth]{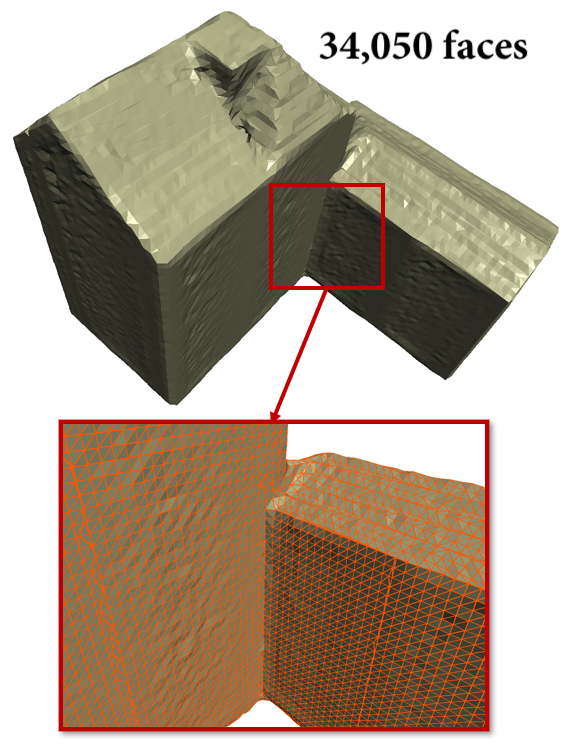}
	\includegraphics[width=0.22\linewidth]{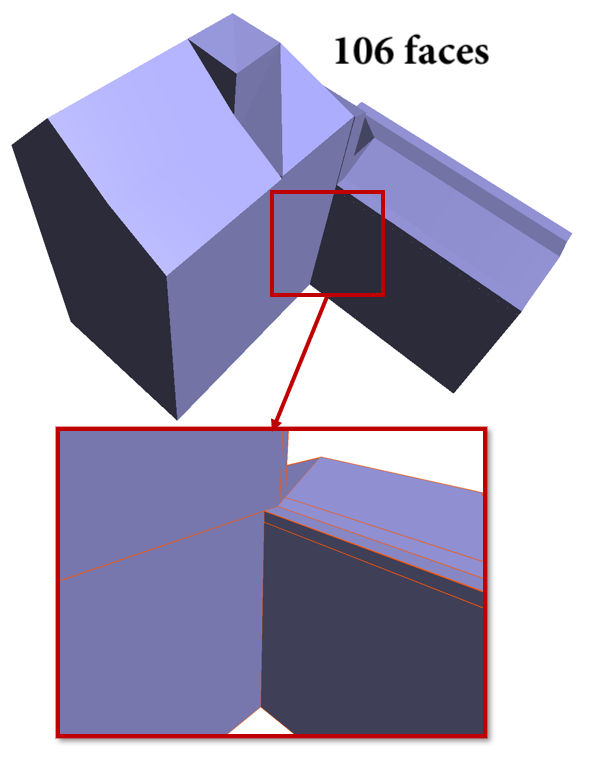}
	
	\subfloat[Input]{
	\vtop{%
  	\vskip-100pt
  	\hbox{%
	\includegraphics[width=0.26\linewidth]{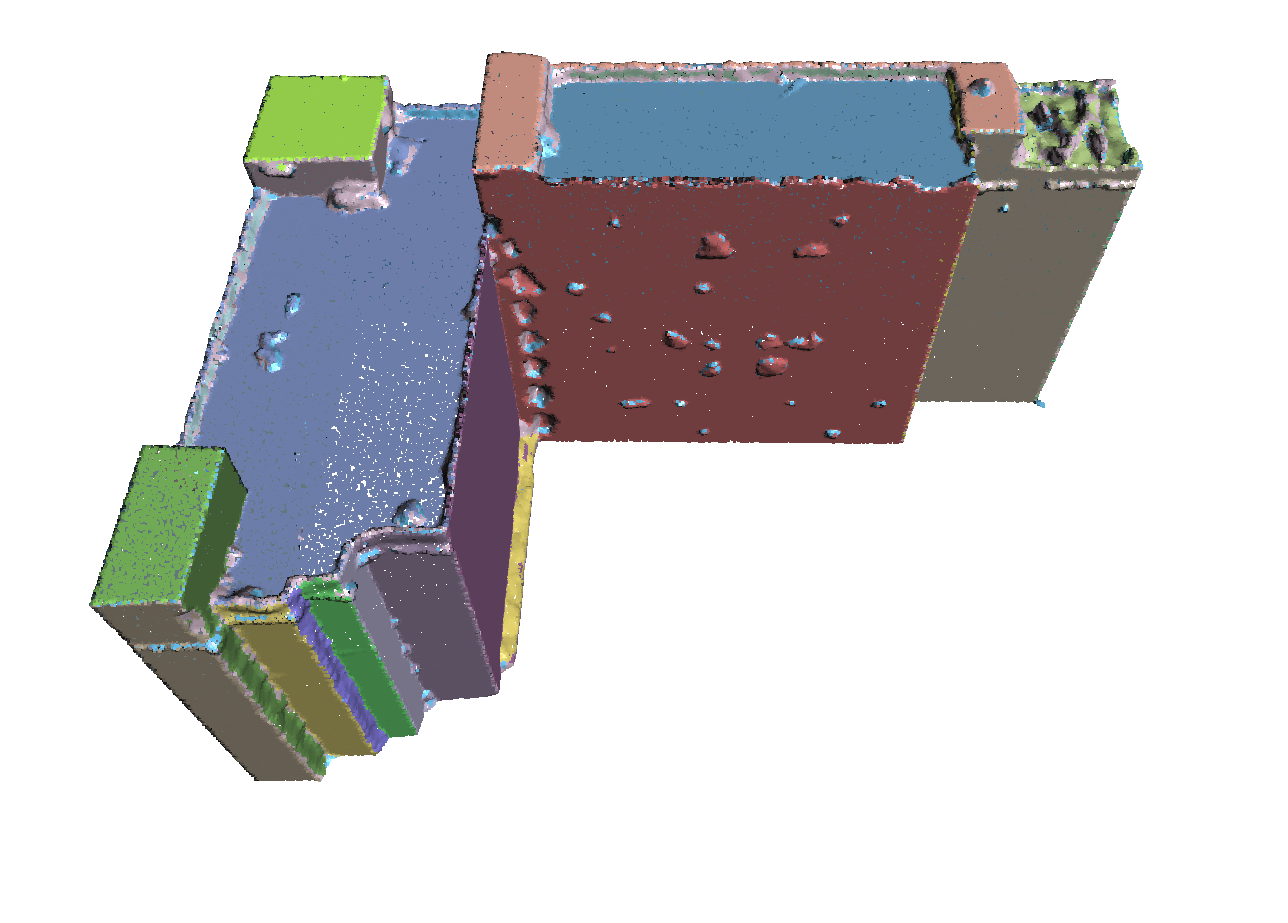}
    }}}
	\subfloat[Poisson]{\includegraphics[width=0.22\linewidth]{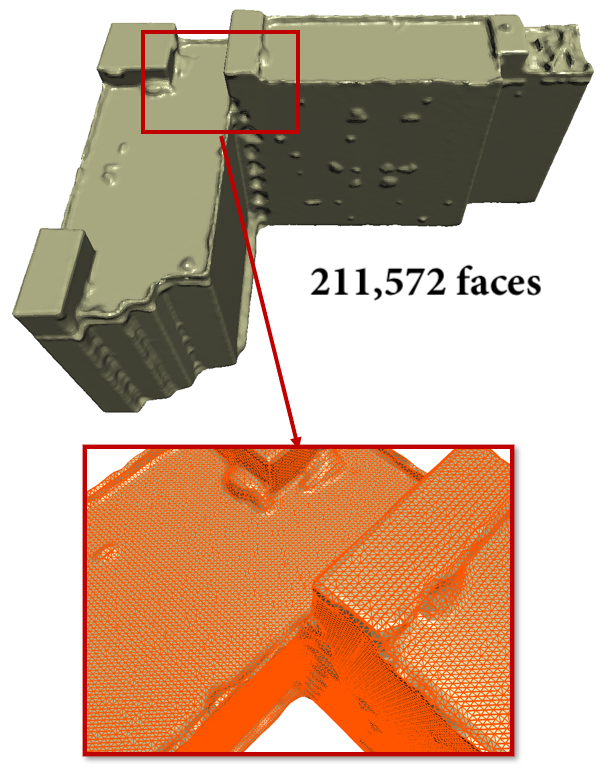}}
	\subfloat[Points2Surf]{\includegraphics[width=0.22\linewidth]{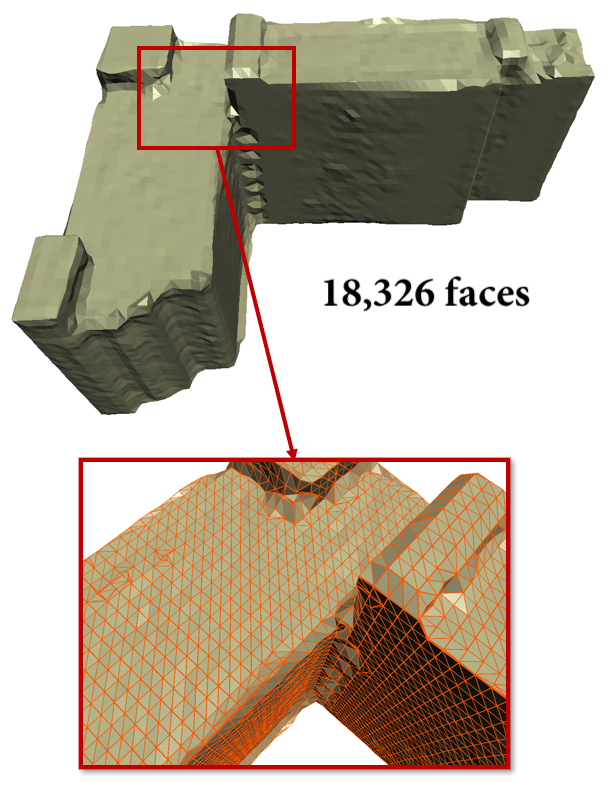}}
	\subfloat[Ours]{\includegraphics[width=0.22\linewidth]{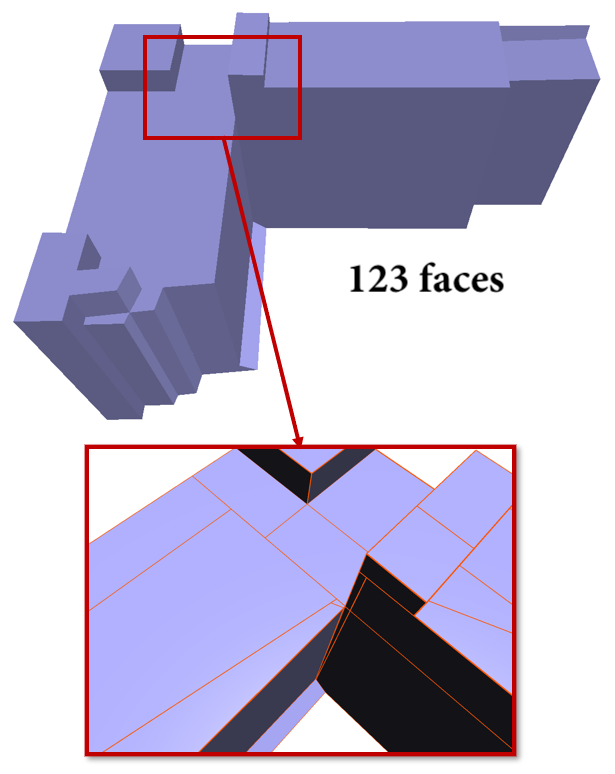}}
	
	\caption{Comparison between our method and two generic reconstruction methods, namely, Poisson surface reconstruction~\citep{kazhdan2006poisson} and Points2Surf~\citep{erler2020points2surf}.}
	\label{fig:comparison_poisson}
\end{figure*}

\autoref{fig:comparison_poisson} demonstrates the comparison with two generic reconstruction methods, namely Poisson surface reconstruction~\citep{kazhdan2006poisson} and the learning-based method Points2Surf~\citep{erler2020points2surf}.
Both methods produce surface models with a massive number of triangle faces and chamfered edges, which not only leads to extra memory consumption but also potentially artifacts such as bumps and holes. In contrast, our approach describes the building geometry as a lightweight watertight polyhedron with sharp edges.

% a bit hacking to align the sub-images vertically

\begin{figure*}[!ht]
	\centering
	\includegraphics[width=0.15\linewidth]{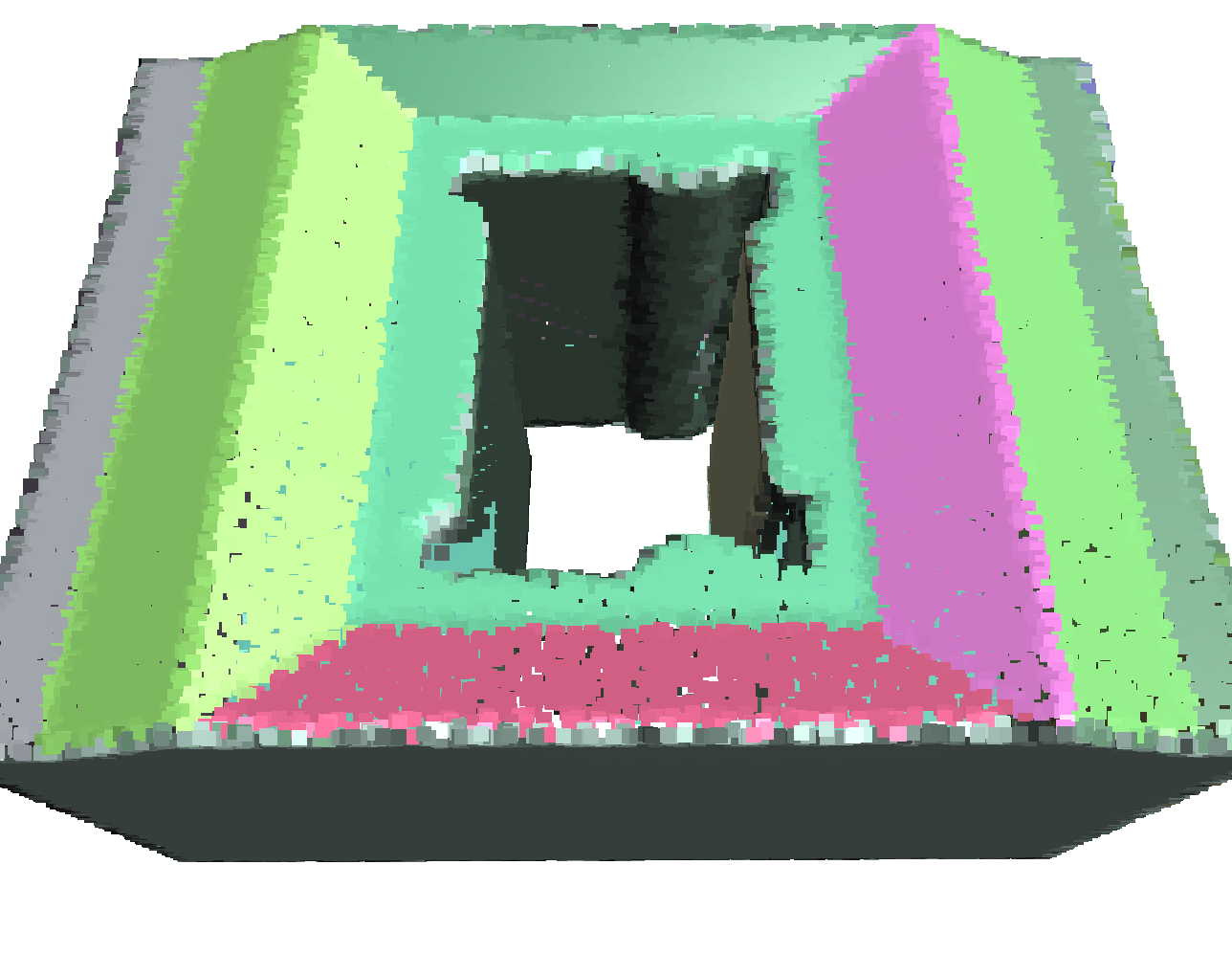}
	\includegraphics[width=0.15\linewidth]{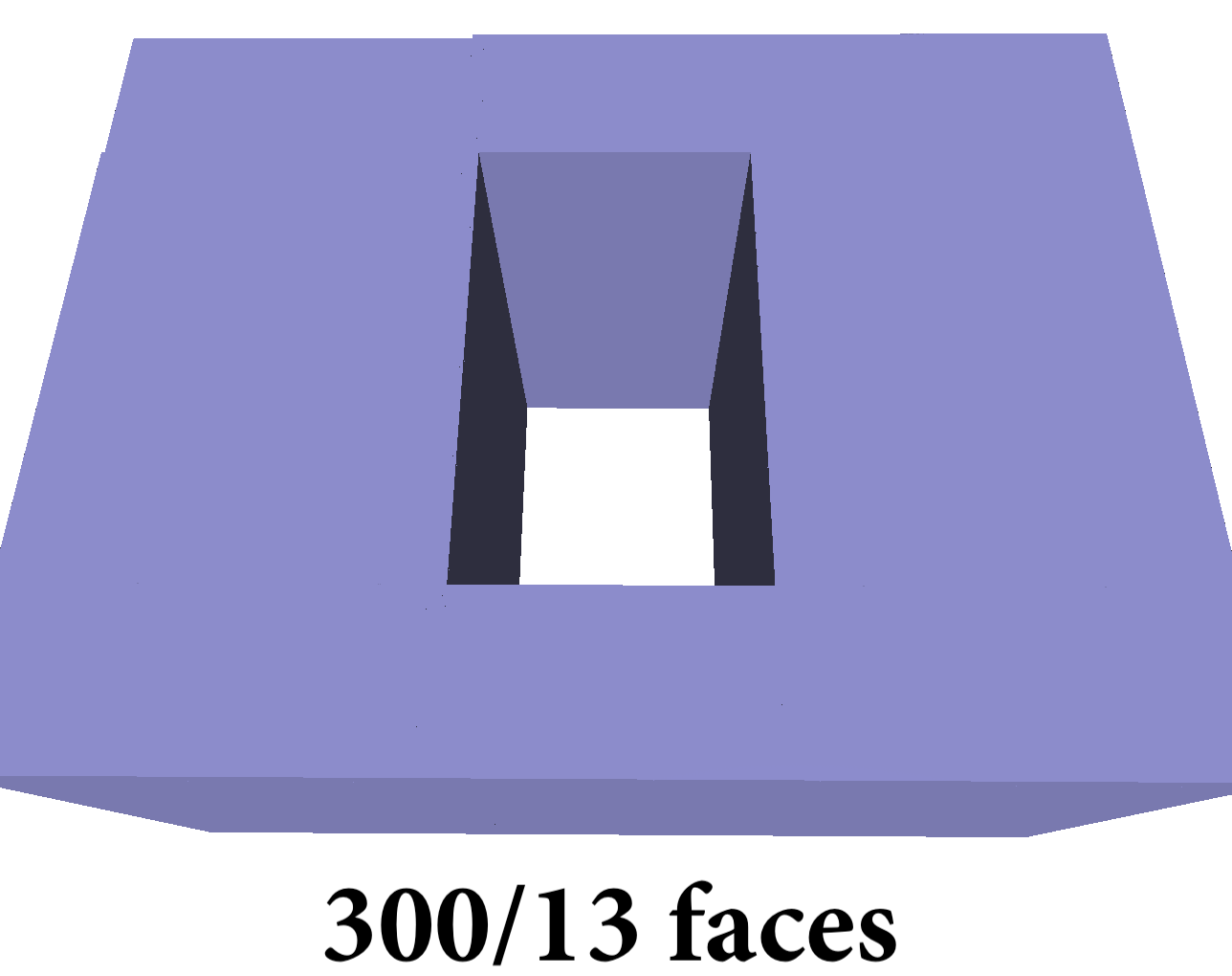}
	\includegraphics[width=0.15\linewidth]{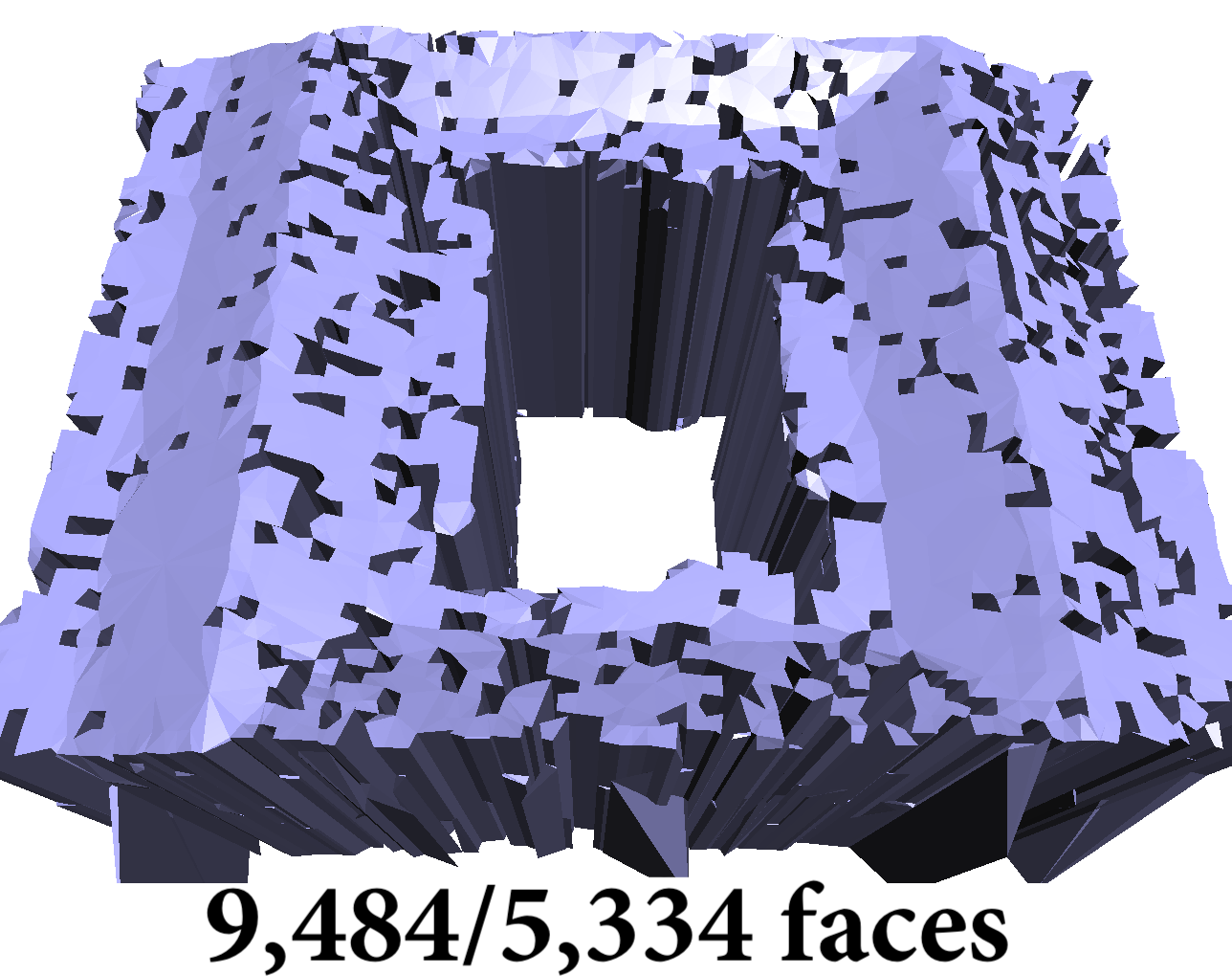}
	\includegraphics[width=0.15\linewidth]{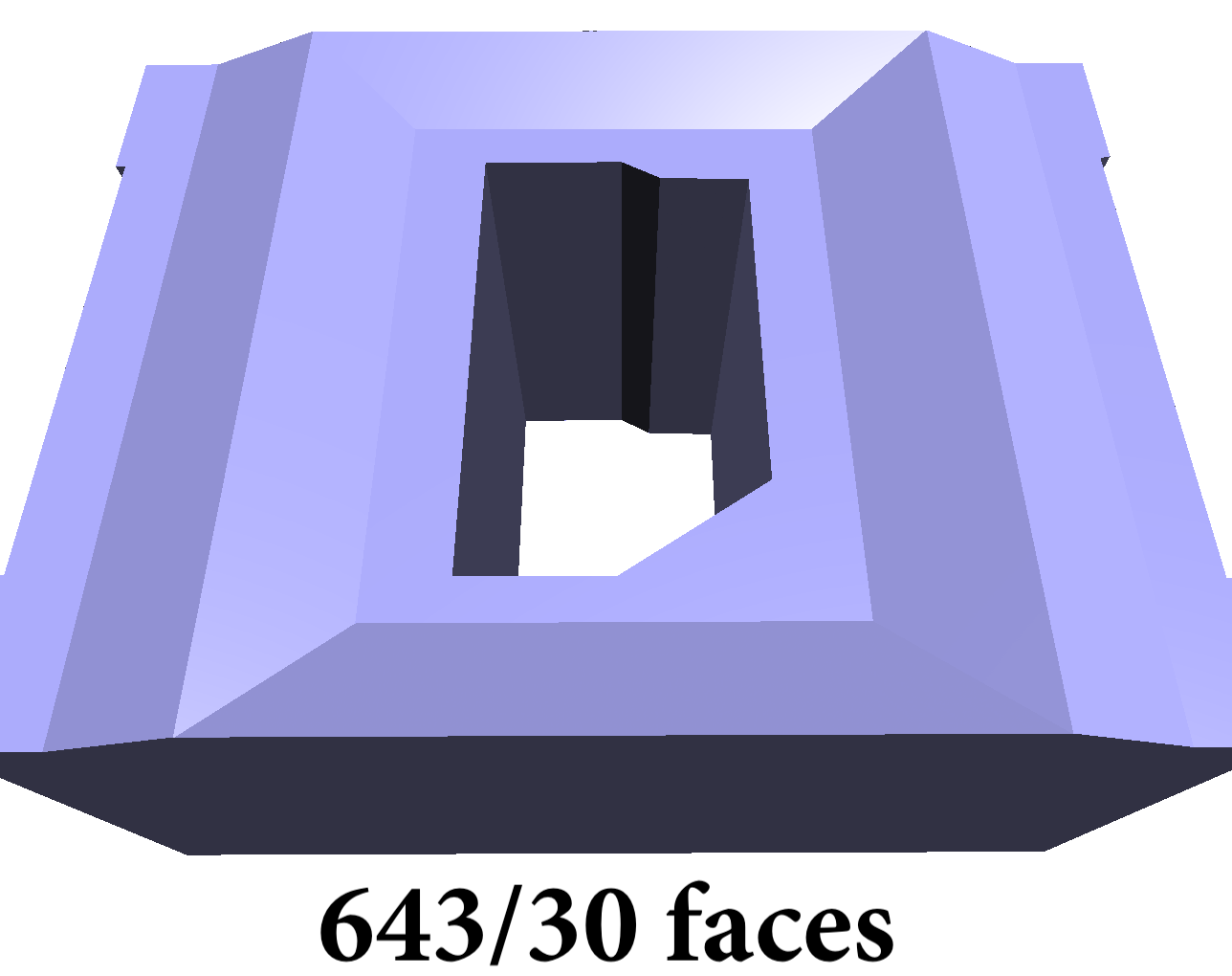}
	\includegraphics[width=0.15\linewidth]{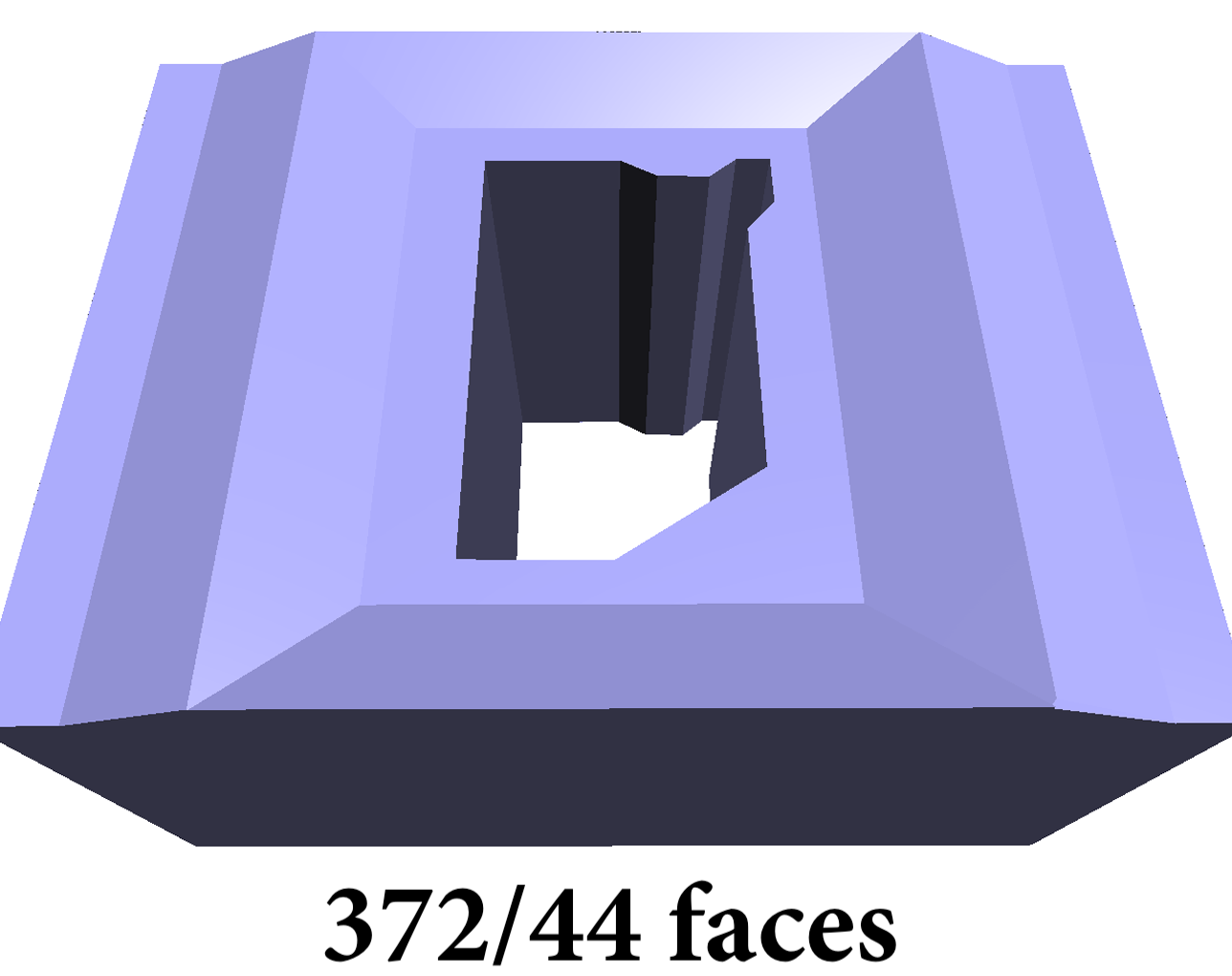}
	\includegraphics[width=0.15\linewidth]{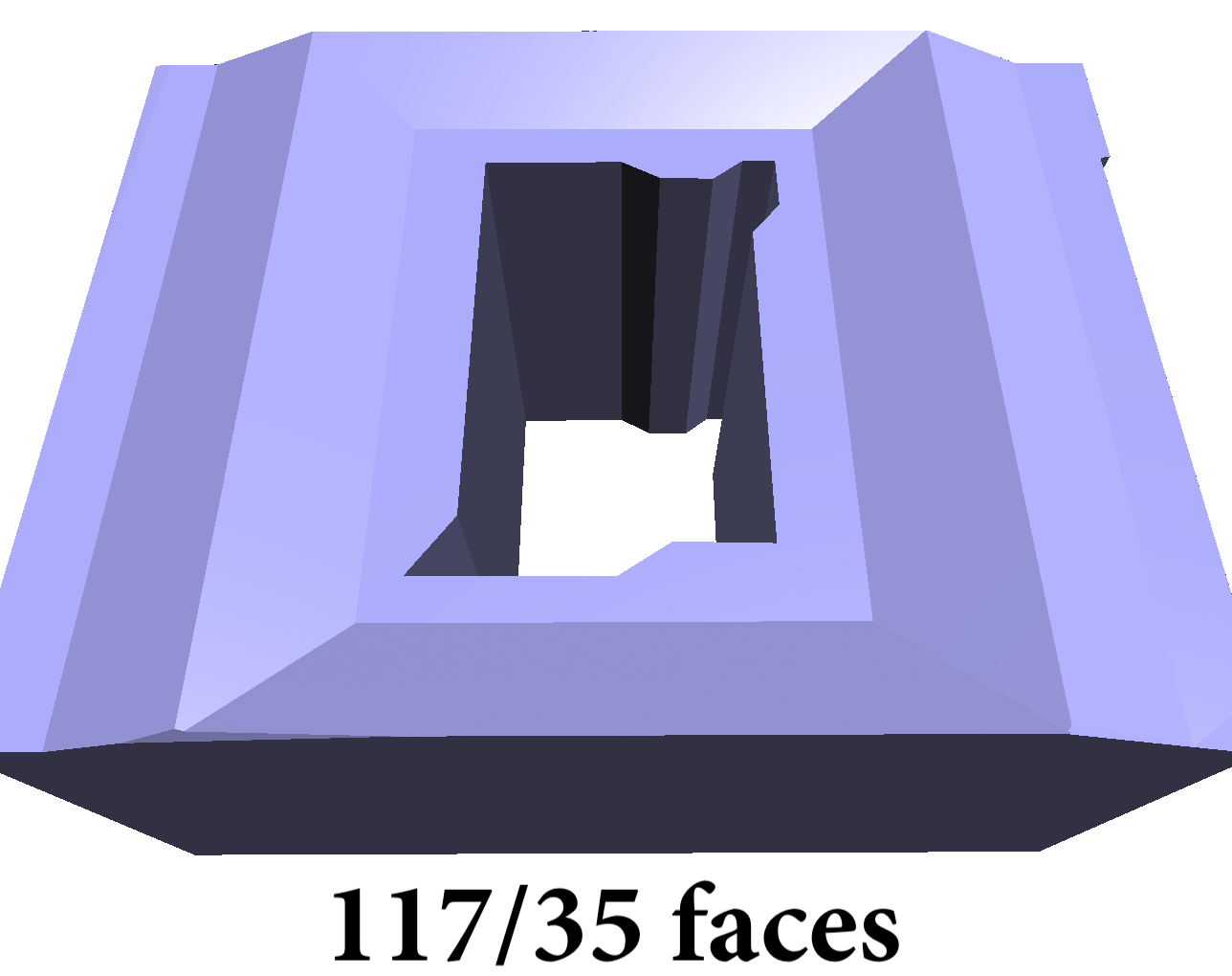}
	
	\subfloat[Input]{\includegraphics[width=0.16\linewidth]{figures/results/comparison/Building_4.png}}
	\subfloat[Manhattan]{\includegraphics[width=0.16\linewidth]{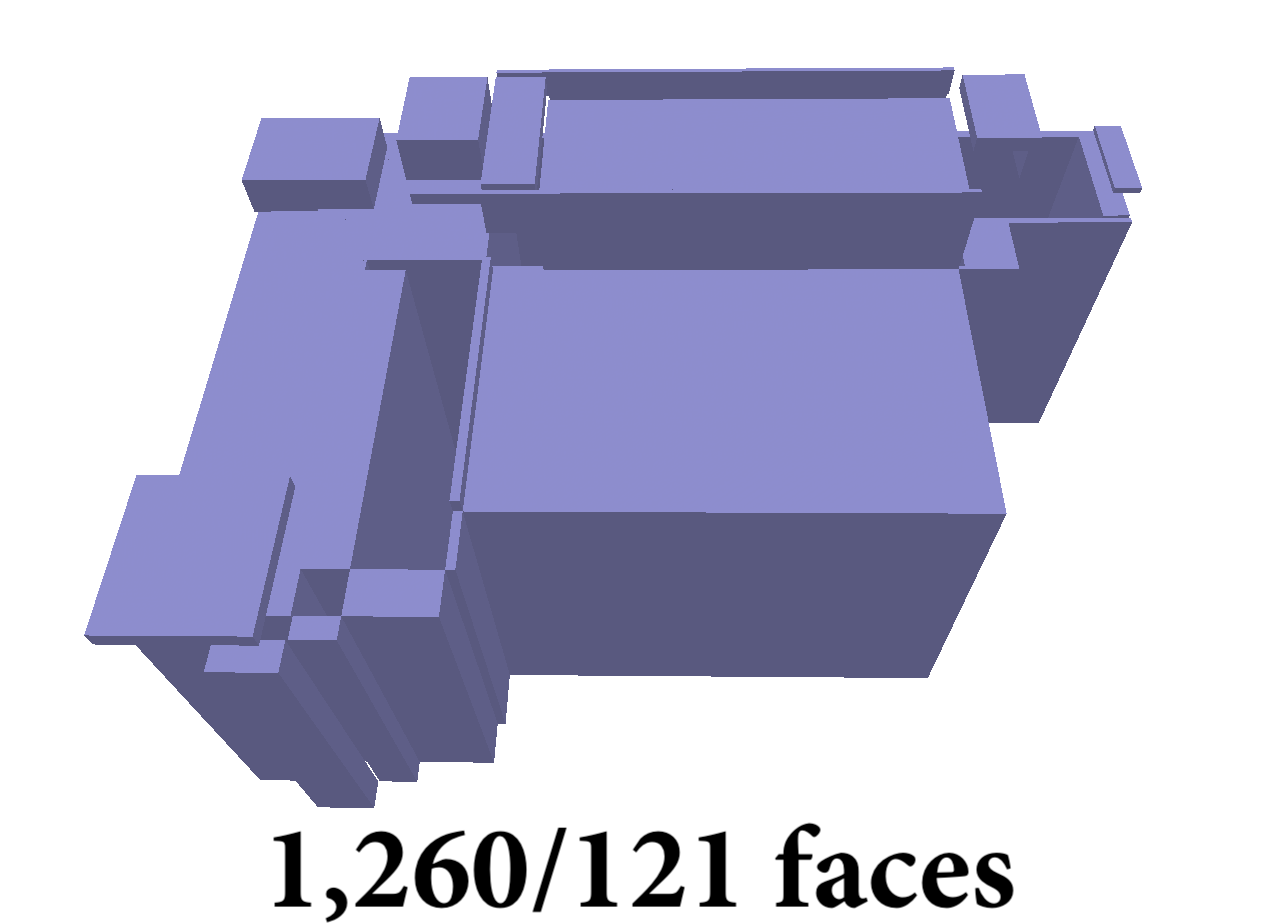}}
	\subfloat[2.5D DC]{\includegraphics[width=0.16\linewidth]{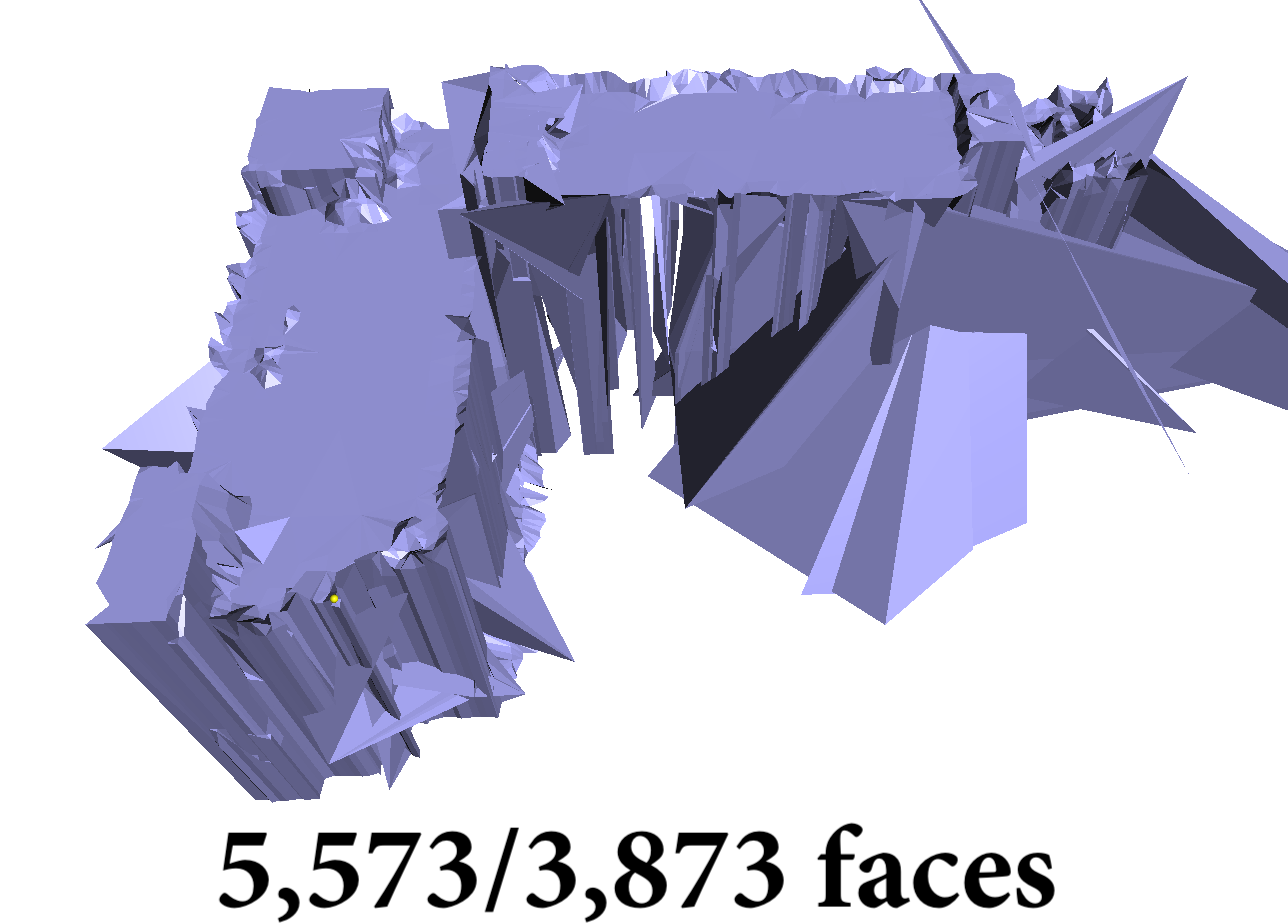}}
	\subfloat[PolyFit]{\includegraphics[width=0.16\linewidth]{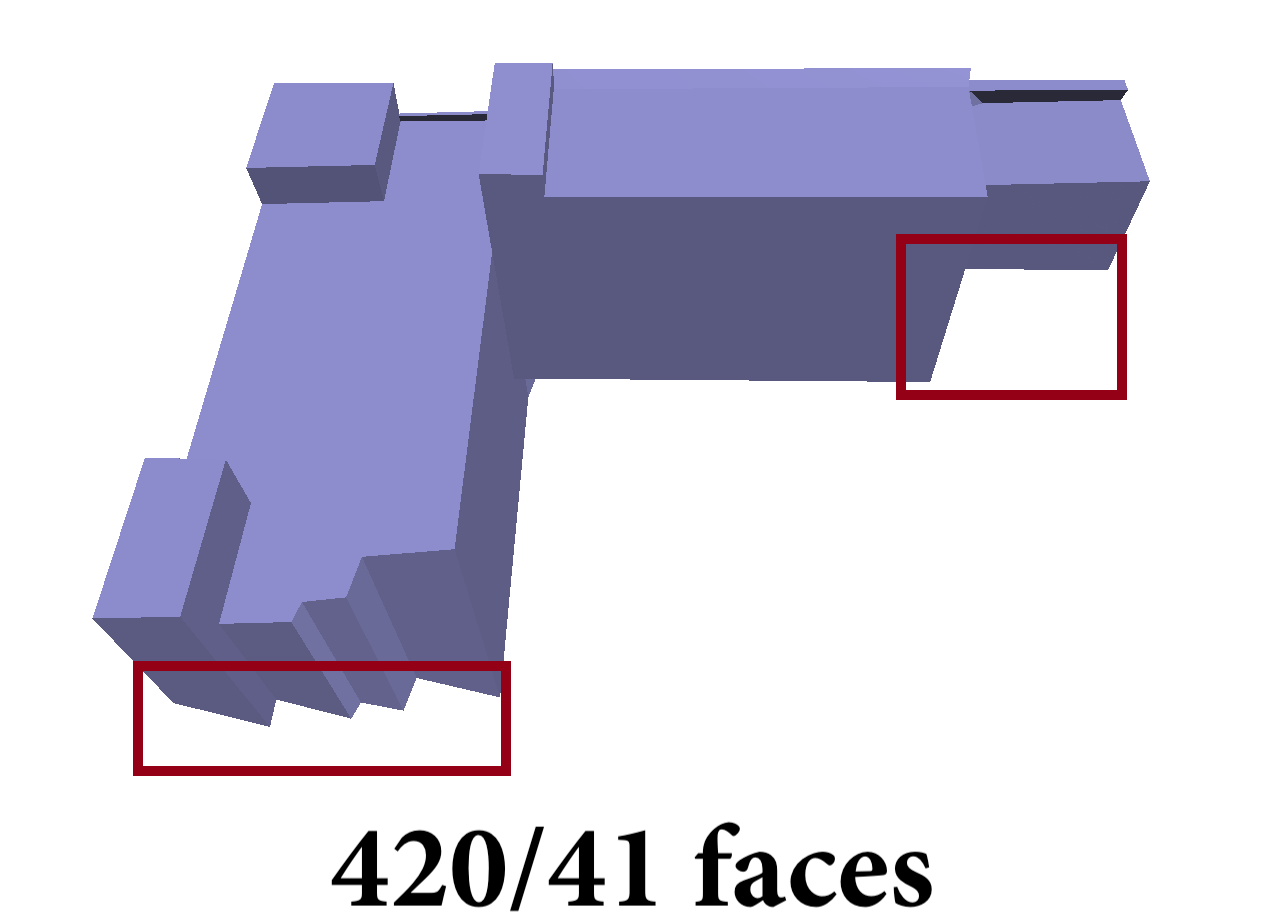}}
	\subfloat[KSR]{\includegraphics[width=0.16\linewidth]{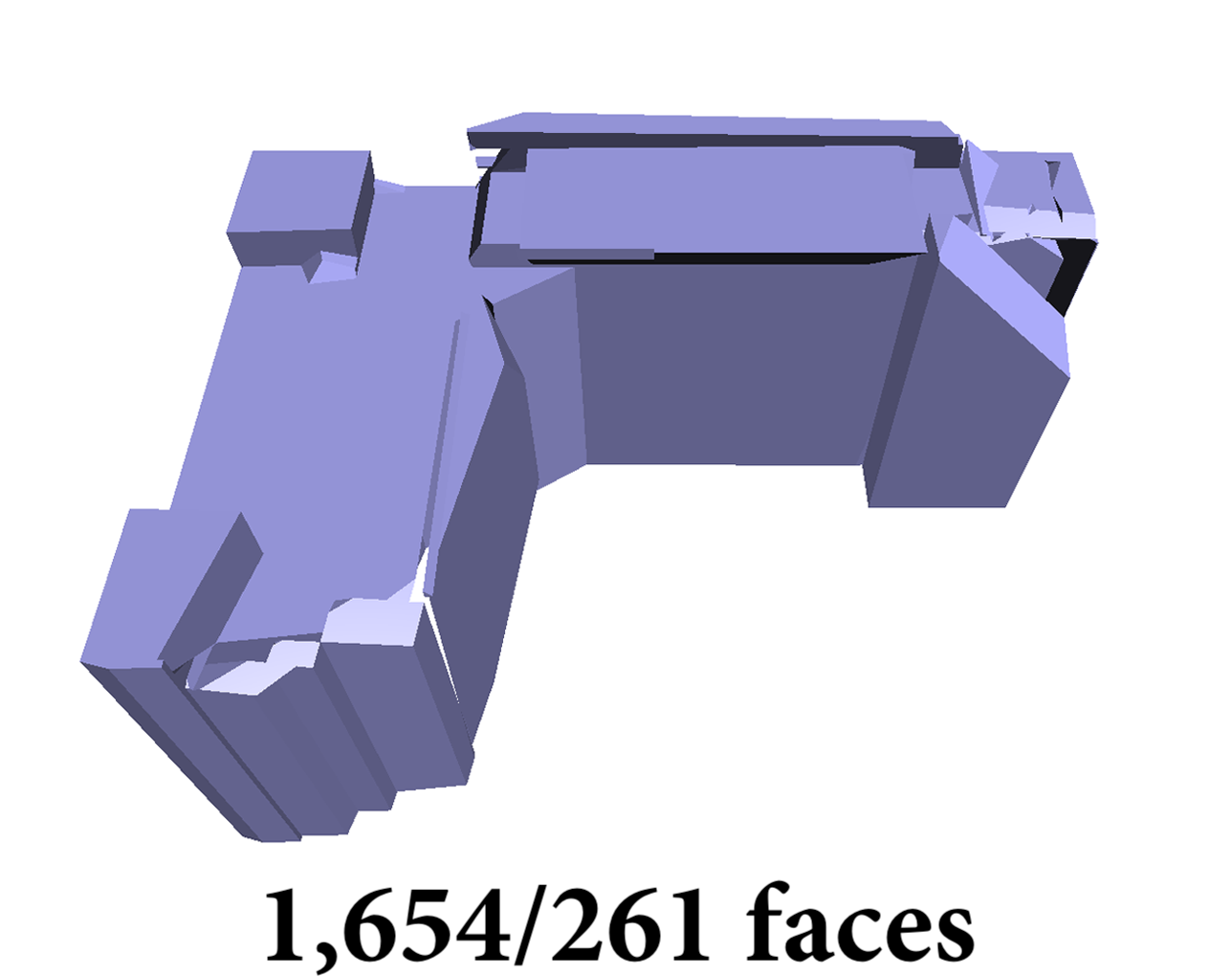}}	
	\subfloat[Ours]{\includegraphics[width=0.16\linewidth]{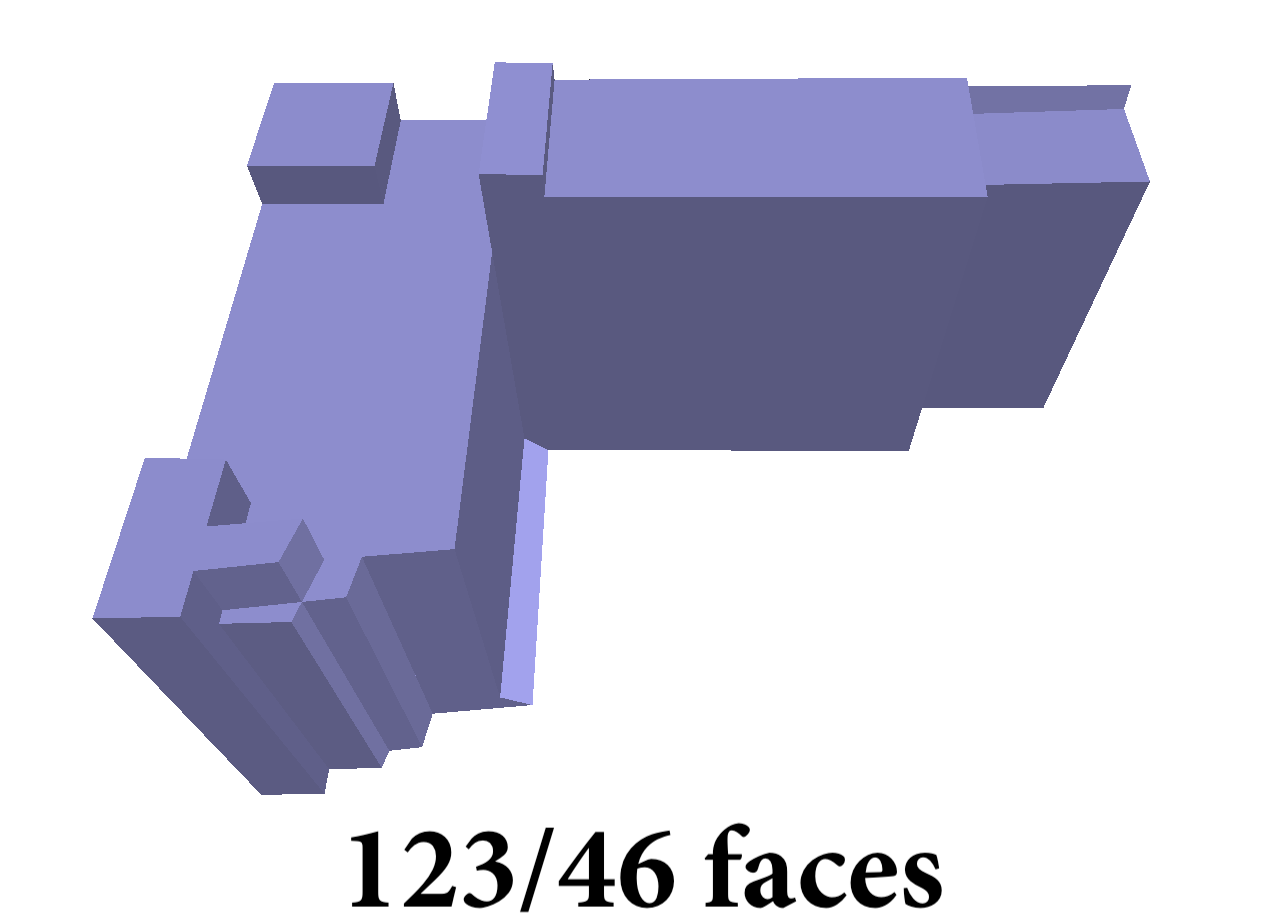}}
	
	\caption{Comparison between our method and two model-based methods, namely, Manhattan-world building reconstruction~\citep{li2016manhattan} and 2.5D Dual Contouring~\citep{zhou20102}, and two primitive assembly based methods PolyFit~\citep{nan2017polyfit} and KSR~\citep{bauchet2020kinetic}. The building in the top row is from the \textit{Helsinki full-view} dataset and the one in the bottom row is from the \textit{Shenzhen} dataset. Two sets of face numbers $(\bullet / \bullet)$ are given, which are the as-is face number from the algorithm and the face number after merging coplanar faces, respectively.
	}
	\label{fig:comparison_reconstruction}
\end{figure*}

In \autoref{fig:comparison_reconstruction}, we demonstrate the comparison of our results with two model-based building reconstruction methods, namely, Manhattan-world reconstruction~\citep{li2016manhattan} and 2.5D Dual Contouring~\citep{zhou20102}, and two primitive assembly methods PolyFit~\citep{nan2017polyfit} and KSR \citep{bauchet2020kinetic}, on two buildings from the \textit{Helsinki full-view} and \textit{Shenzhen} datasets, respectively.
From the results, we can see that neither Manhattan-world reconstruction nor 2.5D Dual Contouring can deliver visually plausible building models. 
The former constrains faces to be axis-aligned, and it thus cannot faithfully recover planar faces of arbitrary orientation, while the latter produces a prohibitive number of faces that do not properly address the piecewise planarity of urban buildings but with undesirable discontinuity. 
Among these methods, the results from PolyFit and KSR are most comparable with ours in terms of compactness.
PolyFit, KSR, and our method fall in the \emph{hypothesizing-and-selection} strategy. Specifically, the three methods tessellate the ambient space into a candidate set with detected planar primitives and then seek a proper arrangement of the candidates with combinatorial analysis. 
The optimization goal in PolyFit is hard coded by the weights for data fitting, surface coverage, and model complexity objectives. 
In contrast, our optimization is essentially guided by the implicit field estimated from the input point cloud using the neural network and further regularized by the MRF in surface extraction.
KSR's voting scheme requires the input point cloud to have high-quality normals associated, while our neural-guided approach enables directly consuming unorganized point clouds, which improves robustness on noisy or incomplete data.

On \textit{Helsinki fullview}, the quantitative analysis presented in \autoref{tab:quantitative} reveals a marginal fidelity advantage of our method while using much fewer faces to represent the buildings. Also, our method can scale to buildings with higher complexity (for those PolyFit fails).

Compared with the exhaustive partitioning strategy adopted by PolyFit, our adaptive space partitioning can produce building models with lower complexity. This can be concluded by comparing the number of faces in the reconstructed surface models from these methods. In fact, the many polygonal faces originating from the same plane can be merged into a single face by a post-processing step but we did not do so in our paper.
It is also worth noting that our adaptive space partitioning cannot be plugged into PolyFit because it would otherwise break PolyFit’s pre-condition that every edge is shared by four candidate faces (except for border edges).
Nevertheless, PolyFit performs pairwise intersection of all input planes and thus the partitioning by the vertical plane is always guaranteed because every pair of the support planes will intersect and result in four candidate faces. In this sense, verticality is not a concern for PolyFit.

\begin{figure}[ht]
	\centering
	\subfloat[Input]{\includegraphics[width=0.3\linewidth]{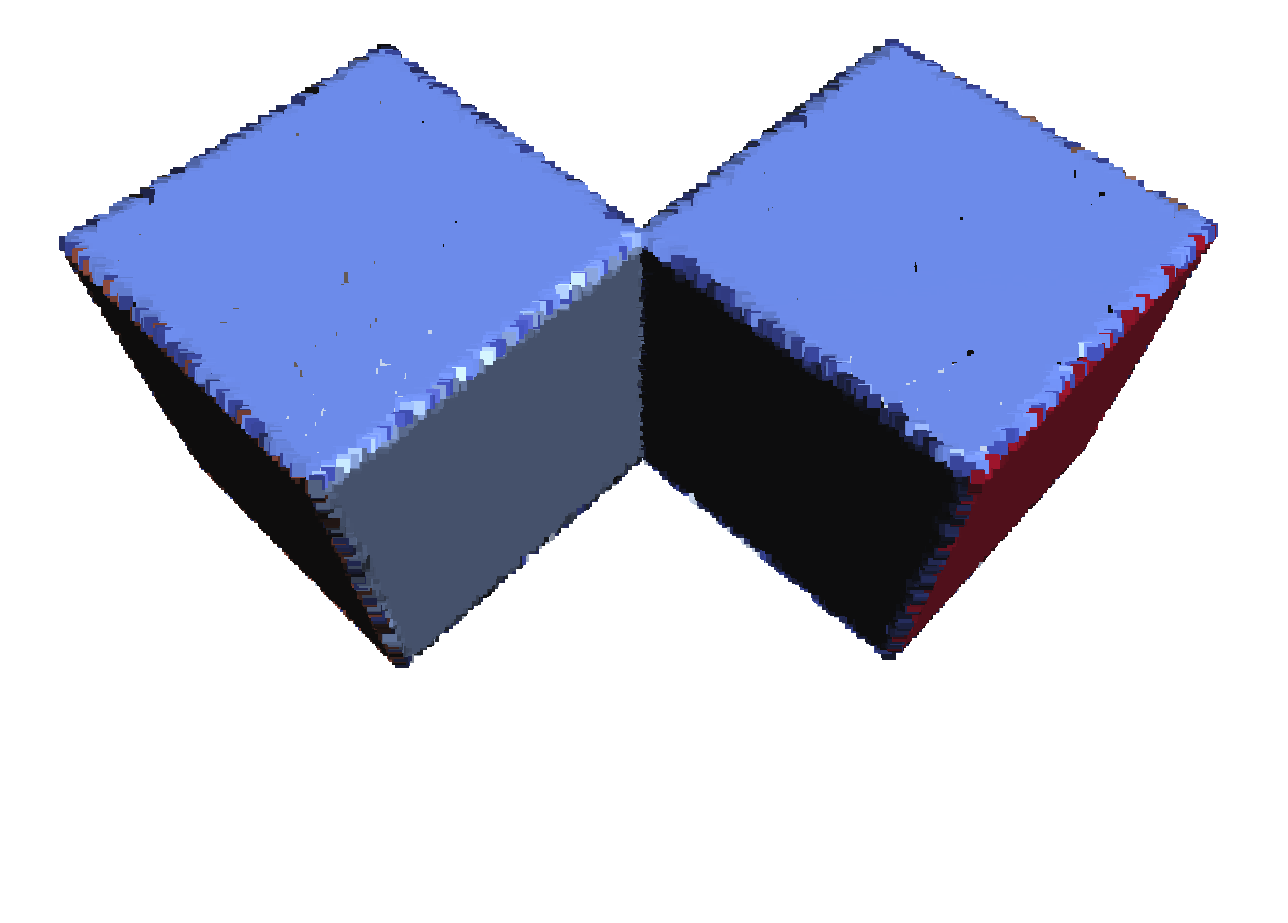}\label{fig:twocubes_input}}
	% \hspace{2em}
	\subfloat[PolyFit]{\includegraphics[width=0.28\linewidth]{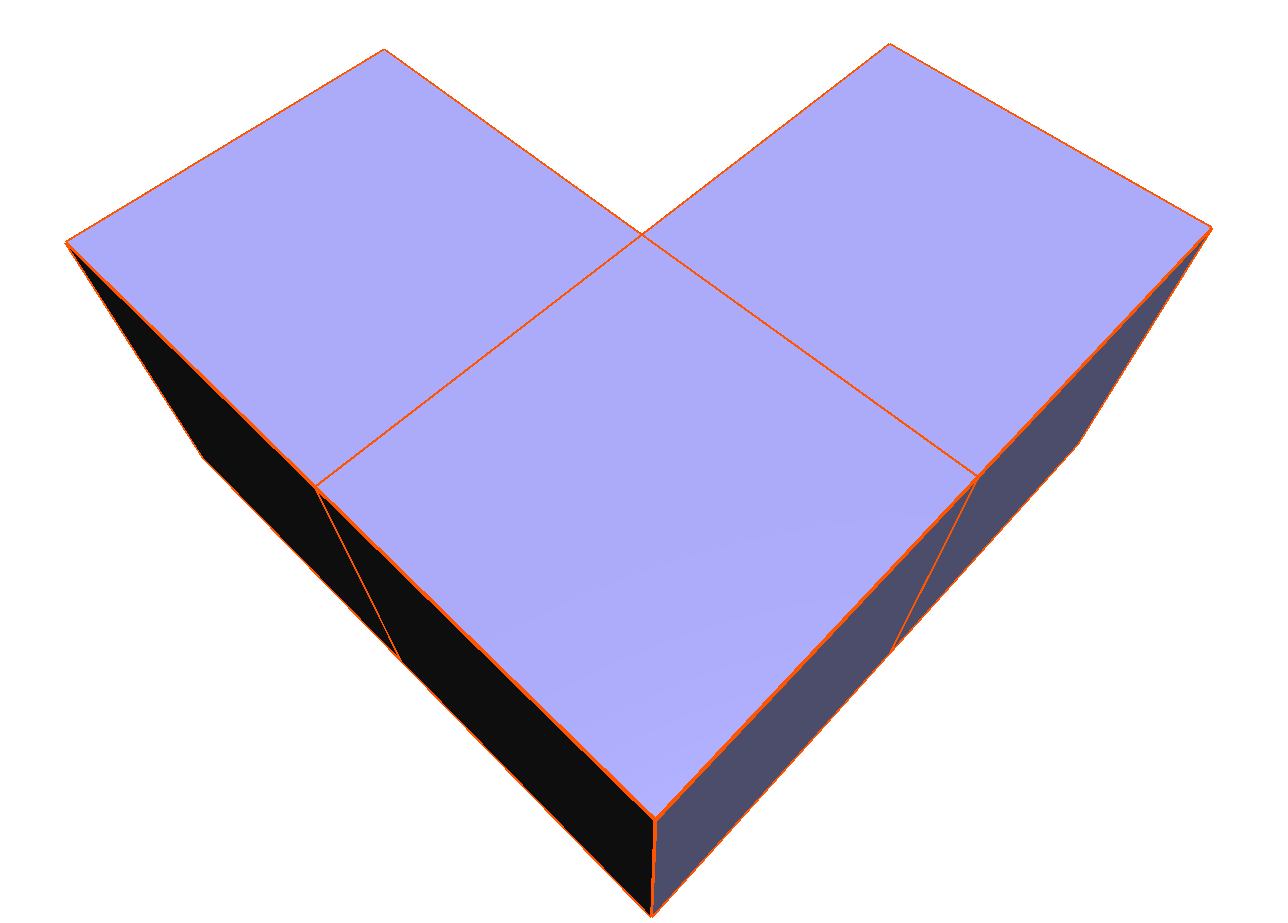}\label{fig:twocubes_polyfit}}
	% \hspace{2em}    
	\subfloat[Ours]{\includegraphics[width=0.3\linewidth]{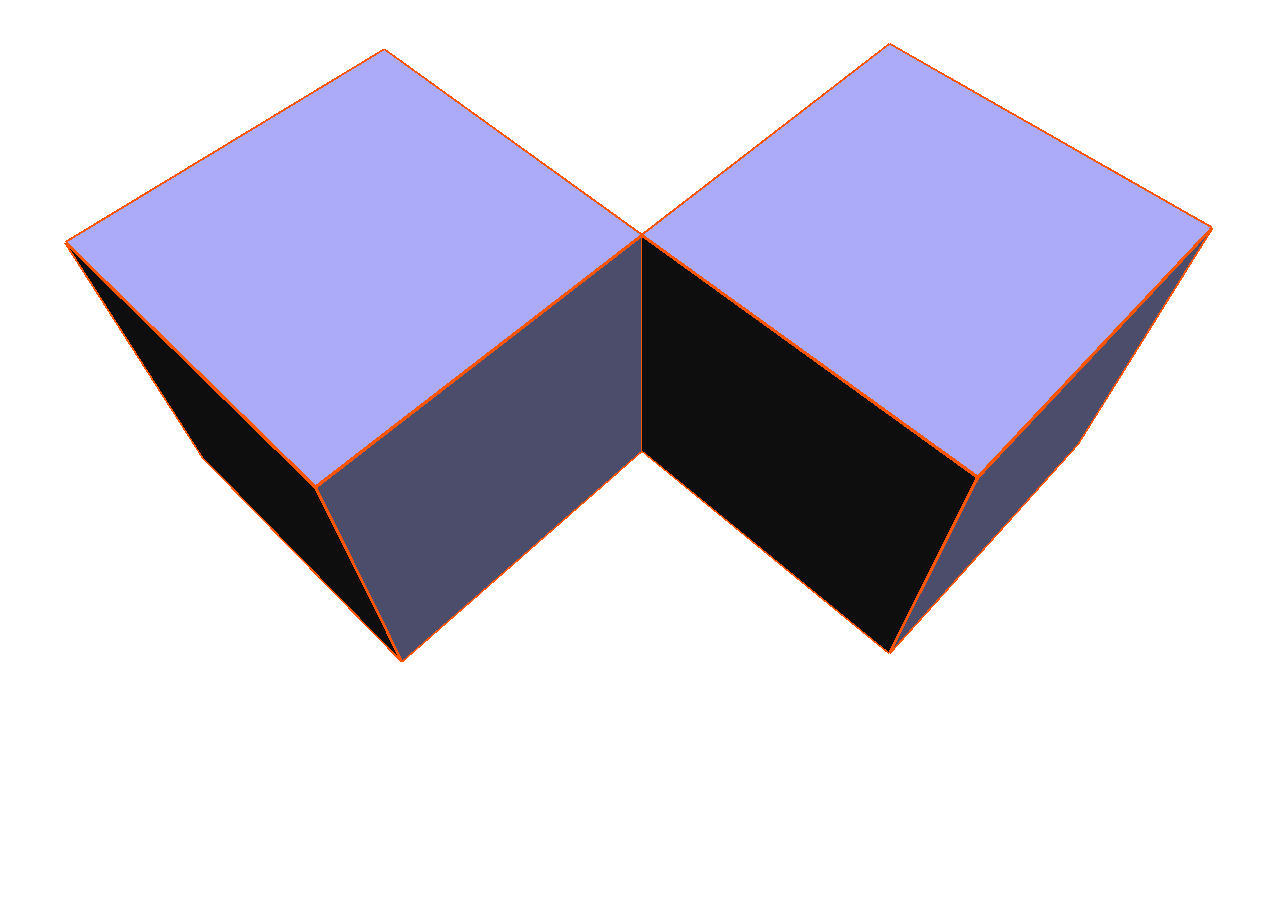}\label{fig:twocubes_ours}}    
	\caption{Comparison between PolyFit~\citep{nan2017polyfit} and our method on a building that demonstrates a non-manifold structure. Our method faithfully reconstructs the surface, while PolyFit impertinently enforces manifoldness, resulting in incorrect reconstruction.}
	\label{fig:twocubes}
\end{figure}

Our experiments reveal that introducing the vertical priority leads to slightly higher accuracy, i.e., 0.9\% less Hausdorff distance on average. We would like to point out that this priority has not been exploited in the previous work.
We have also observed in our experiments that non-manifold structures can significantly affect surface reconstruction, as exemplified by \autoref{fig:twocubes}.
Since our method constrains only water tightness, the reconstructed surface respects faithfully the geometry of the point cloud that exhibits a non-manifold structure, which PolyFit fails due to its manifoldness hard constraint.

\begin{figure*}[!t]
	\centering

	\raisebox{\height-6ex}{
	\includegraphics[width=0.18\linewidth]{figures/results/comparison/5.png}}
	\includegraphics[width=0.14\linewidth]{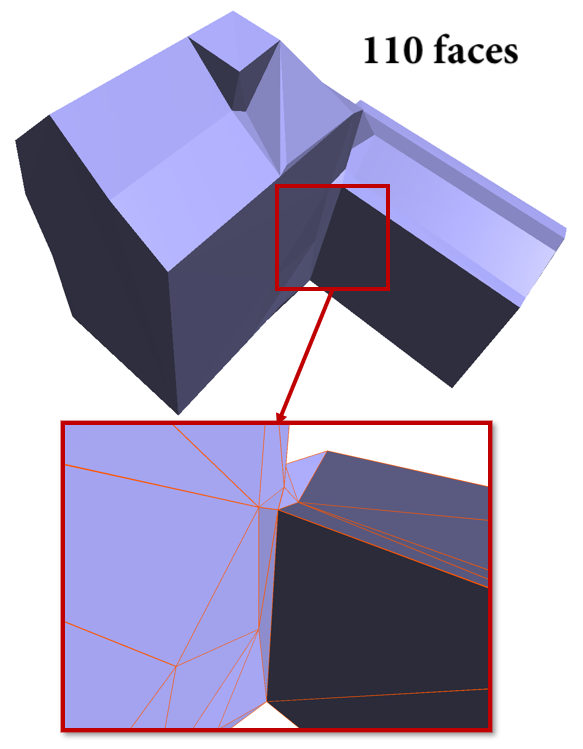}
	\includegraphics[width=0.14\linewidth]{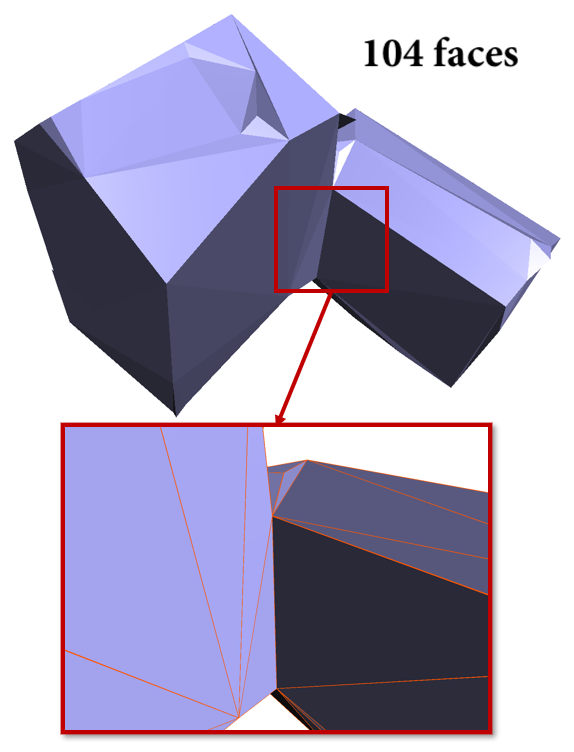}
	\includegraphics[width=0.14\linewidth]{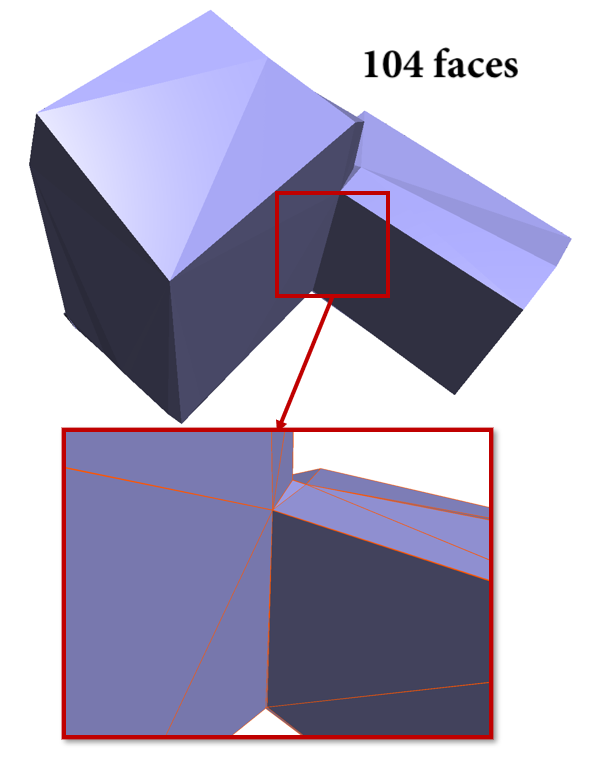}
	\includegraphics[width=0.145\linewidth]{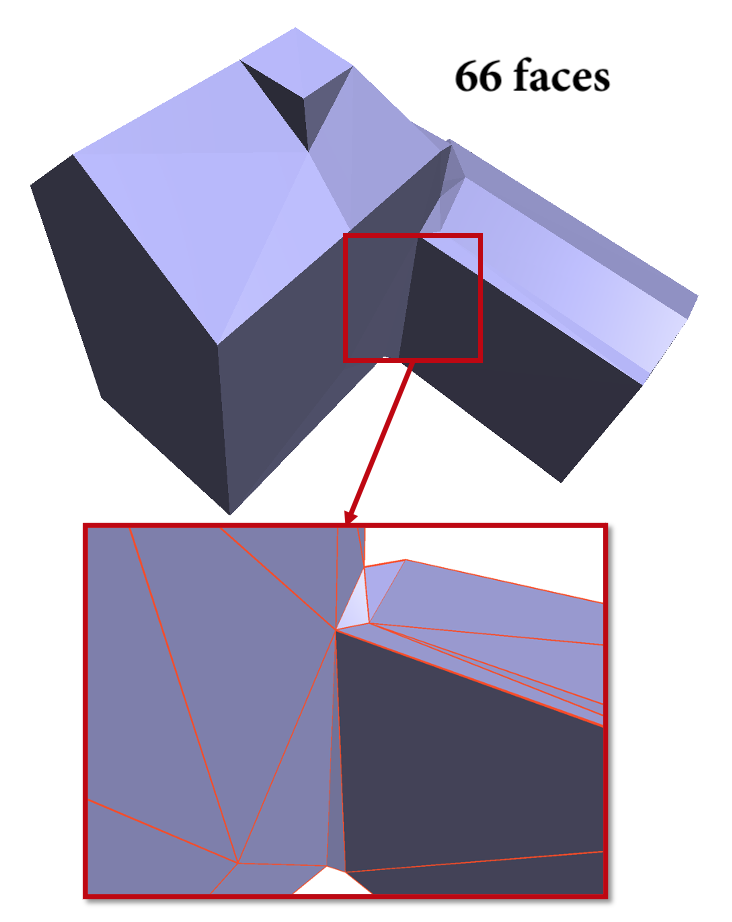}
	\includegraphics[width=0.14\linewidth]{figures/results/comparison/5_ours_with_detail.png}
	
	\subfloat[Input]{
	\centering
	\vtop{%
  	\vskip-80pt
  	\hbox{%
	\includegraphics[width=0.18\linewidth]{figures/results/comparison/Building_4.png}}}}
	\subfloat[QEM]{
	\includegraphics[width=0.14\linewidth]{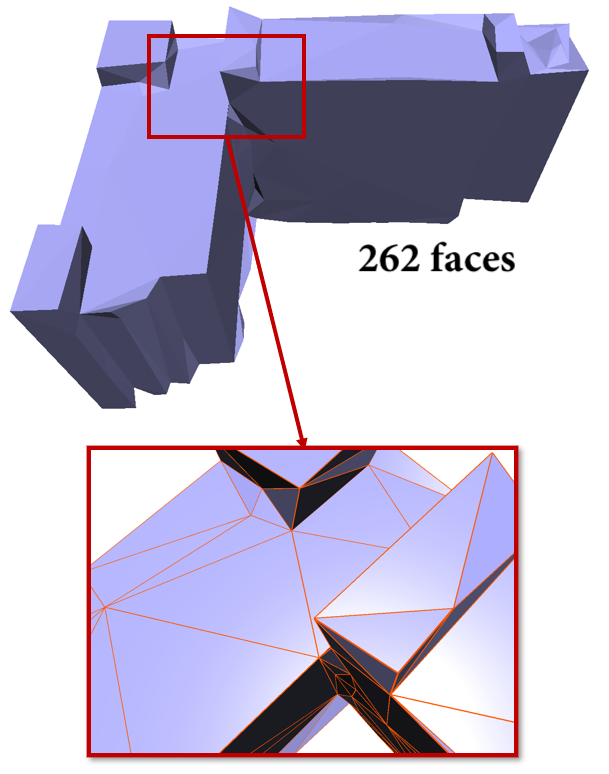}}
	\subfloat[SAMD]{
	\includegraphics[width=0.14\linewidth]{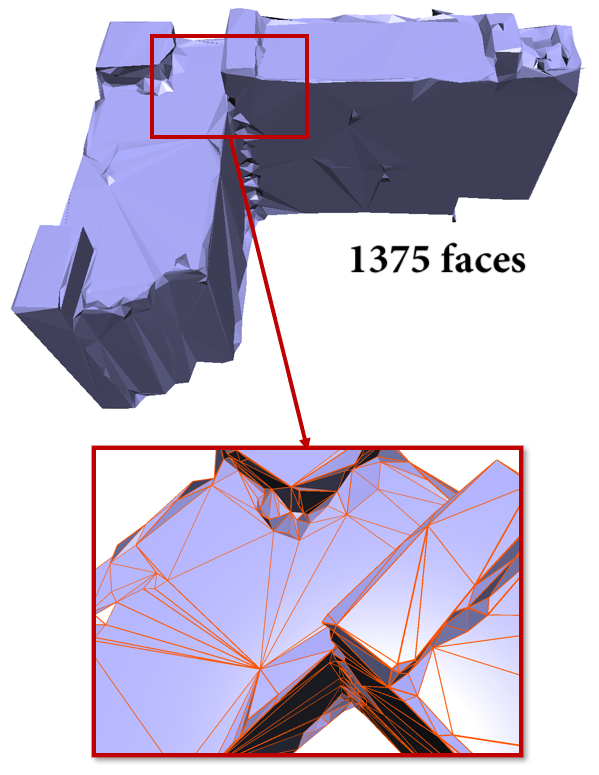}}
	\subfloat[VSA]{
	\includegraphics[width=0.14\linewidth]{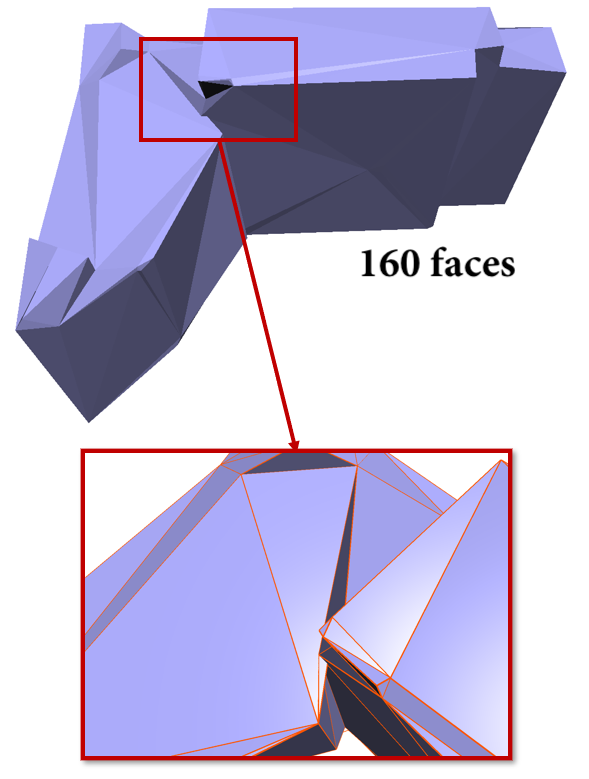}}
	\subfloat[FPS]{
	\includegraphics[width=0.146\linewidth]{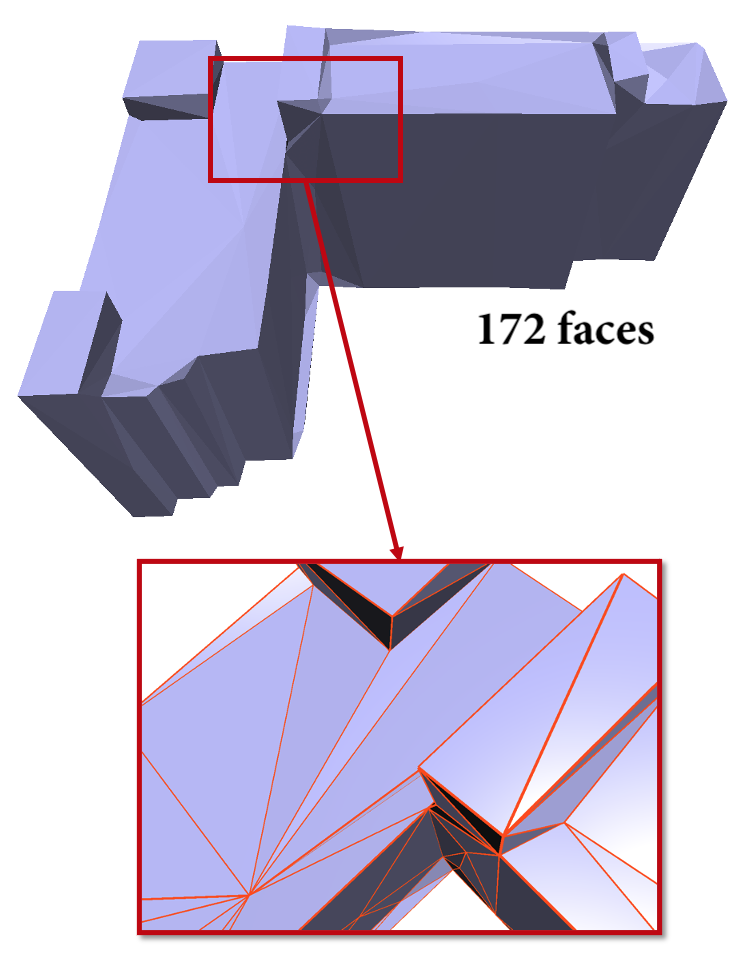}}
	\subfloat[Ours]{
	\includegraphics[width=0.14\linewidth]{figures/results/comparison/Building_4_ours_with_detail.png}}
	
	\caption{Comparison between the results of our method and three geometry simplification methods, namely, QEM~\citep{garland1997surface}, SAMD~\citep{salinas2015structure}, VSA~\citep{cohen2004variational}, and FPS~\citep{li2021feature}. Note that QEM, SAMD, VSA, and FPS all output triangle surfaces while our results are arbitrary-sided polygonal surfaces.}
	\label{fig:comparison_approximation}
\end{figure*}

In addition to the methods from the reconstruction category, we further compared our method with a few commonly used shape simplification methods, namely, QEM~\citep{garland1997surface}, SAMD~\citep{salinas2015structure}, VSA~\citep{cohen2004variational}, and FPS~\citep{li2021feature}.
For these methods, we set their expected number of vertices to be comparable to the models generated by our method, which ensures that the quality of the generated building models is compared at the same complexity level.
Notice that, however, the resulting number of faces is not guaranteed to be identical to ours because no such constraints can be asserted with the three surface approximation methods.
As can be seen from~\autoref{fig:comparison_approximation}, VSA produces surfaces with the worst quality given the same complexity. 
SAMD, though claiming to be structure-aware, fails to deliver surfaces with compact planes, partially because of its strong compliance to the detected primitives, which on the contrary limits its compactness when the primitives are of low quality.
QEM generates fractured surfaces because its quadric error metric contracts edges without attention to the piecewise planarity of the buildings. Nevertheless, the results from QEM are more plausible than VSA and SAMD.
FPS generates the most visually pleasing results among the four methods because of its capability of preserving piecewise-planar structures and sharp features. However, the results are not as regular as ours.
In contrast, our method produces surface models directly from point clouds, which recovers stronger piecewise planarity and meanwhile preserves sharp features.

% Please add the following required packages to your document preamble:
% \usepackage[table,xcdraw]{xcolor}
% If you use beamer only pass "xcolor=table" option, i.e. \documentclass[xcolor=table]{beamer}

\begin{table}[]
\centerline{
\tiny
%\rotatebox{90}{
% Please add the following required packages to your document preamble:
% \usepackage[table,xcdraw]{xcolor}
% If you use beamer only pass "xcolor=table" option, i.e. \documentclass[xcolor=table]{beamer}
\begin{tabular}{|ccccc|ccccc|}
\hline
\textbf{Index} & \textbf{Poisson} & \textbf{PolyFit}       & \textbf{KSR}            & \textbf{Ours}          & \textbf{Index} & \textbf{Poisson} & \textbf{PolyFit}       & \textbf{KSR}            & \textbf{Ours}          \\ \hline
1              & (0.1992, 417546) & \textbf{(0.2004, 33)}  & (0.2007, 60)            & (0.2013, 23)           & 24             & (0.1022, 245028) & -                      & (0.1056, 1382)          & \textbf{(0.0714, 458)} \\
2              & (0.0433, 63440)  & (0.0433, 269)          & (0.0435, 408)           & \textbf{(0.0433, 145)} & 25             & (0.0263, 67000)  & \textbf{(0.0167, 54)}  & (0.0167, 64)            & (0.0180, 41)           \\
3              & (0.0173, 178608) & \textbf{(0.0165, 42)}  & (0.0168, 60)            & (0.0177, 32)           & 26             & (0.1247, 216590) & (0.1247, 45)           & \textbf{(0.1246, 62)}   & (0.1247, 26)           \\
4              & (0.0481, 226278) & (0.2130, 60)           & (0.0483, 74)            & \textbf{(0.0478, 37)}  & 27             & (0.0828, 300112) & (0.0829, 68)           & (0.0830, 68)            & \textbf{(0.0829, 51)}  \\
5              & (0.0573, 163084) & (0.0422, 231)          & (0.0463, 180)           & \textbf{(0.0420, 75)}  & 28             & (0.0938, 215042) & -                      & \textbf{(0.0933, 610)}  & (0.1105, 304)          \\
6              & (0.1518, 405996) & -                      & \textbf{(0.1464, 832)}  & (0.2179, 249)          & 29             & (0.0797, 234884) & (0.1024, 56)           & \textbf{(0.0795, 74)}   & (0.2099, 53)           \\
7              & (0.1184, 302234) & \textbf{(0.1252, 442)} & (0.2016, 270)           & (0.1367, 336)          & 30             & (0.1564, 174826) & \textbf{(0.1573, 25)}  & (0.1574, 38)            & (0.1574, 14)           \\
8              & (0.0889, 133070) & \textbf{(0.0887, 318)} & (0.0888, 208)           & (0.0890, 164)          & 31             & (0.0454, 176860) & (0.0454, 127)          & (0.0464, 136)           & \textbf{(0.0452, 53)}  \\
9              & (0.0746, 277654) & (0.1563, 643)          & (0.2468, 358)           & \textbf{(0.0457, 148)} & 32             & (0.0396, 140286) & (0.0484, 220)          & \textbf{(0.0410, 174)}  & (0.0952, 64)           \\
10             & (0.0611, 198032) & -                      & \textbf{(0.0601, 1088)} & (0.0635, 349)          & 33             & (0.0267, 252528) & \textbf{(0.0518, 344)} & (0.0520, 232)           & (0.1956, 141)          \\
11             & (0.0390, 116550) & -                      & (0.1846, 442)           & \textbf{(0.0370, 137)} & 34             & (0.2245, 287140) & (0.2287, 180)          & (0.2220, 322)           & \textbf{(0.2203, 127)} \\
12             & (0.0622, 214442) & (0.0601, 816)          & (0.1899, 340)           & \textbf{(0.0599, 186)} & 35             & (0.0153, 155978) & (0.0306, 283)          & \textbf{(0.0297, 210)}  & (0.0302, 110)          \\
13             & (0.0167, 107360) & (0.0183, 66)           & (0.0170, 88)            & \textbf{(0.0167, 46)}  & 36             & (0.0472, 458574) & (0.0461, 37)           & (0.0460, 50)            & \textbf{(0.0457, 35)}  \\
14             & (0.1307, 398462) & (0.1313, 311)          & \textbf{(0.1316, 474)}  & (0.1434, 124)          & 37             & (0.0152, 145496) & -                      & \textbf{(0.0610, 6218)} & (0.0970, 918)          \\
15             & (0.0702, 181472) & -                      & (0.0714, 828)           & \textbf{(0.0697, 294)} & 38             & (0.2209, 143716) & (0.2222, 271)          & (0.2220, 222)           & \textbf{(0.2218, 105)} \\
16             & (0.0491, 96932)  & -                      & (0.0499, 706)           & \textbf{(0.0486, 329)} & 39             & (0.1427, 126976) & -                      & \textbf{(0.1402, 848)}  & (0.1445, 315)          \\
17             & (0.0723, 124446) & \textbf{(0.0668, 58)}  & (0.0673, 104)           & (0.0676, 48)           & 40             & (0.0828, 129036) & \textbf{(0.0718, 548)} & (0.0748, 362)           & (0.1849, 99)           \\
18             & (0.0937, 118946) & (0.0935, 25)           & \textbf{(0.0933, 40)}   & (0.0934, 20)           & 41             & (0.0325, 339120) & (0.0285, 71)           & (0.0297, 284)           & \textbf{(0.0280, 36)}  \\
19             & (0.0166, 157064) & \textbf{(0.0277, 90)}  & (0.0286, 124)           & (0.0278, 84)           & 42             & (0.0654, 78496)  & -                      & \textbf{(0.0662, 626)}  & (0.0731, 308)          \\
20             & (0.0845, 148874) & -                      & (0.0822, 300)           & \textbf{(0.0821, 157)} & 43             & (0.0787, 107480) & (0.0785, 64)           & (0.0800, 64)            & \textbf{(0.0780, 35)}  \\
21             & (0.0196, 93014)  & \textbf{(0.0688, 76)}  & (0.0172, 104)           & (0.1456, 44)           & 44             & (0.0164, 168376) & (0.1454, 65)           & \textbf{(0.0187, 64)}   & (0.1631, 37)           \\
22             & (0.0193, 245844) & (0.0216, 852)          & (0.0254, 576)           & \textbf{(0.0213, 228)} & 45             & (0.0432, 104310) & \textbf{(0.0449, 158)} & (0.0453, 150)           & (0.0457, 47)           \\
23             & (0.1023, 160638) & -                      & \textbf{(0.1019, 888)}  & (0.1025, 292)          &                &                  &                        &                         &                        \\ \hline
\end{tabular}
}
	\caption{Quantitative evaluation among Poisson~\citep{kazhdan2006poisson}, PolyFit~\citep{nan2017polyfit}, KSR~\citep{bauchet2020kinetic}, and ours. $(\bullet, \bullet)$ denotes (Hausdorff distance, number of faces of a model from each original algorithm).
	``-'' indicates failure in obtaining the result within 3-hour running of an algorithm.}
\label{tab:quantitative}
\end{table}

\subsubsection{Computational efficiency}

\begin{figure}[!t]
	\centering
	\subfloat[]{\includegraphics[width=0.52\linewidth]{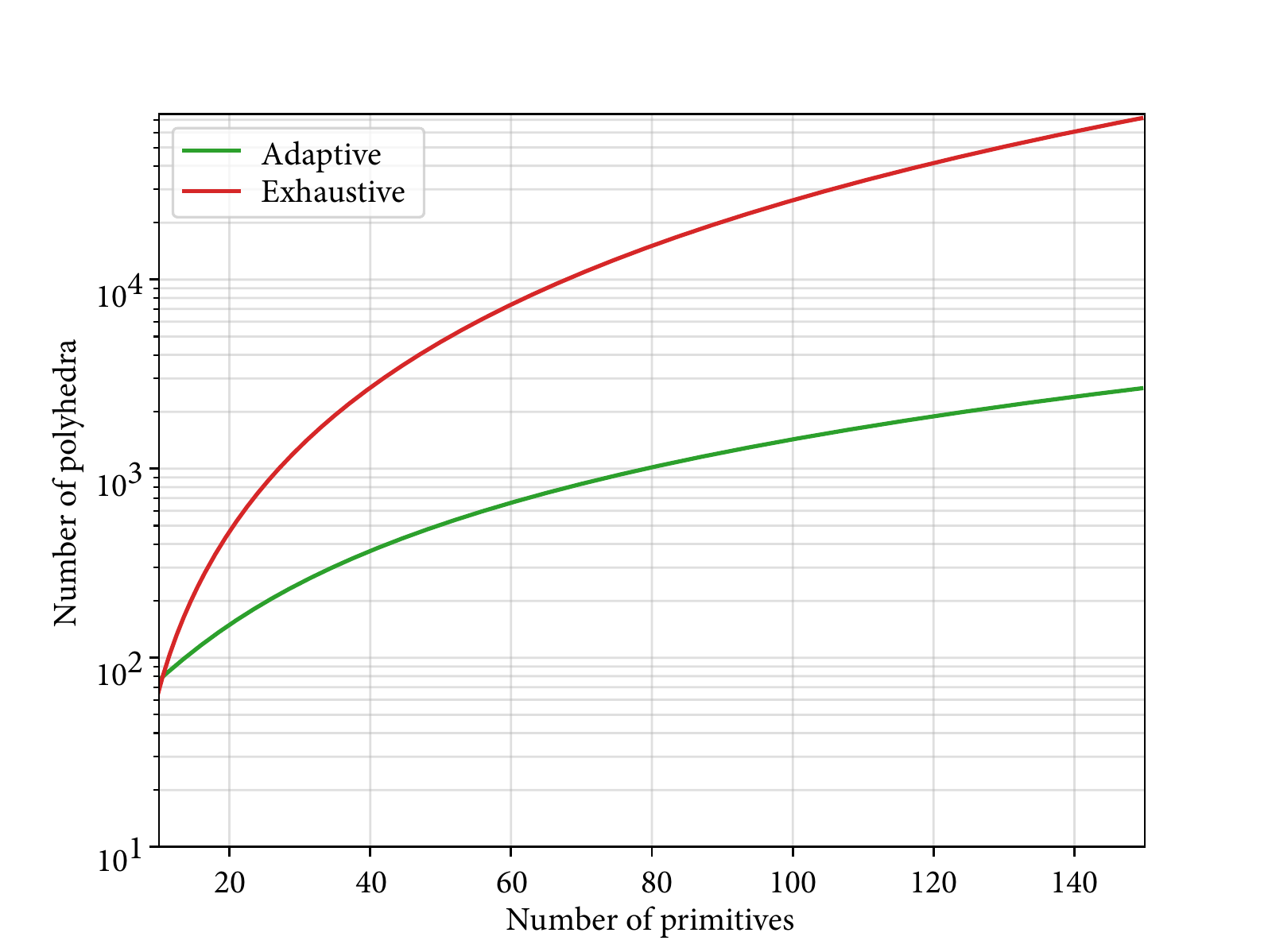}}
	\hspace{-2em}
	\subfloat[]{\includegraphics[width=0.52\linewidth]{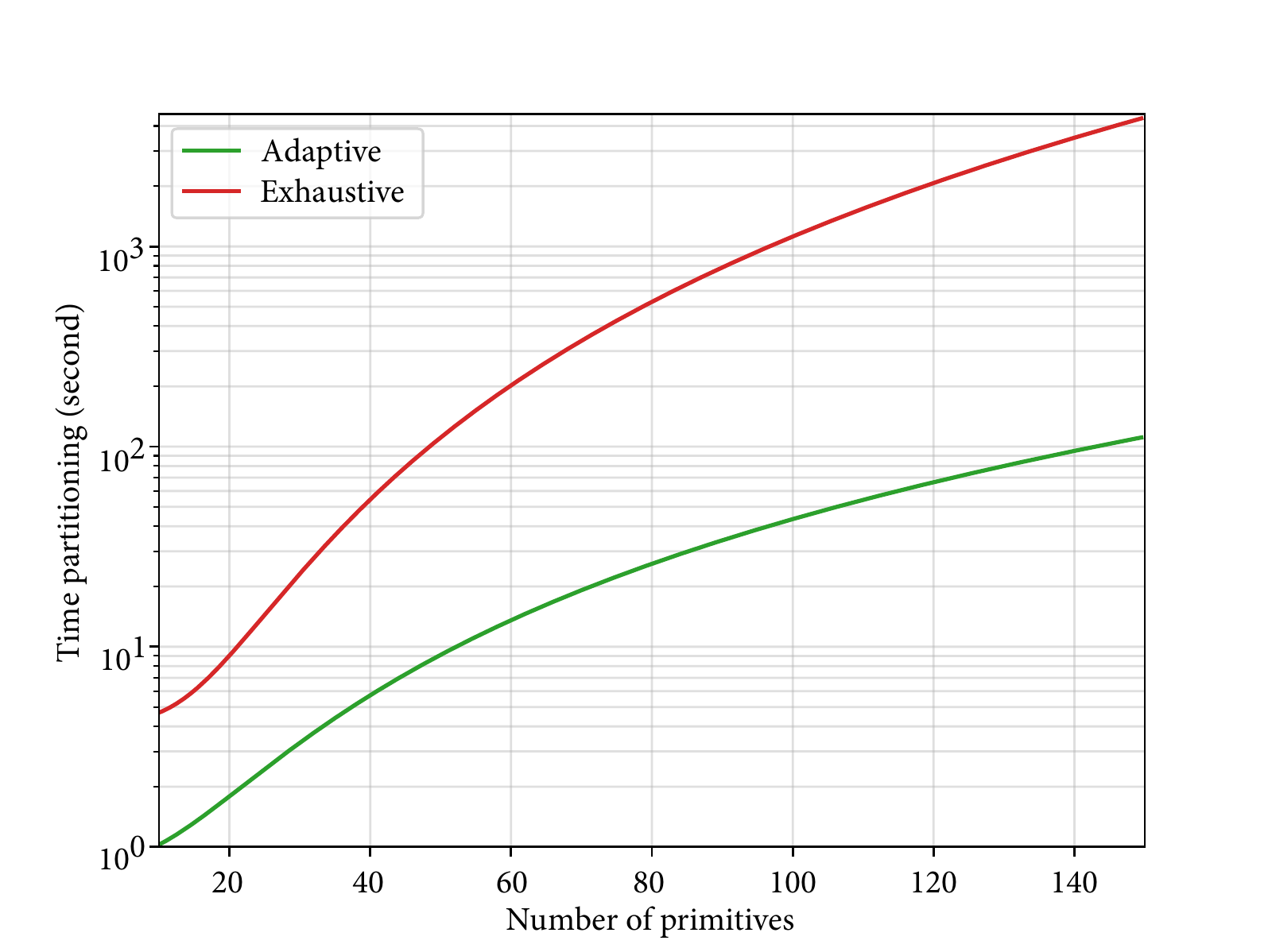}}
	\caption{Comparison between our adaptive space partitioning and exhaustive partitioning in terms of the number of resulting cells (a) and the running time for partitioning (b).}
	\label{fig:computationalefficiency}
\end{figure}

Our adaptive space partitioning can significantly reduce the computations for cell complex creation, compared to an exhaustive partitioning strategy.
With exhaustive partitioning, a massive number of candidate polyhedra are produced and the running time increases accordingly. A comparison between our adaptive space partitioning and the exhaustive partitioning in terms of the number of resulting cells and the running time is presented in~\autoref{fig:computationalefficiency}. The excessive number of cells not only hinders computation but also inclines to defective surfaces on subtle structures where inaccurate labels are more likely to be assigned. 
Instead, the adaptive strategy avoids redundant partitioning and thus produces compact surfaces efficiently.

\begin{figure}[!t]
	\centering
	\subfloat[Performances of reconstruction. The complex building with 260 planar primitives is reconstructed by KSR and ours in 480 seconds and 150 seconds, respectively. Statistics were drawn from ten models with various complexity.]{\includegraphics[width=0.85\linewidth]{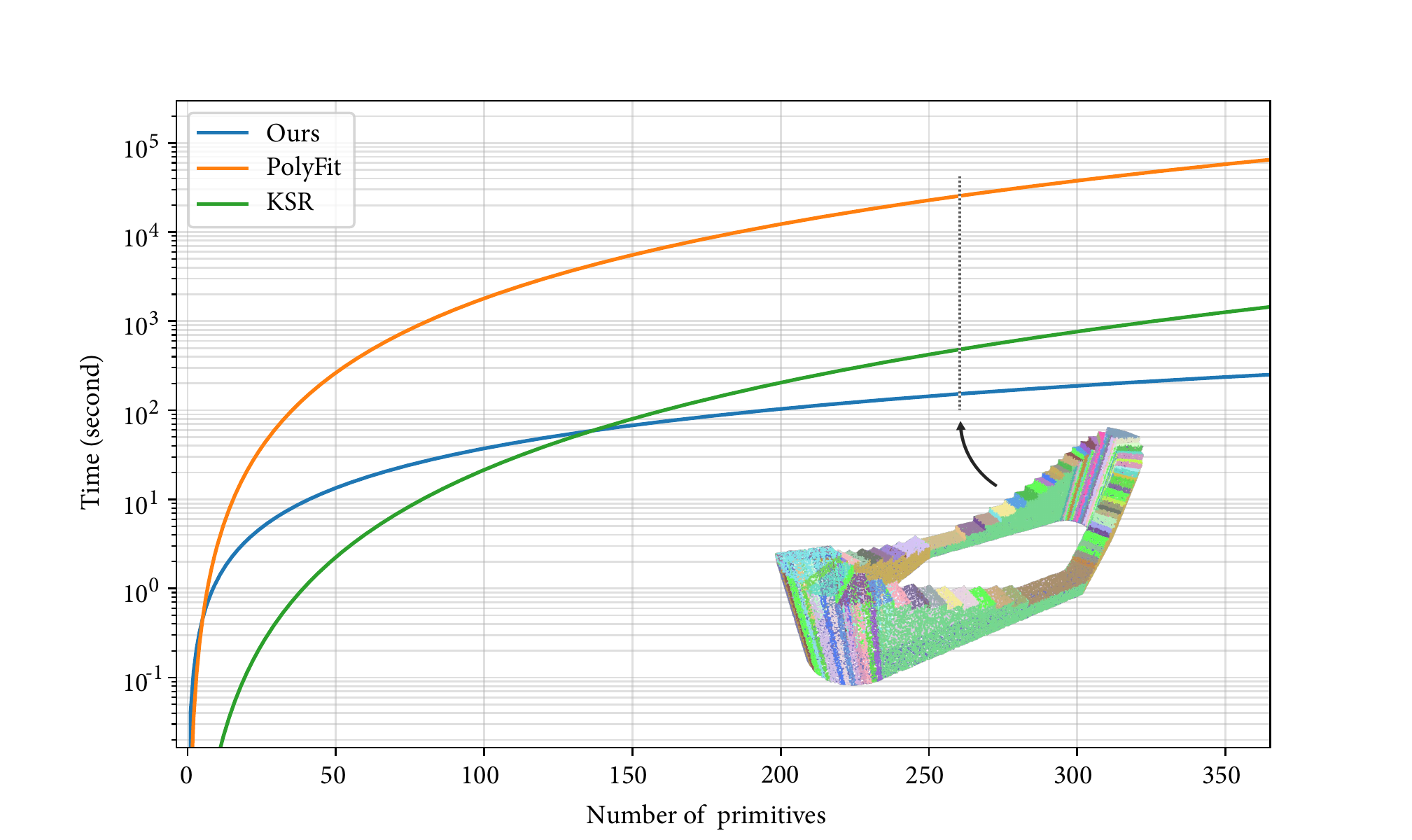}}

	\subfloat[PolyFit generates 762,439 candidate faces and fails to solve the resulting integer program.]{\includegraphics[width=0.29\linewidth]{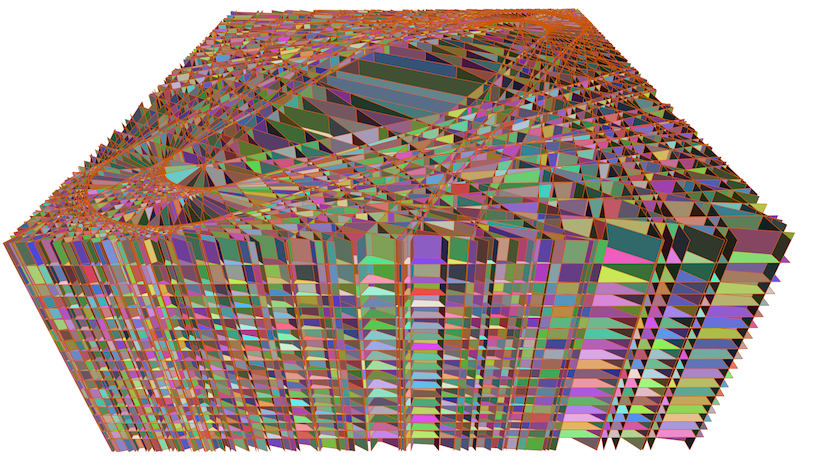}}
	\hspace{1em}
	\subfloat[Our method generates 2,856 candidate cells and produces a faithful reconstruction.]{\includegraphics[width=0.29\linewidth]{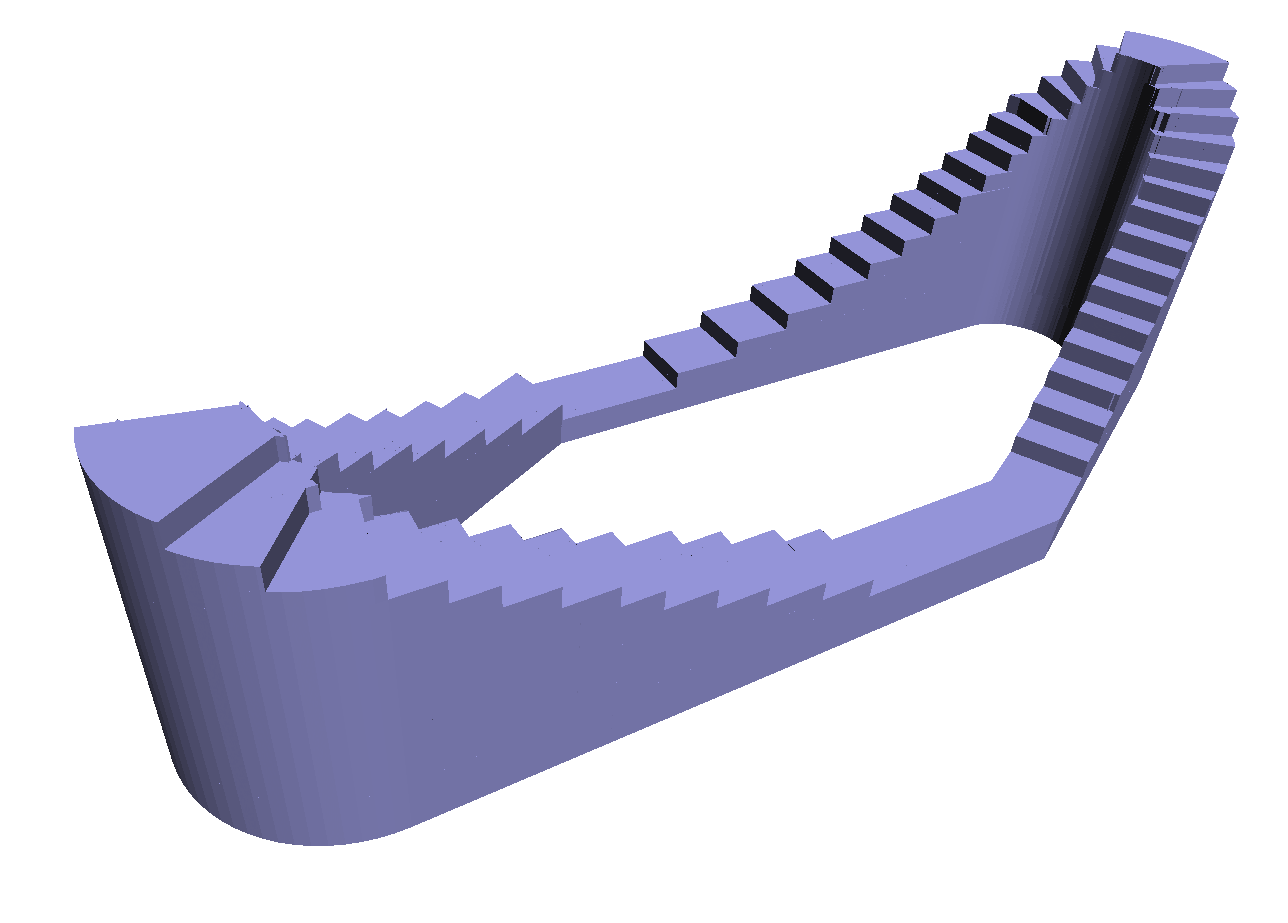}}
	\caption[Scalability evaluation]{Scalability comparison between our method, PolyFit~\citep{nan2017polyfit}, and KSR~\citep{bauchet2020kinetic}.}
	\label{fig:scalability}
\end{figure}

We further compared our method with PolyFit~\citep{nan2017polyfit} and KSR~\citep{bauchet2020kinetic} in terms of scalability. The results are shown in~\autoref{fig:scalability}.
PolyFit is based on the exhaustive partitioning strategy. Specifically, with pairwise intersection from the initial set of planar primitives, PolyFit generates a prohibitively large number of candidate faces, from which an optimal subset of the faces is to be selected using optimization based on its linear integer programming.
In our experiments, PolyFit failed to reconstruct the surface when the number of planar primitives exceeds 100. In contrast, our adaptive binary space partitioning resulted in a significantly smaller number of cells given the same number of input planar primitives, which allows our method to process input point clouds with more than one thousand planar primitives.
For this building with 260 detected primitives, PolyFit generated 762,439 candidate faces from the planar primitives and failed to solve the resulted large linear integer program.
In contrast, our method successfully produced a high-quality surface model in 150 seconds, of which the combinatorial optimization takes only 0.13 seconds.
KSR has higher efficiency than ours when the number of input primitives is smaller than 150. When scaling to inputs of higher complexity, our method excels since the employed adaptive strategy implies a strong locality, as opposed to the globally kinetic structure in KSR. It is also worth noting the potential implementation-wise performance gap, to which the computation overhead may be ascribed especially with a lower number of input primitives\footnote{KSR and PolyFit are implemented in C++, while ours is in Python.}.

\subsubsection{Robustness to noise and incomplete input}

\begin{figure}[!t]
	\centering
	\includegraphics[width=0.18\linewidth]{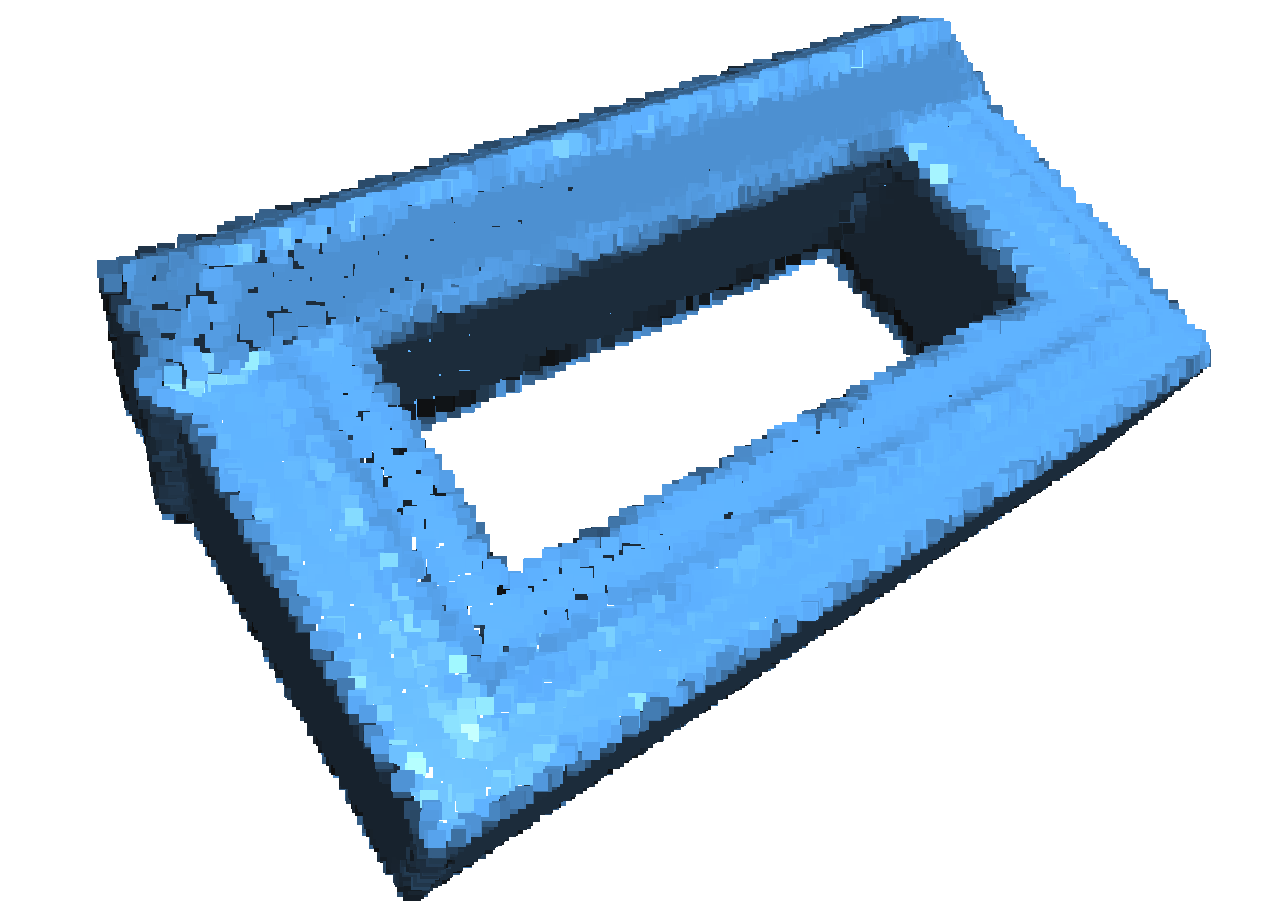}
	\includegraphics[width=0.18\linewidth]{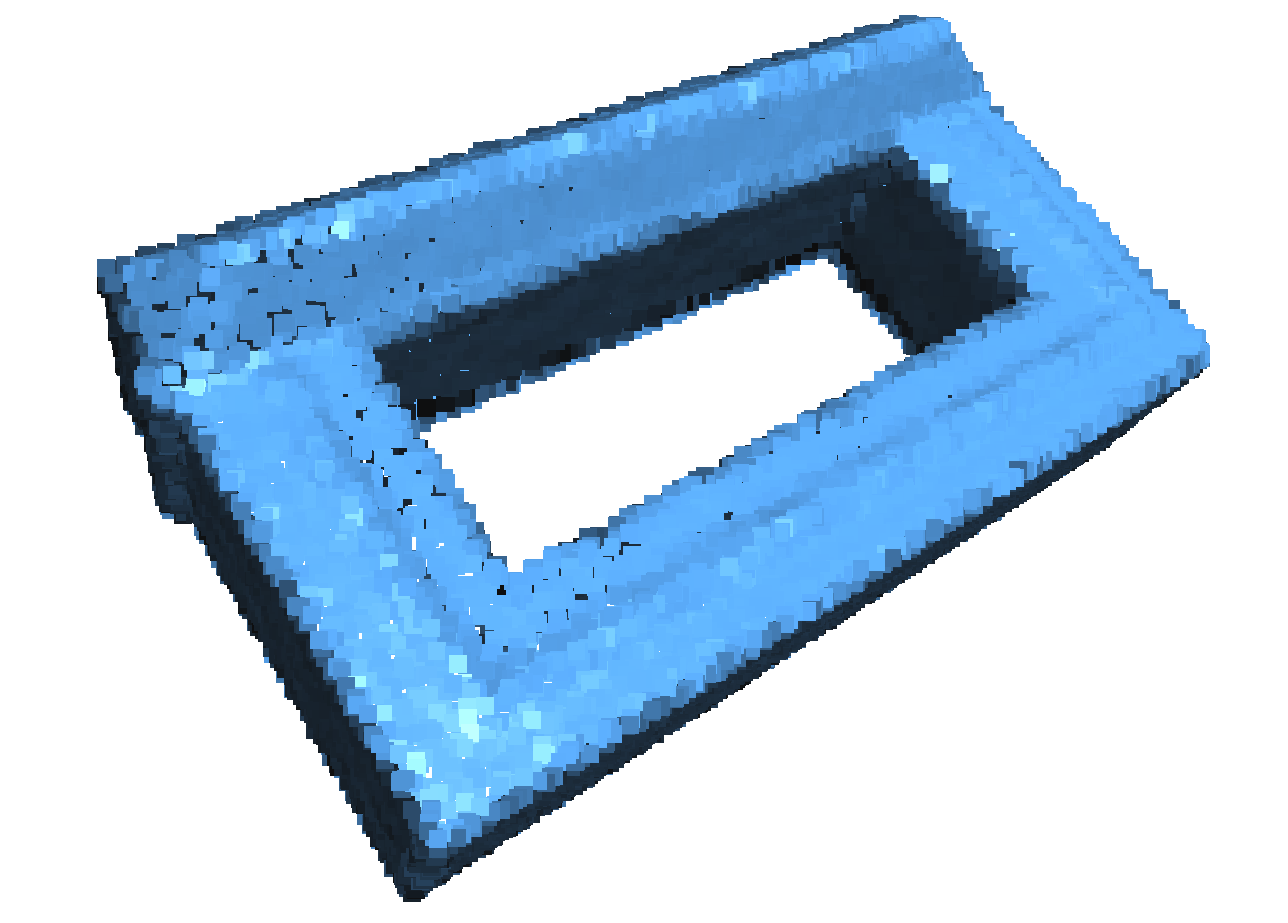}
	\includegraphics[width=0.18\linewidth]{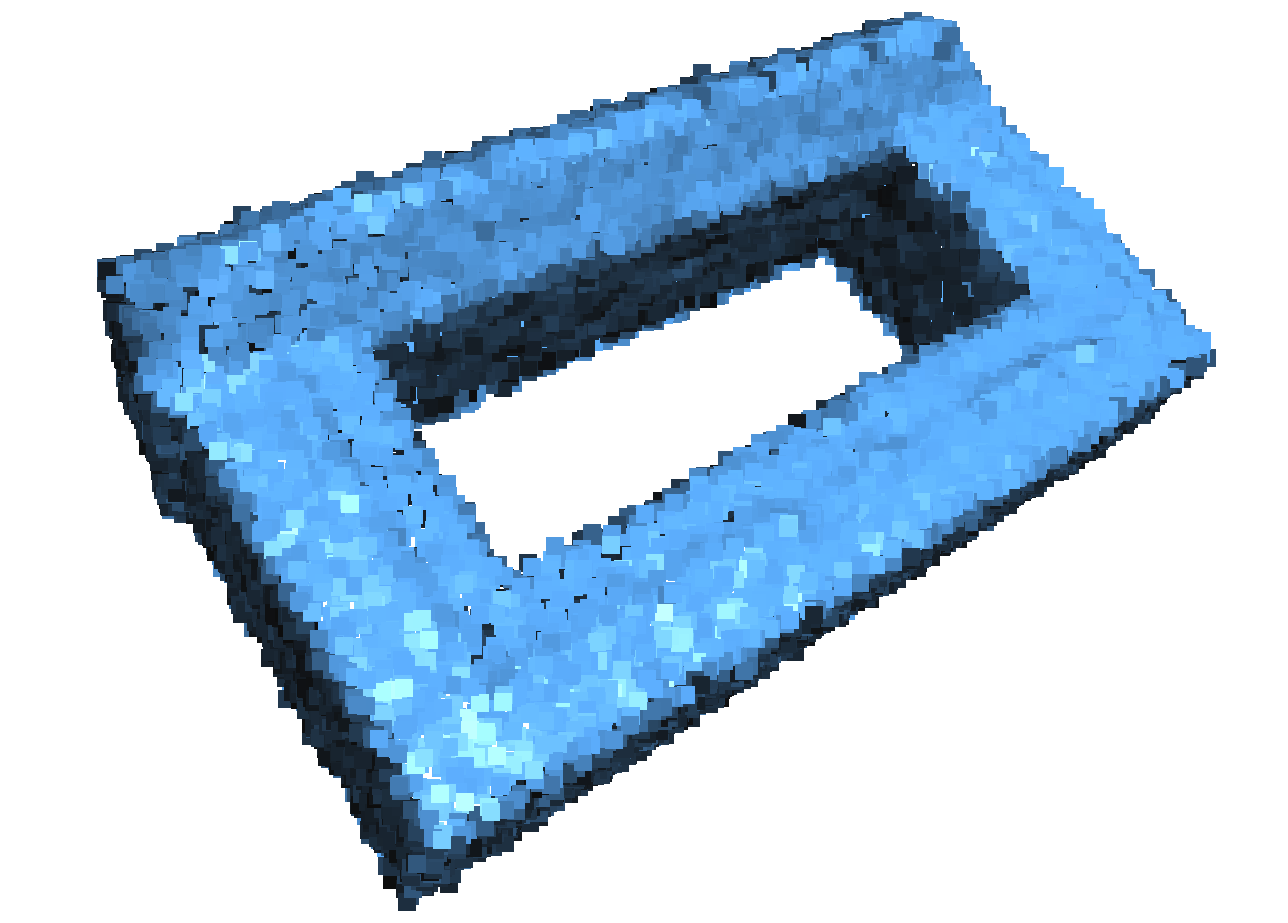}
	\includegraphics[width=0.18\linewidth]{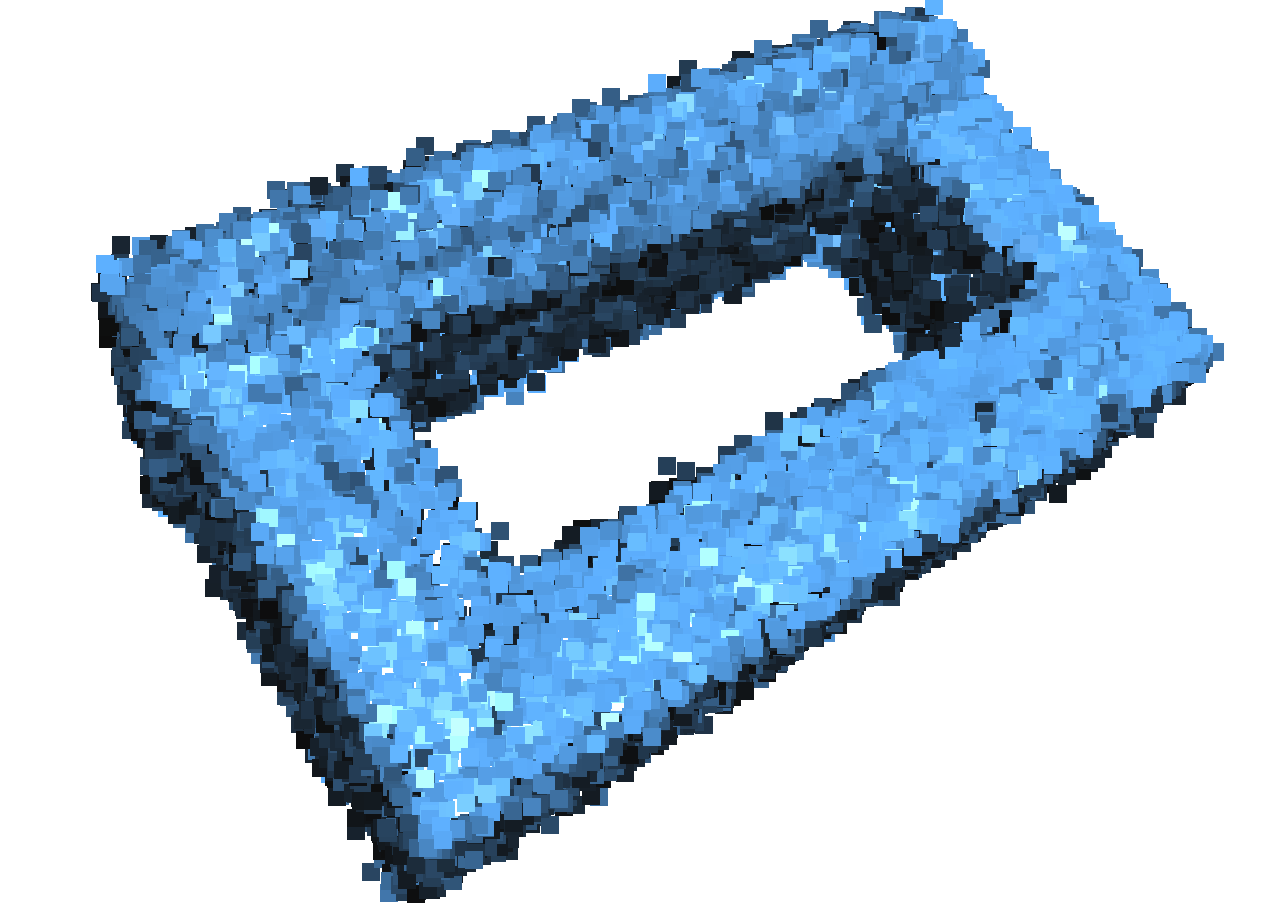}
	\includegraphics[width=0.18\linewidth]{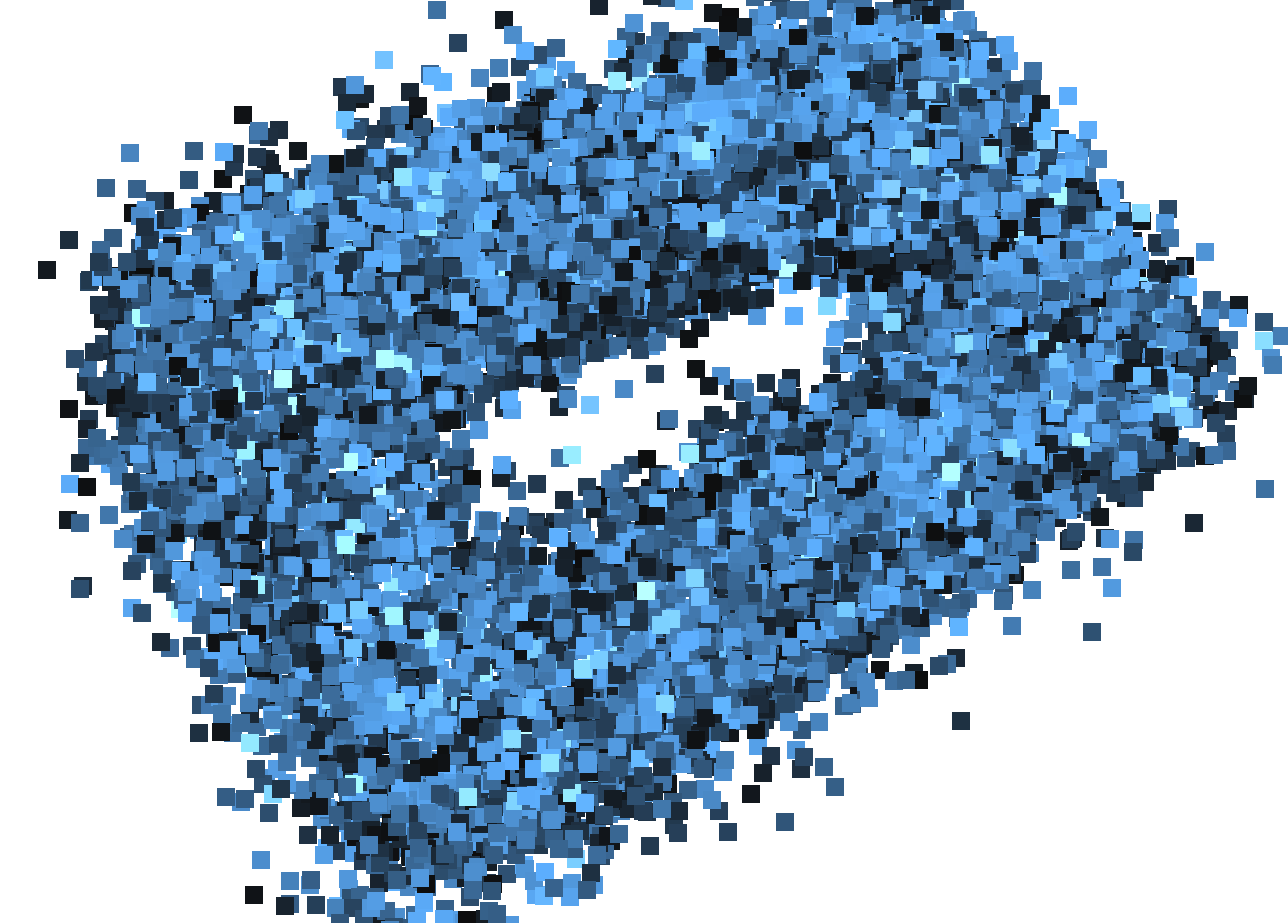}
	
	\subfloat[$0R$]{\includegraphics[width=0.18\linewidth]{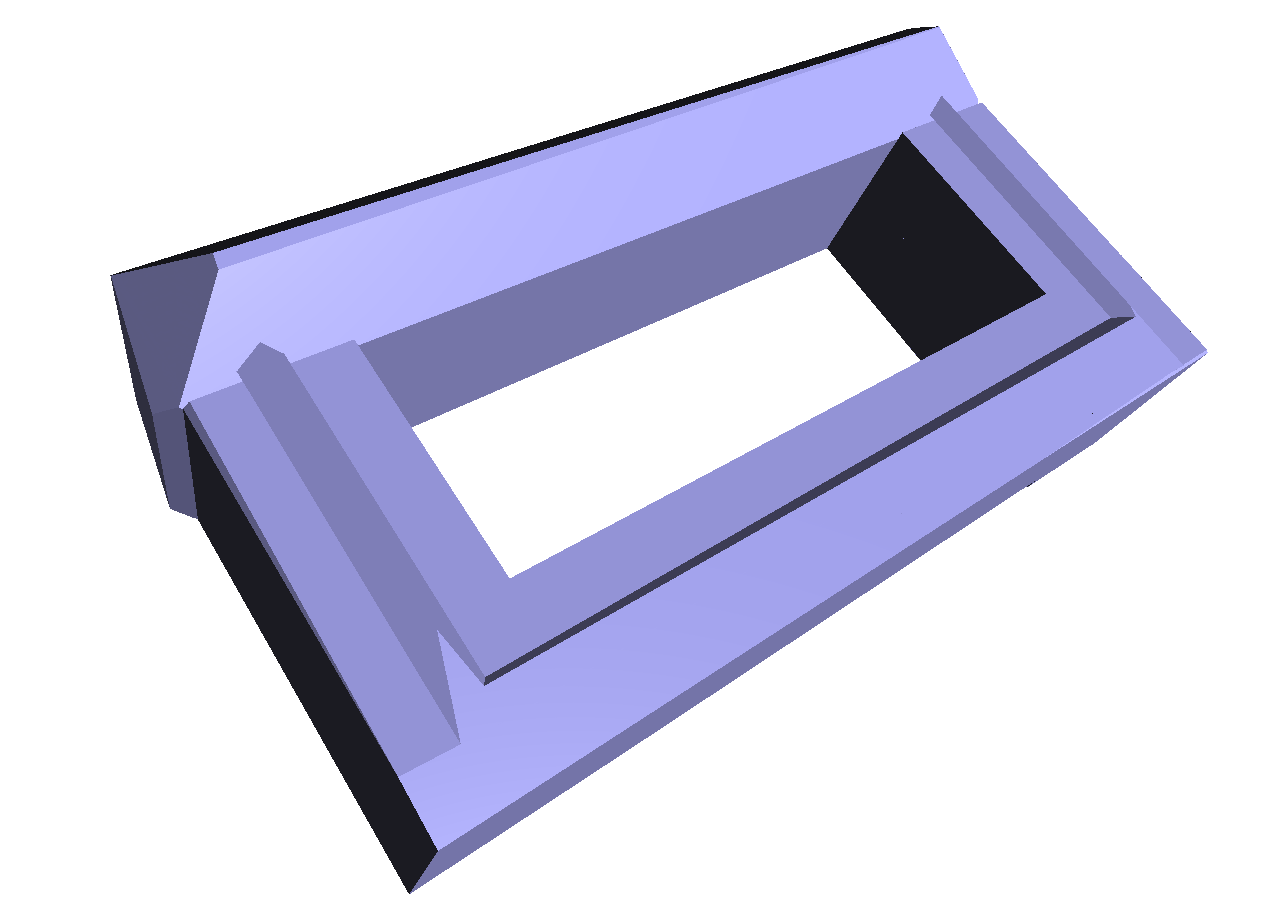}}
	\subfloat[$0.001R$]{\includegraphics[width=0.18\linewidth]{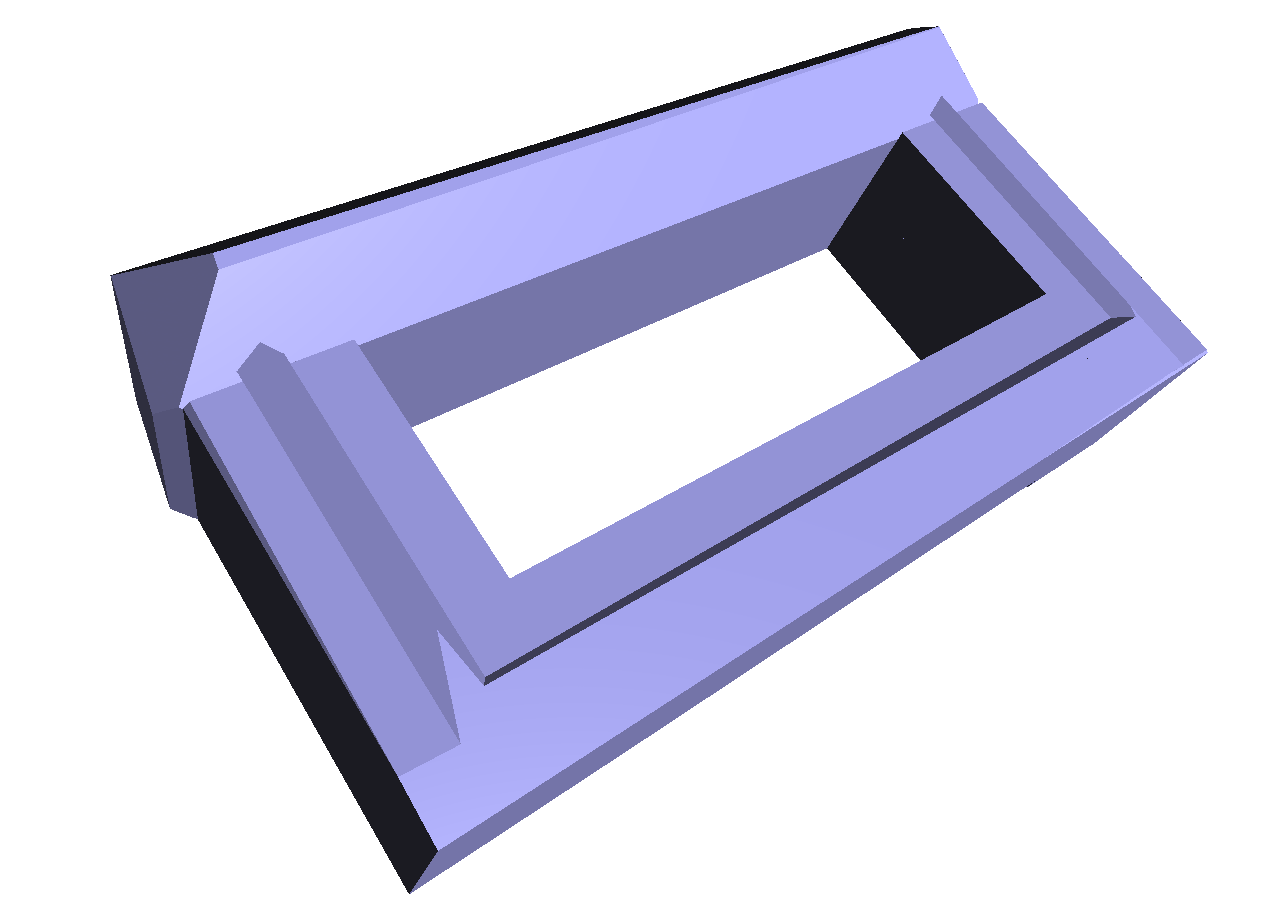}}
	\subfloat[$0.005R$]{\includegraphics[width=0.18\linewidth]{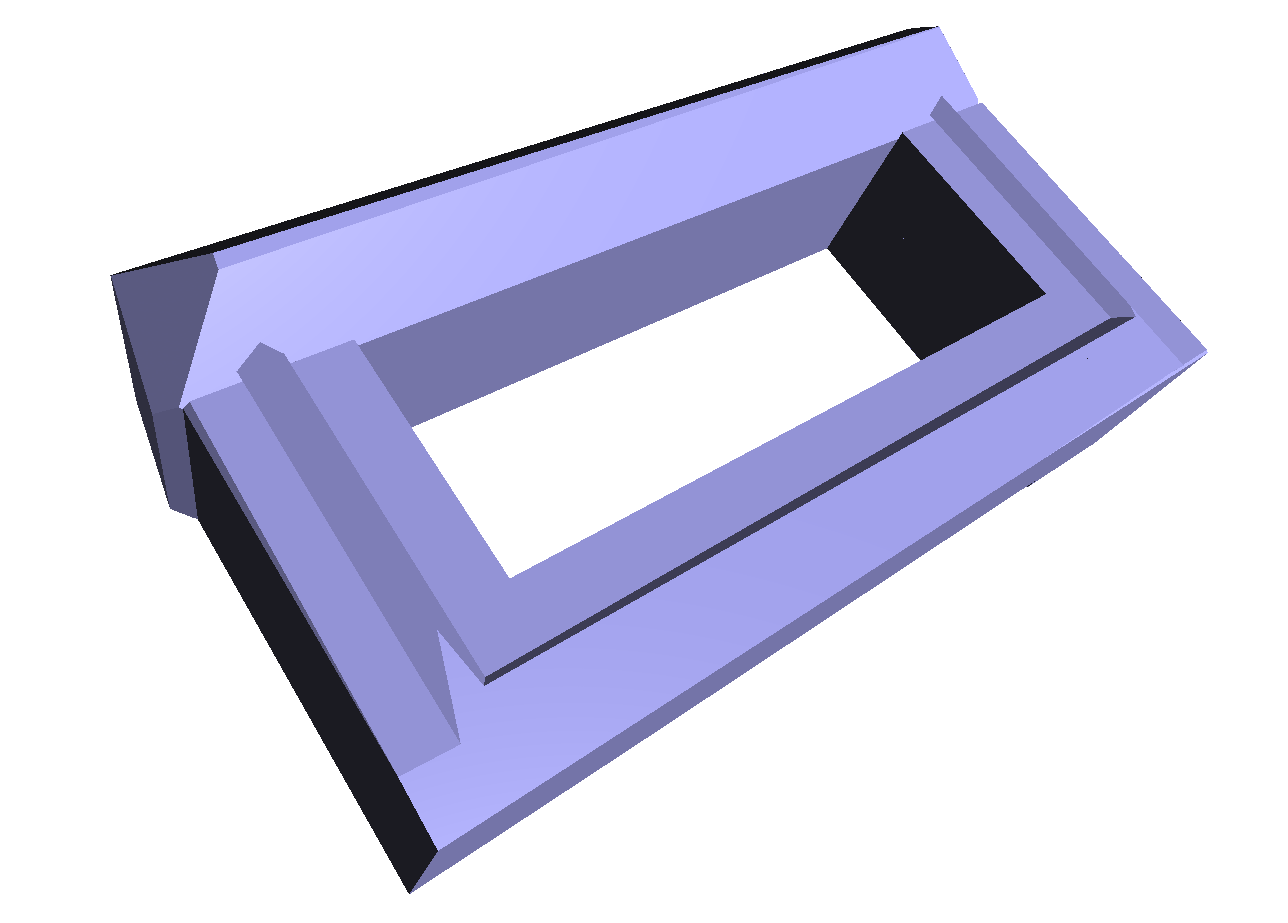}}
	\subfloat[$0.010R$]{\includegraphics[width=0.18\linewidth]{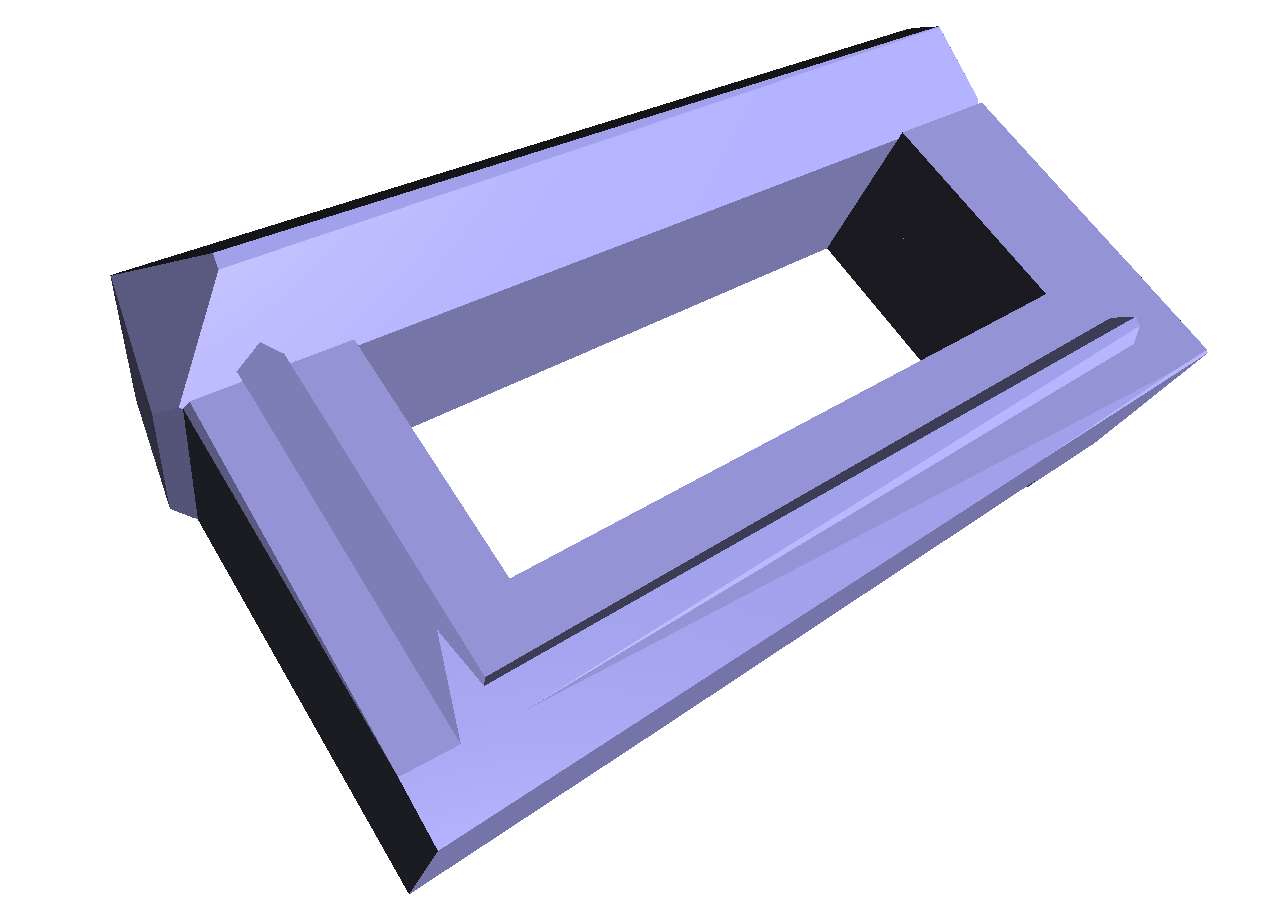}}
	\subfloat[$0.050R$]{\includegraphics[width=0.18\linewidth]{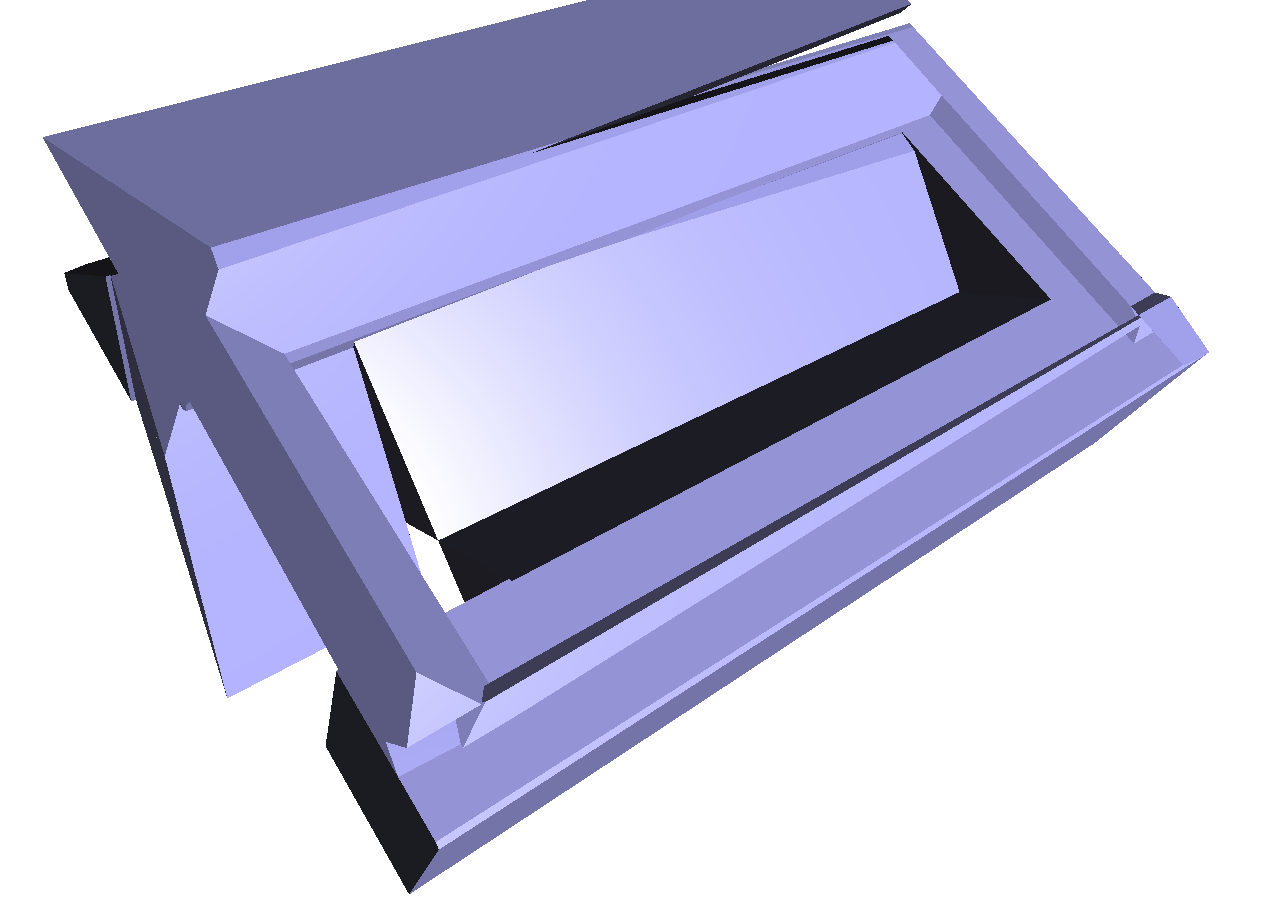}}
	
	\caption{Robustness to noise. Note that our neural network was trained on point clouds with different levels of noise in the range $[0, 0.005R]$, where $R$ denotes the radius of the bounding sphere of the input point cloud.}
	\label{fig:robustness_noise}
\end{figure}

To cope with noisy data, our network was trained on the synthetic point clouds with intentionally added different levels of noise. To evaluate the robustness of our method against noise, we tested our method on point clouds with increasing levels of Gaussian noise. \autoref{fig:robustness_noise} demonstrates such an example.
Though our method was trained on point clouds with low noise levels in the range $[0, 0.005R]$, it produced reliable reconstruction for point clouds with surprisingly higher levels of noise (i.e., until $0.05R$, which is equivalent to measurement errors of as high as 5 meters for a 100 meter-sided building). 

Since our neural network takes the global subsample of the point cloud into account in occupancy learning, it captures the global structures of the buildings, making it robust to incomplete input.
Examples of such reconstruction on the \textit{Helsinki no-bottom} point clouds are shown in~\autoref{fig:results_nobottom} and on the \textit{Shenzhen} point clouds shown in \autoref{fig:results_shenzhen}. In these examples, the ground planes in the input point clouds are completely missing, and the lower parts of some buildings in the \textit{Shenzhen} dataset are also occluded. With such incomplete input, our method successfully produced complete 3D models.

\subsubsection{Effect of parameter $\lambda$}

\begin{figure}[!t]
	\centering
	\includegraphics[width=0.25\linewidth]{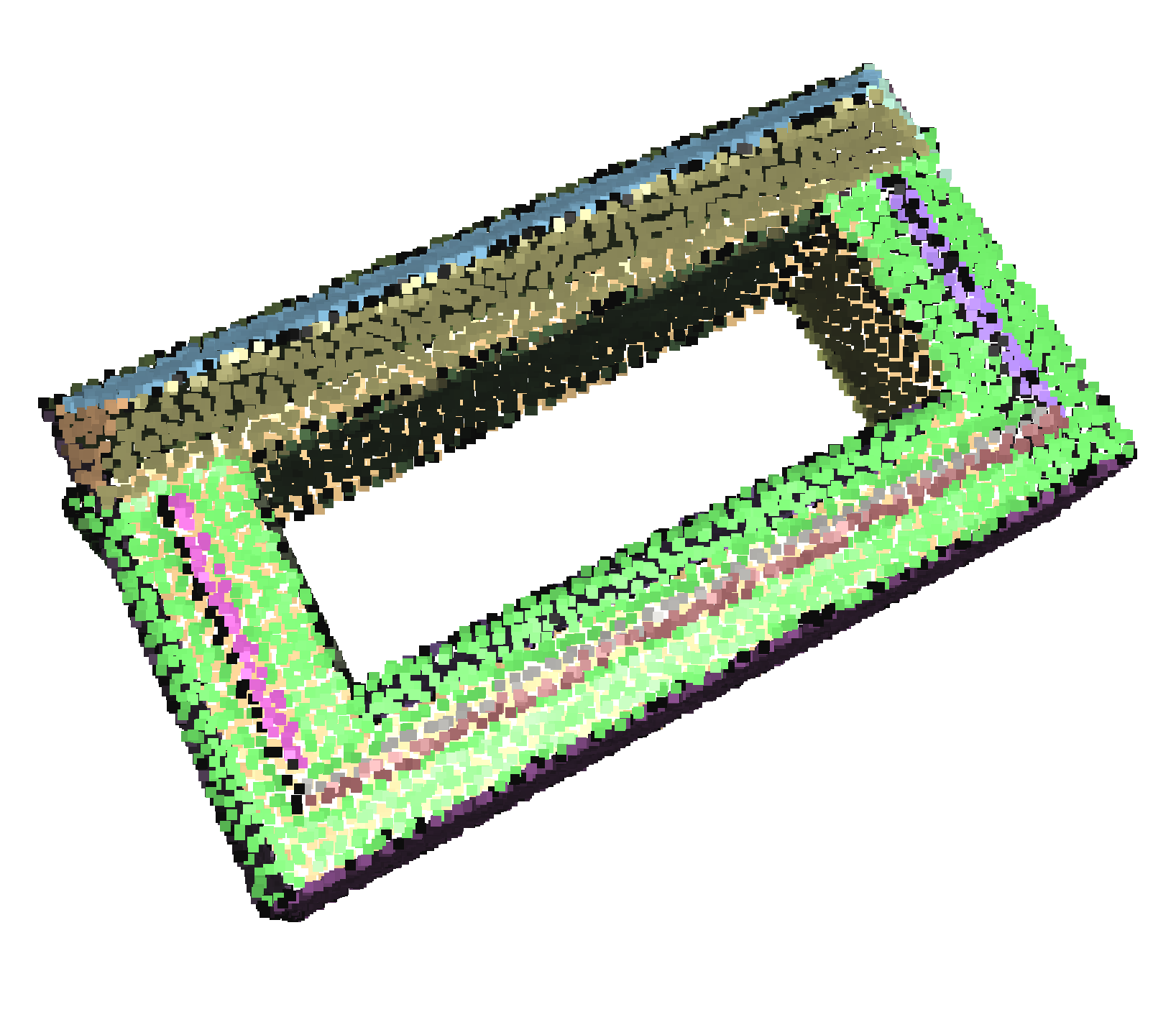}
	\includegraphics[width=0.25\linewidth]{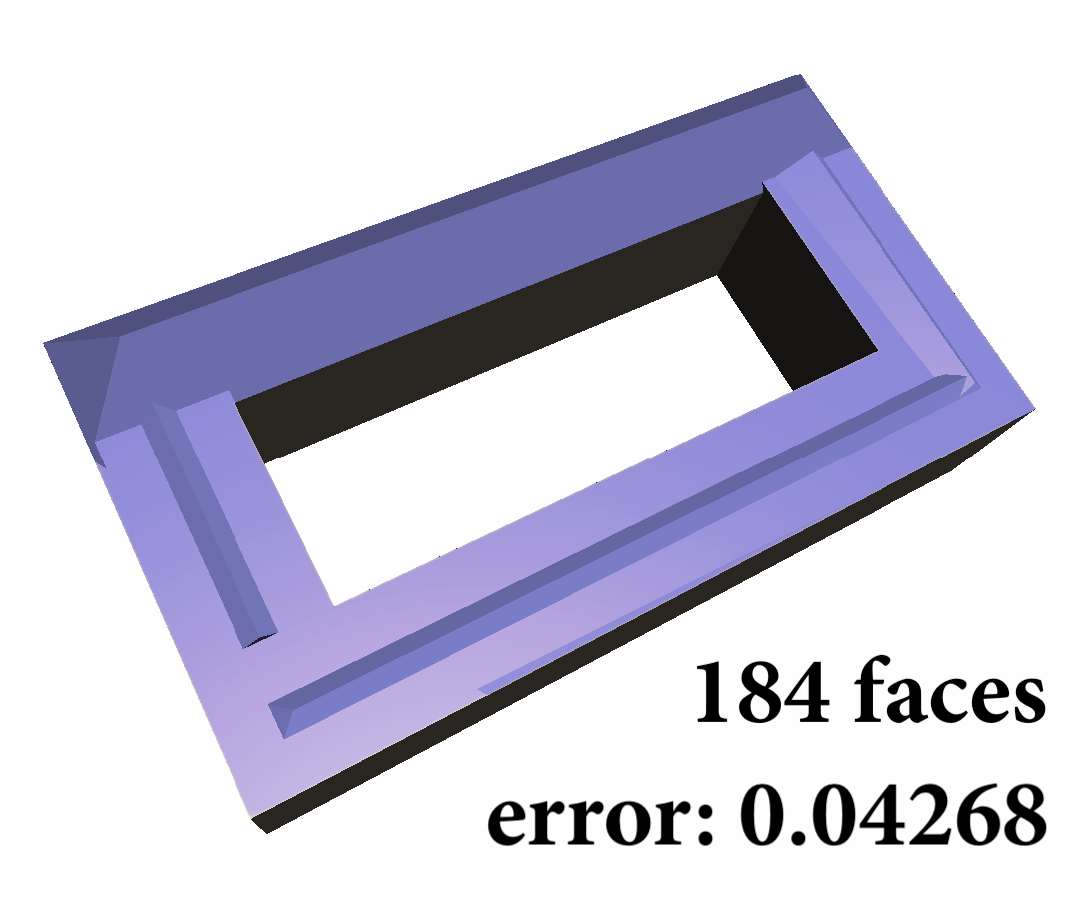}
	\includegraphics[width=0.25\linewidth]{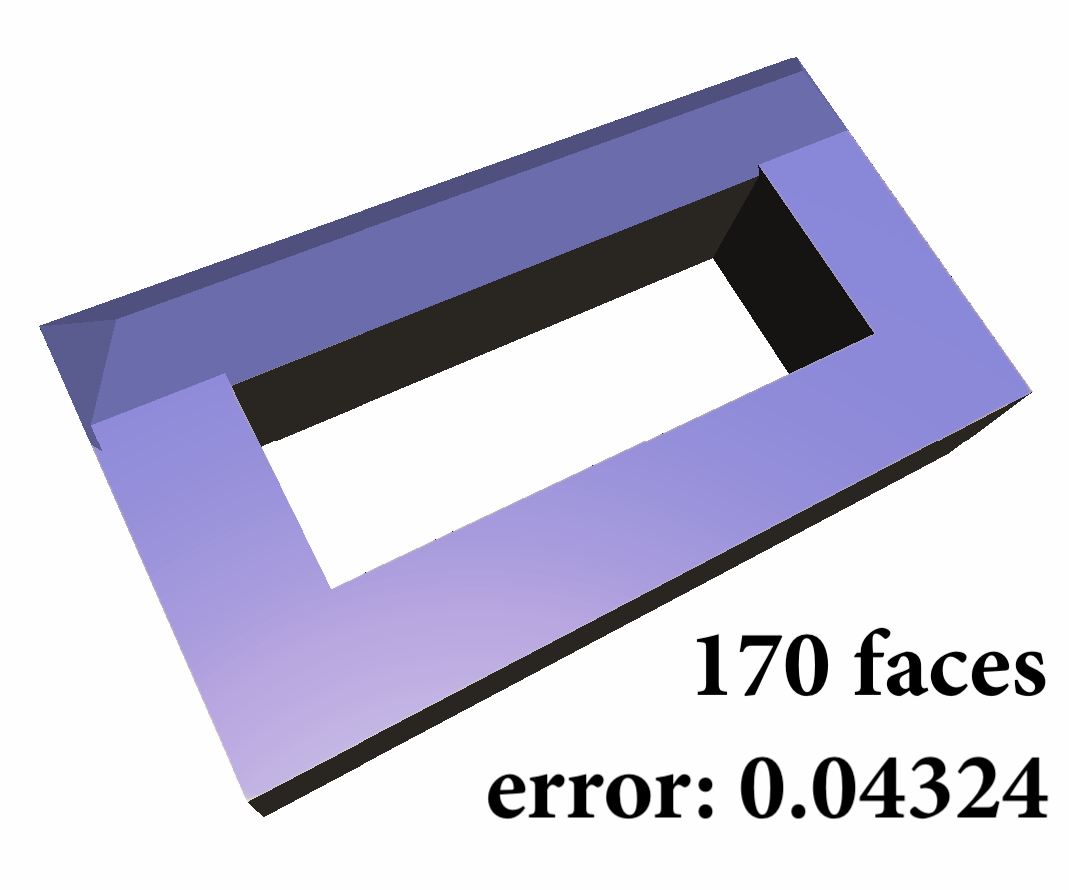}

	\subfloat[Input]{\includegraphics[width=0.25\linewidth]{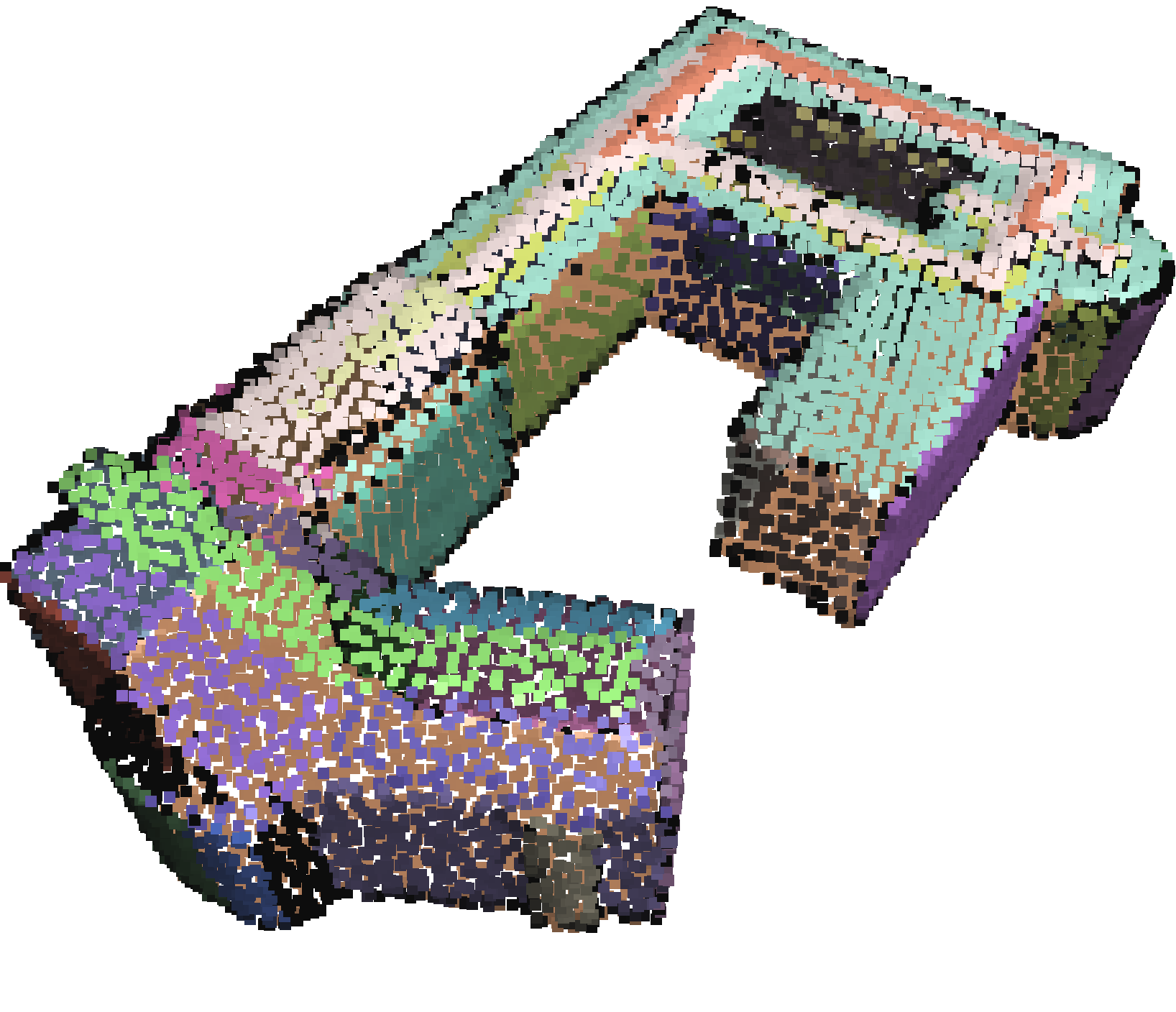}}
	\subfloat[$\lambda = 0.002$]{\includegraphics[width=0.25\linewidth]{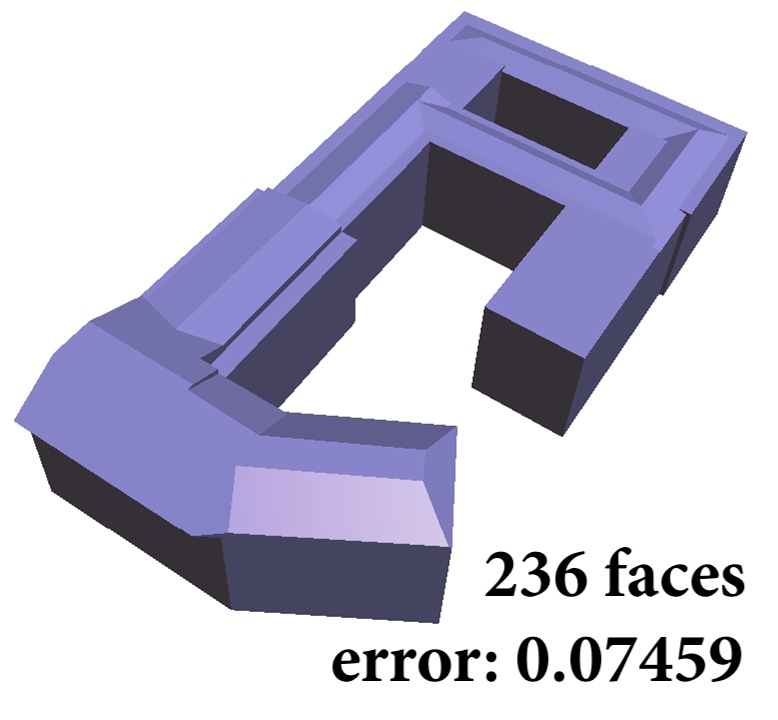}}
	\subfloat[$\lambda = 0.01$]{\includegraphics[width=0.25\linewidth]{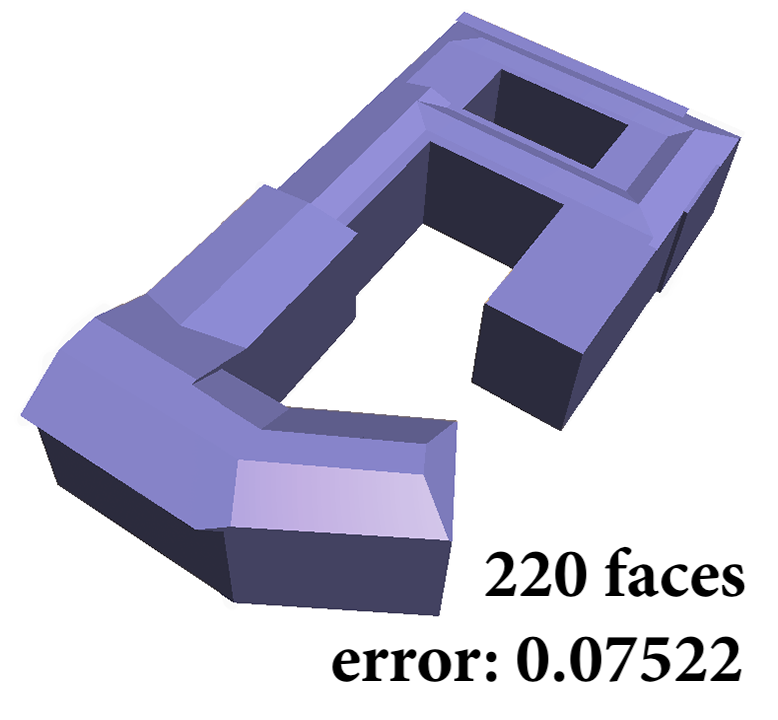}}
	
	\caption{The effect of parameter $\lambda$ on the final reconstruction. Increasing $\lambda$ leads to more compact surface models with fewer faces but potentially larger geometric errors.}
	\label{fig:lambda}
\end{figure}

The parameter $\lambda$ in \autoref{eq:energy} weights the complexity term formulated in the MRF for surface extraction from the cell complex, which controls the complexity of the final surface model. 
Specifically, increasing $\lambda$ leads to a more compact surface with fewer faces, while too large values of $\lambda$ may result in over-simplified surface models. 
\autoref{fig:lambda} demonstrates the effect of $\lambda$ on the final reconstruction. 
In all other experiments, $\lambda$ was set to 0.001, which was chosen by trial and error.

\subsection{Limitations}

\begin{figure}[t]
	\centering
	
	\includegraphics[width=0.22\linewidth]{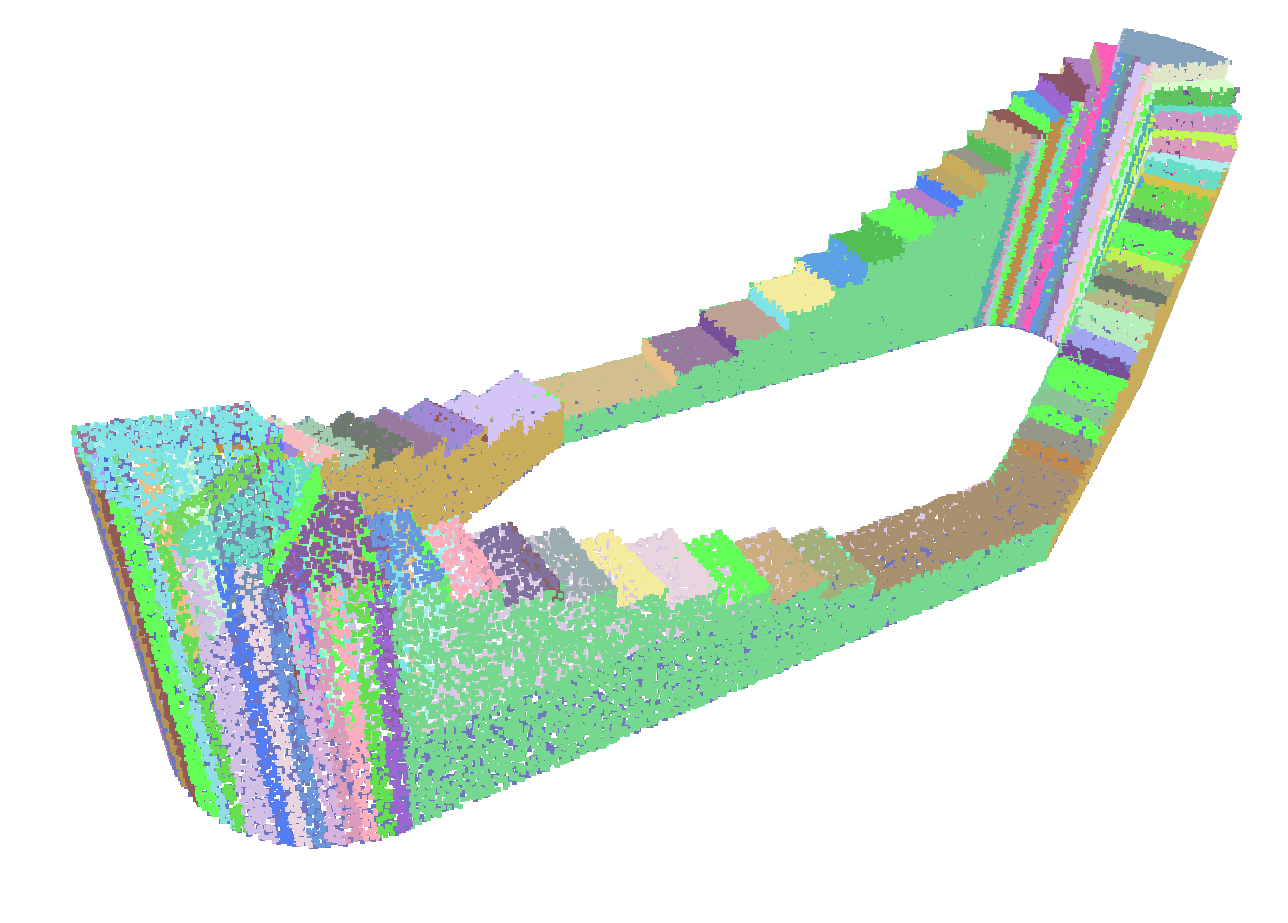}
%	\hspace{-1em}
	\includegraphics[width=0.22\linewidth]{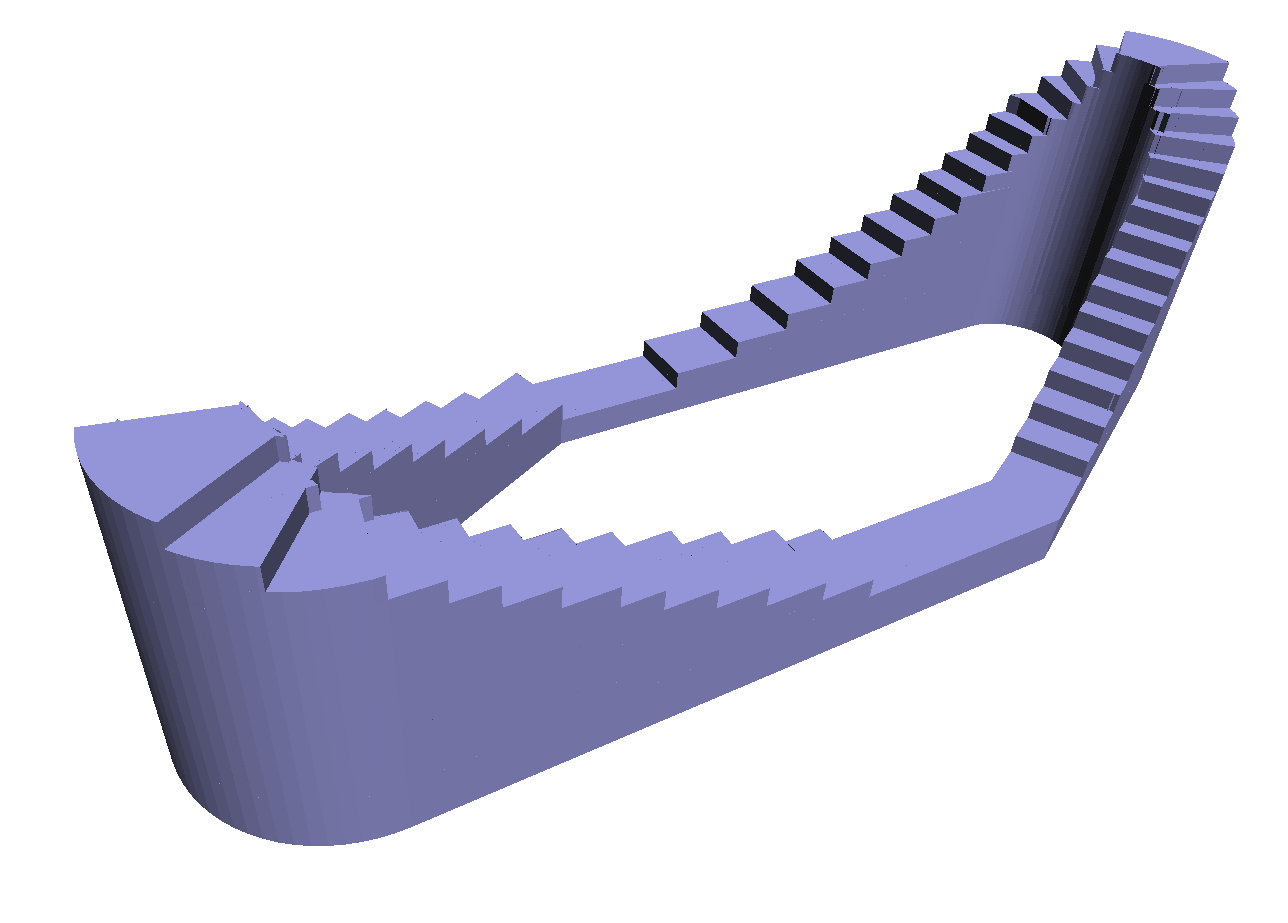}
	\hspace{2em}
	\subfloat{\includegraphics[width=0.22\linewidth]{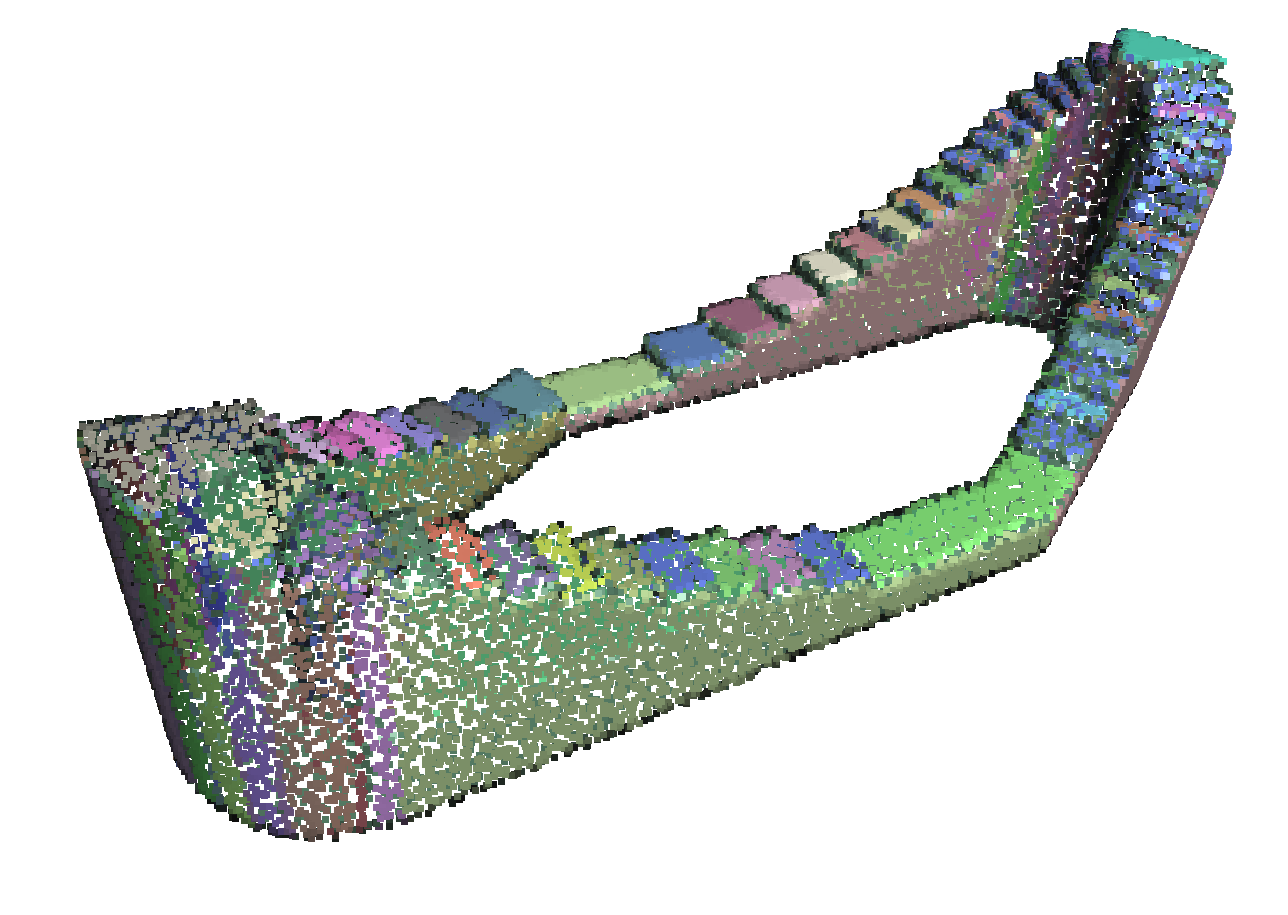}}
%	\hspace{-1em}
	\subfloat{\includegraphics[width=0.22\linewidth]{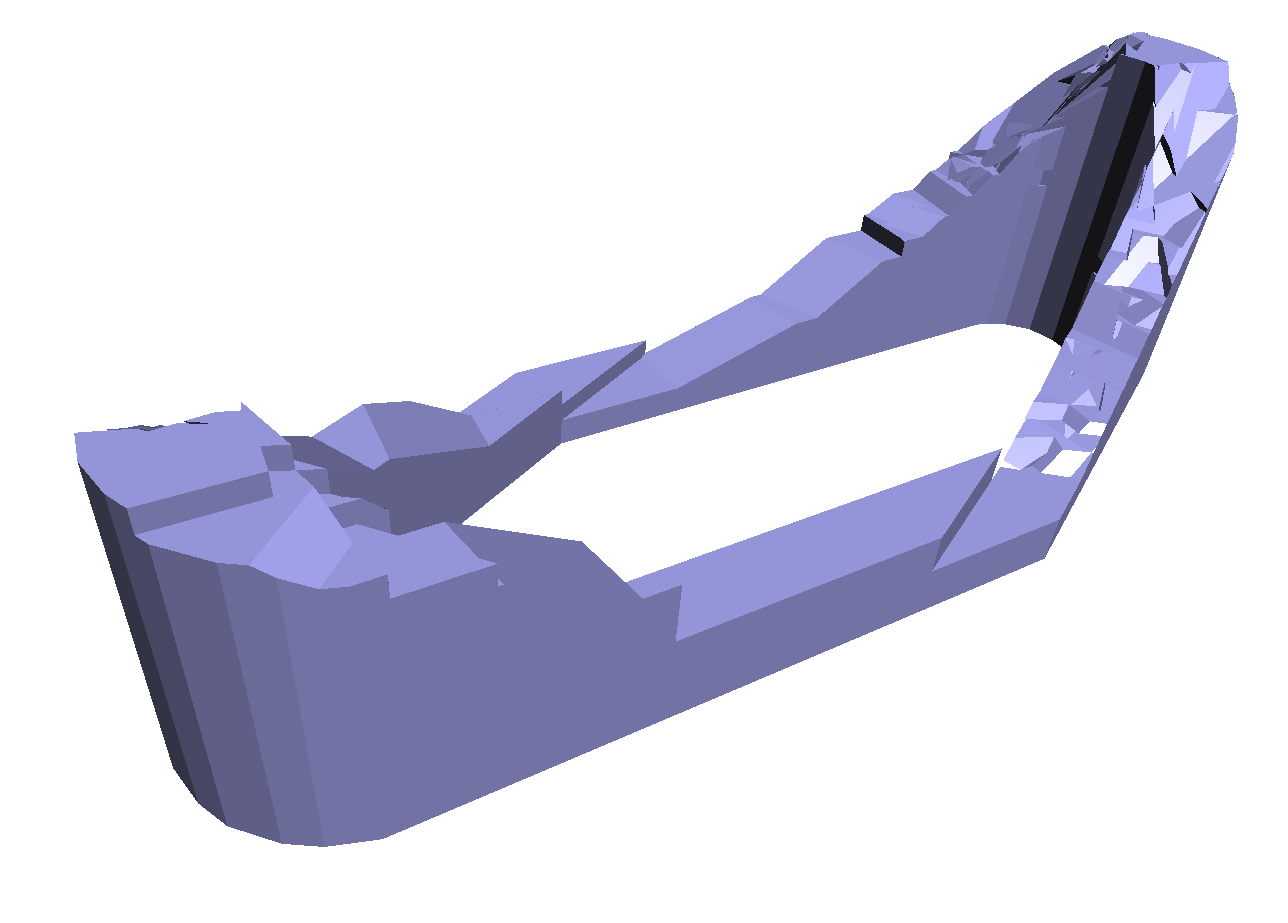}}
	
	\caption{The impact of the quality of detected planar primitives on the final reconstruction. Left: given accurately detected planar primitives, our method reconstructs high-quality building models. 
	Right: the reconstruction result from inaccurately detected planar primitives.}
	\label{fig:stairs_example}
\end{figure}

Similar to existing piecewise-planar object reconstruction methods such as PolyFit~\citep{nan2017polyfit}, our work focuses on assembling the initially detected planar primitives into compact polygonal building models, rather than on detecting the planar primitives. 
We thus have assumed that dominant planes (e.g., walls and roofs) can be detected from input point clouds. 
However, this is not always possible for noisy or incomplete scans. 
In our work, though a few steps are designed to mitigate the inaccuracy from primitive detection, including primitive refinement and the complexity term in the MRF formulation, it may still fail when the provided primitives are not complete or have large errors.
\autoref{fig:stairs_example} shows an example of how the quality of planar primitives impacts the reconstruction.

%\textbf{The `caved' artefact}. Though the complexity term in our MRF formulation penalises irregularity on the surface, the `caved' artefact can remain because a shrunk surface would have a smaller area which is favoured towards the goal of the optimisation (see \autoref{fig:caved}). Due to the formulation of MRF that is only composed of a unary term (i.e., energy defined on each polyhedron) and pairwise term (i.e., energy defined between polyhedra pairs), no global constraints can be imposed such as the total number of faces defined by \citet{nan2017polyfit}. The `caved' artefact can be mitigated with a carefully chosen $\lambda$ value.

%\begin{figure}[ht!]
%	\centering
%	\subfloat[]{\includegraphics[height=0.28\linewidth]{figures/results/limitations/caved_a.png}}
%	\subfloat[]{\includegraphics[height=0.28\linewidth]{figures/results/limitations/caved_b.png}}
%	\caption[Formulation of the `caved' artefact]{Formulation of the `caved' artefact: area of a `caved' surface (b) is smaller than a complete one (a).}
%	\label{fig:caved}
%\end{figure}

\section{Conclusion}

We have presented a novel efficient method for urban building reconstruction by exploiting the learned implicit representation as an occupancy indicator for explicit geometry extraction. 
With our occupancy learning strategy and the MRF formulation, we demonstrate that high-quality building models can be reconstructed with significant advantages in terms of fidelity, compactness, and computational efficiency.
To the best of our knowledge, this is the first work where a deep implicit field is explored for building reconstruction.

In future work, we would like to extend the current method to an end-to-end pipeline by incorporating the construction of explicit geometry into a neural network. 
In addition, our MRF-based surface extraction is efficient enough to allow interactive editing. We plan to integrate user interactions into the reconstruction pipeline to further improve the usability of the method for challenging reconstruction scenarios.

%% The Appendices part is started with the command \appendix;
%% appendix sections are then done as normal sections
%\newpage
%\appendix
%\section*{Appendix}

% \section{Sample Appendix Section}
% \label{sec:sample:appendix}
% Lorem ipsum dolor sit amet, consectetur adipiscing elit, sed do eiusmod tempor section \ref{sec:sample1} incididunt ut labore et dolore magna aliqua. Ut enim ad minim veniam, quis nostrud exercitation ullamco laboris nisi ut aliquip ex ea commodo consequat. Duis aute irure dolor in reprehenderit in voluptate velit esse cillum dolore eu fugiat nulla pariatur. Excepteur sint occaecat cupidatat non proident, sunt in culpa qui officia deserunt mollit anim id est laborum.

%% If you have bibdatabase file and want bibtex to generate the
%% bibitems, please use
%%

\bibliographystyle{elsarticle-harv} 
\bibliography{cas-refs}

%% else use the following coding to input the bibitems directly in the
%% TeX file.

% \begin{thebibliography}{00}

% %% \bibitem[Author(year)]{label}
% %% Text of bibliographic item

% \bibitem[ ()]{}

% \end{thebibliography}
\end{document}